\documentclass{article}

\usepackage{PRIMEarxiv}

\usepackage[authoryear]{natbib}
\usepackage[english]{babel} 
\usepackage{fullpage}
\usepackage{amsmath,amssymb,amscd,amsfonts}
\usepackage{graphicx}
\usepackage{amsthm}
\usepackage{color}
\usepackage{upgreek,enumitem}
\usepackage{blindtext}
\usepackage{tabto}
\usepackage{algorithm}
\usepackage{algorithmic}
\usepackage{caption}
\usepackage{subcaption}
\usepackage{multirow}
\usepackage{booktabs}
\usepackage{float}
\usepackage{hyperref}

\usepackage{xcolor}

\definecolor{myblue}{RGB}{173,216,230} % Define your blue color
\definecolor{mygreen}{RGB}{144,238,144}
\definecolor{myred}{RGB}{255,160,122} % Light salmon

\DeclareMathOperator*{\concat}{%
    \mathchoice%
        {\Big\vert\Big\langle}%
        {\big\vert\big\langle}%
        {\vert\langle}%
        {\vert\langle}%
}

\newtheorem{theorem}{Theorem}[section]
\newtheorem{definition}{Definition}[section]

\newtheorem{lemma}{Lemma}[section]

\usepackage{array}
\newcolumntype{Q}[1]{>{\raggedright\let\newline\\\arraybackslash\hspace{0pt}}m{#1}}
\newcolumntype{U}[1]{>{\centering\let\newline\\\arraybackslash\hspace{0pt}}m{#1}}
\newcolumntype{S}[1]{>{\raggedleft\let\newline\\\arraybackslash\hspace{0pt}}m{#1}}

\usepackage[english]{babel}
\addto\extrasenglish{
  
}

\usepackage[english]{babel}
\addto\extrasenglish{
  
}

\usepackage[english]{babel}
\addto\extrasenglish{
  
}
\usepackage[english]{babel}
\addto\extrasenglish{
  
}

\title{Stabilizing Multimodal Autoencoders: A Theoretical and Empirical Analysis of Fusion Strategies}

\author{
  Diyar Altinses, Andreas Schwung \\
  Automation Technology and Learning Systems \\
  South Westphalia University of Applied Sciences \\
  Soest, 59494 Germany\\
}

\begin{document}
\maketitle

\begin{abstract}
In recent years, the development of multimodal autoencoders has gained significant attention due to their potential to handle multimodal complex data types and improve model performance. Understanding the stability and robustness of these models is crucial for optimizing their training, architecture, and real-world applicability. This paper presents an analysis of Lipschitz properties in multimodal autoencoders, combining both theoretical insights and empirical validation to enhance the training stability of these models. We begin by deriving the theoretical Lipschitz constants for aggregation methods within the multimodal autoencoder framework. We then introduce a regularized attention-based fusion method, developed based on our theoretical analysis, which demonstrates improved stability and performance during training. Through a series of experiments, we empirically validate our theoretical findings by estimating the Lipschitz constants across multiple trials and fusion strategies. Our results demonstrate that our proposed fusion function not only aligns with theoretical predictions but also outperforms existing strategies in terms of consistency, convergence speed, and accuracy. This work provides a solid theoretical foundation for understanding fusion in multimodal autoencoders and contributes a solution for enhancing their performance.
\end{abstract}

% keywords can be removed

\section{Introduction}

In industrial applications, trust and accuracy are paramount since even minor inaccuracies in high-volume production can lead to significant financial losses. In sectors like automotive, aerospace, and pharmaceuticals, for instance, the tolerance for error is minimal, as the cost of production defects or non-conforming products can lead to costly recalls, compliance issues, or even safety risks \citep{marucheck2011product}. Hence, ensuring production quality requires rigorous quality control and precise testing methods with high standards of accuracy and stability \citep{bertolini2021machine}. 

To meet these high standards of accuracy and stability, there is an increasing reliance on machine-learning models capable of processing large volumes of data from various stages of production and environmental factors. These models facilitate real-time anomaly detection, predictive analytics, and automated decision-making in manufacturing processes, enabling faster responses to potential quality issues \citep{gourisaria2021application}. However, achieving such precision with machine learning requires overcoming several challenges. These include training the models on accurate, representative datasets to avoid bias and implementing robust systems \citep{emmanouilidis2019enabling}. Moreover, these machine-learning models must be deployed in environments that are often dynamic and noisy, where variations in temperature, humidity, and equipment wear can impact production consistency.

A promising approach to improve the cost-effectiveness and accuracy of machine learning in industrial settings is the use of multimodal data. Multimodal models integrate data from various sources, such as camera and time-series data, to capture complementary and marginal information that enables a more comprehensive view of the production environment. This combination of diverse data inputs allows the model to detect patterns and anomalies that might be missed when relying on a single data source. This diversity strengthens the model’s ability to make accurate predictions and adapt to varying conditions on the production floor, reducing the risk of errors and enhancing overall system reliability. Consequently, multimodal approaches are particularly beneficial in complex industrial processes where the integration of multiple data sources leads to improved fault detection, predictive maintenance, or quality control capabilities \citep{atrey2010multimodal}.

While researchers are increasingly utilizing multimodal data in machine learning, the datasets they often label as multimodal typically originate from sensors with similar data formats and structures. In real-world applications, however, data sources are far more diverse and require complex heterogeneous data integration techniques. This diversity introduces a need for specialized methods to combine information from varied sensors, each with distinct data types and structures. As a result, datasets featuring heterogeneous data must undergo feature extraction and preprocessing to ensure effective analysis and compatibility. This step is crucial for transforming raw, varied data into formats that models can interpret accurately, thereby enhancing robustness in real-world applications \citep{altinses2023deep}.

A central challenge arises in the effective fusion of different modalities. Many researchers employ straightforward concatenation to retain information from both data sources and to prevent the loss of critical details \citep{ngiam2011multimodal, zeng2019deep}, while others sum data values for a more simplified fusion approach \citep{hazirbas2017fusenet}. Some researchers have proposed advanced fusion strategies, leveraging neural-based approaches to capture inter-modal dependencies dynamically \citep{nagrani2021attention}. Despite the growing use of multimodal data fusion in industrial machine learning, no study has yet performed a detailed mathematical analysis of fusion methods to understand the behavior and properties of fused latent representations in these applications. In this paper, we aim to bridge this gap by conducting a rigorous mathematical investigation of the Lipschitz properties of fusion methods for multimodal machine learning. Our approach leverages a multimodal autoencoder-based model for feature extraction, which allows us to systematically analyze and evaluate how different fusion strategies affect latent space representations. Additionally, we propose a novel regularized attention-based fusion mechanism that maintains the desired Lipschitz properties. We empirically underline our theoretical results on three multimodal industrial datasets of varying complexity, designed to simulate real-world production conditions with diverse data types. These datasets provide insights into the stability, accuracy, and robustness of each fusion method in different scenarios. Our contributions can be summarized as follows:

\begin{enumerate}
    \item We investigate the gradient dynamics of the multimodal autoencoder architecture, emphasizing the Lipschitz constant across fusion setups implemented through concatenation and summation techniques while exploring their broader implications.
    \item Building on our gradient analysis, we introduce a novel fusion strategy aimed at producing enhanced representations, leveraging dot product attention and regularization to effectively integrate multimodal features for improved performance.
    \item We employ three synthetic and one real-world multimodal datasets of robotic tasks within an industrial context to empirically validate our theoretical findings, highlighting improvements in both performance and stability outcomes.
\end{enumerate}

%%% structure
The paper is structured as follows: \autoref{sec:related} provides a comprehensive overview of existing research within the field of multimodal learning. In \autoref{sec:proof}, we present our theoretical setup and present our theoretical results, and propose in \autoref{sec:approach}, based on these results, a new fusion strategy for the multimodal autoencoder architecture. In the next \autoref{sec:experiments} we evaluate our theoretical findings and our approach by using three multimodal datasets and a multimodal autoencoder architecture. Lastly, \autoref{sec:conclusion} summarizes the contents of the paper and provides concluding remarks.

\textbf{Notation:} Key symbols and conventions employed in the mathematical formulations are defined as follows: Let \(\mathbf{a}\) represent a vector, and \(\mathbf{a}_{i}\) denote an indexed vector for a specific purpose. The \(i\)-th entry of the vector \(\mathbf{a}\) is denoted by \(a_{i}\), while \(\mathbf{a}^{(k)}\) refers to the vector \(\mathbf{a}\) associated with the \(k\)-th modality. Similarly, \(\mathbf{A}\) signifies a matrix, and \(\mathbf{A}_{ij}\) is an indexed matrix for particular purposes. The \((i, j)\)-th entry of the matrix \(\mathbf{A}\) is given by \(A_{ij}\), and \(\mathbf{A}^{(k)}\) indicates the matrix \(\mathbf{A}\) corresponding to the \(k\)-th modality. The notation \(\mathbf{A}^\top\) is used to represent the transposed matrix, while \(\mathbf{I}\) stands for the identity matrix. The matrix \(\mathbf{J}^{pq}\) refers to a single-entry matrix, with a value of 1 at the position \((i, j)\) and zeros elsewhere. The matrix norm is denoted by \(||\cdot ||\), with subscripts indicating the specific norm being used. The symbol \(\partial\) represents partial derivatives, and \(\nabla_\mathbf{a}\) denotes the gradient with respect to \(\mathbf{a}\). The supremum of a set is indicated by the term "sup." Additionally, \(\mathcal{D}\) represents the dataset, while \(\mathcal{L}\) signifies the loss function. The variable \(M\) indicates the number of modalities, \(\biguplus\) is used for aggregation of scalars, vectors, or matrices mapping from \(\mathbb{R}^a\) to \(\mathbb{R}\), and \(\concat\) denotes the concatenation of scalars, vectors, or matrices from \(\mathbb{R}^a\) and \(\mathbb{R}^a\) to produce \(\mathbb{R}^{a+b}\). 

%%%%%%%%%%%%%%%%%%%%%%%%%%%%%%%%%%%%%%%%%%%%%%%%%%%%%%%%%%%%%%%%%%%%%%%%%%%%%%%%%%%
%%%%%%%%%%%%%%%%%%%%%%%%%%%%%%%%%%%%%%%%%%%%%%%%%%%%%%%%%%%%%%%%%%%%%%%%%%%%%%%%%%%
%%%%%%%%%%%%%%%%%%%%%%%%%%%%%%%%%%%%%%%%%%%%%%%%%%%%%%%%%%%%%%%%%%%%%%%%%%%%%%%%%%%
%%%%%%%%%%%%%%%%%%%%%%%%%%%%%%%%%%%%%%%%%%%%%%%%%%%%%%%%%%%%%%%%%%%%%%%%%%%%%%%%%%%

\section{Related Work}\label{sec:related}

Recently, the field of machine learning has seen significant advances in diverse application areas, including data reconstruction and multimodal learning. This section is dedicated to a thorough analysis of the relevant literature in the context of multimodal representation learning and multimodal fusion strategies, as well as the areas of robust learning.

\subsection{Robust Learning}

The concept of Lipschitz continuity has gained significant attention in the machine learning community, particularly for its implications in ensuring the robustness of models against perturbations and adversarial attacks. Several studies have explored the role of Lipschitz constants in various learning frameworks to enhance model stability.

Researchers have proposed Lipschitz regularization techniques to improve the robustness of neural networks. For instance, \citep{gouk2021regularisation} demonstrates that incorporating Lipschitz constraints can lead to more stable models that generalize better to unseen data. The researchers of \citep{pauli2021training} have investigated the architecture of Lipschitz neural networks and introduced a method for constructing networks with provable Lipschitz bounds, which enhances the reliability of model predictions. The work \citep{patrini2017making} highlights the role of Lipschitz continuity in making models robust to noisy labels, demonstrating that Lipschitz-regularized loss functions can mitigate the effects of label noise during training. The importance of Lipschitz continuity in unsupervised learning algorithms has been emphasized by \citep{kim2020lipschitz}, who show that enforcing Lipschitz conditions can lead to better clustering and representation learning outcomes.

Apart from the classical neural network architectures, some researchers have dealt with the connection between Lipschitz continuity and adversarial robustness, which has been well-established. Carlini et al. \citep{carlini2017towards} illustrate that models with lower Lipschitz constants exhibit reduced susceptibility to adversarial examples, leading to more secure neural networks. In generative adversarial networks, \citep{zhou2019lipschitz} shows that enforcing Lipschitz continuity in the generator and discriminator networks can lead to improved convergence properties and higher-quality generated samples.

These studies collectively underline the significance of Lipschitz continuity in developing machine learning models that are robust, stable, and capable of generalizing effectively to diverse datasets and applications. However, none of them address the control of Lipschitz properties of multimodal fusion methods for robust representation learning. With this work, we aim to fill this gap.

\subsection{Multimodal Representation Learning}

The integration of information from multiple modalities plays a pivotal role in multimodal learning frameworks, where the goal is to consolidate diverse data into a unified, compact representation that can be used for effective analysis and decision-making \citep{ramachandram2017deep}. Many multimodal learning algorithms employ autoencoders to achieve this compression, representing a substantial step forward in the field by enabling the seamless integration of complex data from varied sources \citep{ma2018deep}. 

In an early application of this concept, \citep{ngiam2011multimodal} developed a deep learning framework to produce joint representations of audio and video data, leveraging deep autoencoders along with Restricted Boltzmann Machines and deep networks. This combination allowed the model to capture inter-modality correlations, resulting in improved accuracy in tasks such as audiovisual speech classification. Expanding upon this approach, \citep{srivastava2012multimodal} adapted the deep Boltzmann machine structure to integrate data from multiple sources, including images and text, which enabled their model to learn shared representations. This shared latent space helped enhance performance on retrieval tasks across modalities, such as image-to-text and text-to-image retrieval. Further extending the multimodal framework, Yang et al. proposed a multimodal autoencoder tailored for pose estimation, designed to process depth maps, heat maps, point clouds, and RGB data concurrently. By learning a joint latent space, their approach captured the unique and complementary features from each modality, improving pose estimation accuracy \citep{yang2019aligning}. Similarly, \citep{suzuki2016joint} introduced variational multimodal autoencoders, which adapt the traditional variational autoencoder to handle multiple modalities. This model generates a probabilistic latent space that aligns with the joint distribution of all input modalities, thereby enhancing both representation learning and the generation of new multimodal data. Lastly, \citep{bouchacourt2018multi} contributed to this domain by focusing on the disentanglement of latent spaces within multimodal autoencoders. By separating shared and modality-specific factors, their model improved both the interpretability and robustness of learned representations, benefiting tasks like cross-modal retrieval and generation by enhancing clarity in how data across modalities are interconnected. 

While these studies showcase the success of multimodal representation learning, there remains a gap in analyzing its mathematical properties. In this paper, we address this gap by conducting a mathematical analysis of multimodal autoencoder stability with the Lipschitz constant. Furthermore, we examine several factors that impact both performance and stability, providing a thorough evaluation of the theoretical foundations to enhance understanding of the underlying mechanisms.

\subsection{Multimodal Fusion Strategies}

In multimodal learning systems, integrating sensor data through fusion plays a crucial role in enabling algorithmic self-correction mechanisms. The specific point in the processing chain where this correction is applied can greatly impact the system's effectiveness. Researchers have outlined three main fusion approaches based on the fusion location: feature-level fusion, decision-level fusion, and intermediate fusion \citep{brena2020choosing} \citep{9190246} \citep{kullu2022deep}. %\citep{bruni2014multimodal} \citep{papadakis2013terrain} \citep{atrey2010multimodal}   \citep{ramachandram2017deep}. 

In addition to these methods, \citep{baltruvsaitis2018multimodal} and \citep{barua2023systematic} proposed a broader taxonomy of multimodal learning strategies, categorizing them as Model-based, Translation-based, Alignment-based, and Co-learning approaches. Our study centers on model-based fusion strategies, particularly in the aggregation of multimodal representations. This aggregation is commonly performed through summation or concatenation \citep{ngiam2011multimodal} \citep{kiela2014learning} \citep{baltruvsaitis2018multimodal}. Summing these representations reduces the dimensionality of the feature space, optimizing computational efficiency while still effectively merging complementary data from each modality to produce an integrated and resilient representation. Alternatively, feature concatenation provides a comprehensive feature vector by preserving the dimensionality of each modality, which is particularly advantageous when combining data sources of differing dimensions. This approach retains unique characteristics from each modality, ensuring that no critical data is averaged out, making it valuable for tasks where each modality brings distinct, useful information to the analysis \citep{liu2018towards} \citep{liu2021comparing}.

To our knowledge, there is currently no research that offers a detailed mathematical analysis of the fundamental properties of aggregation functions used in data-driven methods. Moreover, existing studies have not thoroughly examined the training processes associated with these models. This lack of in-depth exploration underscores the necessity for further research aimed at developing a theoretical framework to inform the design and optimization of aggregation functions within multimodal architectures.

%%%%%%%%%%%%%%%%%%%%%%%%%%%%%%%%%%%%%%%%%%%%%%%%%%%%%%%%%%%%%%%%%%%%%%%%%%%%%%%%%%%
%%%%%%%%%%%%%%%%%%%%%%%%%%%%%%%%%%%%%%%%%%%%%%%%%%%%%%%%%%%%%%%%%%%%%%%%%%%%%%%%%%%
%%%%%%%%%%%%%%%%%%%%%%%%%%%%%%%%%%%%%%%%%%%%%%%%%%%%%%%%%%%%%%%%%%%%%%%%%%%%%%%%%%%
%%%%%%%%%%%%%%%%%%%%%%%%%%%%%%%%%%%%%%%%%%%%%%%%%%%%%%%%%%%%%%%%%%%%%%%%%%%%%%%%%%%

\section{Lipschitz properties of Multimodal Autoencoders} \label{sec:proof}

Analyzing the Lipschitz continuity of the gradients is essential for several reasons. It guarantees the convergence of gradient-based methods like gradient descent, allows for better parameter tuning, and ensures the stability and robustness of the optimization process \citep{patel2024gradient}. In the following sections, we investigate the Lipschitz properties of multimodal autoencoder architectures. Initially, we focus on the derivatives of the architectures. Subsequently, we analyze the sensitivity of the gradients of these models.

\subsection{Preliminaries} \label{sec:methods}

The Lipschitz constant $L$ quantifies the maximum rate of change of the function, with smaller values indicating smoother changes and greater stability. In constrained optimization, Lipschitz continuity helps maintain feasible solutions close to the optimum. Overall, it ensures efficient, stable, and robust optimization and is defined as follows \citep{mai2021stability}:

\begin{definition}
A function $ f $ is Lipschitz continuous if there exists a constant $ L $ such that for all $ x_1, x_2 $ in the domain:
\begin{align}
    \| f(x_1) - f(x_2) \| \leq L \| x_1 - x_2 \|
\end{align}
\end{definition}

The triangle inequality is crucial in proving Lipschitz continuity because it allows the decomposition of the difference between function values into manageable parts. In a normed space $\left(X,\|{\cdot }\|\right)$, the triangle inequality in the form $\|x+y\|\leq \|x\|+\|y\|$ is required as one of the properties that the norm must satisfy for all $x,y\in X$. Specifically, it also follows that ${\Big |}\|x\|-\|y\|{\Big |}\leq \|x\pm y\|\leq \|x\|+\|y\|$ as well as $\left\|\sum _{i=1}^{n}x_{i}\right\|\leq \sum _{i=1}^{n}\|x_{i}\|$ for all $x_{i}\in X$. In the special case of $L^p$ spaces, the triangle inequality is called Minkowski's inequality and is proven using Hölder's inequality. This decomposition allows us to relate the difference in function values directly to the difference in inputs, thereby effectively establishing the Lipschitz condition. Here, we assess the sensitivity of the autoencoder's gradients to variations in the input based on $\| \nabla f(x_1) - \nabla f(x_2) \| \leq L \| x_1 - x_2 \|$.

The multimodal autoencoder fusion paradigm is composed of $M \in \mathbb{N}$ individual autoencoders, each tailored for a specific modality, designed to facilitate resilient reconstruction. The primary objective of this architecture is to transform the $k$-th multimodal input $\mathbf{x}^{(i)}_k \in \mathbb{R}^p$ into a unified latent representation and subsequently reconstruct the original input by aggregating all modality representations while minimizing information loss and redundancy. Specifically, we aim to project the data points into a $q$-dimensional subspace $\mathbb{R}^q$ (where $q < Mp$) such that the reconstruction error is minimized. Formally, this involves determining the mappings defined by the encoders $E^{(i)}: \mathbb{R}^p \rightarrow \mathbb{R}^q$, which are combined using an aggregation function $\biguplus$, as well as the decoders $D^{(i)}: \mathbb{R}^q \rightarrow \mathbb{R}^p$. The goal is to minimize the reconstruction loss, expressed as follows:
\begin{align}
    \min \mathcal{L}(\mathbf{x}_k) = \min_{D, E} \sum_{k=1}^N  \sum_{i=1}^M \left|\left| \mathbf{x}_k^{(i)} - D^{(i)} \left(\biguplus_{n=1}^M  \left( E^{(n)} (\mathbf{x}_k^{(n)}) \right)\right) \right|\right|.
\end{align}

\subsection{Derivatives of Autoencoders Architectures}

% These derivations build on the foundational work of Altinses et al. \citep{altinses2024convex}. 
In this section, we derive the gradients for our multimodal autoencoder architectures, including two specific aggregation methods: summation and concatenation. These derivatives are then used to determine the Lipschitz constant to analyze the stability of the convergence process. Since an autoencoder comprises two parameter sets, the encoder and the decoder, we analyze both separately.

The derivative of the multimodal autoencoder w.r.t. to the parameters of the decoder $\theta_D^{(k)}$ is given by:
\begin{lemma}
    The loss function $\mathcal{L}(\mathbf{x}_k) = \sum_{i=1}^M \left|\left| \mathbf{x}_k^{(i)} - D^{(i)} \left(\biguplus_{n=1}^M  \left( E^{(n)} (\mathbf{x}_k^{(n)}) \right)\right) \right|\right|$, defined as the squared Euclidean reconstruction error of a multimodal autoencoder with $M$ encoder $E^{(i)}: \mathbb{R}^{p} \rightarrow \mathbb{R}^{q}$ and decoder $D^{(i)}: \mathbb{R}^{q} \rightarrow \mathbb{R}^{p}$, has the derivative with respect to any decoder parameters of 
    \begin{align*}
        \frac{\partial}{\partial \theta_D^{(k)}} \left\| \mathbf{x}^{(k)} - D^{(k)}(\mathbf{u}) \right\|^2 = -2 \cdot \left( \mathbf{x}^{(k)} - D^{(k)}(\mathbf{u}) \right)^\top \cdot \frac{\partial D^{(k)}(\mathbf{u})}{\partial \theta_D^{(k)}},
    \end{align*}
    for any $M\in \mathbb{N}$ and with $\mathbf{u} = \left(\biguplus_{n=1}^M  E^{(n)}( \mathbf{x}^{(n)}) \right)$. 
\label{lemma:derivative_decoder}
\end{lemma}
The sum vanishes because the only remaining term is the one involving $ D^{(k)} $. Thus, the derivative reduces to the same expression as in the unimodal case. Similarly, we calculate the derivative w.r.t. the encoder parameters:

\begin{lemma}
    The loss function $\mathcal{L}(\mathbf{x}_k) = \sum_{i=1}^M \left|\left| \mathbf{x}_k^{(i)} - D^{(i)} \left(\biguplus_{n=1}^M  \left( E^{(n)} (\mathbf{x}_k^{(n)}) \right)\right) \right|\right|$, defined as the squared Euclidean reconstruction error of a multimodal autoencoder with $M$ encoder $E^{(i)}: \mathbb{R}^{p} \rightarrow \mathbb{R}^{q}$ and decoder $D^{(i)}: \mathbb{R}^{q} \rightarrow \mathbb{R}^{p}$, has the derivative with respect to any encoder parameters of 
    \begin{align*}
        \frac{\partial \mathcal{L}(\mathbf{x})}{\partial \theta_E^{(k)}} = -2 \sum_{i=1}^M \left( \mathbf{x}^{(i)} - D^{(i)}(\mathbf{u}) \right)^\top \cdot \frac{\partial D^{(i)}(\mathbf{u})}{\partial \mathbf{u}} \cdot \frac{\partial \mathbf{u}}{\partial \theta_E^{(k)}}
    \end{align*}
    for any $M\in \mathbb{N}$ and with $\mathbf{u} = \left(\biguplus_{n=1}^M  E^{(n)}( \mathbf{x}^{(n)}) \right)$. 
\label{lemma:derivative_encoder}
\end{lemma}

This formulation reveals a key structural difference between encoder and decoder gradients: while the derivative w.r.t. decoder parameters ($ \theta_D^{(k)} $) is local, depending only on the $k$-th decoder's reconstruction error and its parameters, allowing independent updates per decoder, the encoder gradient (with respect to $\theta_E^{(k)}$) is influenced by all modalities simultaneously, since every decoder output depends on the shared latent representation $\mathbf{u}$, which is computed from all encoder outputs. Consequently, the encoder gradient captures the global reconstruction sensitivity of the entire model, making it inherently more entangled and complex than the decoder gradient. This also implies that ensuring stability or Lipschitz continuity in the encoder pathway is more challenging, as it must account for its indirect effect on every reconstruction pathway through the latent aggregation.

Given that \autoref{lemma:derivative_encoder} characterizes both the aggregated latent representation $\mathbf{u}$ and its derivative term appearing in $\frac{\partial D^{(i)}(\mathbf{u})}{\partial \mathbf{u}}$, we first analyze the fusion gradients by considering summation-based aggregation. Therefore, let $\mathbf{u} = \sum_{n=1}^M E^{(n)}(\mathbf{x}^{(n)})$, where each $E^{(n)}: \mathbb{R}^{d_n} \to \mathbb{R}^D$ is a differentiable encoder function parameterized by $\theta_E^{(n)}$, and $\mathbf{x}^{(n)}$ is the input to the $n$-th encoder. Then, for any $k \in \{1, \dots, M\}$, the derivative of $\mathbf{u}$ with respect to $\theta_E^{(k)}$ is:  
\begin{align}
    \frac{\partial \mathbf{u}}{\partial \theta_E^{(k)}} = \frac{\partial E^{(k)}(\mathbf{x}^{(k)})}{\partial \theta_E^{(k)}},
\end{align}
and all other derivatives $\frac{\partial \mathbf{u}}{\partial \theta_E^{(n)}}$ for $n \neq k$ are zero.   

For concatenation-based fusion, we analogously derive the gradient terms with respect to the encoder parameters $\theta_E^{(k)}$ in \autoref{lemma:derivative_concat}. This allows direct comparison of gradient behavior between summation and concatenation fusion approaches.

\begin{lemma}
    Let $\mathbf{u} = \left(\concat_{n=1}^M  E^{(n)}( \mathbf{x}^{(n)}) \right)$, where each $E^{(n)}: \mathbb{R}^{d_n} \to \mathbb{R}^{D_n}$ is a differentiable encoder function parameterized by $\theta_E^{(n)}$, and $\mathbf{x}^{(n)}$ is the input to the $n$-th encoder. The concatenated output $\mathbf{u}$ has dimension $\sum_{n=1}^M D_n$. Then, for any $k \in \{1, \dots, M\}$, the derivative of $\mathbf{u}$ with respect to $\theta_E^{(k)}$ is:  
    \begin{align*}
        \frac{\partial \mathbf{u}}{\partial \theta_E^{(k)}} = \left(\mathbf{0}_{D_1}, \dots, \mathbf{0}_{D_{k-1}}, \frac{\partial E^{(k)}(\mathbf{x}^{(k)})}{\partial \theta_E^{(k)}}, \mathbf{0}_{D_{k+1}}, \dots, \mathbf{0}_{D_M}\right),
    \end{align*}
where $\frac{\partial E^{(k)}}{\partial \theta_E^{(k)}}$ is a Jacobian of shape $D_k \times |\theta_E^{(k)}|$ (for vector $\theta_E^{(k)}$),  $\mathbf{0}_{D_n}$ denotes a zero matrix of shape $D_n \times |\theta_E^{(k)}|$, and the block structure aligns with the concatenation order.  
\label{lemma:derivative_concat}
\end{lemma}

The derivatives of multimodal encoder aggregations exhibit distinct structural properties depending on the aggregation method. For summation $\mathbf{u} = \sum_{n=1}^M E^{(n)}(\mathbf{x}^{(n)})$, the gradient with respect to $\theta_E^{(k)}$ isolates the $k$-th encoder’s Jacobian $\frac{\partial E^{(k)}}{\partial \theta_E^{(k)}}$, as other terms vanish due to parameter independence. This reflects local gradient flow, where updates to $\theta_E^{(k)}$ depend solely on $E^{(k)}$’s contribution. In contrast, concatenation $\mathbf{u} = \left(\concat_{n=1}^M  E^{(n)}( \mathbf{x}^{(n)}) \right)$ produces a block-sparse Jacobian, where $\frac{\partial \mathbf{u}}{\partial \theta_E^{(k)}}$ is zero everywhere except for the $k$-th block, preserving encoder-specific gradients in disjoint subspaces.

\subsection{Lipschitz Analysis of Autoencoder Gradients}\label{ssec:lipschitz}

Having a Lipschitz constant ensures that a machine learning model responds predictably and stably to input changes, which supports convergence during training and enhances robustness. A low Lipschitz constant further improves smoothness, noise resistance, and adversarial robustness by limiting how much the output can change with small input variations. This leads to better generalization, numerical stability, and safer deployment in sensitive applications. In the following \autoref{theorem:lipschitz_multimodal_autoencoder_D} and \autoref{theorem:lipschitz_multimodal_autoencoder_E}, we analyze the Lipschitz continuity of the autoencoder gradients to form a foundational step toward optimizing and deploying effective autoencoder architectures in practical scenarios.

\begin{theorem}
The gradient of the loss function $\mathcal{L}(\mathbf{x}_k) = \sum_{i=1}^M \left|\left| \mathbf{x}_k^{(i)} - D^{(i)} \left(\biguplus_{n=1}^M  \left( E^{(n)} (\mathbf{x}_k^{(n)}) \right)\right) \right|\right|$ with respect to the decoder parameters $\theta_D^{(k)}$, defined as the squared Euclidean reconstruction error of a multimodal autoencoder with $M$ encoder $E^{(i)}: \mathbb{R}^{p} \rightarrow \mathbb{R}^{q}$ and decoder $D^{(i)}: \mathbb{R}^{q} \rightarrow \mathbb{R}^{p}$, has the Lipschitz constant
\begin{align*}
    L_D^{(k)} = 2 \cdot \left[ B_{\text{grad}}^{(k)} + B_{\text{grad}}^{(k)} \cdot L_{D^{(k)}}^{(\text{func})} + C \cdot L_{D^{(k)}}^{(\text{grad})} \right] \cdot L_{\text{agg}} \cdot L_{E^{(i)}}^{(\text{func})}
\end{align*}
where $B_{\text{grad}}^{(k)}$ is an upper bound on the decoder gradient norm, $L_{D^{(k)}}^{(\text{func})}$ and $L_{E^{(i)}}^{(\text{func})}$ is the Lipschitz constant of the decoder and encoder function, $L_{D^{(k)}}^{(\text{grad})}$ is the Lipschitz constant of the decoder gradient, $C$ is a bound on the input norm $\| \mathbf{x}^{(k)} \|$, $L_{\text{agg}}$ is the Lipschitz constant of the aggregation method.
\label{theorem:lipschitz_multimodal_autoencoder_D}
\end{theorem}

\begin{theorem}
The gradient of the loss function $\mathcal{L}(\mathbf{x}_k) = \sum_{i=1}^M \left|\left| \mathbf{x}_k^{(i)} - D^{(i)} \left(\biguplus_{n=1}^M  \left( E^{(n)} (\mathbf{x}_k^{(n)}) \right)\right) \right|\right|$ with respect to the encoder parameters $\theta_E^{(k)}$, defined as the squared Euclidean reconstruction error of a multimodal autoencoder with $M$ encoder $E^{(i)}: \mathbb{R}^{p} \rightarrow \mathbb{R}^{q}$ and decoder $D^{(i)}: \mathbb{R}^{q} \rightarrow \mathbb{R}^{p}$, has the Lipschitz constant
\begin{align*}
    L_E^{(k)} = 2 \sum_{i=1}^M \left[L_{D^{(i)}}^{(\text{func})} \cdot B_{\text{agg}}^{(k)} +\left( L_{D^{(i)}}^{(\text{func})} \right)^2 \cdot B_{\text{agg}}^{(k)} \cdot L_{\text{agg}}^{(\text{func})} + C \cdot L_{D^{(i)}}^{(\text{grad})} \cdot B_{\text{agg}}^{(k)} + C \cdot L_{D^{(i)}}^{(\text{func})} \cdot L_{\text{agg}}^{(\text{grad},k)} \right]
\end{align*}
where $L_{D^{(i)}}^{(\text{func})}$ is Lipschitz constant of $D^{(i)}$, $L_{D^{(i)}}^{(\text{grad})}$ the Lipschitz constant of the gradients, $B_{\text{agg}}^{(k)}$ the norm of the derivative of the latent code with respect to the $k$-th parameter and $L_{\text{agg}}^{(\text{grad}, k)}$ its sensitivity to input changes, and $C$ the bounded norm of the input vectors $\mathbf{x}^{(i)}$.
\label{theorem:lipschitz_multimodal_autoencoder_E}
\end{theorem}

While the decoder Lipschitz constant $L_D^{(k)}$ involves the sum over all encoder Lipschitz constants, it only concerns a single decoder $D^{(k)}$. In contrast, the encoder Lipschitz constant $L_E^{(k)}$ aggregates over all decoders $D^{(i)}$, making the encoder responsible for reconstructing every modality. This means $L_E^{(k)}$ includes terms for all $M$ decoders, while $L_D^{(k)}$ depends on just one. Moreover, the encoder gradient involves derivatives of both the encoder and decoder, so its Lipschitz constant captures a deeper interaction between the encoder parameters and the entire reconstruction pipeline. As a result, $L_E^{(k)}$ is typically higher, and encoder training becomes more sensitive and potentially more unstable, especially in settings with many modalities. This makes the encoder the more challenging part to train.

The choice of aggregation method significantly impacts the Lipschitz properties of the resulting structure. Two common aggregation techniques are summation $\left\|\sum_k E^{(k)}(\mathbf{x}^{(k)})\right\|$ and vertical concatenation $\left\|\concat_k E^{(k)}(\mathbf{x}^{(k)})\right\|$, which are analyzed in the following section.

\subsection{Lipschitz Analysis of Aggregation Methods}

The impact of the aggregation methods is based on the gradients of the aggregation and on the aggregation itself. Comparing the Frobenius norms of the aggregated results under these two methods provides insights into their Lipschitz characteristics. This comparison leads to the following theorem, which establishes an inequality between the Frobenius norm of the summation and vertical concatenation.

\begin{theorem}
Let $\{E^{(i)} : \mathbb{R}^d \to \mathbb{R}^m\}_{i=1}^n$ be encoder functions with functional Lipschitz constants $L_{E^{(i)}}^{(\text{func})} \geq 0$. Define the concatenated function $E_{\mathrm{concat}}(x) = [E^{(1)}(x^{(1)}), \dots, E^{(n)}(x)^{(n)}] \in \mathbb{R}^{nm}$ and the summed function $E_{\mathrm{sum}}(x) = \sum_{i=1}^n E^{(i)}(x^{(i)}) \in \mathbb{R}^m$. Then their Lipschitz constants satisfy
\begin{align*}
    L_{\mathrm{concat}} = \left( \sum_{i=1}^n (L_{E^{(i)}}^{(\text{func})})^2 \right)^{1/2} \leq \sum_{i=1}^n L_{E^{(i)}}^{(\text{func})} = L_{\mathrm{sum}},
\end{align*}
with equality if and only if at most one $L_{E^{(i)}}^{(\text{func})}$ is nonzero.
\label{theorem: lipschitz_aggregation}
\end{theorem}

This implies that the Lipschitz constant of a multimodal autoencoder architecture that uses concatenation for aggregation is lower than that of an architecture that uses summation for aggregation unless at most one $\mathbf{x}_k$ is non-zero. Note that this assumes the matrix sums are identical when training with sum and concat as aggregations. As a result, the concatenation approach yields a more stable model with a lower potential for overfitting due to its lower Lipschitz constant. In contrast, the summation approach, with its higher Lipschitz constant, promotes less stability and robustness in the model, making it more sensitive to variations in the input data and potentially reducing its generalization capabilities. 

For both fusion methods, summation $\mathbf{u}_{\text{sum}} = \sum_{n=1}^M E^{(n)}(\mathbf{x}^{(n)})$ and concatenation\newline $\mathbf{u}_{\text{cat}} = \concat_{n=1}^M E^{(n)}(\mathbf{x}^{(n)})$, the gradient with respect to encoder parameters $\theta_E^{(k)}$ exhibits identical Lipschitz continuity. Specifically, when each encoder $E^{(n)}$ is $L_n$-Lipschitz in $\theta_E^{(k)}$, the gradient of either aggregated output satisfies:  
\begin{align*}
    \left\| \frac{\partial \mathbf{u}_{\text{agg}}}{\partial \theta_E^{(k)}}(\theta_1) - \frac{\partial \mathbf{u}_{\text{agg}}}{\partial \theta_E^{(k)}}(\theta_2) \right\| \leq L_k \|\theta_1 - \theta_2\|, 
\end{align*}
where $\text{agg} \in \{\text{sum}, \text{cat}\}$. This equivalence arises because the Jacobian $\frac{\partial \mathbf{u}_{\text{cat}}}{\partial \theta_E^{(k)}}$ reduces to $\frac{\partial E^{(k)}}{\partial \theta_E^{(k)}}$ (with zero-padding for other encoders), preserving the spectral norm. Thus, the local Lipschitz constant $L_k$ depends solely on the $k$-th encoder’s smoothness, irrespective of the aggregation method.  

Consequently, when designing multimodal autoencoders, the choice of fusion method significantly impacts the model's optimization and should be carefully considered based on the specific requirements and constraints of the application. Using concatenation to merge multimodal latent representations combines joint and marginal information but does not explicitly model interactions between modalities. Since the latent features remain unlinked, the decoder must implicitly infer these relationships, which can limit its ability to effectively leverage cross-modal dependencies. Additionally, traditional approaches to multimodal fusion, such as concatenation or sums, often fail to adequately capture complex cross-modal relationships. These methods typically treat modalities as independent and lack the ability to dynamically focus on the relevant information within and across modalities, which can lead to suboptimal performance.

%%%%%%%%%%%%%%%%%%%%%%%%%%%%%%%%%%%%%%%%%%%%%%%%%%%%%%%%%%%%%%%%%%%%%%%%%%%%%%%%%%%
%%%%%%%%%%%%%%%%%%%%%%%%%%%%%%%%%%%%%%%%%%%%%%%%%%%%%%%%%%%%%%%%%%%%%%%%%%%%%%%%%%%
%%%%%%%%%%%%%%%%%%%%%%%%%%%%%%%%%%%%%%%%%%%%%%%%%%%%%%%%%%%%%%%%%%%%%%%%%%%%%%%%%%%
%%%%%%%%%%%%%%%%%%%%%%%%%%%%%%%%%%%%%%%%%%%%%%%%%%%%%%%%%%%%%%%%%%%%%%%%%%%%%%%%%%%

\section{Lipschitz-Regularized Multimodal Attention Fusion}\label{sec:approach}

Unlike passive merging techniques, attention dynamically models cross-modal interactions by learning adaptive weights, ensuring that the fusion process prioritizes the most relevant features across modalities. While this adaptability enhances expressivity, it also raises critical questions about the stability of the fusion process, particularly how sensitive the output is to perturbations in multimodal inputs. A deeper mathematical understanding of the Lipschitz continuity of multimodal attention mechanisms provides insights into their sensitivity to input changes and guides the design of more stable models. 

\subsection{Multimodal Attention}

In this study, we employ a pairwise-averaged multimodal bilinear attention mechanism to model the interactions between $M$ input vectors, \(\mathbf{v}_1, \mathbf{v}_2, \dots, \mathbf{v}_M  \in \mathbb{R}^d\), which represent the flattened latents generated by the encoders $E^{(n)}$. This mechanism operates by projecting the input vectors into a shared representation space, calculating an attention score based on their similarity, and finally scaling the original vectors accordingly. The pairwise-averaged multimodal bilinear attention is defined as follows:

\begin{definition}
    \label{def:multimodal_attention}
    Let \(d, n \in \mathbb{N}\), and let \(\mathbf{v}_1, \ldots, \mathbf{v}_n \in \mathbb{R}^d\) be input vectors corresponding to \(n\) modalities, with learnable weight matrices \(\mathbf{W}_1, \ldots, \mathbf{W}_n \in \mathbb{R}^{d \times d}\).  
    For each pair of modalities \(i \neq j\), define the bilinear attention score $a_{ij} := (\mathbf{W}_i \mathbf{v}_i)^\top (\mathbf{W}_j \mathbf{v}_j) \in \mathbb{R}$. The pairwise-averaged multimodal bilinear attention mechanism is the mapping $\Phi: (\mathbb{R}^d)^n \to \mathbb{R}^{nd}$ with $\Phi(\mathbf{v}_1, \ldots, \mathbf{v}_n) = \left[ \alpha_1 \mathbf{v}_1 \,;\, \alpha_2 \mathbf{v}_2 \,;\, \ldots \,;\, \alpha_n \mathbf{v}_n \right]$, where each modality-specific attention coefficient \(\alpha_i \in \mathbb{R}\) is given by the average of its pairwise interactions $\alpha_i := \frac{1}{n-1} \sum_{\substack{j=1 \\ j \neq i}}^{n} a_{ij}$.
\end{definition}

In the pairwise-averaged bilinear attention mechanism, the input vectors
\(\mathbf{v}_1,\dots,\mathbf{v}_n\in\mathbb{R}^d\) are first linearly projected
into a shared representation space via learnable weight matrices
\(\mathbf{W}_1,\dots,\mathbf{W}_n\in\mathbb{R}^{d\times d}\). For each ordered
pair \(i\neq j\), the mechanism computes a bilinear interaction score
\(a_{ij}=(\mathbf{W}_i\mathbf{v}_i)^\top(\mathbf{W}_j\mathbf{v}_j)\), which
is the dot product of the two transformed vectors in the shared space and
plays the role analogous to a \(\mathbf{Q}\mathbf{K}^\top\) term in
dot-product attention. Each modality \(i\) then aggregates its pairwise
interactions by taking the average
\(\alpha_i=\tfrac{1}{n-1}\sum_{j\neq i} a_{ij}\), producing a scalar
coefficient that quantifies how strongly modality \(i\) aligns with the
other modalities on average. The coefficient \(\alpha_i\) is used to scale
the original input \(\mathbf{v}_i\), and the scaled vectors
\(\alpha_1\mathbf{v}_1,\dots,\alpha_n\mathbf{v}_n\) are concatenated to form
the fused output \(\mathbf{z}=[\alpha_1\mathbf{v}_1;\dots;\alpha_n\mathbf{v}_n]\).
This fused representation encapsulates the pairwise relationships between
modalities while preserving each modality's original representation. To
examine stability and robustness properties, we first derive a Lipschitz
bound for the pairwise-averaged multimodal attention map and then obtain
corresponding bounds for its gradients.

\subsection{Lipschitz Constant of Multimodal Attention}

We now analyze the Lipschitz continuity of the pairwise-averaged multimodal bilinear attention mechanism. Specifically, we study the spectral norms of the Jacobian matrices of the fused output \(\mathbf{z} = [\alpha_1 \mathbf{v}_1;\dots;\alpha_n \mathbf{v}_n]\) with respect to the individual inputs \(\mathbf{v}_1, \dots, \mathbf{v}_n \in \mathbb{R}^d\). The spectral norm of each Jacobian block quantifies how changes in a specific input vector affect the fused representation, providing an upper bound on the mechanism’s local sensitivity to perturbations. This analysis allows us to characterize the stability of the attention fusion process as a function of the operator norms of the projection matrices \(\mathbf{W}_1,\dots,\mathbf{W}_n\) and the magnitudes of the input vectors.

\begin{theorem}
    \label{theorem:lipschitz_attention}
    Let \(d,n \in \mathbb{N}\), and let matrices \(\mathbf{W}_1,\dots,\mathbf{W}_n \in \mathbb{R}^{d\times d}\) be given. Define the map $\Phi(\mathbf{v}_1,\dots,\mathbf{v}_n) = \big[ \alpha_1 \mathbf{v}_1 \,;\, \alpha_2 \mathbf{v}_2 \,;\, \ldots \,;\, \alpha_n \mathbf{v}_n \big]$, where $\alpha_i := \frac{1}{n-1}\sum_{\substack{j=1\\j\neq i}}^n a_{ij}$ with $a_{ij} := (\mathbf{W}_i\mathbf{v}_i)^\top(\mathbf{W}_j\mathbf{v}_j)\in\mathbb{R}$ for \(i\neq j\). Let \(\mathcal{K}\subseteq (\mathbb{R}^d)^n\) be compact and set $R := \sup_{(\mathbf{v}_1,\dots,\mathbf{v}_n)\in\mathcal K}\max_{1\le i\le n}\|\mathbf{v}_i\|$ and $M := \max_{1\le i\le n}\|\mathbf{W}_i\|$. Then \(\Phi\) is Lipschitz continuous on \(\mathcal K\) with the Lipschitz constant satisfying the explicit bound $L \le 4\, M^2 R^2$.
\end{theorem}

Notably, the Lipschitz constant \(L\) of the pairwise-averaged bilinear attention mechanism grows quadratically with the input magnitude (scaling with \(R^2\)) and linearly with the spectral norms of the weight matrices \(\mathbf{W}_1,\dots,\mathbf{W}_n\). To better understand the sensitivity of this mechanism, we compare its Lipschitz constant with that of standard vector summation and concatenation operations. Assuming the input norms satisfy \(R \leq 1\), two limiting behaviors can be observed. First, when the product of the largest spectral norms satisfies \(M^2 < \tfrac{1}{4}\), where \(M = \max_i \|\mathbf{W}_i\|\), the Lipschitz constant of the pairwise-averaged attention mapping remains strictly smaller than that of linear operations such as summation or concatenation. In this regime, the attention fusion is smoother and less sensitive to input perturbations. Conversely, when \(M^2 > \tfrac{\sqrt{2}}{4}\), the attention mechanism becomes more sensitive than simple linear combinations, exhibiting stronger amplification of variations in the input space. These observations indicate that when the projection matrices are spectrally bounded or inputs are normalized, the pairwise-averaged bilinear attention retains controlled, nearly linear behavior. However, as the weight norms increase or the inputs become large, the mechanism may transition to a highly sensitive regime, amplifying perturbations and potentially affecting the model’s stability during training.

\subsection{Gradients of the Pairwise-Averaged Multimodal Attention}

To estimate the Lipschitz constant of the multimodal attention gradients, we analyze how variations in the input vectors \(\mathbf{v}_1,\dots,\mathbf{v}_n\) propagate through the pairwise-averaged mechanism and affect the fused output. This involves computing the Jacobian of the fused vector \(\mathbf{z} = [\alpha_1\mathbf{v}_1;\dots;\alpha_n\mathbf{v}_n]\) with respect to each input \(\mathbf{v}_k\), capturing both the direct dependence (through the scaling of \(\mathbf{v}_i\)) and the indirect dependence (through \(\alpha_i\)). The chain rule yields explicit block formulas for every Jacobian block.

\begin{lemma}
\label{lemma:attention_gradients}
    Under the setting of \autoref{def:multimodal_attention}, let \(\Phi:(\mathbb{R}^d)^n\to\mathbb{R}^{nd}\) be the fused map \(\Phi(\mathbf v_1,\dots,\mathbf v_n)=[\alpha_1\mathbf v_1;\dots;\alpha_n\mathbf v_n]\)
    with $\alpha_i=\frac{1}{n-1}\sum_{\substack{j=1\\j\neq i}}^n a_{ij}$ and $a_{ij}=(\mathbf W_i\mathbf v_i)^\top(\mathbf W_j\mathbf v_j)$. Then the Jacobian matrix \(\dfrac{\partial\Phi}{\partial\mathbf v_k}\in\mathbb{R}^{nd\times d}\)
    (derivative of the concatenated output with respect to \(\mathbf v_k\)) has the block structure $\frac{\partial\Phi}{\partial\mathbf v_k}
    = \begin{bmatrix} B_{1,k}; B_{2,k}; \dots; B_{n,k}\end{bmatrix}^\top$ with $B_{i,k}\in\mathbb{R}^{d\times d}$, where for each \(i\in\{1,\dots,n\}\) and fixed \(k\in\{1,\dots,n\}\),
    \[
    B_{i,k} =
    \begin{cases}
    \displaystyle
    \alpha_i I_d \;+\; \Big(\frac{1}{n-1}\sum_{\substack{j=1\\j\neq i}}^n (\mathbf W_i)^\top(\mathbf W_j\mathbf v_j)\Big)\mathbf v_i^\top,
    & i = k, \\[10pt]
    \displaystyle
    \frac{1}{n-1} \Big((\mathbf W_k)^\top(\mathbf W_i\mathbf v_i)\Big)\mathbf v_i^\top,
    & i \neq k.
    \end{cases}
    \]
\end{lemma}

Each Jacobian \(\dfrac{\partial\Phi}{\partial\mathbf v_k}\in\mathbb{R}^{nd\times d}\) has an \(n\times 1\) block structure of \(d\times d\) blocks \(B_{i,k}\) indexed by the output modality \(i\) and the input modality \(k\). Every block \(B_{i,k}\) decomposes into two components. The first component is a diagonal scaling term \(\alpha_i I_d\) (present only when \(i=k\)), representing self-scaling of modality \(i\) by its aggregated attention coefficient. The second component is an outer-product-like modulation term arising from the dependence of \(\alpha_i\) on the inputs. It has the form $\frac{\partial\alpha_i}{\partial\mathbf v_k}\,\mathbf v_i^\top$, where, by the pairwise-averaged construction,
\begin{align}
    \frac{\partial\alpha_i}{\partial\mathbf v_k}=
    \begin{cases}
    \displaystyle \frac{1}{n-1}\sum_{j\ne i} (\mathbf W_i)^\top(\mathbf W_j\mathbf v_j), & k=i,\\
    \displaystyle \frac{1}{n-1}(\mathbf W_k)^\top(\mathbf W_i\mathbf v_i), & k\ne i.
    \end{cases}
\end{align}
Thus, each block is either
\begin{align}
    B_{i,i}=\alpha_i I_d + \Big(\frac{1}{n-1}\sum_{j\ne i} (\mathbf W_i)^\top(\mathbf W_j\mathbf v_j)\Big)\mathbf v_i^\top
\end{align}
(for the self-block), or
\begin{align}
    B_{i,k}=\frac{1}{n-1}\big((\mathbf W_k)^\top(\mathbf W_i\mathbf v_i)\big)\mathbf v_i^\top
\quad (i\ne k)
\end{align}
(for cross-blocks). The outer-product structure makes explicit two mechanisms of modulation: self-modulation (terms involving the sum over \(j\neq i\) that couple modality \(i\) to its own vector \(\mathbf v_i\)) and cross-modulation (terms proportional to \((\mathbf W_k)^\top(\mathbf W_i\mathbf v_i)\,\mathbf v_i^\top\) that capture how perturbations in modality \(k\) influence the output block for modality \(i\)). The factor \(1/(n-1)\) appearing in the modulation terms stems from the averaging in the definition of \(\alpha_i\).

\subsection{Lipschitz Continuity of Pairwise-Averaged Attention Gradients}

In the case of simple aggregation mechanisms such as vector summation or concatenation, the overall smoothness of the aggregation gradients is entirely determined by the smoothness of the $k$-th encoder. These operations are Lipschitz-preserving and do not introduce additional sensitivity in the gradient computations. Consequently, the Lipschitz constant of the gradient of the aggregated representation is unaffected by the aggregation itself and depends solely on the encoder parameters.

In contrast, the pairwise-averaged bilinear attention mechanism introduces a multiplicative coupling between all modality pairs through the bilinear terms \(a_{ij} = (\mathbf{W}_i \mathbf{v}_i)^\top (\mathbf{W}_j \mathbf{v}_j)\). This coupling causes the gradients of the fused representation to exhibit cross-modal dependencies, so that small perturbations in one modality can influence the gradients associated with others. As shown in \autoref{theorem:attention_lipschitz_gradients}, this interaction increases the sensitivity of the gradient map, with a Lipschitz constant that depends cubically on the operator norms of the weight matrices.

\begin{theorem}
\label{theorem:attention_lipschitz_gradients}
Let $\Phi: (\mathbb{R}^d)^n \to \mathbb{R}^{nd}$ be the pairwise-averaged multimodal bilinear attention mechanism defined in \autoref{def:multimodal_attention}. Assume the matrices $\mathbf{W}_1,\dots,\mathbf{W}_n \in \mathbb{R}^{d \times d}$ are fixed, and denote
\[
M := \max_{i \in \{1,\dots,n\}} \|\mathbf{W}_i\|, 
\qquad
R := \sup_{(\mathbf{v}_1,\dots,\mathbf{v}_n)\in\mathcal K} \max_i \|\mathbf{v}_i\|,
\]
for some compact set $\mathcal K \subseteq (\mathbb{R}^d)^n$.  
Then the gradient map $\nabla \Phi: (\mathbb{R}^d)^n \to \mathbb{R}^{nd \times d}$ is Lipschitz continuous on $\mathcal K$, with Lipschitz constant bounded by
\begin{align*}
    L_{\mathrm{grad}}
    \;\leq\;
    C_n \, M^3 R,
\end{align*}
where $C_n = \mathcal{O}(n)$ is a constant depending linearly on the number of modalities $n$.
\end{theorem}

This result shows that the gradients of the pairwise-averaged bilinear attention mechanism are Lipschitz continuous, but with a constant that scales cubically in the operator norms of the projection matrices. Unlike simple linear aggregations, the pairwise multiplicative structure introduces higher-order coupling terms across modalities, increasing the sensitivity of the gradient to perturbations in the input space. Controlling the spectral norms of the matrices $\mathbf{W}_i$ and bounding the input magnitudes $R$ are thus essential to ensure stable and smooth optimization dynamics.

\subsection{Lipschitz Regularization}

The results presented in \autoref{theorem:lipschitz_attention} and \autoref{theorem:attention_lipschitz_gradients} highlight that the sensitivity to input perturbations is closely tied to the magnitude of its learnable parameters of the multimodal attention. In particular, the gradient behavior can become highly unstable when weight matrices are large or unregularized, leading to sharp fluctuations that can hinder effective optimization. This sensitivity can significantly impact convergence and generalization in practice. To promote smoother gradients and more robust training dynamics, the following two subsections introduce strategies designed to explicitly control the Lipschitz constant of the multimodal attention mechanism by directly regulating the influence of the attention parameters.

\subsubsection{Normalization and Spectral Control for Stability}
\label{subsub:scaling}

To ensure numerical stability and control the Lipschitz constant of the multimodal attention mechanism, we extend the model to handle \( n \) modalities while maintaining bounded output variations and minimizing sensitivity to outliers. Large input magnitudes can lead to inflated attention scores and unstable gradients; thus, each input vector \(\mathbf{v}_i \in \mathbb{R}^d\) is first normalized to unit norm before the linear transformation:
\[
\mathbf{v}_i' = \frac{\mathbf{v}_i}{\|\mathbf{v}_i\|_2}, \quad i = 1, \dots, n.
\]
This normalization step ensures consistent scaling across modalities and prevents any single modality from disproportionately influencing the attention computation.

The Lipschitz constant of the pairwise-averaged bilinear attention mechanism depends on the spectral norms of the projection matrices \(\mathbf{W}_1, \dots, \mathbf{W}_n\), as established in \autoref{theorem:lipschitz_attention} and \autoref{theorem:attention_lipschitz_gradients}. To prevent excessive amplification of input magnitudes, we apply spectral normalization to each weight matrix:
\[
\mathbf{W}_i' = \frac{\mathbf{W}_i}{\|\mathbf{W}_i\|_2}, \quad i = 1, \dots, n.
\]
Spectral normalization constrains the largest singular value \(\sigma_{\text{max}}(\mathbf{W}_i)\) of each matrix, effectively bounding the operator norm of the transformation. If the initial spectral norm satisfies \(\sigma_{\text{max}}(\mathbf{W}_i) > 1\), the normalization rescales it to 1, thereby reducing the Lipschitz constant by a factor of \(\frac{1}{\sigma_{\text{max}}(\mathbf{W}_i)}\). Conversely, if \(\sigma_{\text{max}}(\mathbf{W}_i) \le 1\), this step preserves the stability of the mapping without introducing additional amplification.

The normalized transformations are then applied to the normalized inputs:
\[
\tilde{\mathbf{v}}_i = \mathbf{W}_i' \mathbf{v}_i', \quad i = 1, \dots, n.
\]
The pairwise attention scores are subsequently computed between all modality pairs:
\[
a_{ij} = \tilde{\mathbf{v}}_i^\top \tilde{\mathbf{v}}_j, \quad i,j = 1, \dots, n,\; i \neq j.
\]
To ensure these dot products remain well-scaled as the dimensionality \(d\) increases, each score is divided by \(\sqrt{d}\):
\[
\mathbf{A}_{ij} = \frac{\tilde{\mathbf{v}}_i^\top \tilde{\mathbf{v}}_j}{\sqrt{d}}.
\]
This scaling factor, originally introduced by \citep{vaswani2017attention}, prevents the attention logits from growing with dimension and helps maintain bounded gradient magnitudes. By controlling both the input and operator norms, this extended normalization and scaling scheme directly constrains the overall Lipschitz constant of the multimodal fusion mechanism, ensuring smoother optimization dynamics and improved numerical stability across modalities.

\subsubsection{Regularized multimodal attention}

Spectral normalization operates by explicitly controlling the largest singular value of each weight matrix, ensuring that no single layer can amplify gradients beyond a fixed limit. This hard constraint guarantees that the Lipschitz constant of each layer remains at most one, which is particularly valuable for maintaining stable training in neural architectures. However, spectral normalization alone does not directly influence the overall scale or distribution of the weights beyond this singular value constraint. 

When we additionally introduce L2 regularization to the multimodal attention parameters through a loss term \( \mathcal{L}_{\text{reg}} = \lambda \|\mathbf{W}\|_F^2 \), where \( \|\mathbf{W}\|_F^2 = \sum_{i=1}^m \sum_{j=1}^n w_{ij}^2 = \sum_{i=1}^{\min(m,n)} \sigma_i(\mathbf{W})^2 \) represents the squared Frobenius norm, the gradient updates during optimization take the form 
\begin{align}
    \mathbf{W} \leftarrow \mathbf{W} - \eta (\nabla_{\mathbf{W}} \mathcal{L}_{\text{task}} + 2\lambda \mathbf{W}).
\end{align}
The effect of L2 regularization can be understood by examining its influence on the singular values. Since \( \|\mathbf{W}\|_F^2 = \sum_i \sigma_i^2 \), the gradient term \( -2\lambda \mathbf{W} \) pushes all singular values \( \sigma_i \) toward zero. However, the subsequent spectral normalization rescales the matrix such that the largest singular value becomes exactly one. This creates a dynamic where the relative magnitudes of the singular values are preserved, but their absolute scale is reduced. 

% \[
% \mathcal{L}_{\text{reg}} = \lambda \sum^M_m \|\mathbf{W}_m\|_2^2 .
% \]

The combination of spectral normalization and L2 regularization leads to weight matrices that are not only Lipschitz-constrained but also have suppressed smaller singular values. The strength of this effect is controlled by the regularization parameter \( \lambda \), which determines how aggressively the smaller singular values are pushed down relative to the largest one. Proper tuning of \( \lambda \) ensures that we strike a balance between maintaining sufficient model capacity and achieving the desired regularization effects.

\subsection{Lipschitz Constant Estimation}

To estimate the Lipschitz constant, we focus on the sub-models \( f(\mathbf{x})_i \) of the multimodal autoencoder, where each encoder or decoder constitutes a sub-model. For the \(i\)-th sub-model with input \(\mathbf{x}\) and output \( f(\mathbf{x})_i \), the Lipschitz constant \(L_i\) represents the smallest value that bounds the maximum rate of change in the model's output relative to changes in its input. As demonstrated by \citep{virmaux2018lipschitz}, computing this constant precisely is an NP-hard problem. Therefore, following the approach of \citep{virmaux2018lipschitz}, we approximate it by assessing the maximum Frobenius norm of the model's gradient with respect to the input \(\mathbf{x}\):
\begin{align}  
    L \approx \max_{\mathbf{x} \in \mathcal{X}} \|\nabla_{\mathbf{x}} f(\mathbf{x})\|_F  
\end{align} 
Since we aim to analyze the variation in gradients, the Lipschitz constant \( L \) of the gradient of a function \( f \) is defined as:  
\[
\| \nabla_{\mathbf{x}} f(\mathbf{x}) - \nabla_{\mathbf{y}} f(\mathbf{y}) \| \leq L \| \mathbf{x} - \mathbf{y} \|, \quad \forall \mathbf{x}, \mathbf{y} \in \mathbb{R}^n.
\]
To estimate \( L \) empirically, we compute the gradients \( \nabla f(\mathbf{x}) \) and \( \nabla f(\mathbf{y}) \) for sampled pairs of points \( (\mathbf{x}, \mathbf{y}) \), then evaluate the ratio:  
\[
L \approx \max_{\mathbf{x}, \mathbf{y}} \frac{\| \nabla_{\mathbf{x}} f(\mathbf{x}) - \nabla_{\mathbf{y}} f(\mathbf{y}) \|}{\| \mathbf{x} - \mathbf{y} \|}.
\]
Due to the large size of the datasets, we use random samples instead of the full dataset to reduce computational costs and memory usage while ensuring a representative estimate. Similarly, to compute the Lipschitz constants, we use random subsets of output gradients rather than the complete set of gradients. Additionally, we add a small threshold \( \epsilon \) to avoid division by zero when \( \| x - y \| \) is very small. The overall Lipschitz estimation is described in \autoref{alg:lipschitz_nogradfunc}.

\begin{algorithm}[htbp]
\caption{Estimate Lipschitz Constant of Gradients}
\label{alg:lipschitz_nogradfunc}
\begin{algorithmic}
\REQUIRE Function \( f: \mathbb{R}^n \rightarrow \mathbb{R} \), dimension \( n \), number of samples \( N \), domain range \( [a, b] \).
\ENSURE Estimated Lipschitz constant \( L \).

\STATE Initialize \( L_{\text{max}} \gets 0 \).

\FOR{$k = 1$ to $N$}
    \STATE Sample \( x \sim \text{Uniform}([a, b]^n) \) and \( y \sim \text{Uniform}([a, b]^n) \).
    \IF{\( \| x - y \| > \epsilon \)}
        \STATE Compute \( \nabla f(x) \) and \( \nabla f(y) \) using backpropagation (or any gradient computation method).
        \STATE Compute \( L_k \gets \frac{\| \nabla f(x) - \nabla f(y) \|}{\| x - y \|} \).
        \STATE Update \( L_{\text{max}} \gets \max(L_{\text{max}}, L_k) \).
    \ENDIF
\ENDFOR
\STATE \RETURN \( L_{\text{max}} \)
\end{algorithmic}
\end{algorithm}

Alternative approaches, such as full optimization-based or semidefinite programming (SDP) formulations, or exact layer-wise global bound propagation, are theoretically sound but computationally infeasible for high-dimensional neural networks. Therefore, our strategy offers a practical and robust approximation of the Lipschitz behavior during training.

%%%%%%%%%%%%%%%%%%%%%%%%%%%%%%%%%%%%%%%%%%%%%%%%%%%%%%%%%%%%%%%%%%%%%%%%%%%%%%%%%%%
%%%%%%%%%%%%%%%%%%%%%%%%%%%%%%%%%%%%%%%%%%%%%%%%%%%%%%%%%%%%%%%%%%%%%%%%%%%%%%%%%%%
%%%%%%%%%%%%%%%%%%%%%%%%%%%%%%%%%%%%%%%%%%%%%%%%%%%%%%%%%%%%%%%%%%%%%%%%%%%%%%%%%%%
%%%%%%%%%%%%%%%%%%%%%%%%%%%%%%%%%%%%%%%%%%%%%%%%%%%%%%%%%%%%%%%%%%%%%%%%%%%%%%%%%%%

\section{Experiments}\label{sec:experiments}

In the following subsections, we present our experimental setup and evaluation of our methodology. Building on the theoretical Lipschitz properties of various aggregation strategies in multimodal autoencoders, we aim to validate these findings through empirical examples. Furthermore, we analyze the training outcomes by examining the convergence behavior when using summation and concatenation, as well as our proposed approach as an aggregation method. This analysis highlights the alignment between our theoretical results and empirical observations, demonstrating the effectiveness and robustness of our approach in practical scenarios.

\subsection{Synthetic Multimodal Heterogeneous Datasets}

\begin{figure}
     \centering
     \begin{subfigure}[b]{0.24\linewidth}
         \centering
         \includegraphics[width=\linewidth]{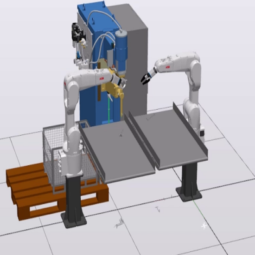}
         \caption{Camera 1}
     \end{subfigure}
     \hfill
     \begin{subfigure}[b]{0.24\linewidth}
         \centering
         \includegraphics[width=\linewidth]{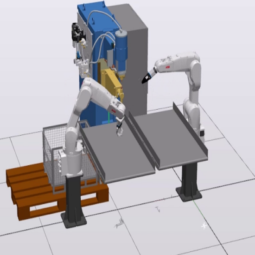}
         \caption{Camera 2}
     \end{subfigure}
     \hfill
     \begin{subfigure}[b]{0.24\linewidth}
         \centering
         \includegraphics[width=\linewidth]{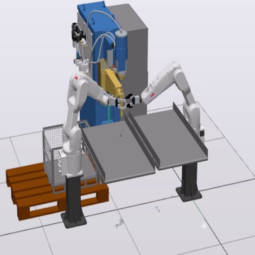}
         \caption{Camera 3}
     \end{subfigure}
     \hfill
     \begin{subfigure}[b]{0.24\linewidth}
         \centering
         \includegraphics[width=\linewidth]{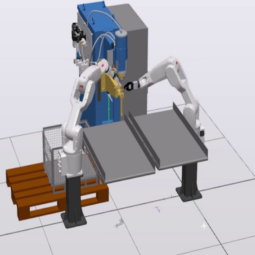}
         \caption{Camera 4}
     \end{subfigure}
    \caption{Four camera samples of the ABB robots within the RobotStudio cooperative welding simulation environment.}
    \label{fig:m_different_cams_abb2}
\end{figure}

In our research, we utilize three distinct datasets with a focus on industrial scenarios, particularly emphasizing the use of robotic systems. The first dataset is generated using the MuJoCo simulation environment, which includes an end effector along with a detailed representation of the surrounding environment in a pick-and-place task. The second and third datasets are created using ABB RobotStudio, which simulated a single robot and dual robot welding station scenario. All datasets are based on the work of \citep{altinses2024benchmarking}, where the authors introduce three simulated industrial multimodal datasets for benchmarking purposes. In \autoref{fig:m_different_cams_abb2}, we present four samples of the image modality of the dual robot welding station dataset.

The primary objective of this study is to reconstruct corrupted bimodal information. The inputs utilized for this purpose encompass static images $S$ and time series sensory data $T$, expressed as the tuple $m \in {s, t}$. To effectively capture temporal dependencies within the time series data, a segmentation approach is employed. The sensory data is divided into sequential segments with overlapping windows, each window containing a specific number of data points arranged in chronological order. For each static image, historical samples are organized into a sequence with a length of $w_l=20$ and a step size of $w_s=1$. Consequently, the dataset undergoes a reduction to $n = \frac{n-w_l}{w_s}$, accounting for the exclusion of initial samples up to the sequence length in the second modality. The temporal modality, post-segmentation, adopts a matrix structure denoted as $T\in \mathbb {R}^{w_l\times d_t}$, where $T$ comprises a collection of $n$ samples within a $d_t$-dimensional space. Conversely, the spatial aspect of the camera modality manifests as a higher-dimensional matrix $S\in \mathbb {R}^{3\times s_{r}\times s_{c}}$, with $s_{r}$ and $s_{c}$ set to 256, defining the number of rows and columns, respectively.

\subsection{Real-world Multimodal Heterogeneous Datasets}

The RoboMNIST dataset, based on the work of \citep{behzad2025robomnist} serves as a real-world dataset for evaluating multimodal representation learning in robotic perception tasks. It consists of recordings from a robotic setup performing digit manipulation and recognition under varying lighting, pose, and background conditions. Unlike purely synthetic datasets, RoboMNIST captures the inherent sensor noise, drift, and variability found in real-world industrial environments, thereby providing a more realistic evaluation scenario. Each sample in the dataset is represented by a trimodal observation:
\begin{enumerate}
    \item Vision modality: RGB image frames captured from three distinct camera perspectives.
    \item Kinematic modality: Joint positions, velocities, and end-effector poses of the robot arm.
    \item Signal modality: WiFi Channel State Information describing the wireless propagation characteristics between the transmitter and receiver during task execution.
\end{enumerate}

\begin{figure}
     \centering
     \begin{subfigure}[b]{0.24\linewidth}
         \centering
         \includegraphics[width=\linewidth]{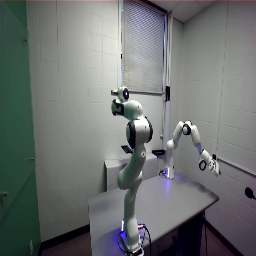}
         \caption{Sample 1}
         \end{subfigure}
     \hfill
     \begin{subfigure}[b]{0.24\linewidth}
         \centering
         \includegraphics[width=\linewidth]{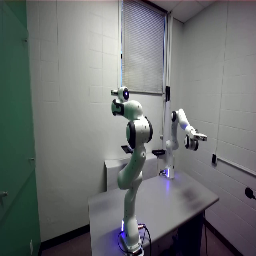}
         \caption{Sample 2}
     \end{subfigure}
     \hfill
     \begin{subfigure}[b]{0.24\linewidth}
         \centering
         \includegraphics[width=\linewidth]{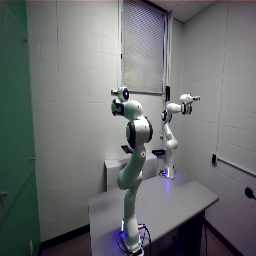}
         \caption{Sample 3}
     \end{subfigure}
     \hfill
     \begin{subfigure}[b]{0.24\linewidth}
         \centering
         \includegraphics[width=\linewidth]{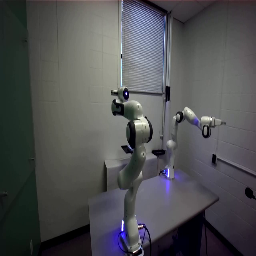}
         \caption{Sample 4}
     \end{subfigure}
    \caption{Four camera samples of the RoboMNIST dataset.}
    \label{fig:robomnist}
\end{figure}

In this work, we use only one of the three camera perspectives to limit the model complexity. The visual data are preprocessed by resizing to ($256 \times 256$) pixels and normalizing pixel intensities. All continuous variables from the kinematic and signal modalities are standardized by subtracting the mean and dividing by the standard deviation computed over the training set. This ensures balanced feature scaling across modalities and facilitates stable multimodal fusion. \autoref{fig:robomnist} shows four representative examples from the RoboMNIST dataset, illustrating the environmental context. Each sample includes synchronized measurements from all three modalities.

\subsection{Experimental Setup}

The dataset comprises valuable information encompassing the joint angles of the robot's pose and a single camera perspective, with the underlying assumption that these data sources emulate real-world camera and sensor acquisition. Our data preprocessing pipeline incorporates modality-specific normalization. 

\begin{figure}
    \centering
    \includegraphics[width=1\linewidth]{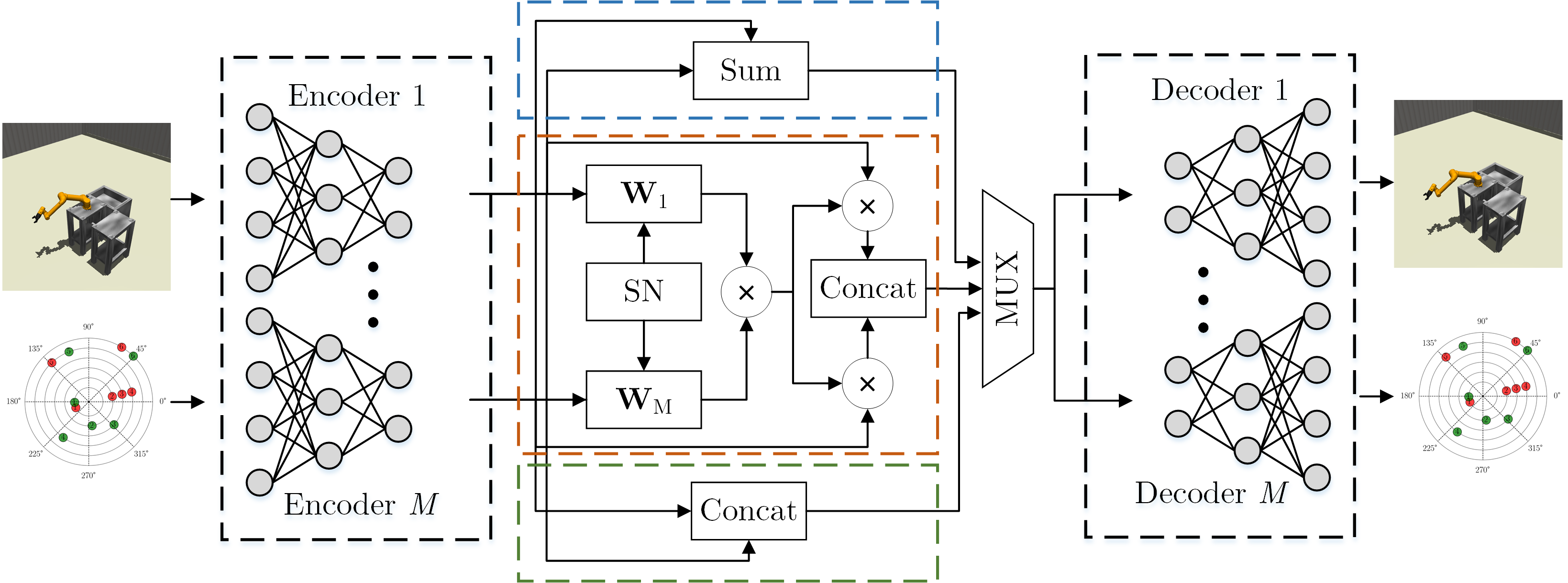}
    \caption{Multimodal autoencoder including the three aggregation methods. The sum aggregation is represented in blue, concatenation in green, and multimodal attention-based fusion in orange. The SN block represents the spectral normalization, and the MUX block represents a multiplexer.}
    \label{fig:multimodal_autoencoder}
\end{figure}

The multimodal fusion architecture, illustrated in \autoref{fig:multimodal_autoencoder}, contains a set of two Autoencoders, each corresponding to a distinct modality, and a computation model responsible for the data fusion process. The compute element includes an aggregation function that fuses the latent features, resulting in a more reliable and robust representation of all modalities. The target is to improve the reconstruction using a sparse vector of highly expressive combinations.

The base network architecture for all aggregation methods consists of two encoders and two decoders. The image modality encoder follows a structure of 3-256-128-64-32-32-32, where each number represents the input dimensions of a 2D convolutional neural network. The convolutional layers use a kernel size of 5, a stride of 2, and a padding of 1, with ReLU activation applied between layers. The decoder for the image modality mirrors this structure but employs transposed convolutional layers. For the temporal modality, the encoder is composed of fully connected layers structured as $n_i$-64-128-128-256-256-288, with ReLU activation applied after each layer. The corresponding decoder follows a mirrored architecture. For computing attention scores during feature fusion, we employ a two-layer attention network architecture that processes and transforms the input vectors while preserving their dimensionality. Specifically, the network maintains 288 input and output features throughout both layers.

The training procedure for the architecture spans a total of $2\cdot 10^2$ epochs, employing a batch size of 64 instances. The optimization process is executed using the ADAM optimizer with a learning rate of $\lambda = 10^{-3}$, which was chosen for its efficiency in facilitating updates during training. To comprehensively evaluate and compare our approach, we employ the mean squared error loss metric as a measure of the quality of the estimations generated by the model.

\subsection{Performance Evaluation - MuJoCo}

\begin{figure}
     \centering
     \begin{subfigure}[b]{0.323\textwidth}
         \centering
         \includegraphics[width=\textwidth]{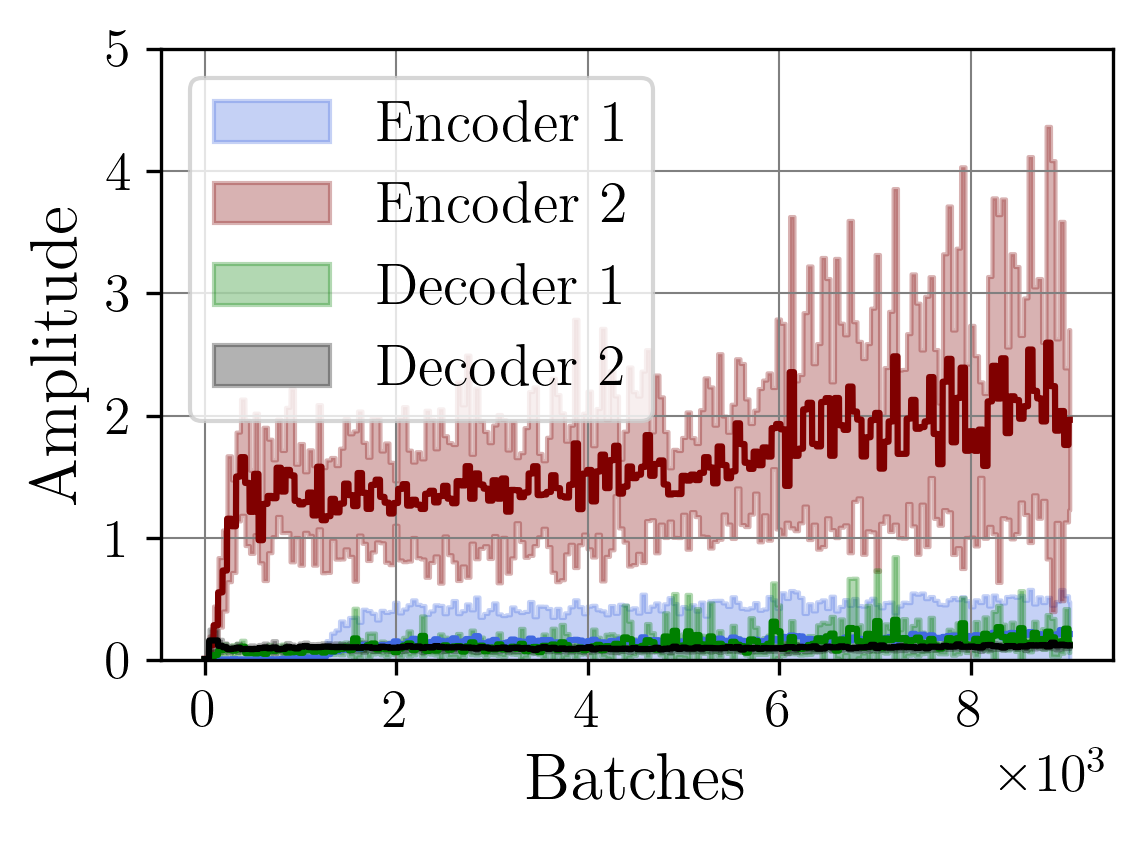}
         \caption{Summation}
     \end{subfigure}
     \hfill
     \begin{subfigure}[b]{0.323\textwidth}
         \centering
         \includegraphics[width=\textwidth]{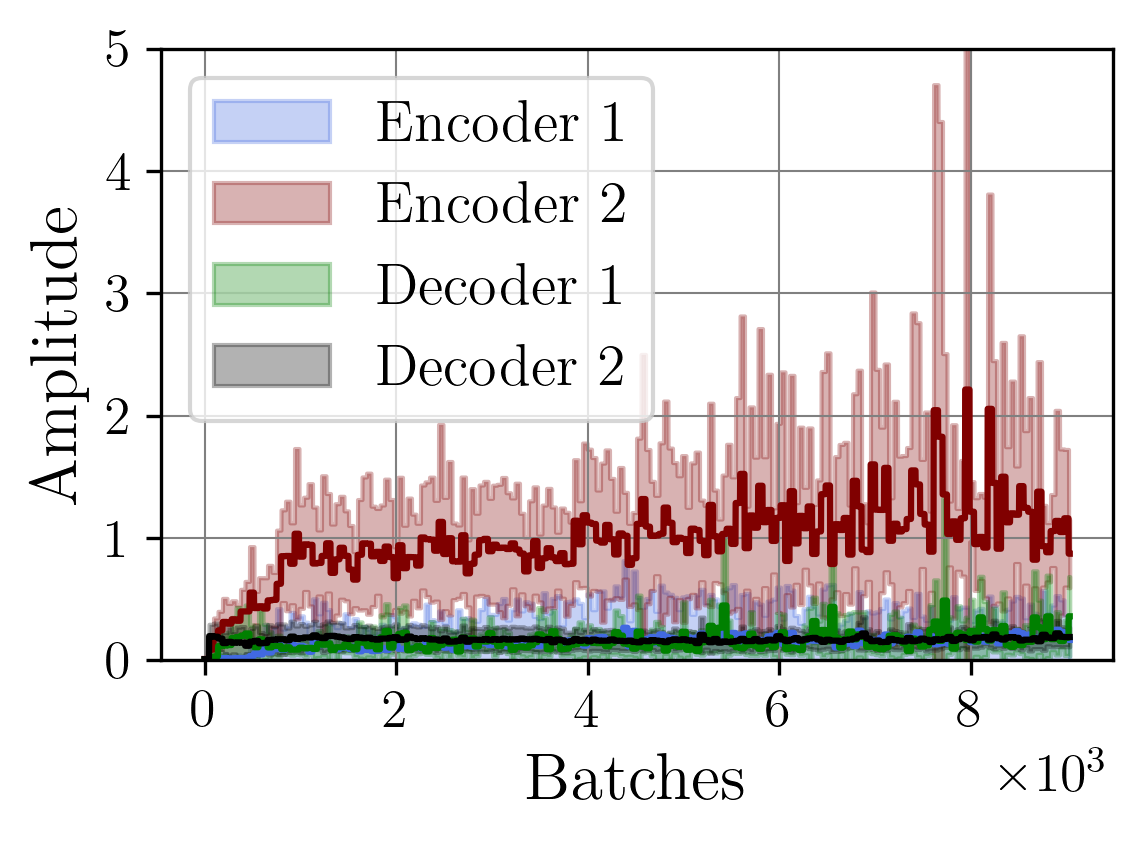}
         \caption{Concatenation}
     \end{subfigure}
     \hfill
     \begin{subfigure}[b]{0.323\textwidth}
         \centering
         \includegraphics[width=\textwidth]{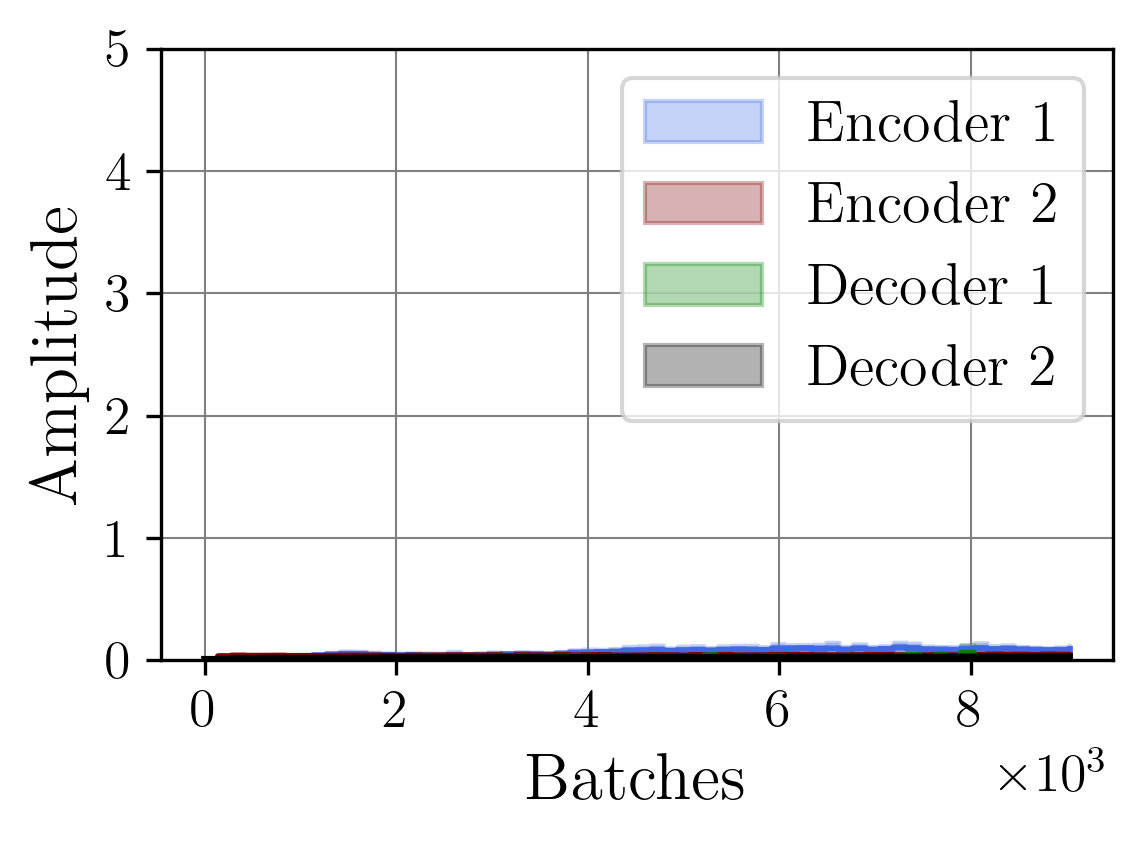}
         \caption{Attention}
     \end{subfigure}
    \caption{Estimated Lipschitz constants of each submodel in the multimodal autoencoder during training on the MuJoCo UR5 dataset, using summation, concatenation, and attention. Solid lines show the mean across trials; shaded areas indicate variation.}
    \label{fig:mujoco_lipschitz_comparison}
\end{figure}

In this section, we assess the performance of the three fusion methods by examining the estimated Lipschitz constants for each sub-model alongside their training performance across 10 independent trials using the MuJoCo dataset. As shown in \autoref{fig:mujoco_lipschitz_comparison}, the Lipschitz constants vary significantly depending on the aggregation method employed. Through this comparative analysis, we highlight key differences in model stability and sensitivity, reinforcing the consistency between our theoretical framework and experimental results.

\begin{figure}
     \centering
     \begin{subfigure}[b]{0.24\textwidth}
         \centering
         \includegraphics[width=\textwidth]{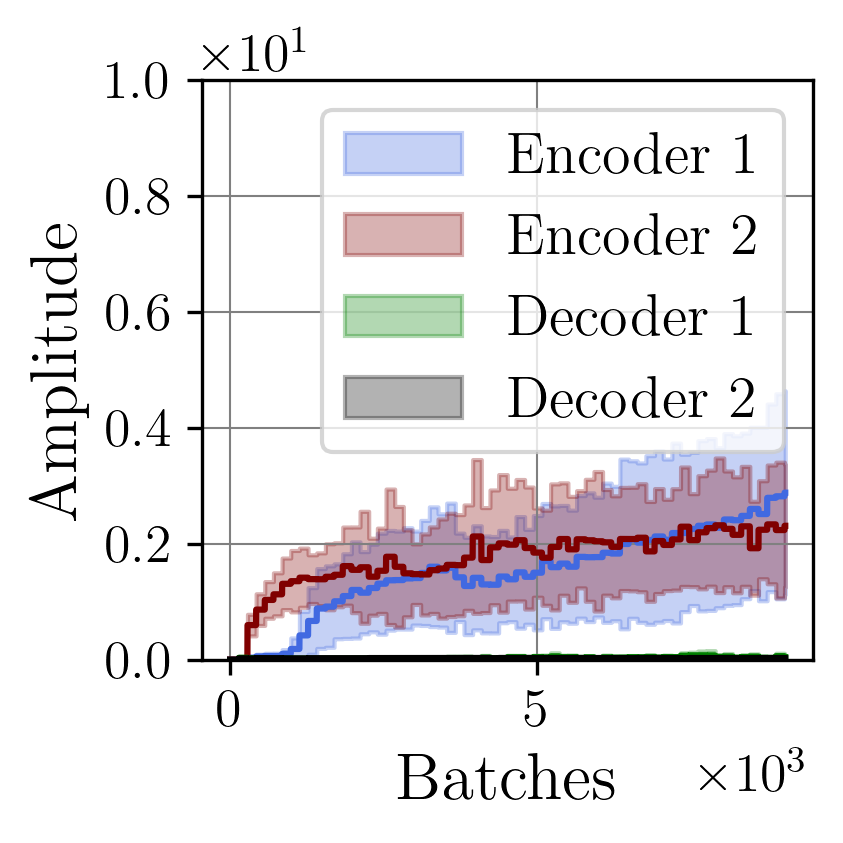}
         \caption{Normal}
     \end{subfigure}
     \hfill
     \begin{subfigure}[b]{0.24\textwidth}
         \centering
         \includegraphics[width=\textwidth]{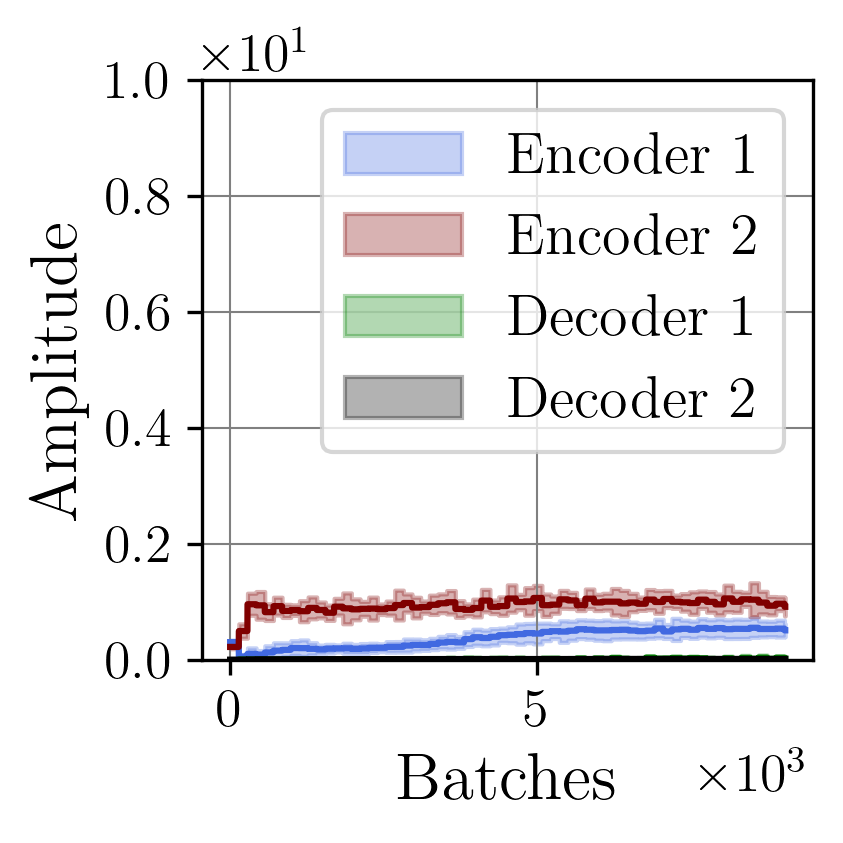}
         \caption{Scaling}
     \end{subfigure}
     \hfill
     \begin{subfigure}[b]{0.24\textwidth}
         \centering
         \includegraphics[width=\textwidth]{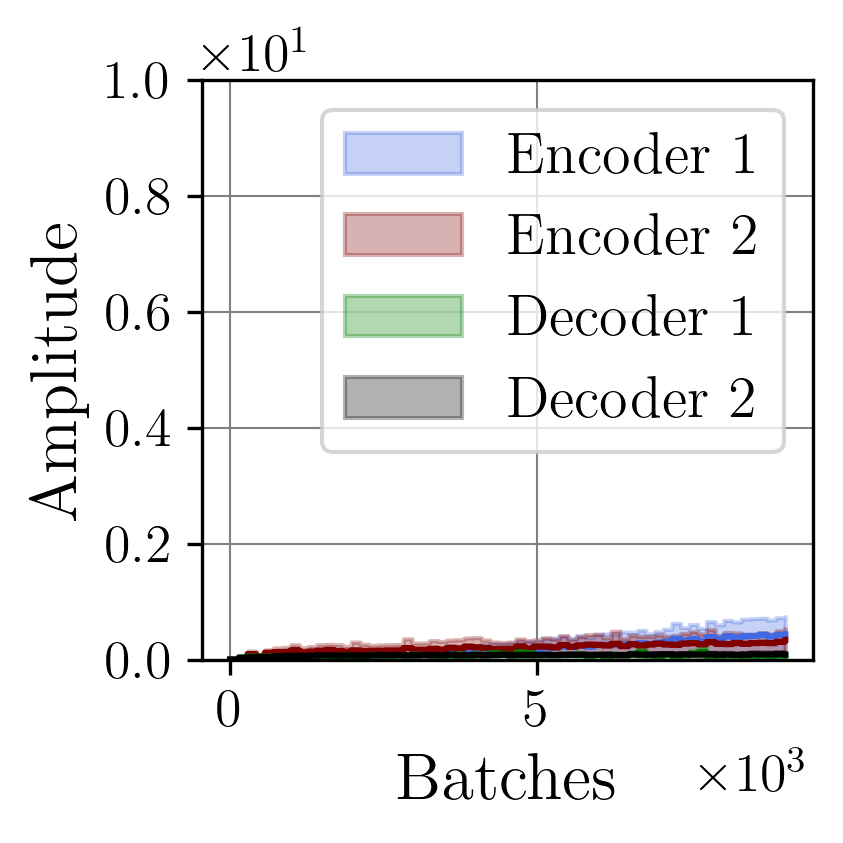}
         \caption{Regularization}
     \end{subfigure}
     \hfill
     \begin{subfigure}[b]{0.24\textwidth}
         \centering
         \includegraphics[width=\textwidth]{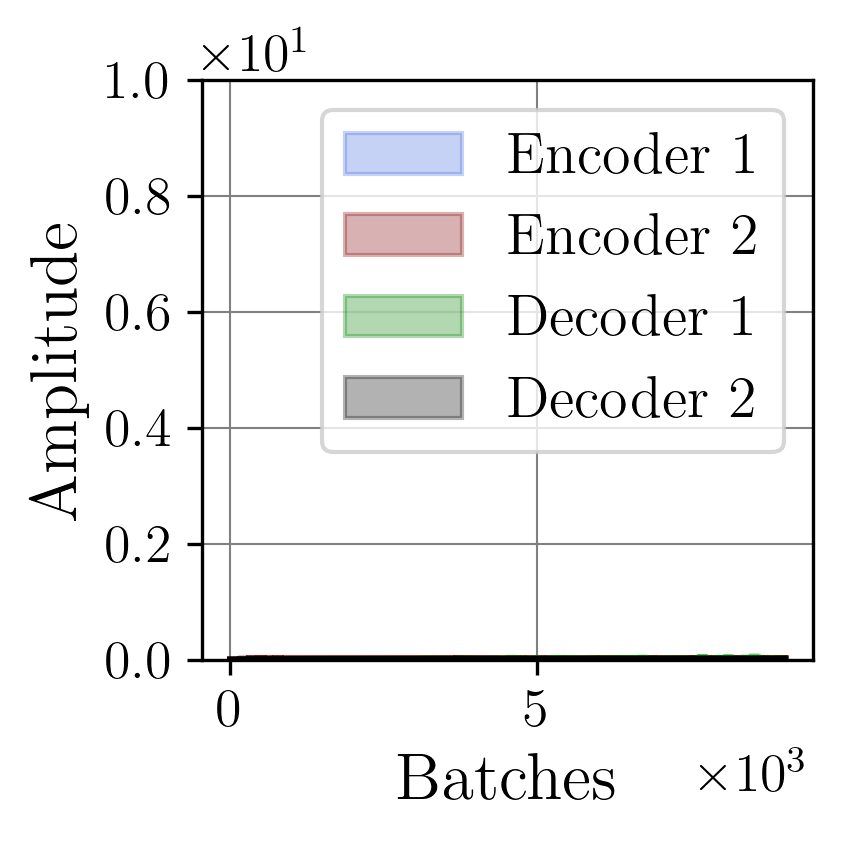}
         \caption{Scaling + Reg.}
     \end{subfigure}
    \caption{Impact of norm scaling and regularization on the Lipschitz constant in the attention-based aggregation of the multimodal autoencoder on the MuJoCo UR5 dataset, averaged over 10 trials. Solid lines indicate the mean and shaded areas show variation.}
    \label{fig:impact}
\end{figure}

The observed Lipschitz constants differ across fusion strategies due to their distinct noise sensitivity and gradient propagation properties. Consistent with our theoretical analysis in \autoref{theorem:lipschitz_multimodal_autoencoder_D} and \autoref{theorem:lipschitz_multimodal_autoencoder_E}, we observe that the encoder architecture contributes more significantly to the Lipschitz constant than other network components. The summation or concatenation of features amplifies the worst-case input perturbations. For high-dimensional image data, small input variations may propagate nonlinearly through the encoder, amplifying this effect. For attention-based fusion, regularization and scaling constrain the Lipschitz constant growth, steering the model toward low Lipschitz constants. In \autoref{fig:impact}, we present the individual impact of scaling and regularization in the attention-based aggregation. 

\begin{figure}
    \centering
    \includegraphics[width=0.5\linewidth]{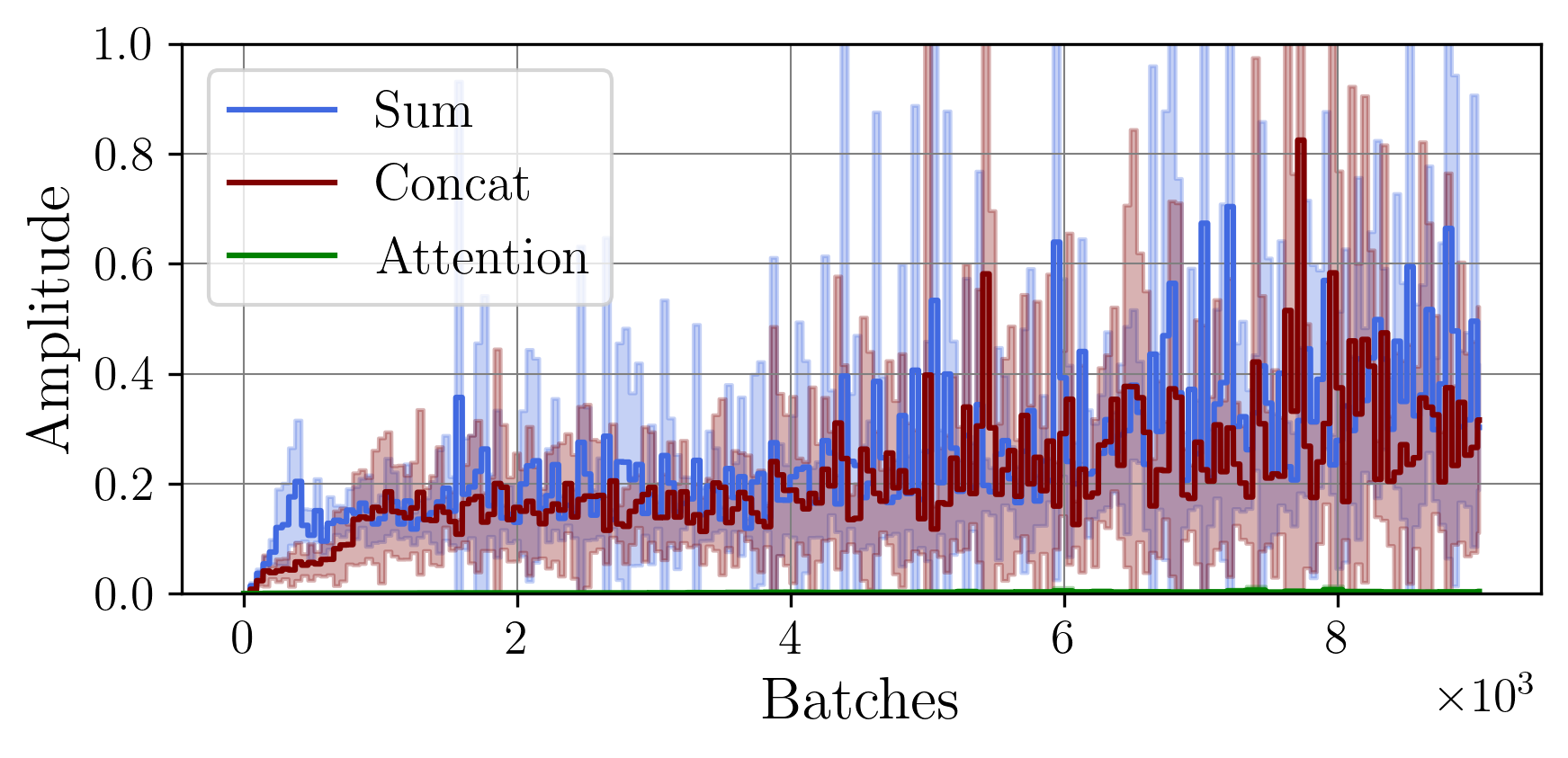}
    \caption{Mean (solid) and variations (transparent) of the Lipschitz constants across 10 trials using summation, concatenation, and attention in the multimodal autoencoder on the MuJoCo UR5 dataset.}
    \label{fig:overall_lipschitz}
\end{figure}

The scaling strategy effectively stabilizes the overall Lipschitz constant around 1, which is particularly important given that the encoders are among the most sensitive components of the architecture. Since they process and transmit the full input signal through all subsequent layers, their contribution to the model's sensitivity is significant. By constraining their output magnitude, scaling directly limits the potential for uncontrolled gradient growth. In parallel, the applied regularization further reduces the Lipschitz constant by approximately a factor of 10, mitigating the effect of large parameter values and enhancing smoothness. When combined, these two mechanisms also reinforce each other, resulting in a compounded effect that drives the overall Lipschitz constant toward values close to zero. Having analyzed the Lipschitz behavior of individual submodules, we now turn to the overall model in \autoref{fig:overall_lipschitz}.

The increasing Lipschitz constants during training follow a typical trajectory shaped by optimization dynamics. In the initial phase, the Lipschitz constant rises due to random weight initialization, which induces unregularized feature transformations with poor gradient coherence, resulting in large gradient norms as the model reacts aggressively to early batches. This aligns with the sharp loss drops characteristic of early training. During the optimization process, the Lipschitz constant stabilizes as modality-specific encoders learn to suppress cross-modal noise, stabilizing gradients. However, in the final fine-tuning phase, the Lipschitz constant increases again as the model begins to fit subtle noise or dataset-specific artifacts, increasing sensitivity to adversarial perturbations. This phenomenon is consistent with findings in \citep{zhang2021understanding}, where Lipschitz constants rise near the interpolation threshold. 

\begin{figure}
     \centering
     \begin{subfigure}[b]{0.325\textwidth}
         \centering
         \includegraphics[width=\textwidth]{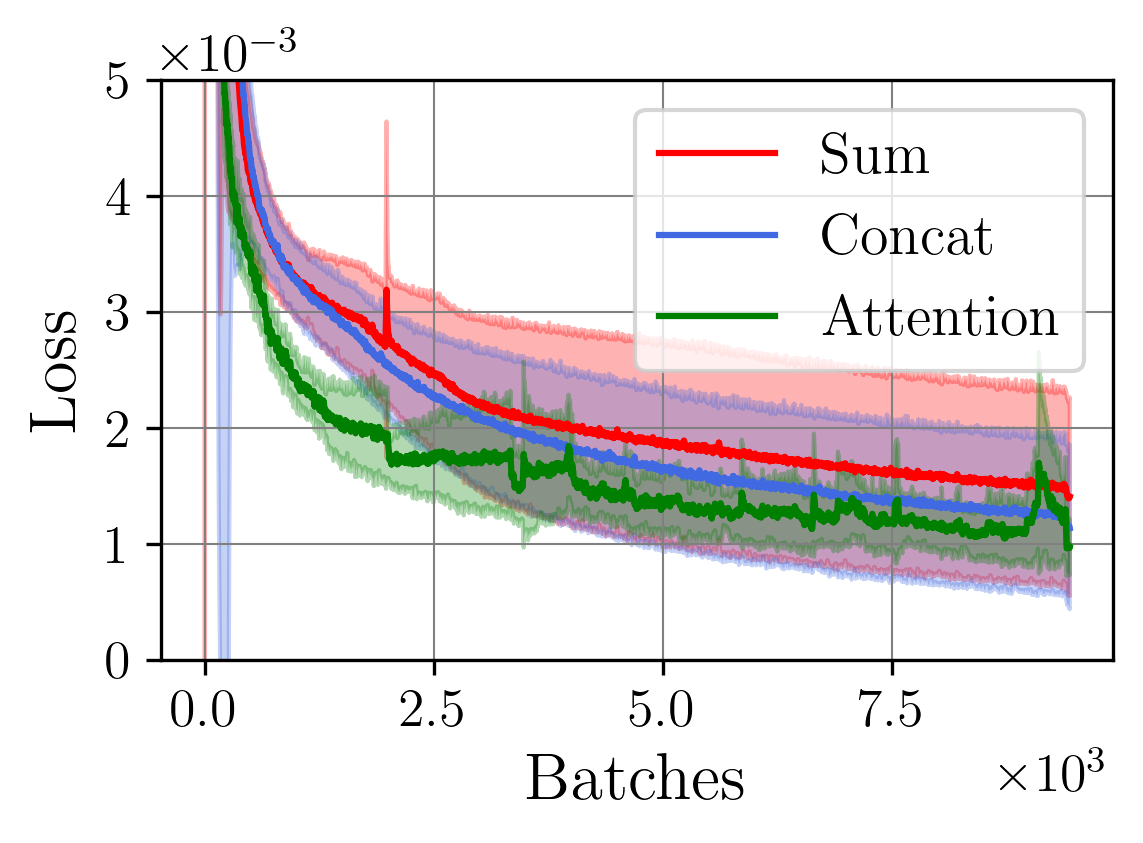}
         \caption{Camera}
     \end{subfigure}
     \hfill
     \begin{subfigure}[b]{0.325\textwidth}
         \centering
         \includegraphics[width=\textwidth]{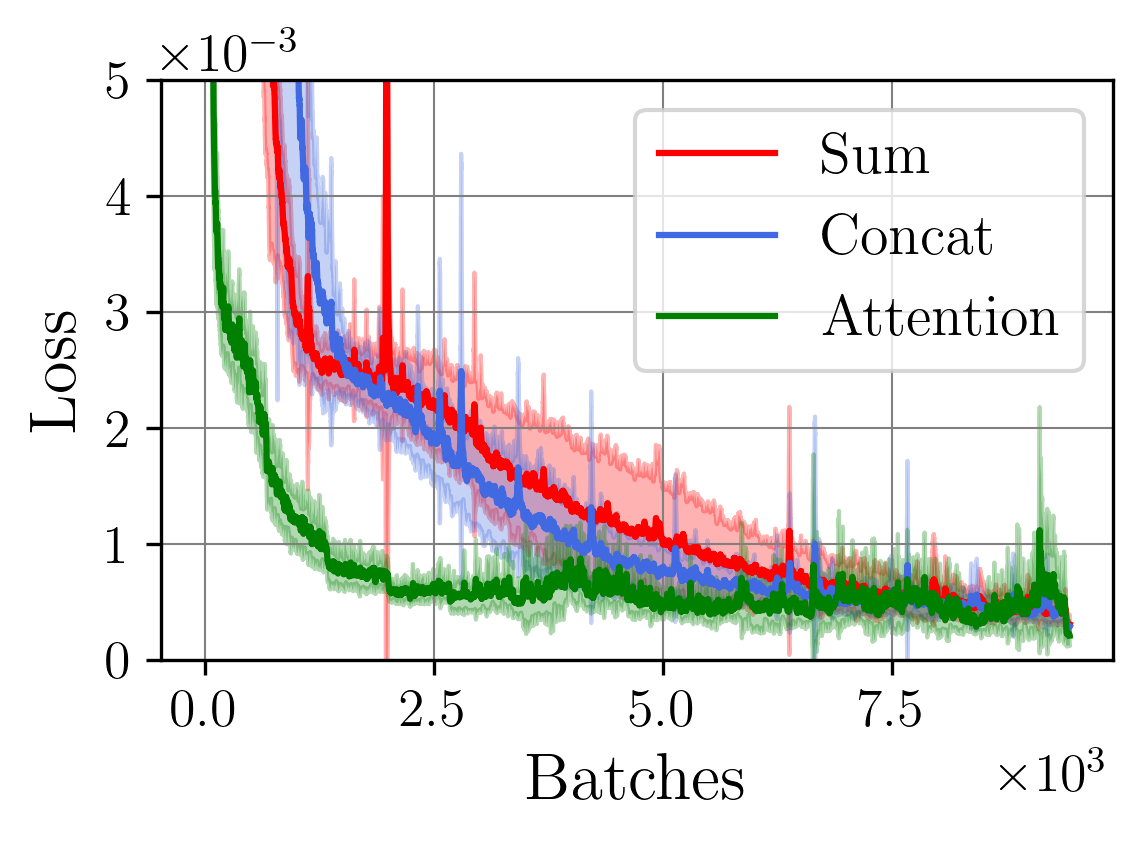}
         \caption{Sensor}
     \end{subfigure}
     \hfill
     \begin{subfigure}[b]{0.325\textwidth}
         \centering
         \includegraphics[width=\textwidth]{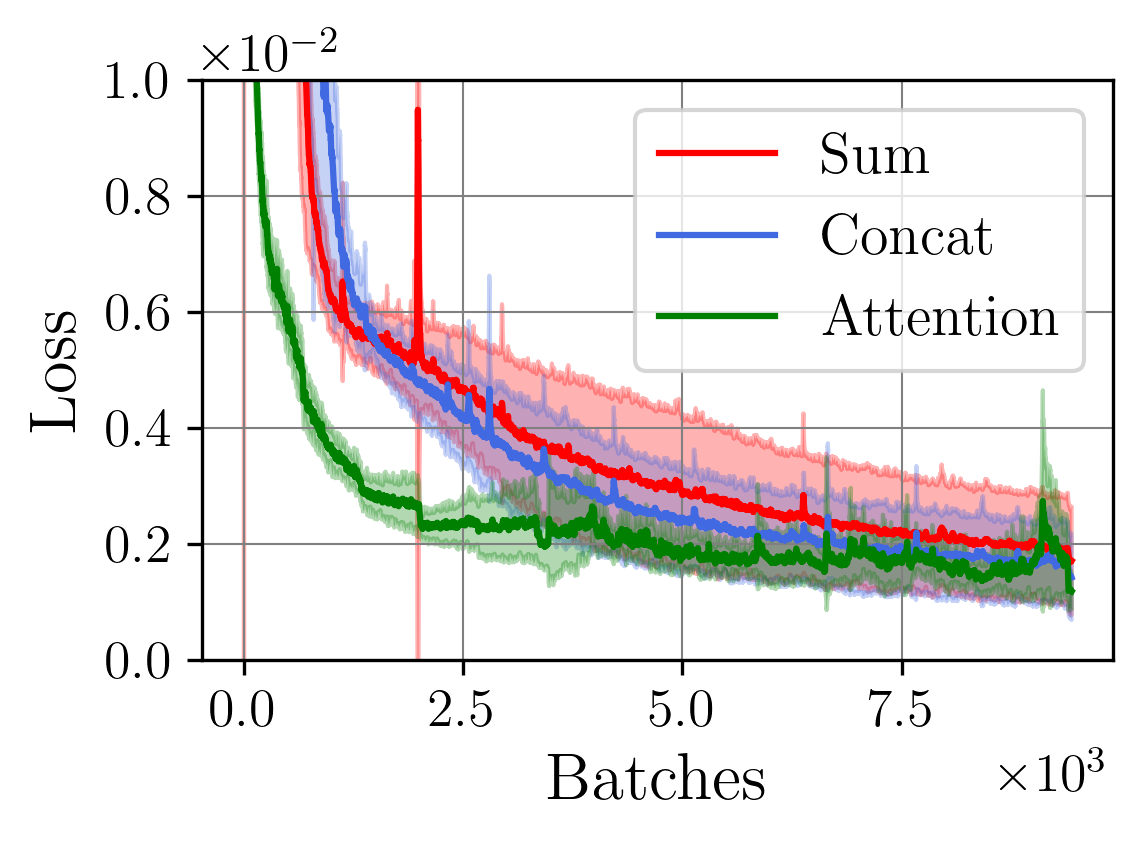}
         \caption{Overall}
     \end{subfigure}
    \caption{Training loss of the multimodal autoencoder on the MuJoCo UR5 dataset using summation, concatenation, and attention aggregation. The solid line shows the mean over 10 trials and the shaded area indicates variance.}
    \label{fig:mujoco_loss_comparison}
\end{figure}

\begin{figure}
     \centering
     \begin{subfigure}[b]{0.325\textwidth}
         \centering
         \includegraphics[width=\textwidth]{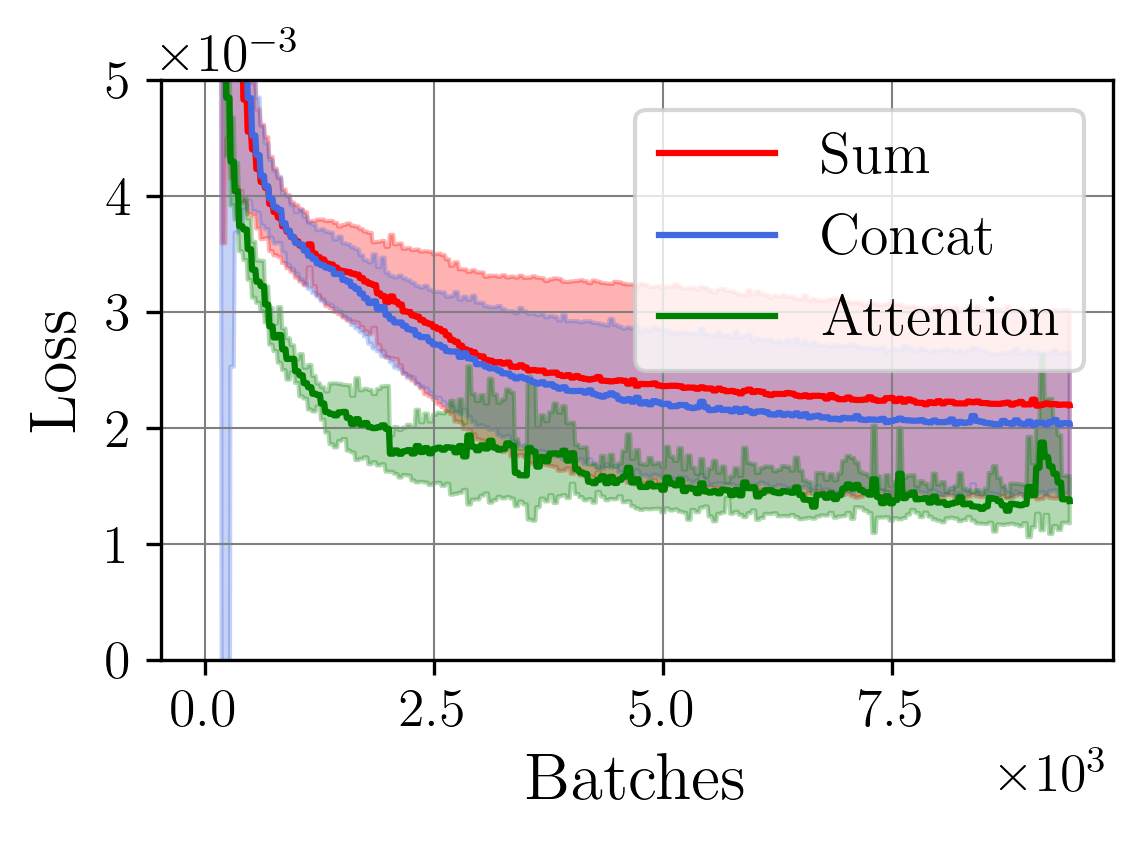}
         \caption{Camera}
     \end{subfigure}
     \hfill
     \begin{subfigure}[b]{0.325\textwidth}
         \centering
         \includegraphics[width=\textwidth]{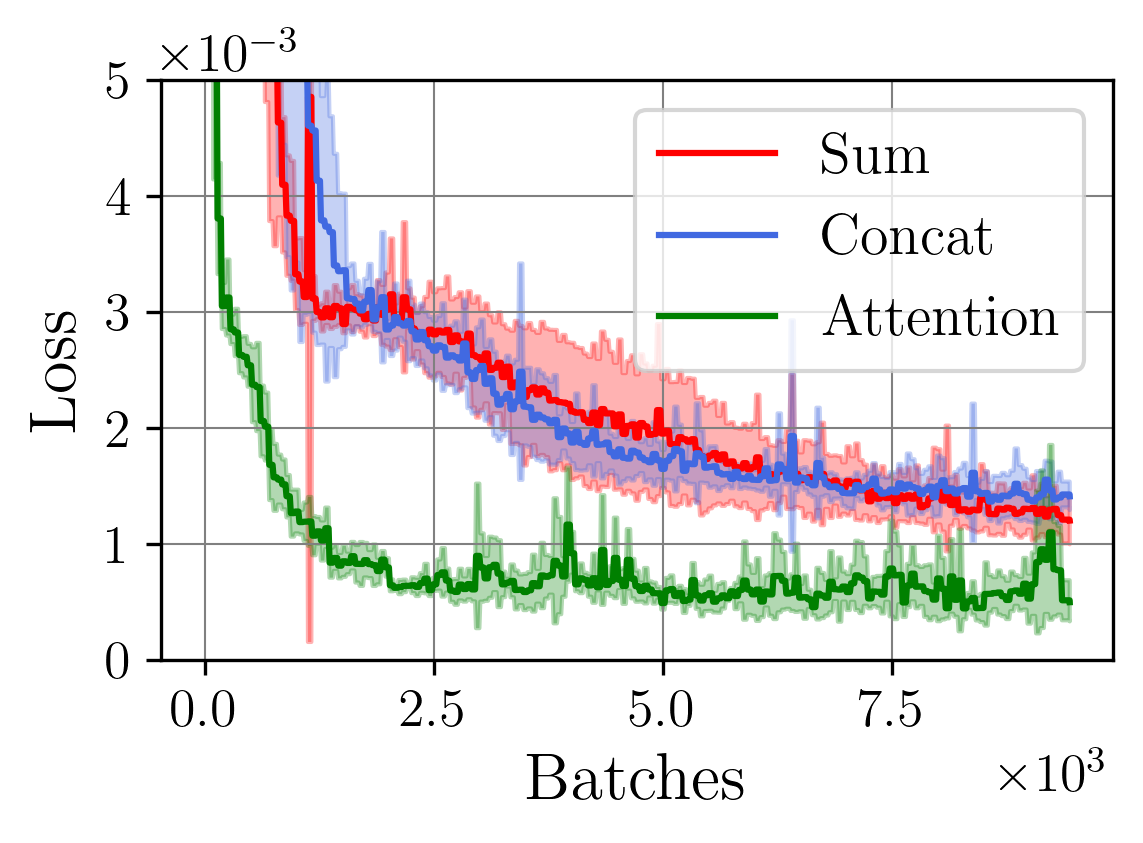}
         \caption{Sensor}
     \end{subfigure}
     \hfill
     \begin{subfigure}[b]{0.325\textwidth}
         \centering
         \includegraphics[width=\textwidth]{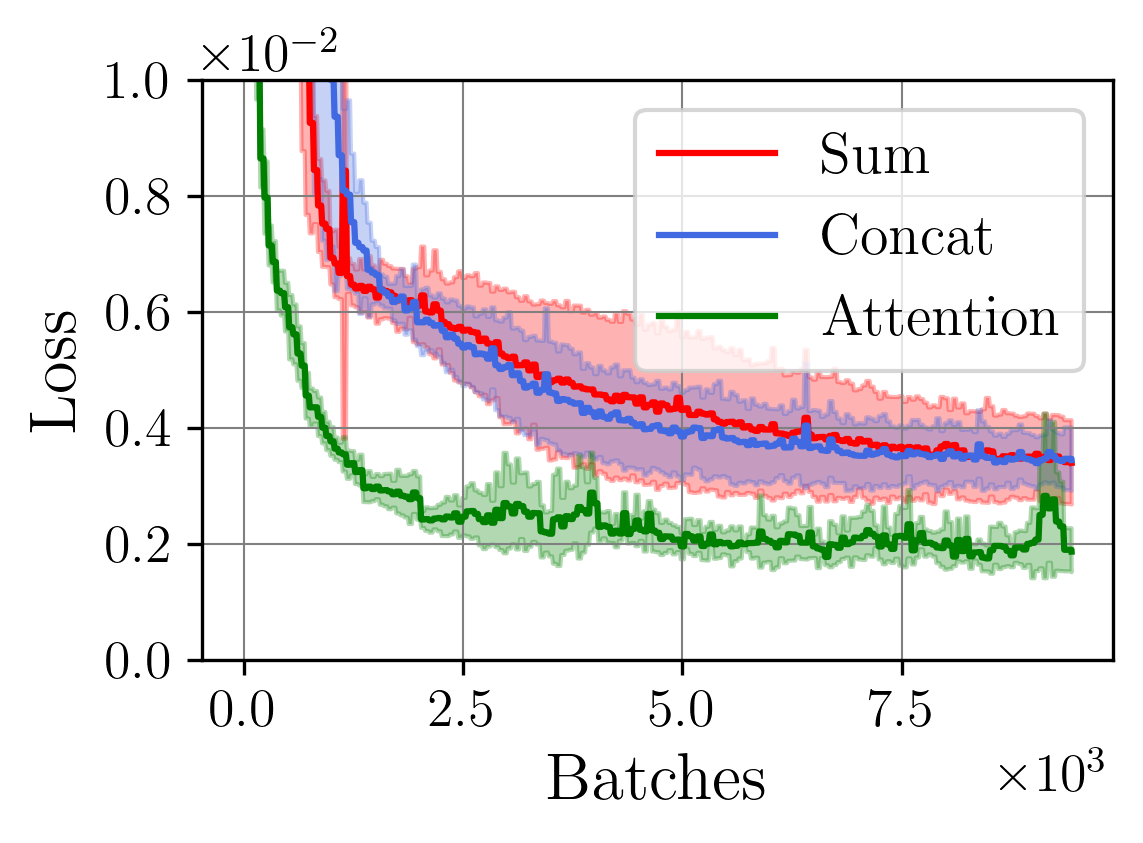}
         \caption{Overall}
     \end{subfigure}
    \caption{Testing loss of the multimodal autoencoder on the MuJoCo UR5 dataset using summation, concatenation, and attention aggregation. The solid line shows the mean over 10 trials and the shaded area indicates variance.}
    \label{fig:mujoco_loss_test_comparison}
\end{figure}

Following the Lipschitz constant estimation, we aim to examine the impact of the aggregation method on the training performance. To this end, we present the mean and variability of the training and testing dynamics for the aggregation methods in \autoref{fig:mujoco_loss_comparison} and \autoref{fig:mujoco_loss_test_comparison}. This analysis allows us to evaluate how different aggregation strategies influence the stability and convergence behavior.

\begin{figure}
     \centering
     \begin{subfigure}[b]{0.49\textwidth}
         \centering
         \includegraphics[width=\textwidth]{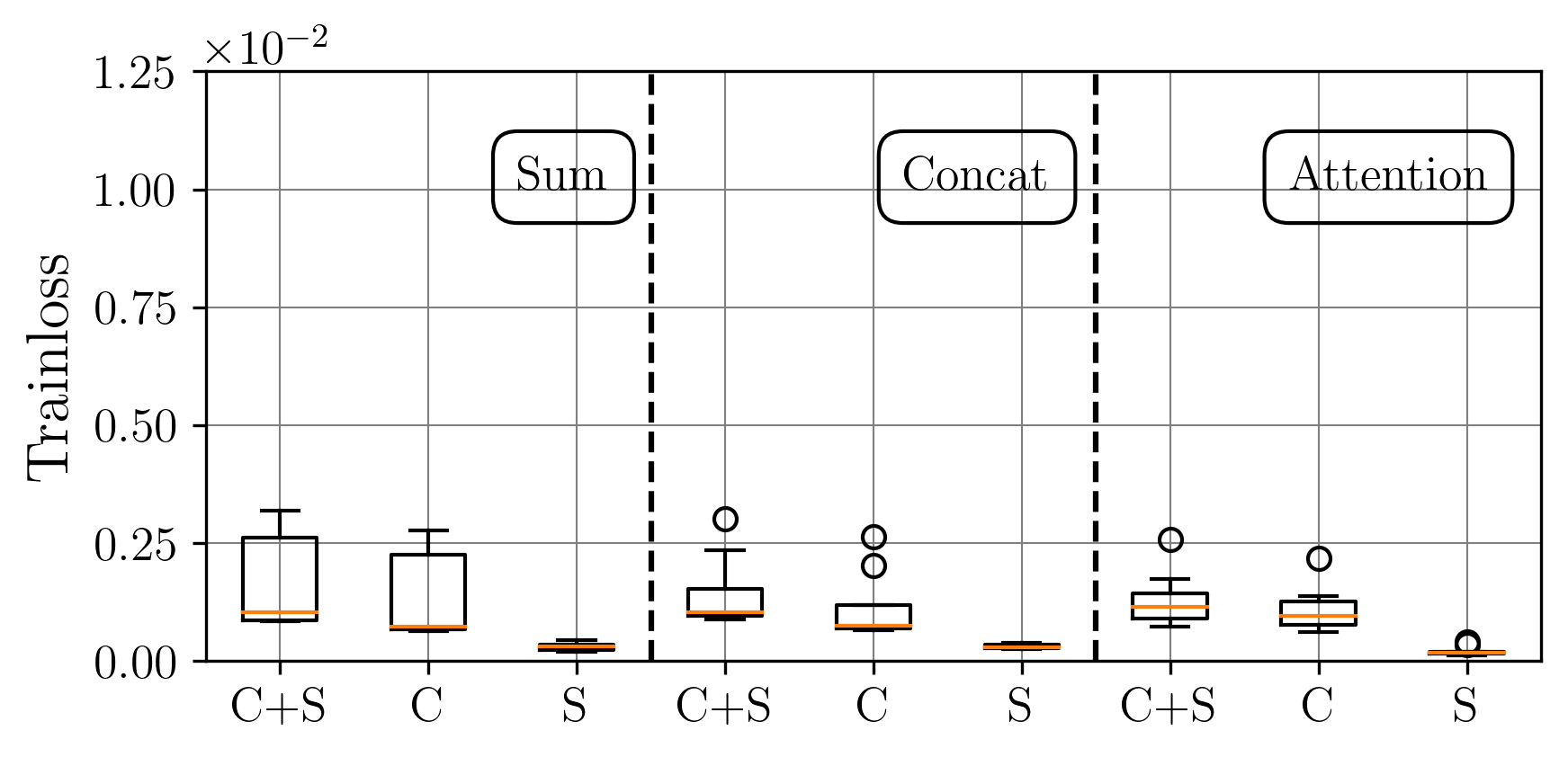}
         \caption{Training}
     \end{subfigure}
     \hfill
     \begin{subfigure}[b]{0.49\textwidth}
         \centering
         \includegraphics[width=\textwidth]{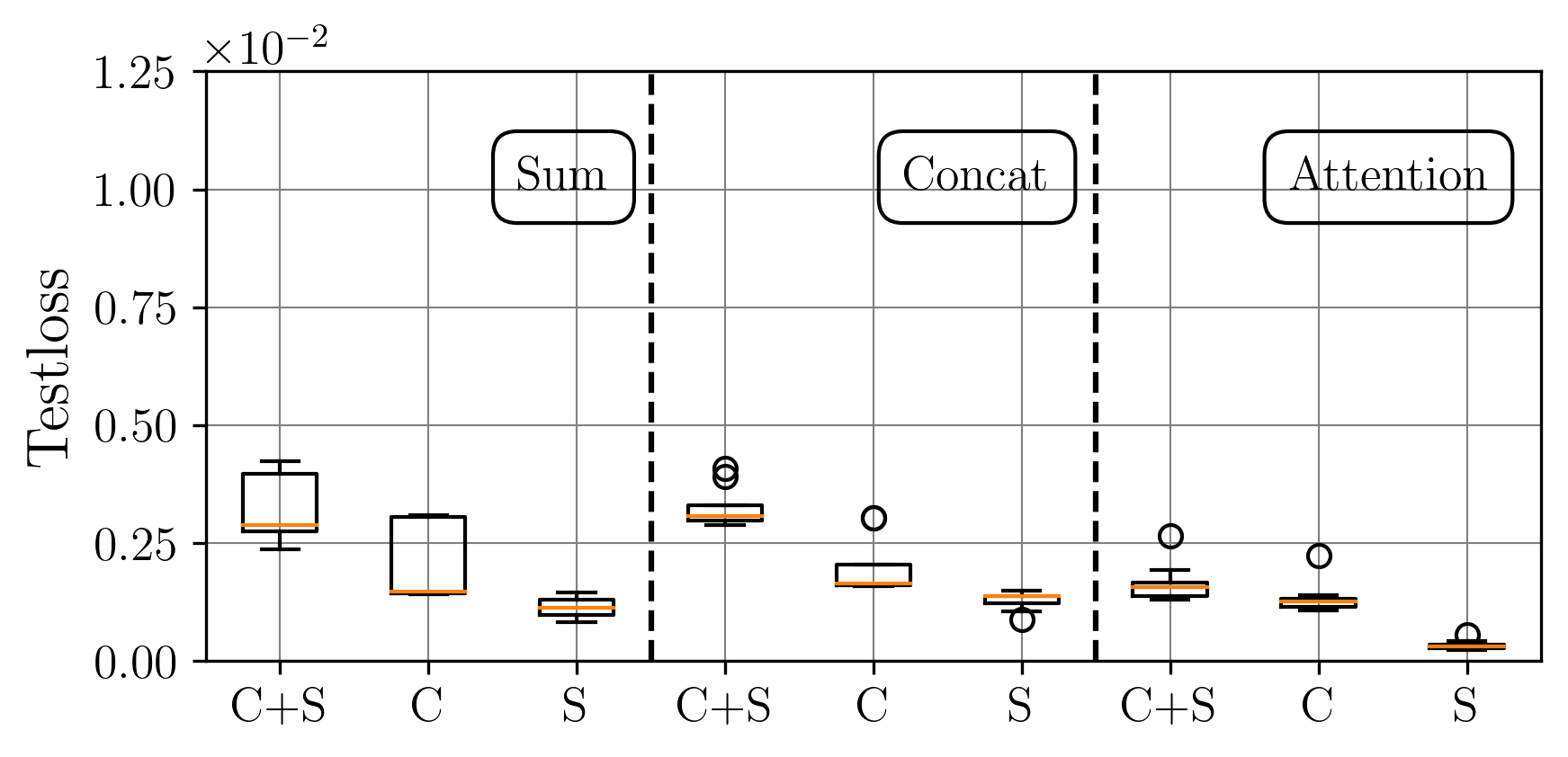}
         \caption{Testing}
     \end{subfigure}
    \caption{Boxplot analysis of loss distributions on the MuJoCo UR5 dataset for different fusion strategies: summation, concatenation, and attention. The label S represents the sensor loss, C the camera loss, and C+S the overall loss.}
    \label{fig:mujoco_loss_boxplot_comparison}
\end{figure}

In \autoref{fig:mujoco_loss_test_comparison}, we observe that the concatenation method, as demonstrated theoretically in \autoref{ssec:lipschitz}, exhibits lower variation in both training and test loss across the 10 trials compared to the sum. This suggests greater stability and consistency in the performance of the model when using concatenation as the aggregation strategy. The attention-based aggregation shows the best results in terms of consistency and performance. To further verify this observation, we provide a statistical evaluation in \autoref{fig:mujoco_loss_boxplot_comparison} using boxplots. These boxplots offer a detailed visual representation of the distribution, spread, and variability of the training and test losses for each aggregation method. 

\subsection{Performance Evaluation - Single Robot Welding Station}

\begin{figure}
     \centering
     \begin{subfigure}[b]{0.323\textwidth}
         \centering
         \includegraphics[width=\textwidth]{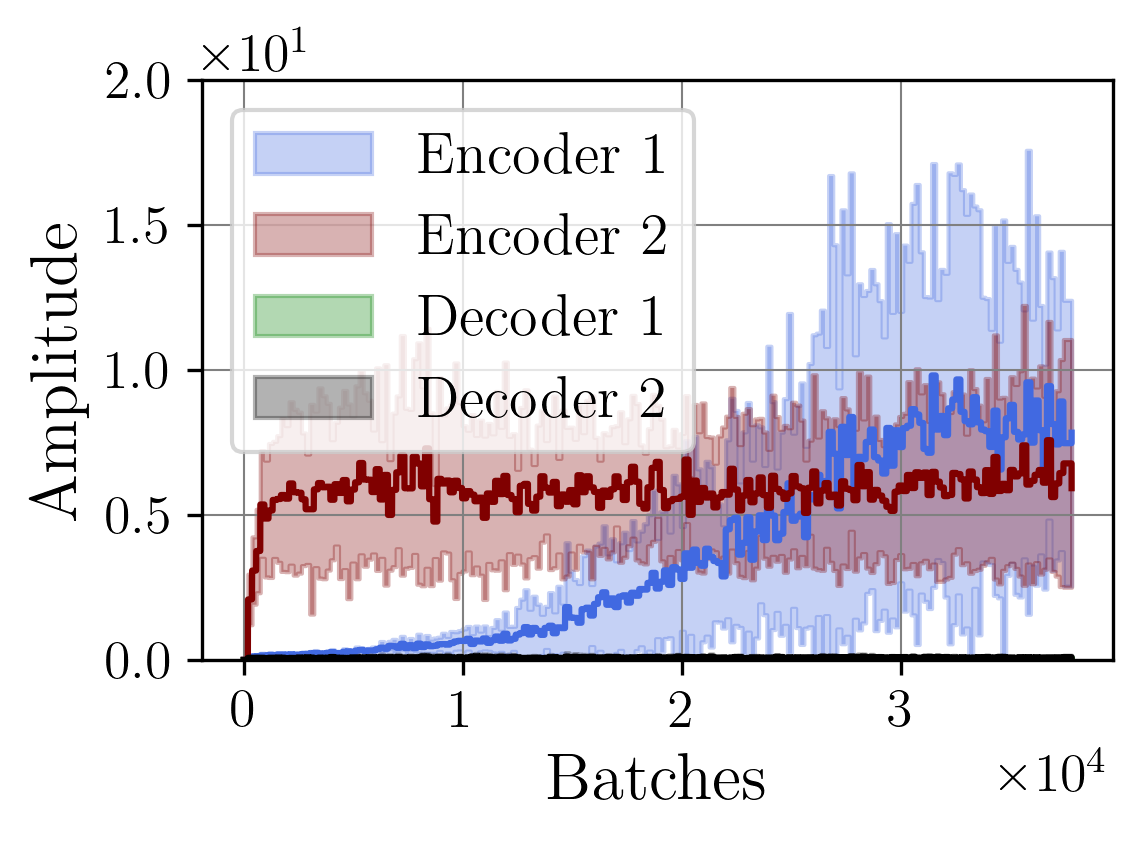}
         \caption{Summation}
     \end{subfigure}
     \hfill
     \begin{subfigure}[b]{0.323\textwidth}
         \centering
         \includegraphics[width=\textwidth]{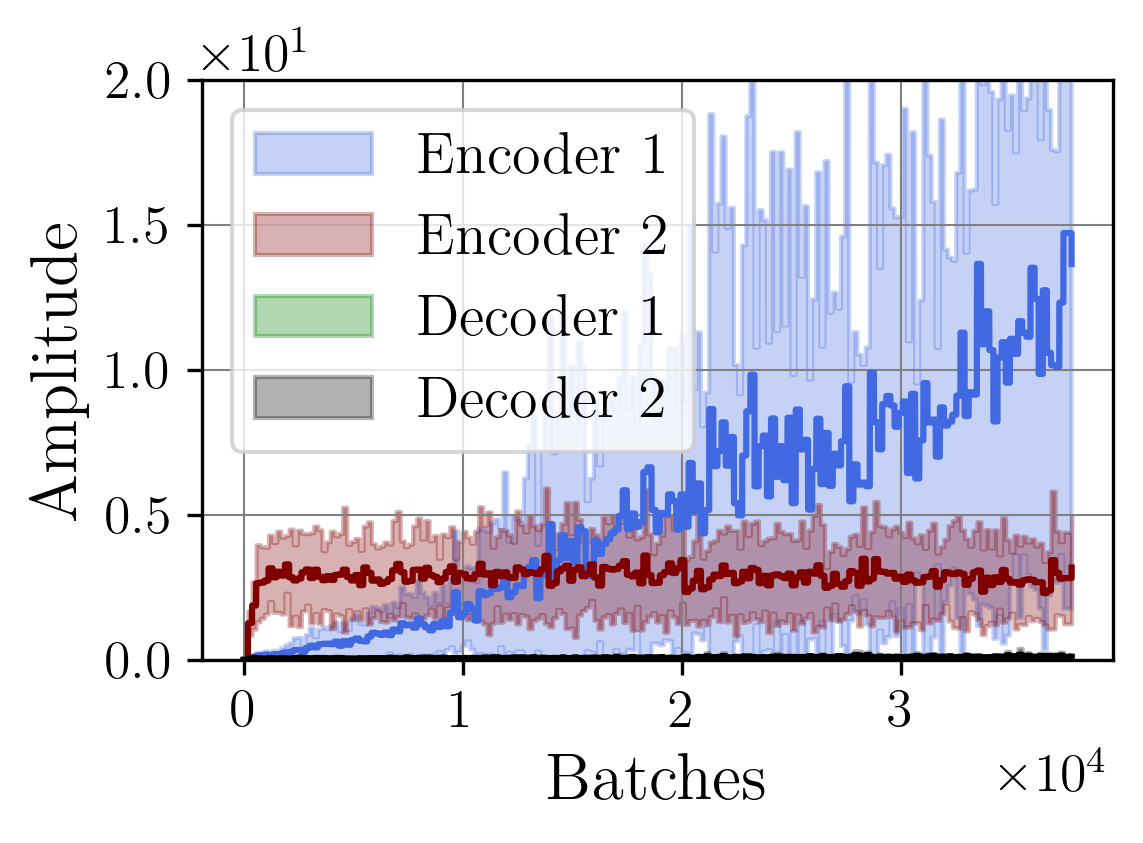}
         \caption{Concatenation}
     \end{subfigure}
     \hfill
     \begin{subfigure}[b]{0.323\textwidth}
         \centering
         \includegraphics[width=\textwidth]{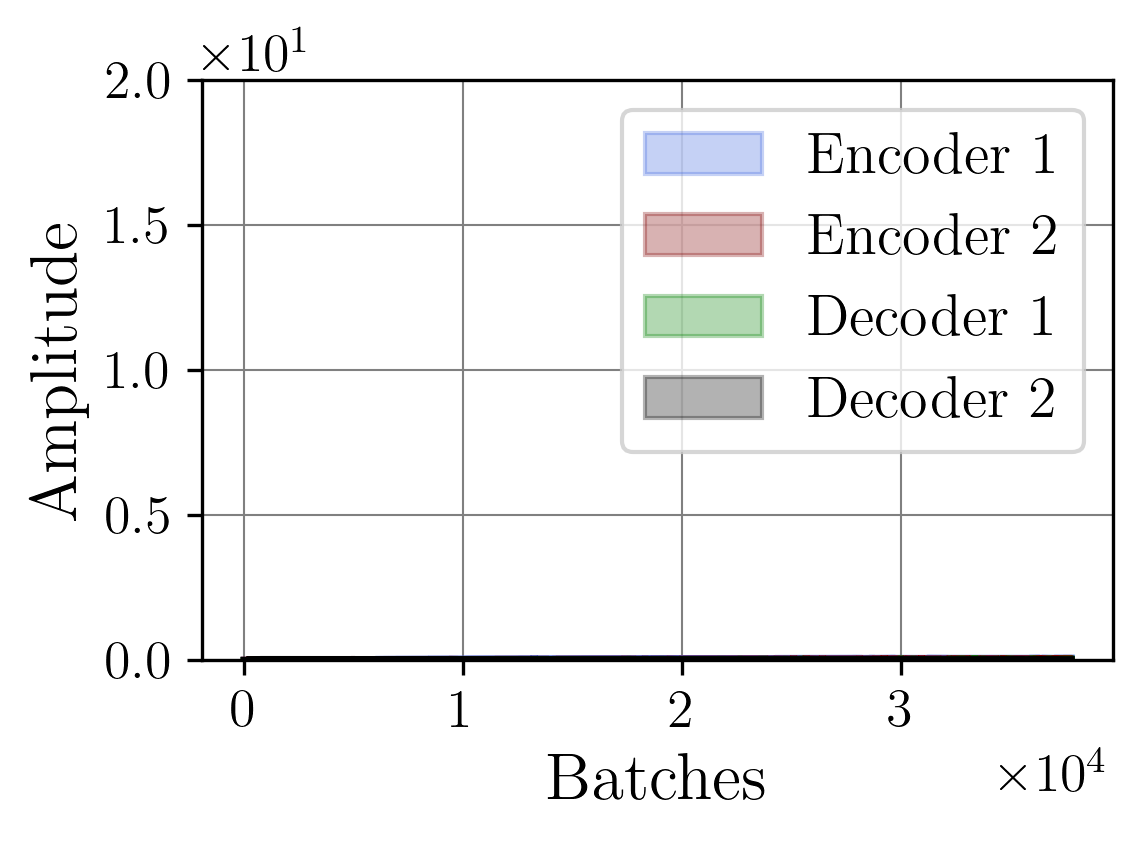}
         \caption{Attention}
     \end{subfigure}
    \caption{Estimated Lipschitz constants of each submodel in the multimodal autoencoder during training on the single-robot welding station dataset, using summation, concatenation, and attention. Solid lines show the mean across trials; shaded areas indicate variation.}
    \label{fig:abb1_lipschitz_comparison}
\end{figure}

We extend the same evaluation methodology to the single-robot welding station dataset. The results are more clear due to two key factors: (1) the larger visual representation of the robot in the images, enhancing dynamic feature visibility, and (2) the robot's constrained action range, potentially reducing motion variability. A comparative analysis of Lipschitz continuity metrics in \autoref{fig:abb1_lipschitz_comparison}, both methods (sum and concatenation) produce values more than ten times higher than those obtained with regularized attention, suggesting that attention mechanisms contribute to greater stability in gradient propagation.  

In the summation-based architecture (Figure 9a), we observe a consistent increase in the Lipschitz constant, particularly in Encoder 1, which is in charge of the image modality. The decoders exhibit lower Lipschitz values, but still show a rising trend, indicating a general instability in this formulation. The concatenation strategy (Figure 9b) demonstrates slightly improved stability. Although Encoder 1 still shows increasing Lipschitz values, the growth is more gradual compared to the summation approach. Encoders remain more sensitive than decoders, but the overall amplitude is reduced, implying a more controlled propagation of gradients. In contrast, the regularized attention-based model (Figure 9c) maintains exceptionally low Lipschitz constants across all submodules throughout training. This behavior is a direct consequence of the architectural choice to scale attention weights by their norm and to apply explicit regularization. This design effectively constrains the model's sensitivity to input changes, promoting both robustness and smooth optimization dynamics.

\begin{figure}
     \centering
     \begin{subfigure}[b]{0.325\textwidth}
         \centering
         \includegraphics[width=\textwidth]{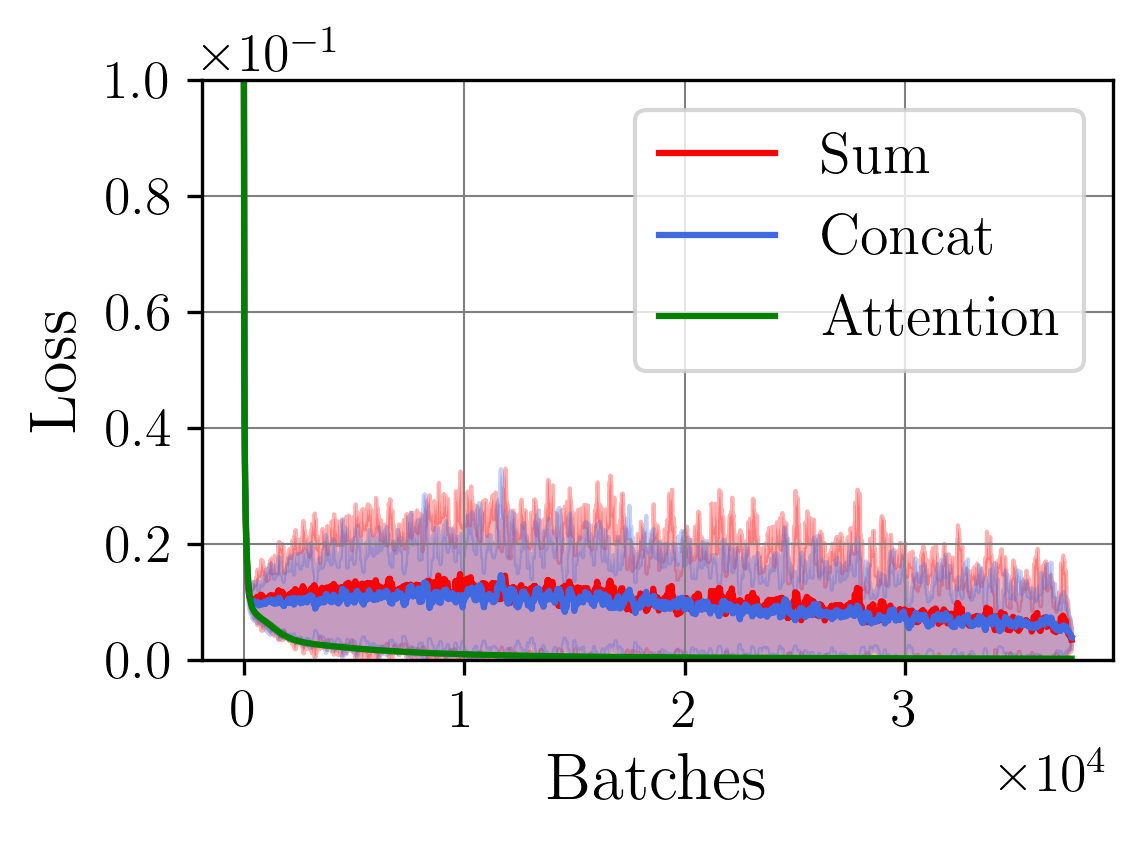}
         \caption{Camera}
     \end{subfigure}
     \hfill
     \begin{subfigure}[b]{0.325\textwidth}
         \centering
         \includegraphics[width=\textwidth]{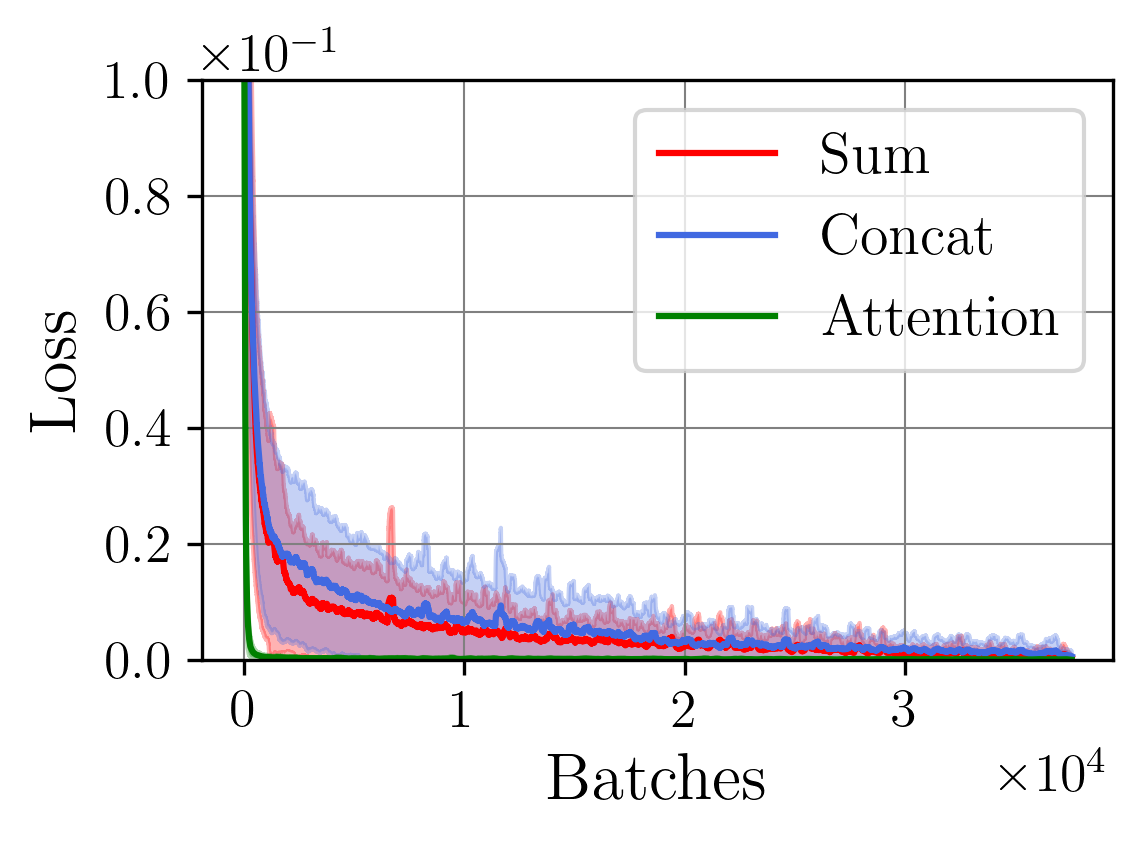}
         \caption{Sensor}
     \end{subfigure}
     \hfill
     \begin{subfigure}[b]{0.325\textwidth}
         \centering
         \includegraphics[width=\textwidth]{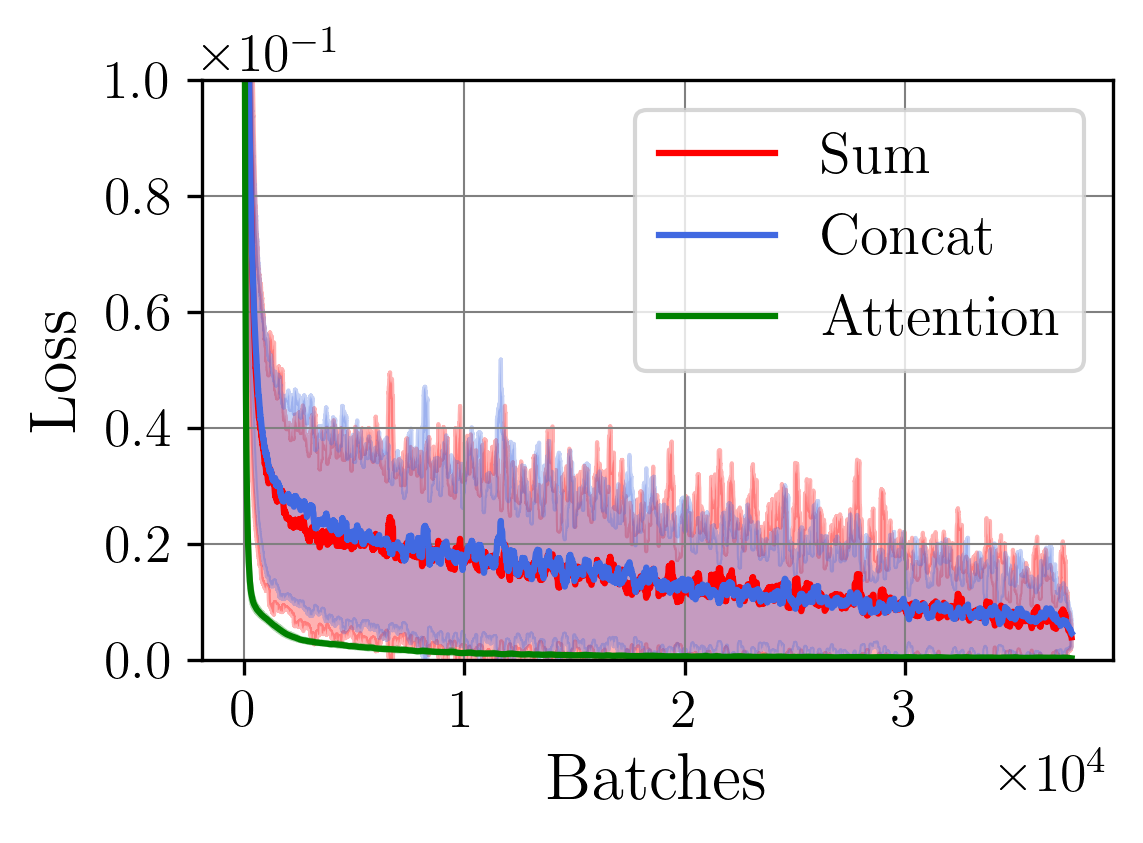}
         \caption{Overall}
     \end{subfigure}
    \caption{Training loss of the multimodal autoencoder on the single-robot welding station dataset using summation, concatenation, and attention aggregation. The solid line shows the mean over 10 trials and the shaded area indicates variance.}
    \label{fig:abb1_loss_comparison}
\end{figure}

\begin{figure}
     \centering
     \begin{subfigure}[b]{0.325\textwidth}
         \centering
         \includegraphics[width=\textwidth]{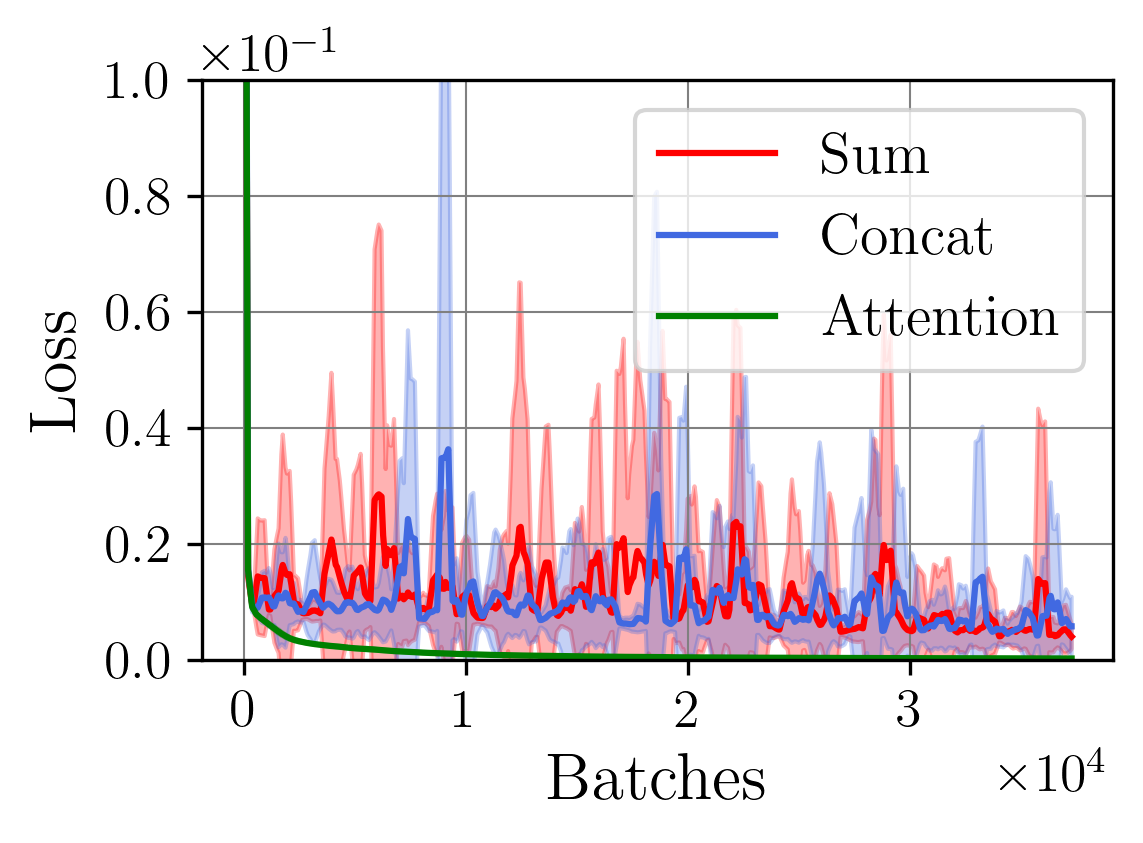}
         \caption{Camera}
     \end{subfigure}
     \hfill
     \begin{subfigure}[b]{0.325\textwidth}
         \centering
         \includegraphics[width=\textwidth]{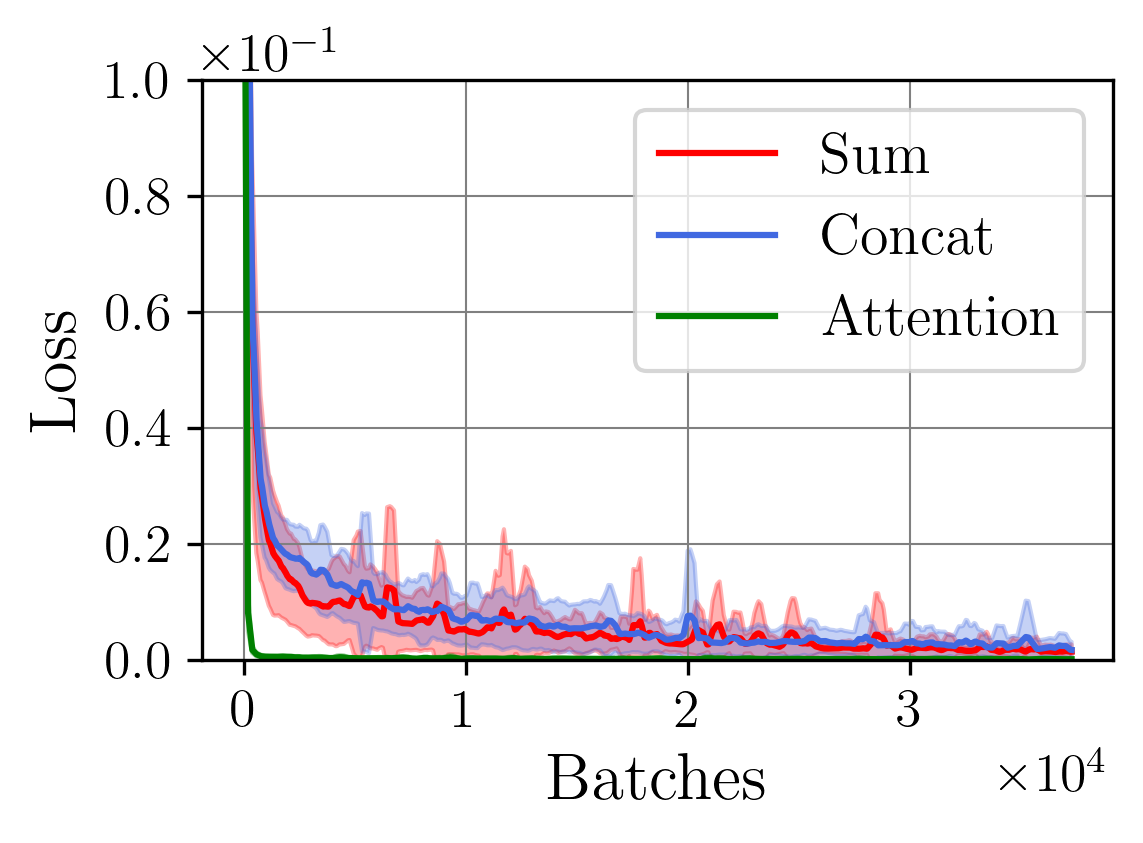}
         \caption{Sensor}
     \end{subfigure}
     \hfill
     \begin{subfigure}[b]{0.325\textwidth}
         \centering
         \includegraphics[width=\textwidth]{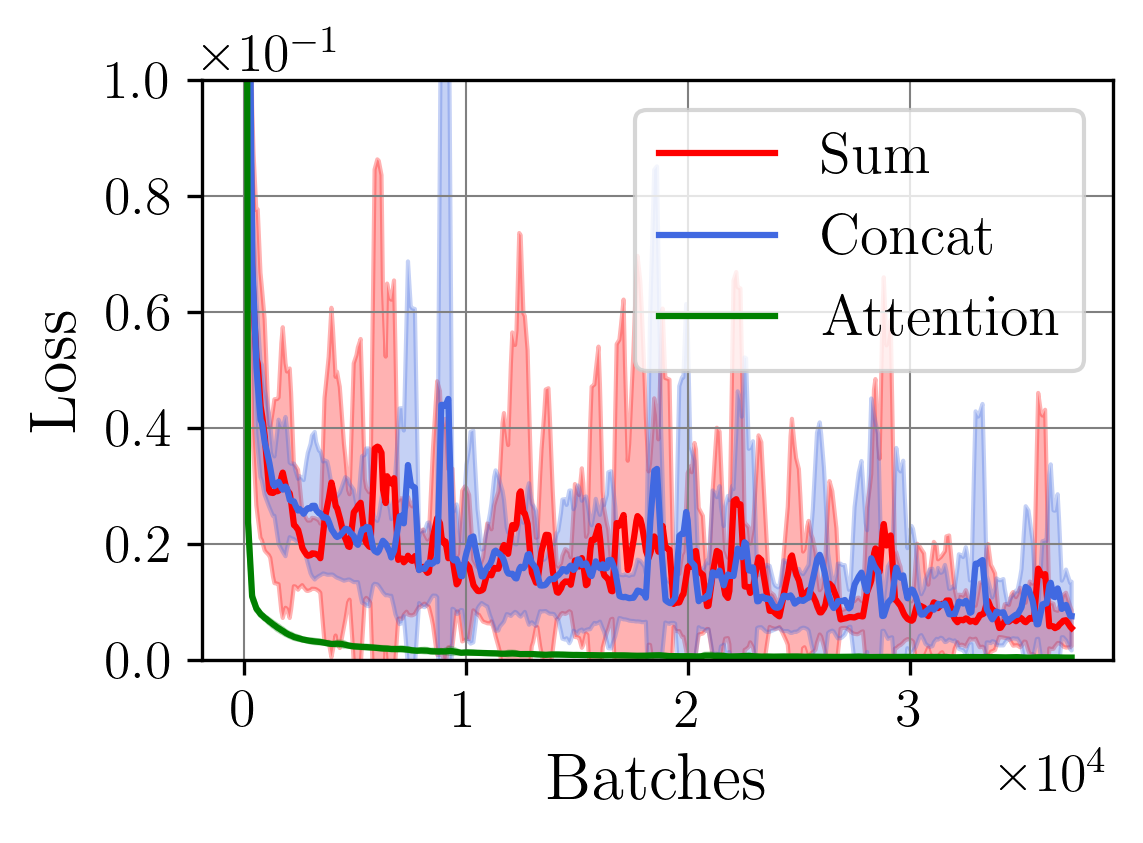}
         \caption{Overall}
     \end{subfigure}
    \caption{Testing loss of the multimodal autoencoder on the single-robot welding station dataset using summation, concatenation, and attention aggregation. The solid line shows the mean over 10 trials and the shaded area indicates variance.}
    \label{fig:abb1_loss_test_comparison}
\end{figure}

\autoref{fig:abb1_loss_comparison} and \autoref{fig:abb1_loss_test_comparison} compare the training and testing losses, respectively, across the three aggregation methods: summation, concatenation, and attention, evaluated on camera, sensor, and overall data modalities. Each curve represents the mean over 10 trials, with shaded regions indicating batch-wise variation using the single robot welding station dataset.

During training, all methods demonstrate a general trend of decreasing loss, but important differences emerge in both convergence speed and stability. The attention-based model (green curve) exhibits the fastest convergence and reaches the lowest loss across all input types. Its loss curve quickly flattens near zero, with minimal variation, indicating both efficient learning and robust generalization during optimization. In contrast, the summation (red) and concatenation (blue) approaches converge more slowly and with significantly more variance, particularly in the early phases of training. Notably, the classic fusion approaches tend to a higher loss level, especially in the camera modality, suggesting poorer optimization dynamics. The attention mechanism consistently achieves the lowest test loss across all modalities, with minimal fluctuations, indicating strong generalization.

\subsection{Performance Evaluation - Dual Robot Welding Station}

\begin{figure}
     \centering
     \begin{subfigure}[b]{0.323\textwidth}
         \centering
         \includegraphics[width=\textwidth]{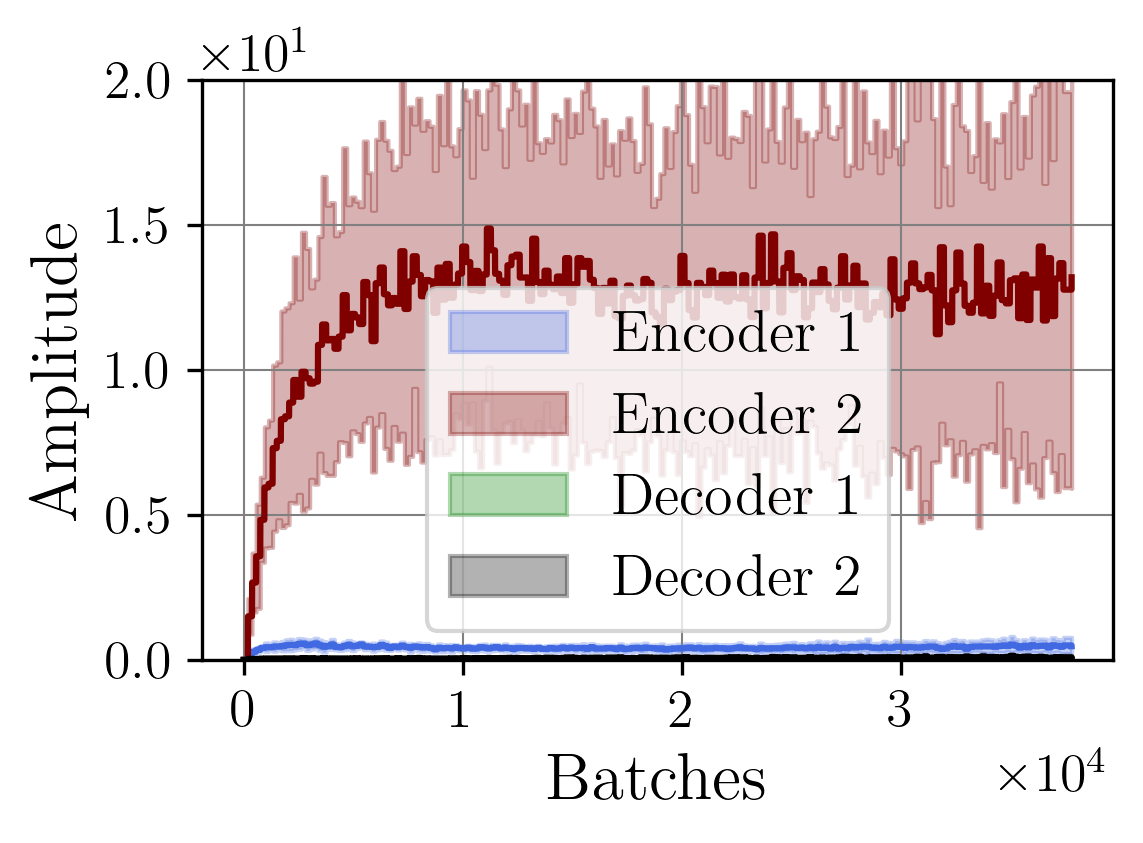}
         \caption{Summation}
     \end{subfigure}
     \hfill
     \begin{subfigure}[b]{0.323\textwidth}
         \centering
         \includegraphics[width=\textwidth]{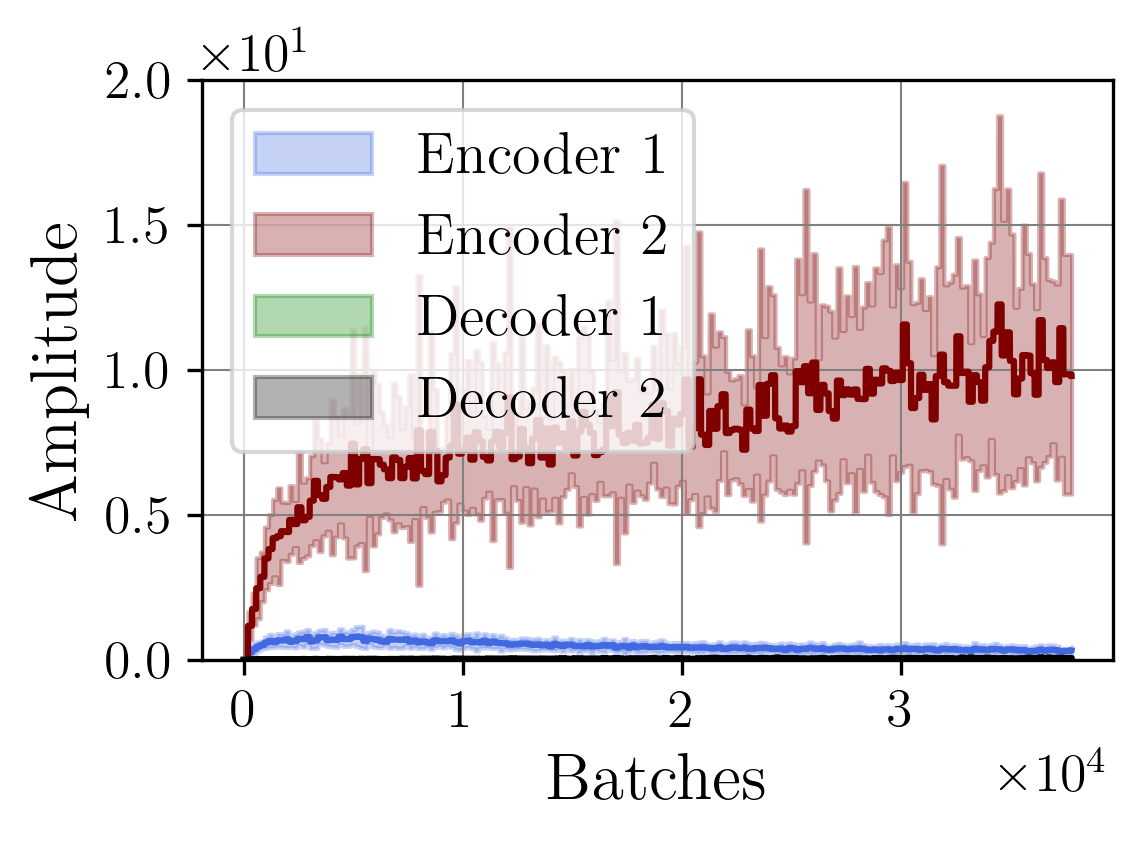}
         \caption{Concatenation}
     \end{subfigure}
     \hfill
     \begin{subfigure}[b]{0.323\textwidth}
         \centering
         \includegraphics[width=\textwidth]{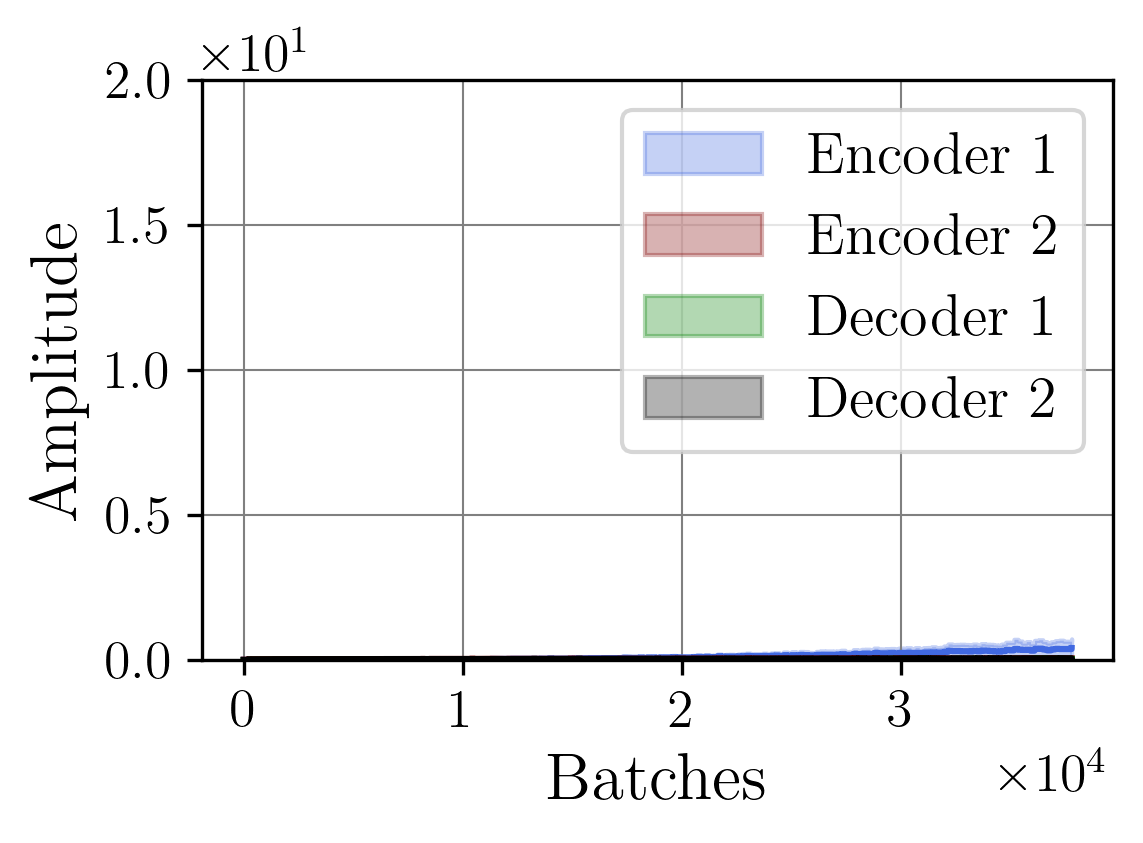}
         \caption{Attention}
     \end{subfigure}
    \caption{Estimated Lipschitz constants across 10 trials (mean: solid; variation: shaded areas) of each submodel during training on the dual-robot welding station dataset, using different aggregation methods.}
    \label{fig:abb2_lipschitz_comparison}
\end{figure}

Extending our evaluation to a cooperative dual-robot welding task, we observe consistent trends in aggregation performance despite key differences in data characteristics. In this scenario, two robots operate in a similarly constrained action space but are represented at a smaller scale within the images, resulting in reduced motion dynamics compared to the previous dataset. Remarkably, concatenation-based aggregation again yields lower Lipschitz constants than summation, with both methods exceeding regularized attention, which underscores attention’s inherent stability advantages. The persistence of this trend, even under weaker motion saliency, suggests that attention’s robustness stems not merely from motion clarity but from its adaptive feature weighting. \autoref{fig:abb2_lipschitz_comparison} illustrates the estimated Lipschitz constants for each submodule, two encoders and two decoders, during training across the three aggregation strategies: summation, concatenation, and attention.

In the summation-based architecture (Figure 12a), Encoder 2 exhibits a sharp rise in the Lipschitz constant early in training, stabilizing at a high amplitude. This suggests that this submodule becomes increasingly sensitive to input perturbations over time. The other components, particularly Encoder 1 and both decoders, maintain lower but still non-negligible values. The significant and persistent amplitude of Encoder 2 indicates that summation can lead to overamplification of certain pathways, potentially reducing robustness. The concatenation-based mode (Figure 12b) shows a similar pattern, though with slightly more controlled growth in Encoder 2. This implies that concatenation may mitigate some of the instability seen in summation, but still lacks sufficient regularization to enforce Lipschitz continuity effectively. In stark contrast, the attention-based architecture (Figure 12c) maintains remarkably low Lipschitz constants across all submodules throughout training. All curves remain near zero, suggesting strong stability and minimal sensitivity to input variations.

\begin{figure}
     \centering
     \begin{subfigure}[b]{0.325\textwidth}
         \centering
         \includegraphics[width=\textwidth]{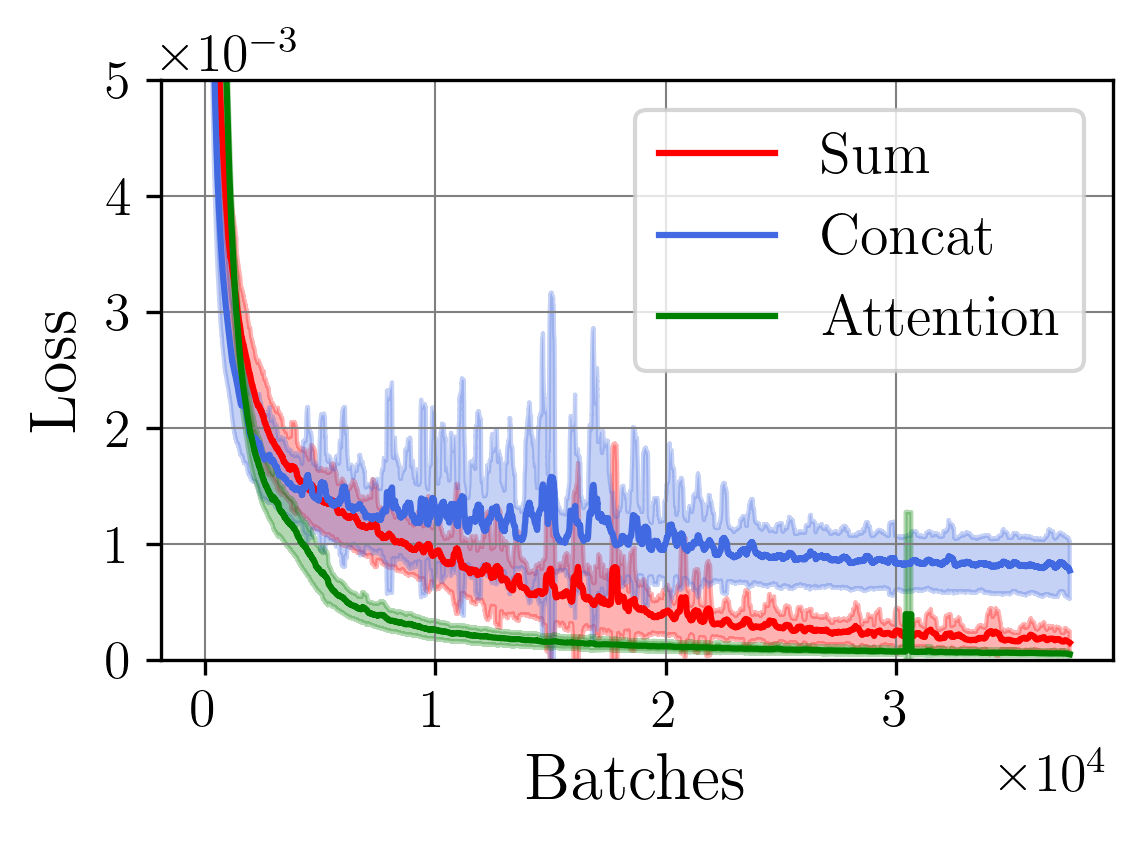}
         \caption{Camera}
     \end{subfigure}
     \hfill
     \begin{subfigure}[b]{0.325\textwidth}
         \centering
         \includegraphics[width=\textwidth]{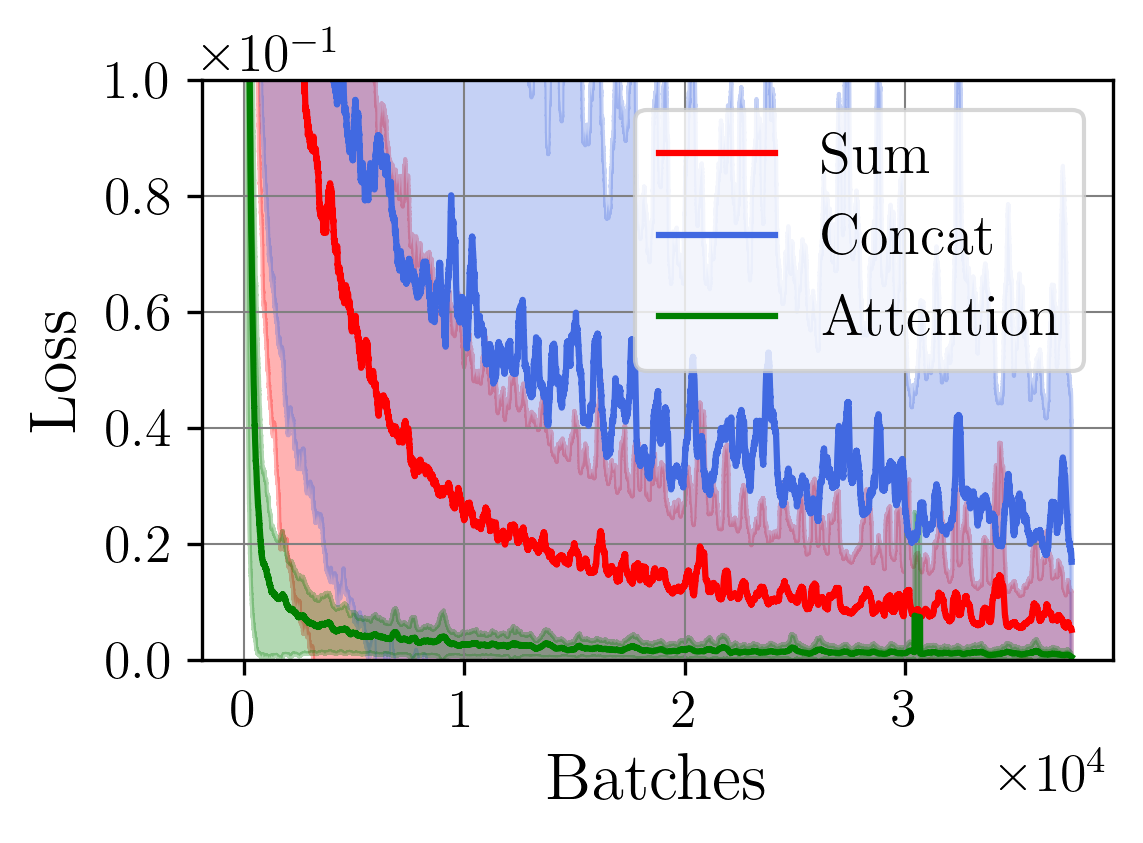}
         \caption{Sensor}
     \end{subfigure}
     \hfill
     \begin{subfigure}[b]{0.325\textwidth}
         \centering
         \includegraphics[width=\textwidth]{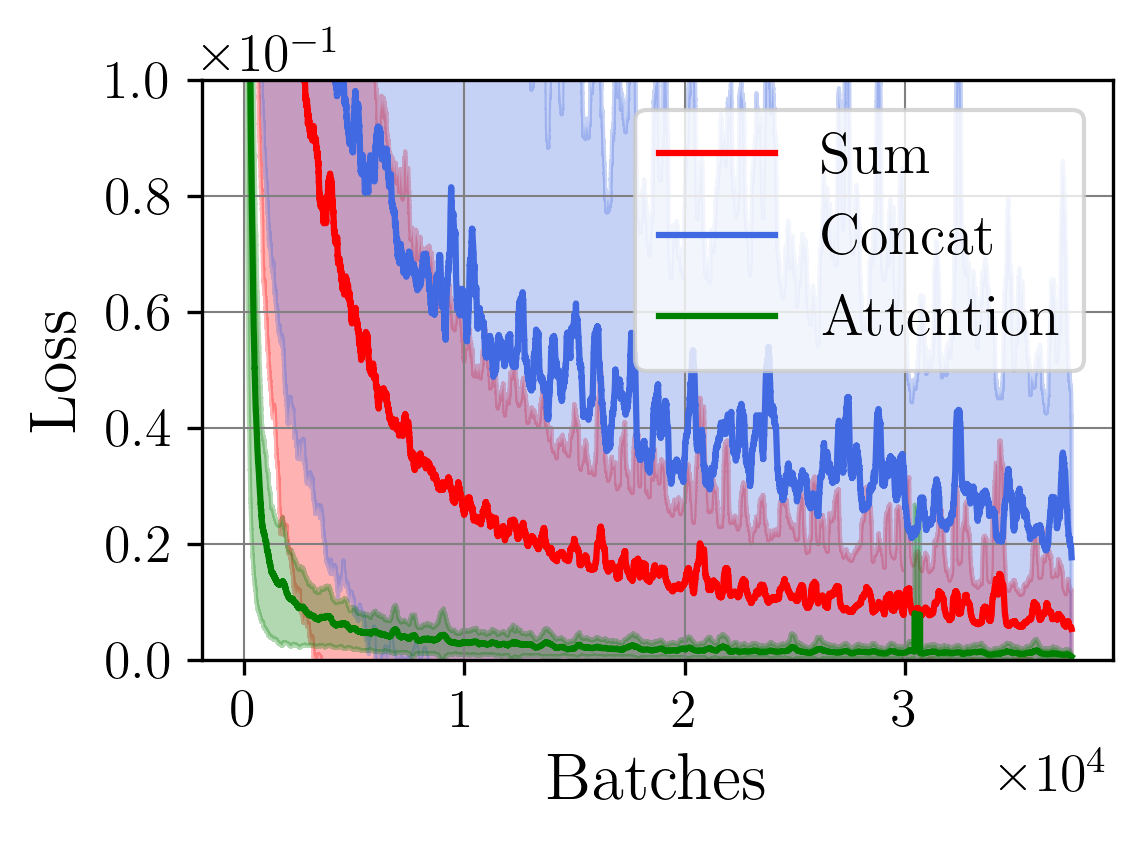}
         \caption{Overall}
     \end{subfigure}
    \caption{Training loss over 10 trials (mean: solid; variance: shaded area) of the multimodal autoencoder on the dual-robot welding station dataset using summation, concatenation, and attention aggregation.}
    \label{fig:abb2_loss_comparison}
\end{figure}

\begin{figure}
     \centering
     \begin{subfigure}[b]{0.325\textwidth}
         \centering
         \includegraphics[width=\textwidth]{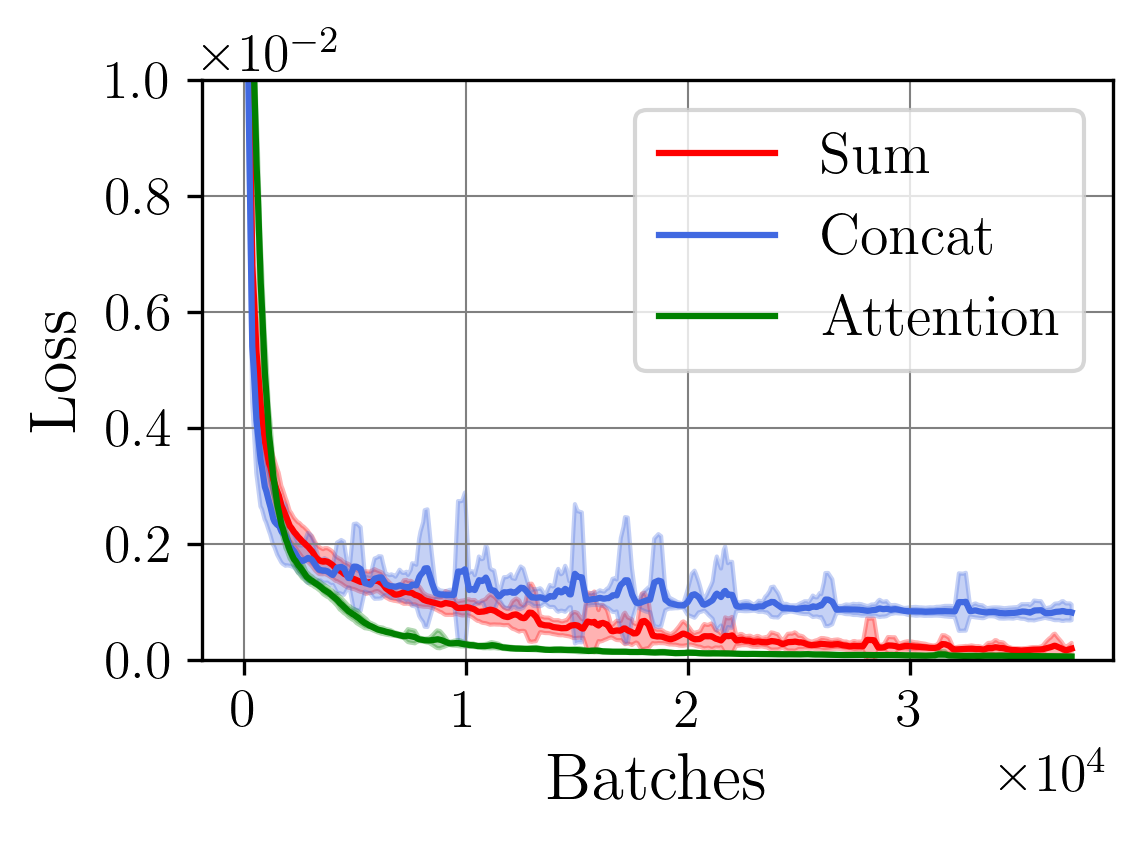}
         \caption{Camera}
     \end{subfigure}
     \hfill
     \begin{subfigure}[b]{0.325\textwidth}
         \centering
         \includegraphics[width=\textwidth]{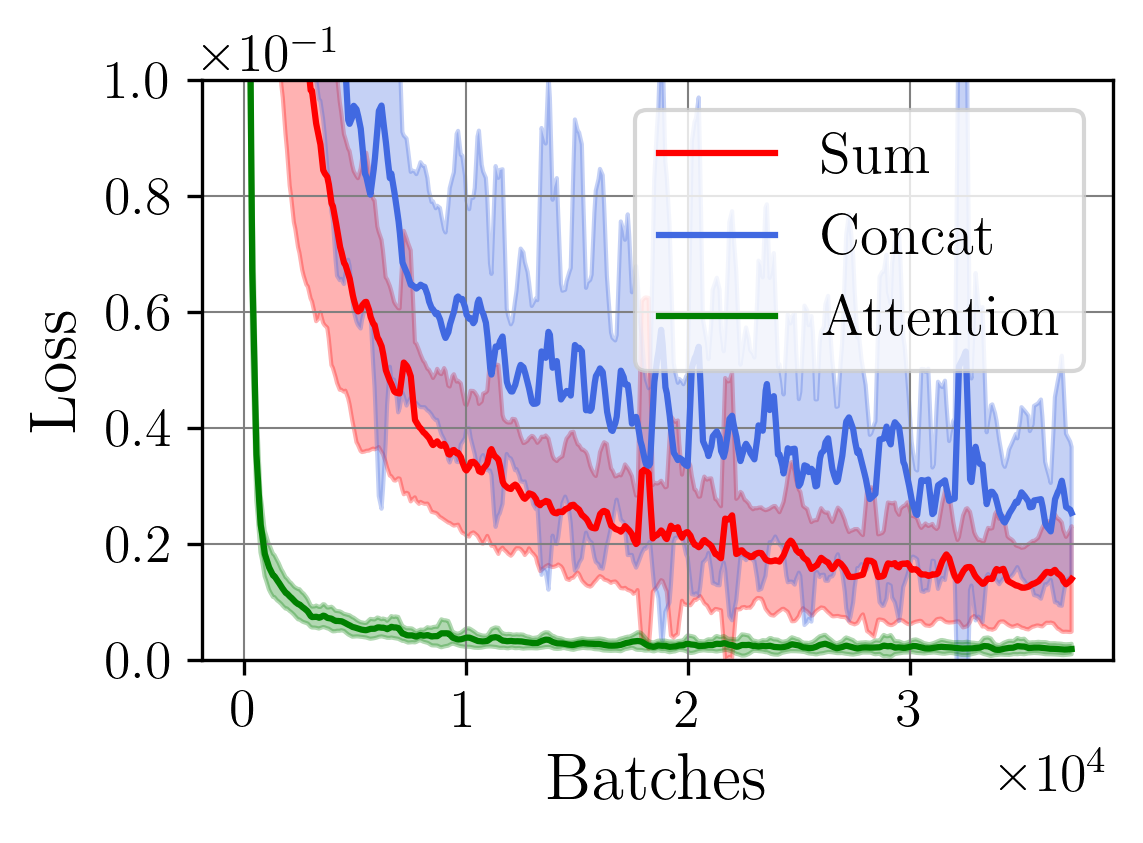}
         \caption{Sensor}
     \end{subfigure}
     \hfill
     \begin{subfigure}[b]{0.325\textwidth}
         \centering
         \includegraphics[width=\textwidth]{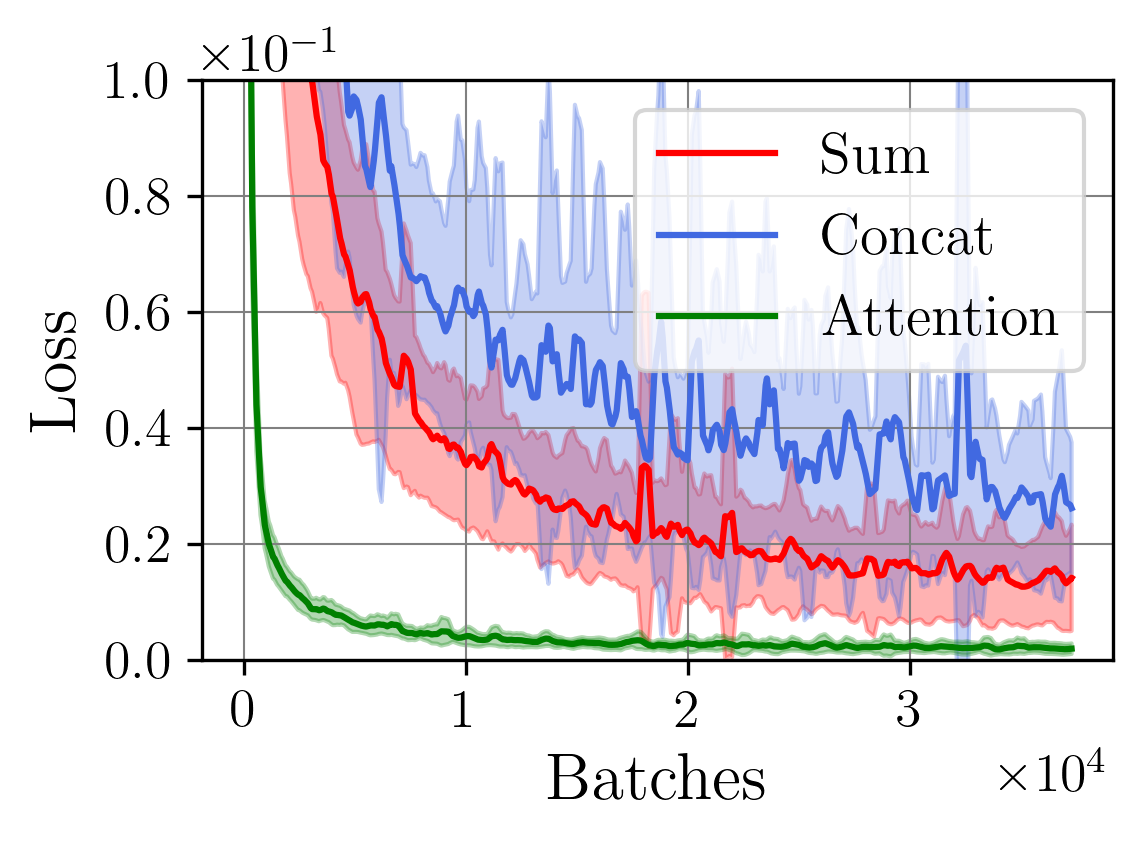}
         \caption{Overall}
     \end{subfigure}
    \caption{Testing loss over 10 trials (mean: solid; variance: shaded area) of the multimodal autoencoder on the dual-robot welding station dataset using summation, concatenation, and attention aggregation.}
    \label{fig:abb2_loss_test_comparison}
\end{figure}

Training dynamics in \autoref{fig:abb2_loss_comparison} and \autoref{fig:abb2_loss_test_comparison} further support this interpretation: while sum and concatenation exhibit pronounced loss fluctuations, attention maintains stable convergence. This implies that spatial-coherence priors in attention mechanisms compensate for diminished motion cues, enabling reliable optimization even when robotic motions are less visually prominent. Such consistency across datasets reinforces the suitability for industrial vision tasks where object scale and motion variability may vary substantially. Although concatenation offers a widely used method for fusing multimodal latent representations, it inherently treats each modality as an independent stream rather than components of a jointly structured system. By appending features side by side, this approach delegates the cross-modal interactions to the decoder. As a result, the fused representation lacks explicit alignment or interaction between modalities, which can limit reconstruction quality.


\begin{figure}
     \centering
     \begin{subfigure}[b]{0.24\textwidth}
         \centering
         \includegraphics[trim={0 1.cm 0 0},clip,width=\textwidth]{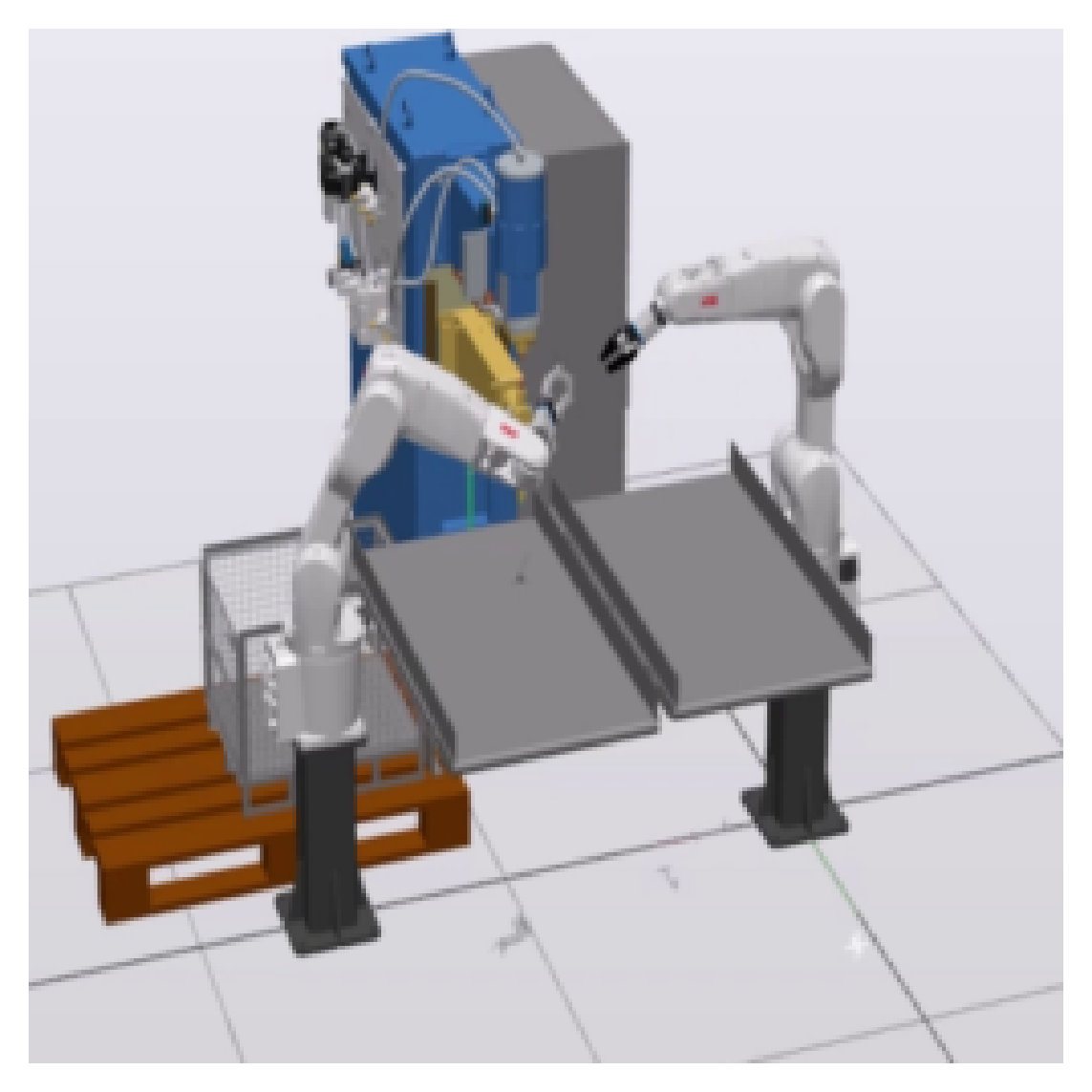}
     \end{subfigure}
     \hfill
     \begin{subfigure}[b]{0.24\textwidth}
         \centering
         \includegraphics[trim={0 1.cm 0 0},clip,width=\textwidth]{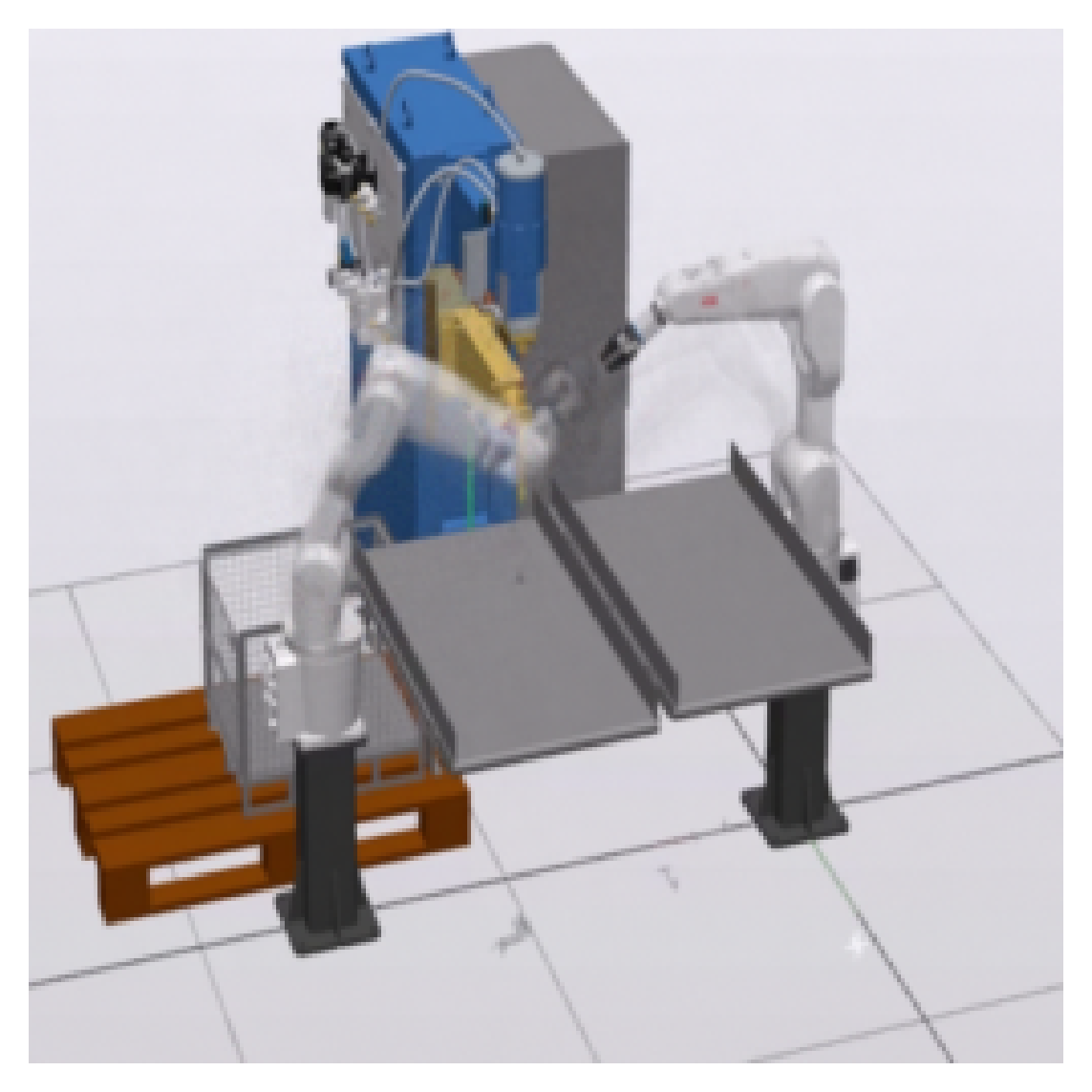}
     \end{subfigure}
     \hfill
     \begin{subfigure}[b]{0.24\textwidth}
         \centering
         \includegraphics[trim={0 1.cm 0 0},clip,width=\textwidth]{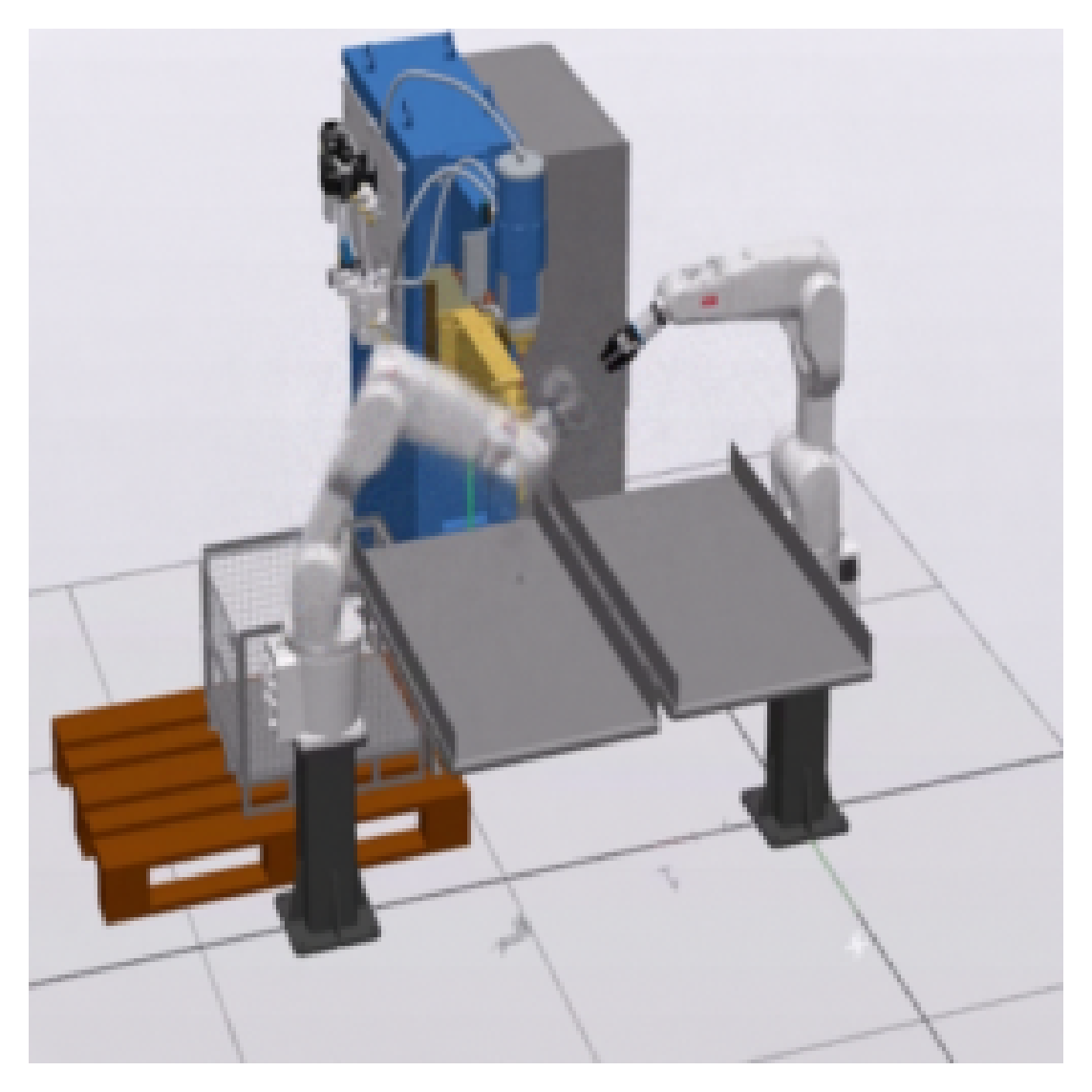}
     \end{subfigure}
    \hfill
     \begin{subfigure}[b]{0.24\textwidth}
         \centering
         \includegraphics[trim={0 1.cm 0 0},clip,width=\textwidth]{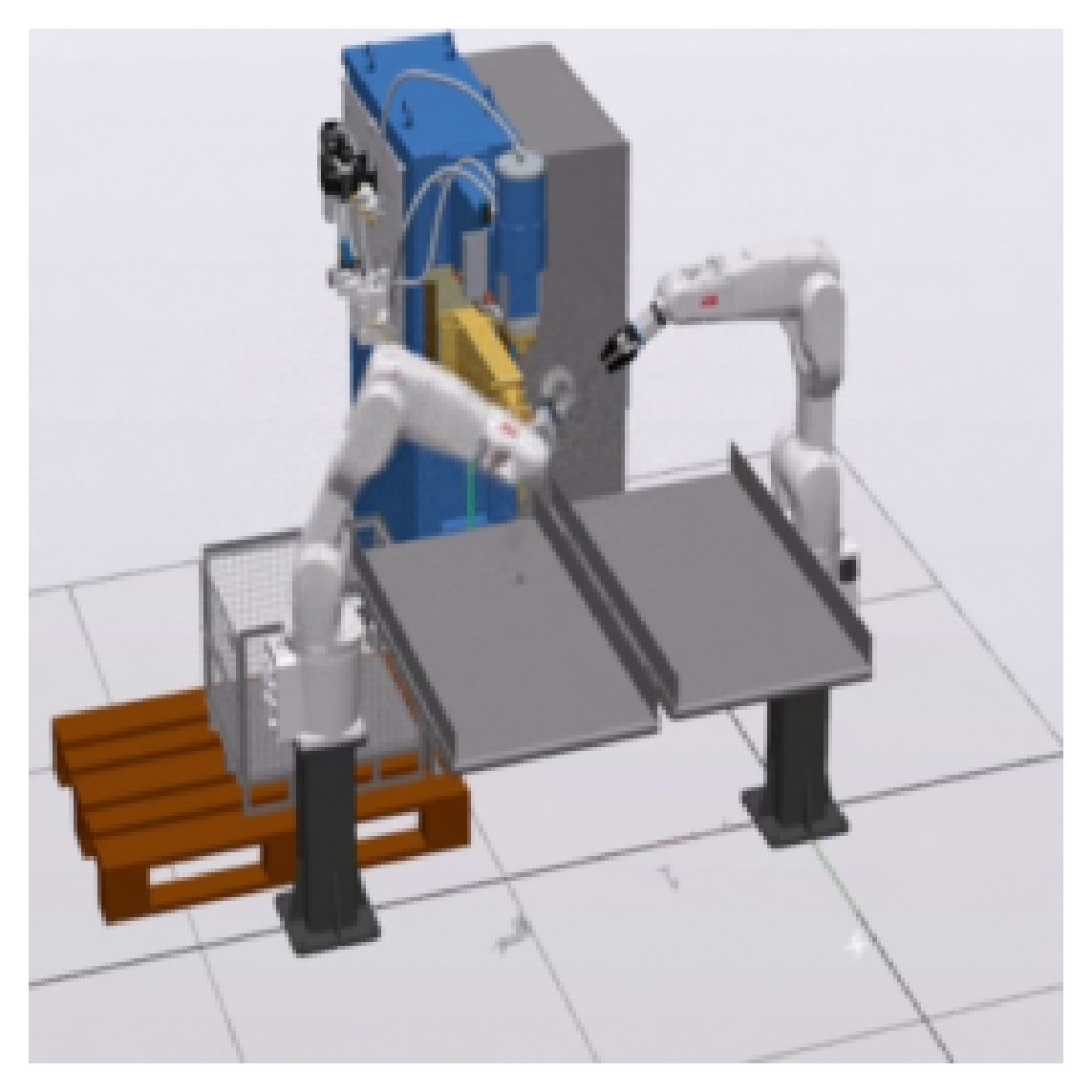}
     \end{subfigure}

          \begin{subfigure}[b]{0.24\textwidth}
         \centering
         \includegraphics[trim={0 1.cm 0 0},clip,width=\textwidth]{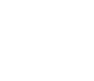}
     \end{subfigure}
     \hfill
     \begin{subfigure}[b]{0.24\textwidth}
         \centering
         \includegraphics[width=\textwidth]{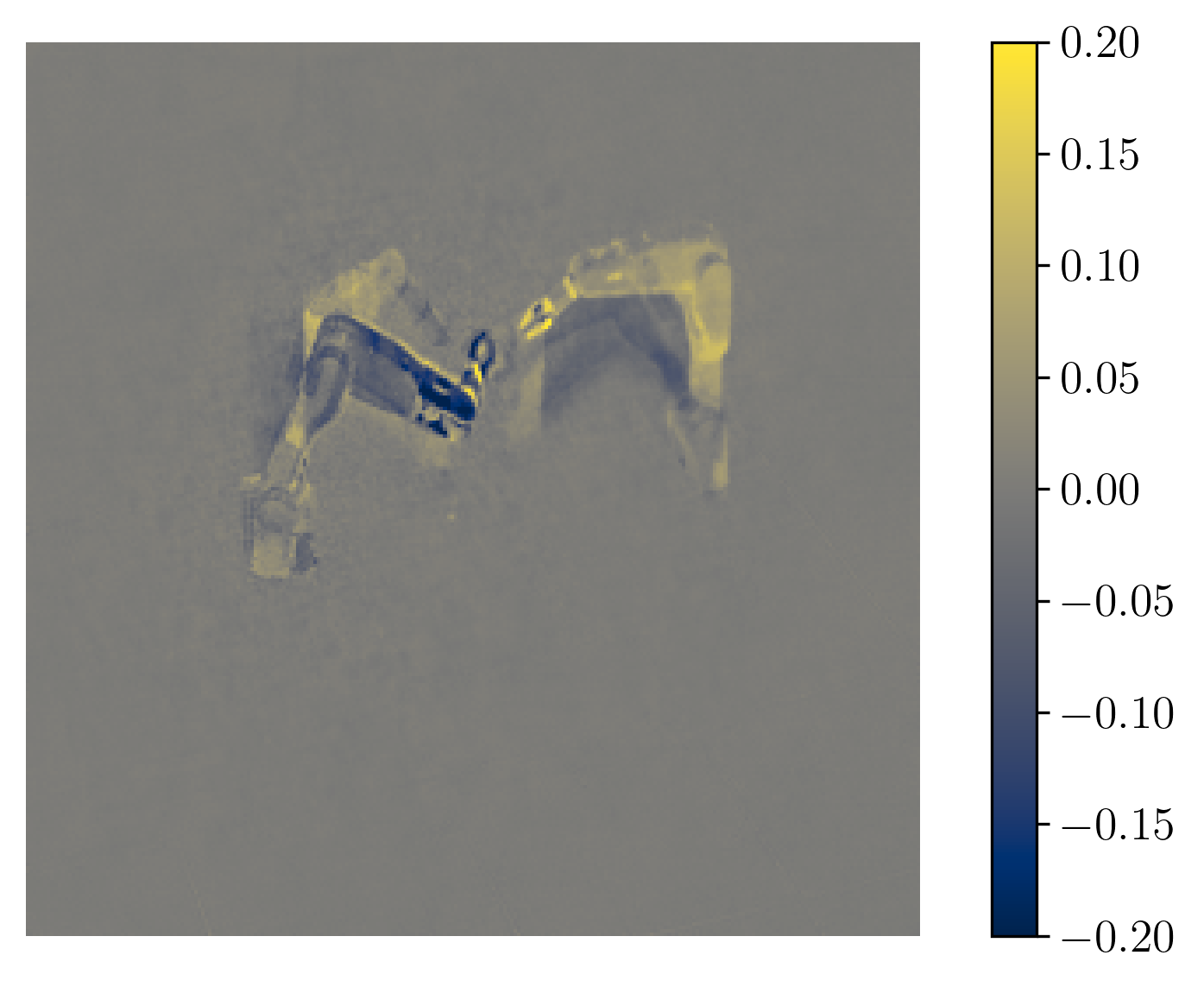}
     \end{subfigure}
     \hfill
     \begin{subfigure}[b]{0.24\textwidth}
         \centering
         \includegraphics[width=\textwidth]{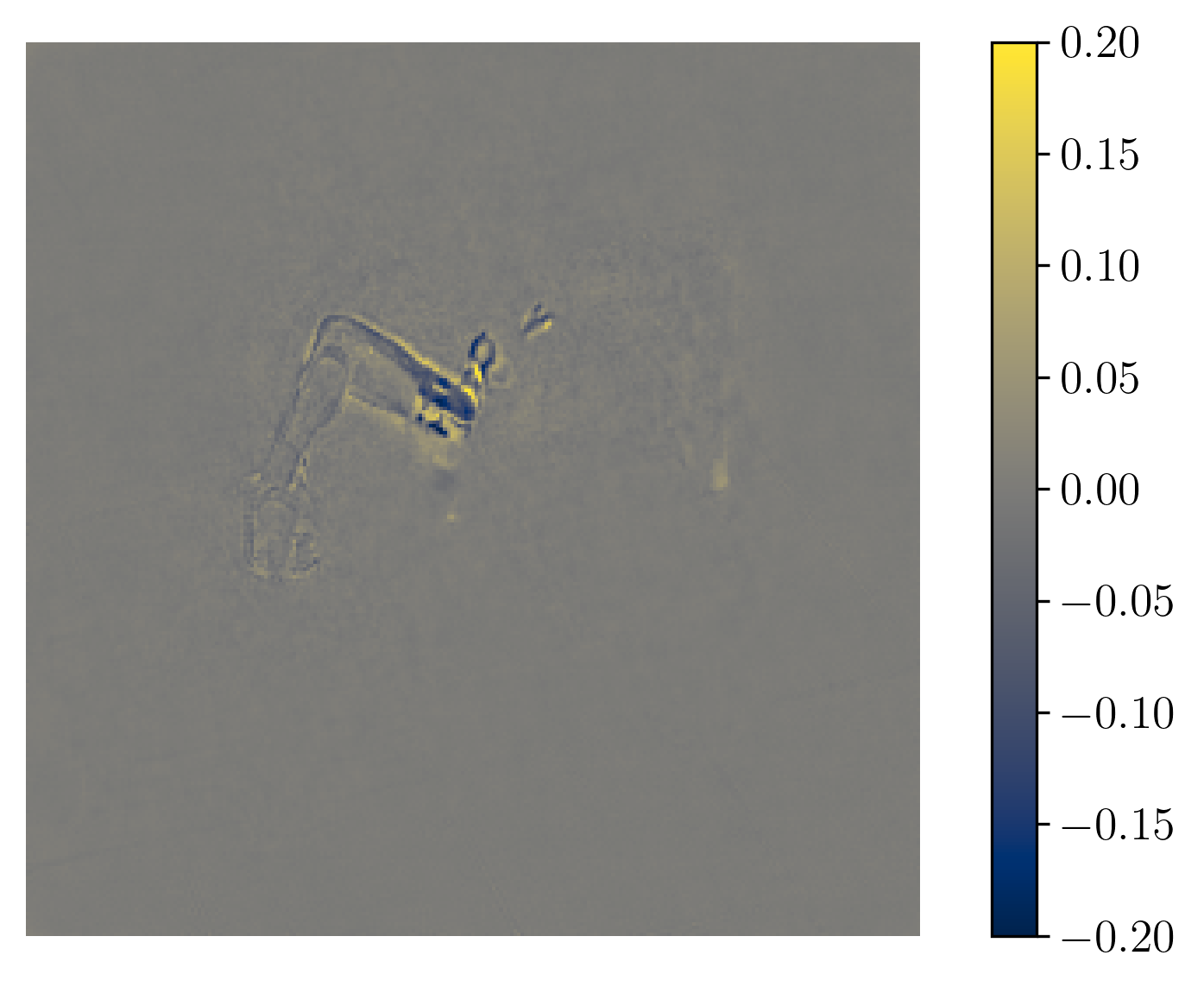}
     \end{subfigure}
    \hfill
     \begin{subfigure}[b]{0.24\textwidth}
         \centering
         \includegraphics[width=\textwidth]{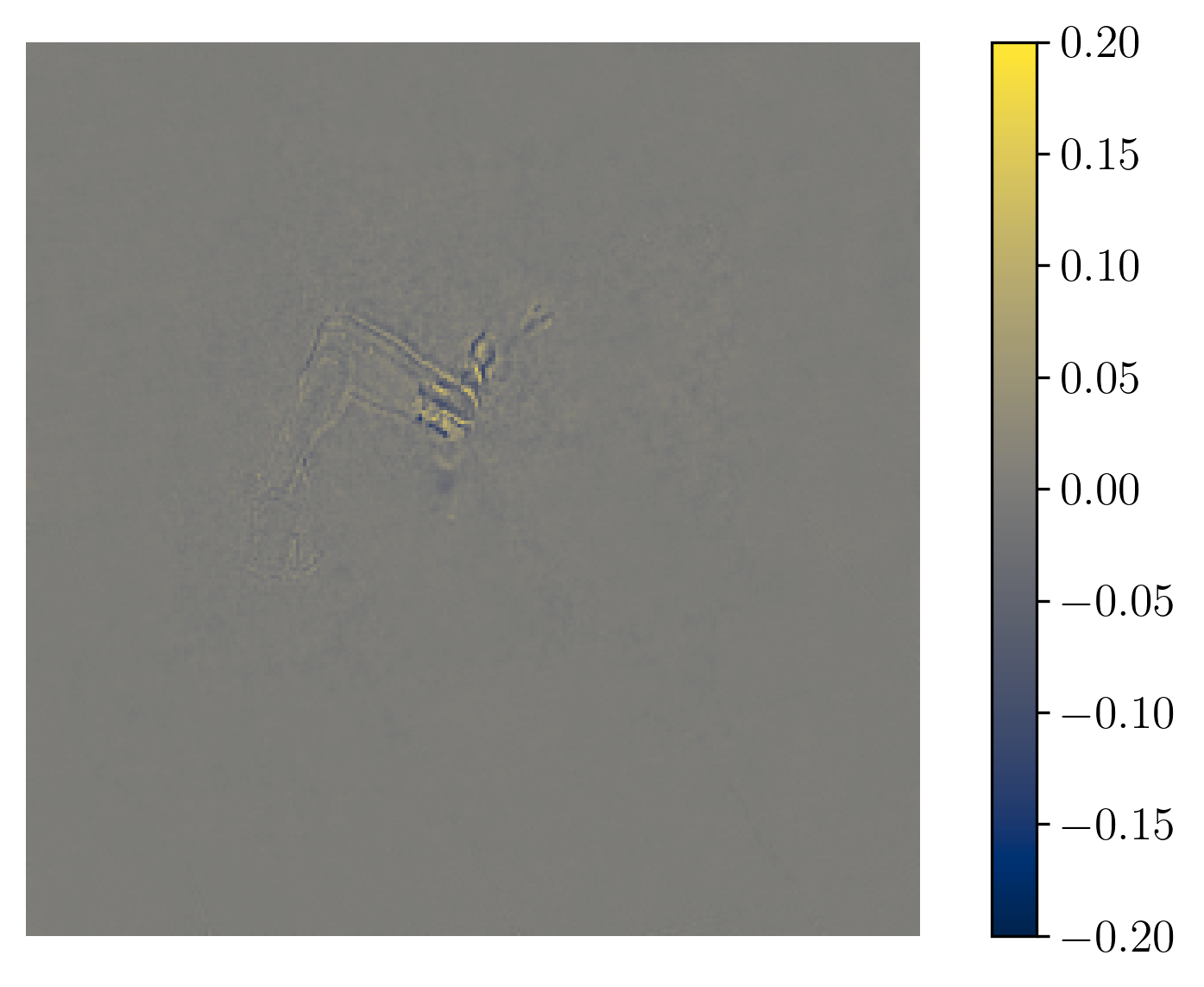}
     \end{subfigure}

          \begin{subfigure}[b]{0.24\textwidth}
         \centering
         \includegraphics[trim={0 1.cm 0 0},clip,width=\textwidth]{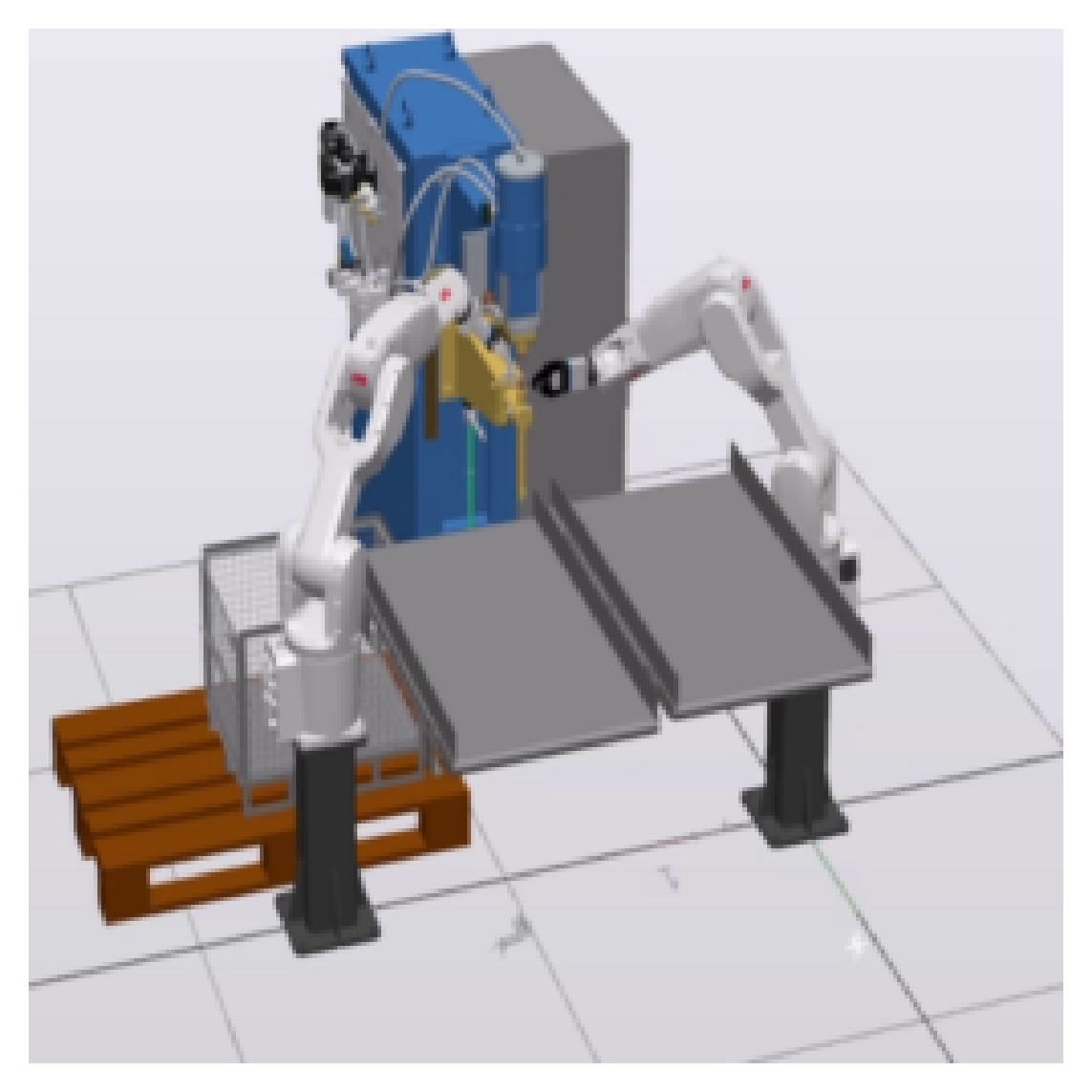}
     \end{subfigure}
     \hfill
     \begin{subfigure}[b]{0.24\textwidth}
         \centering
         \includegraphics[trim={0 1.cm 0 0},clip,width=\textwidth]{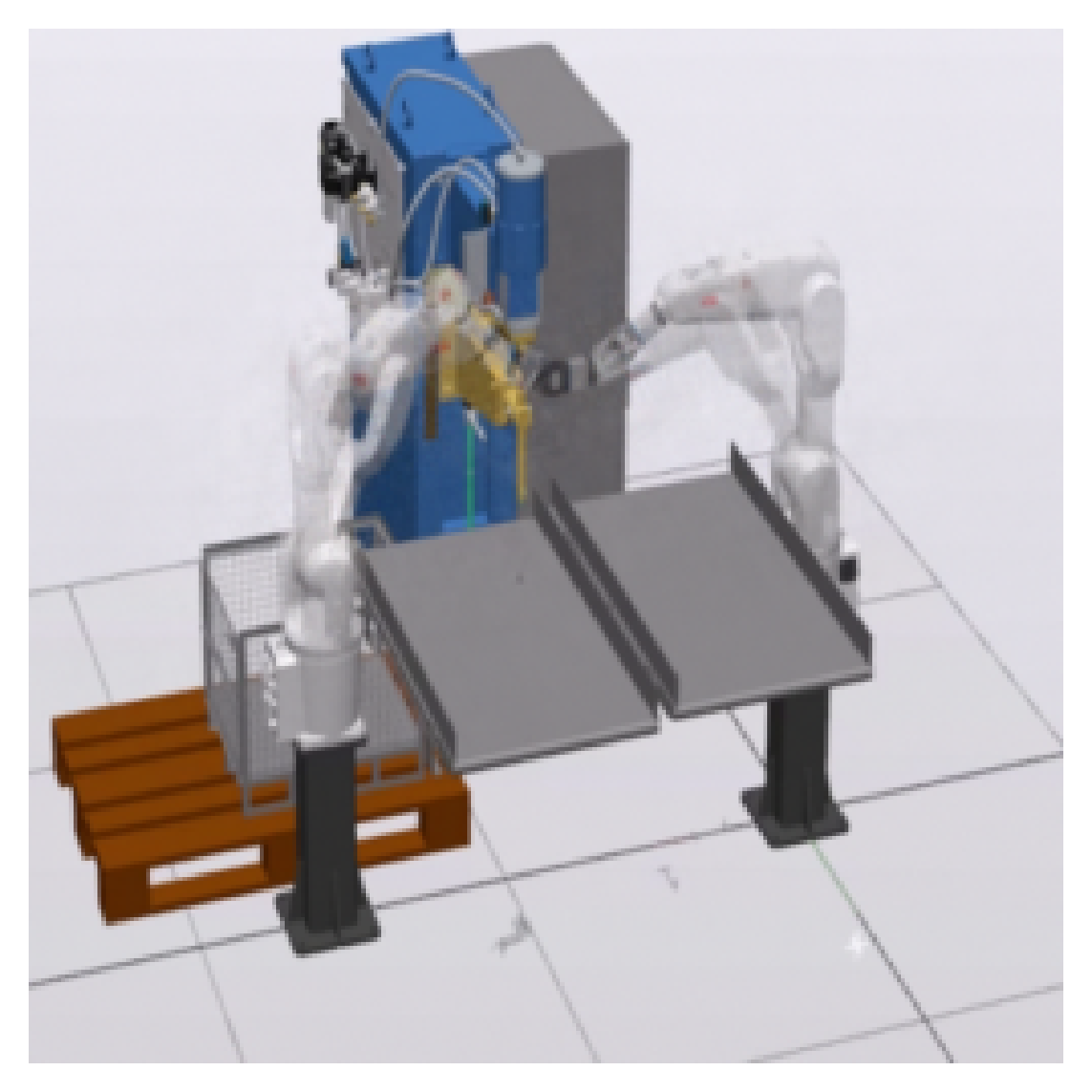}
     \end{subfigure}
     \hfill
     \begin{subfigure}[b]{0.24\textwidth}
         \centering
         \includegraphics[trim={0 1.cm 0 0},clip,width=\textwidth]{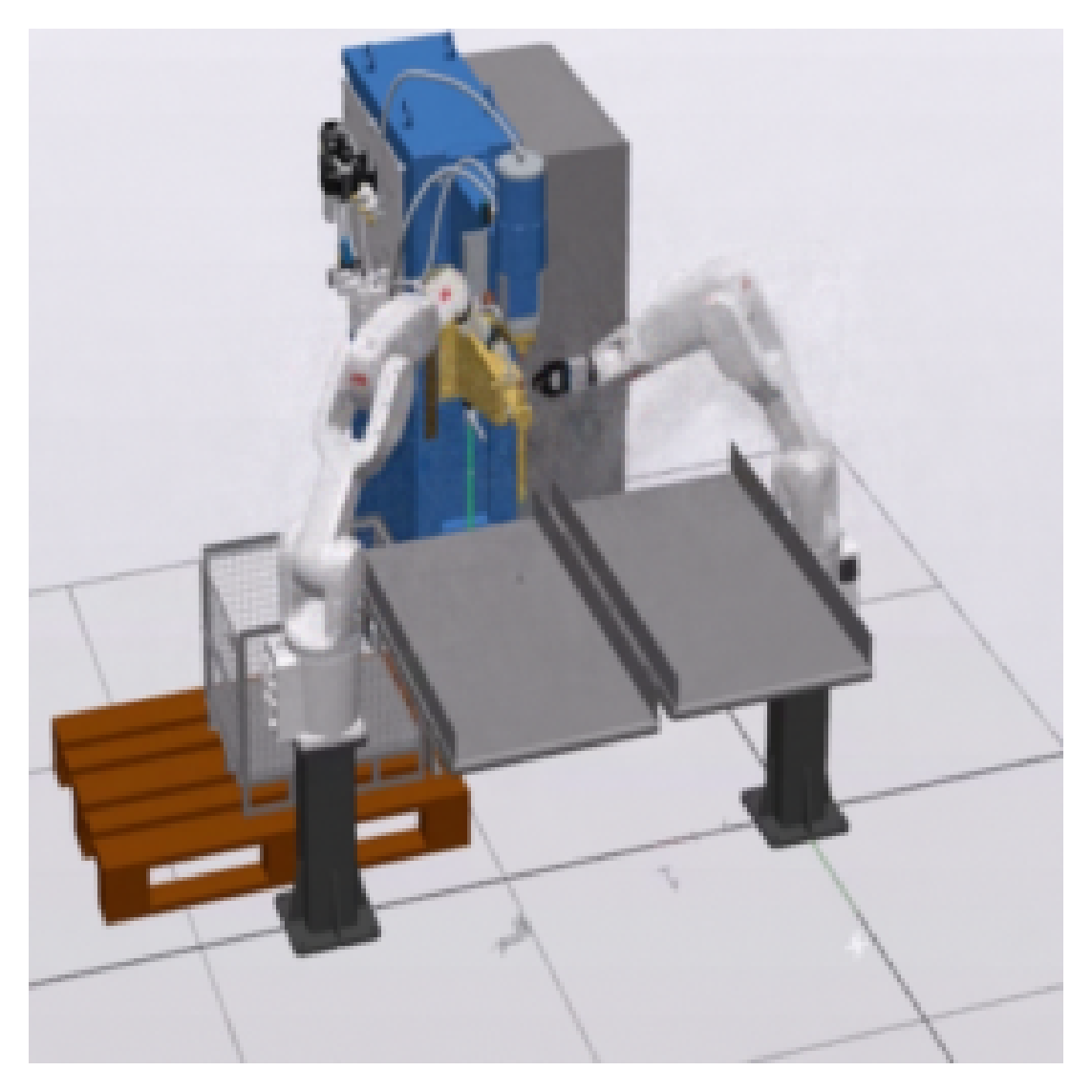}
     \end{subfigure}
    \hfill
     \begin{subfigure}[b]{0.24\textwidth}
         \centering
         \includegraphics[trim={0 1.cm 0 0},clip,width=\textwidth]{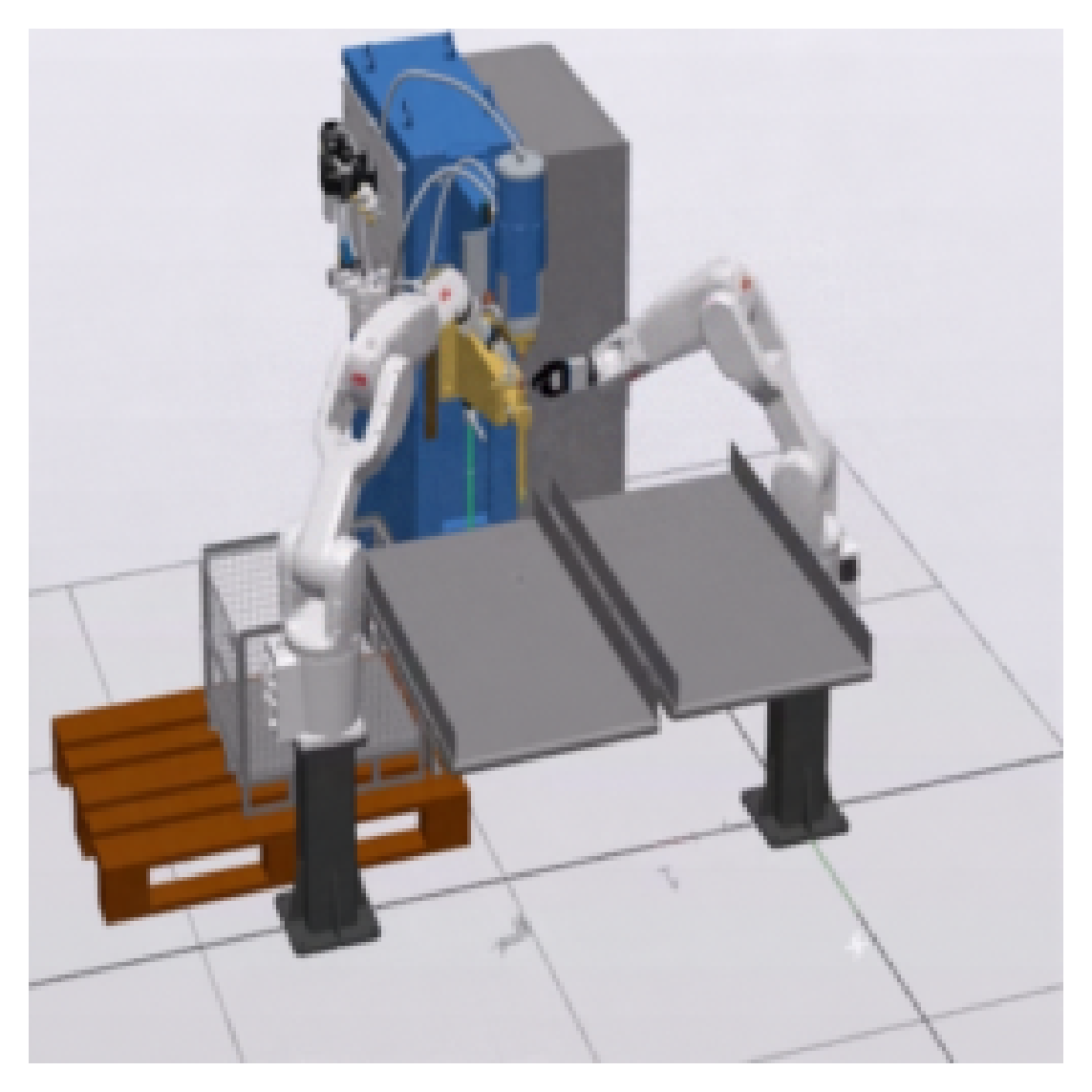}
     \end{subfigure}

          \begin{subfigure}[b]{0.24\textwidth}
         \centering
         \includegraphics[trim={0 1.cm 0 0},clip,width=\textwidth]{blank.png}
     \end{subfigure}
     \hfill
     \begin{subfigure}[b]{0.24\textwidth}
         \centering
         \includegraphics[width=\textwidth]{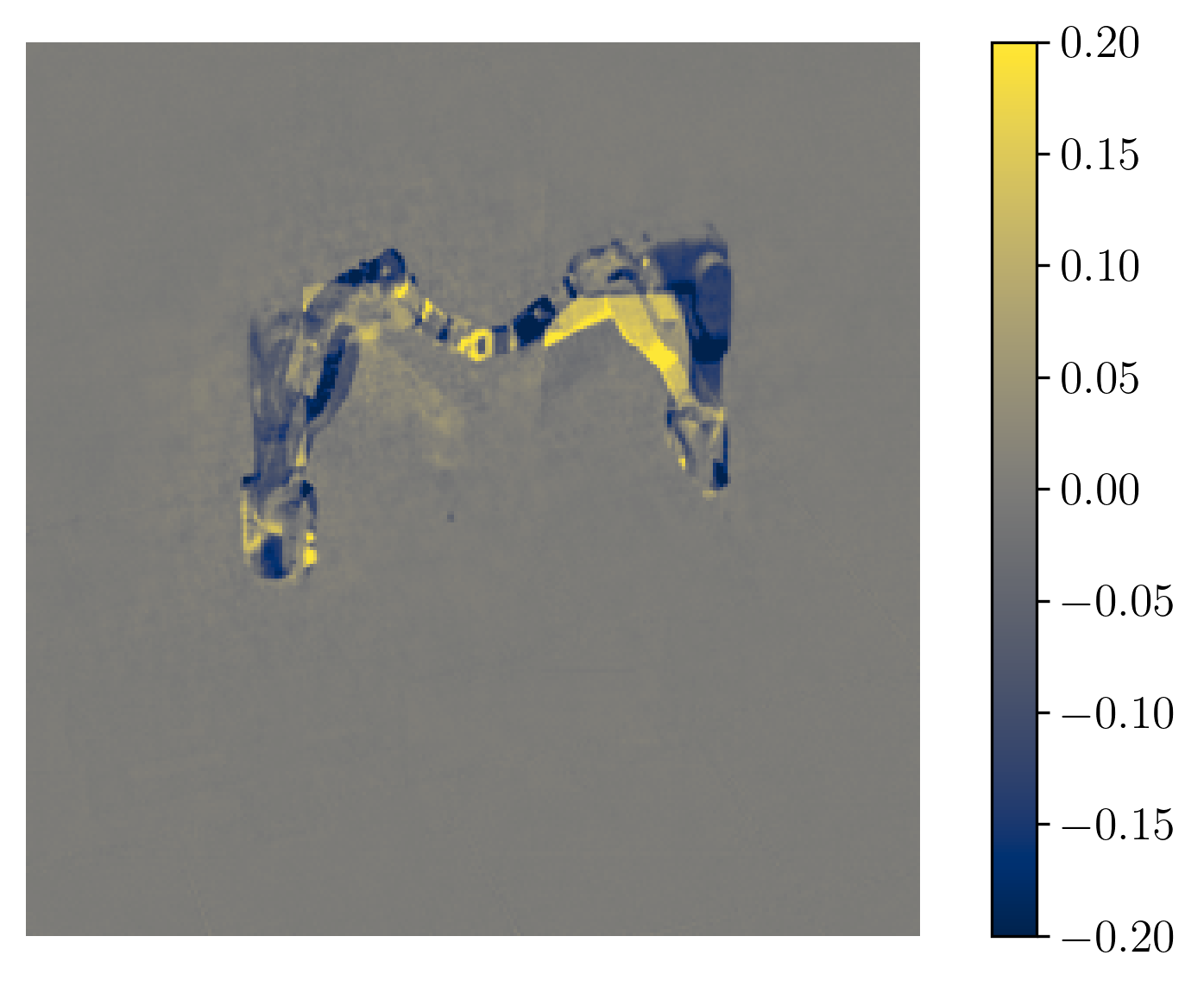}
     \end{subfigure}
     \hfill
     \begin{subfigure}[b]{0.24\textwidth}
         \centering
         \includegraphics[width=\textwidth]{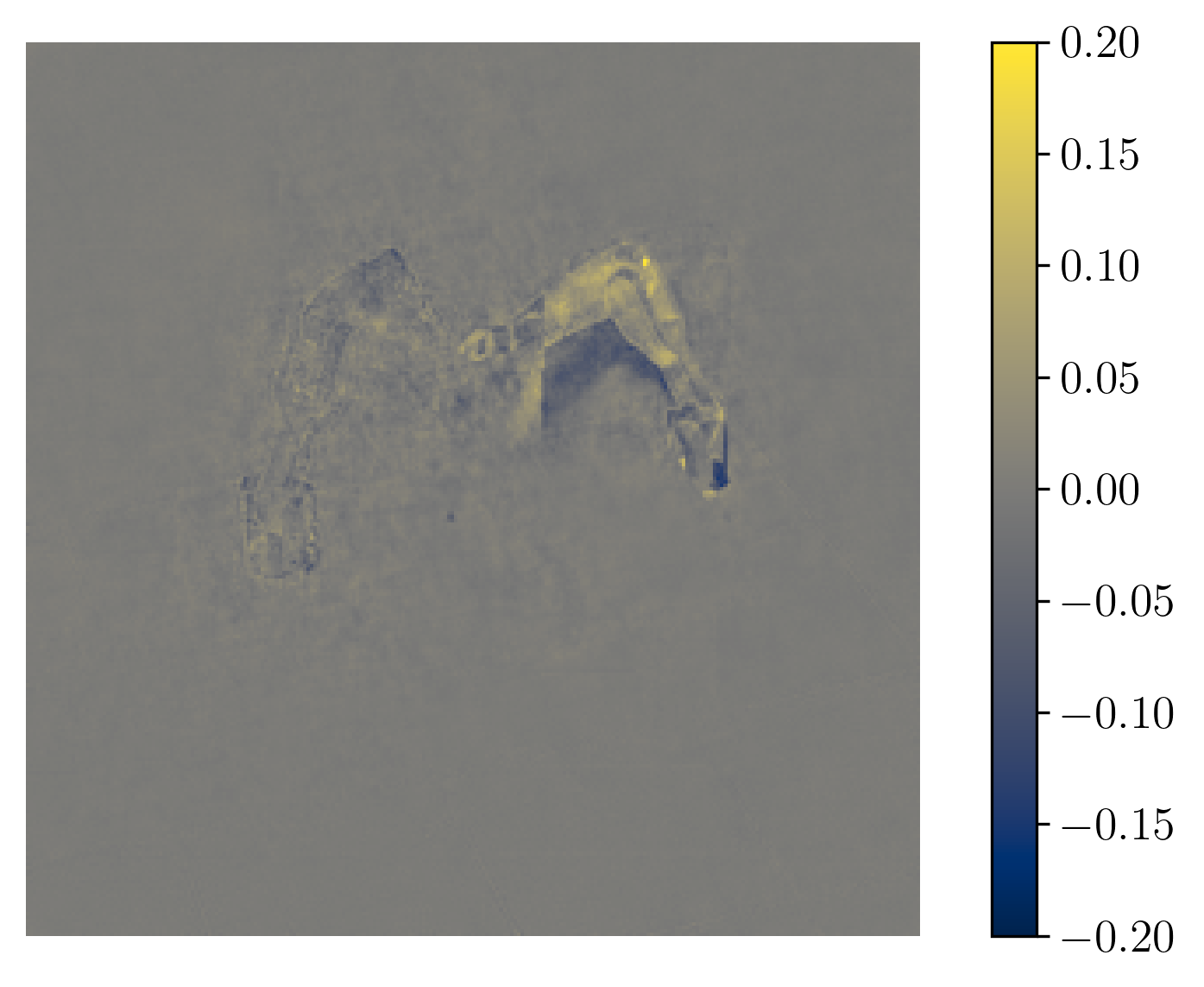}
     \end{subfigure}
    \hfill
     \begin{subfigure}[b]{0.24\textwidth}
         \centering
         \includegraphics[width=\textwidth]{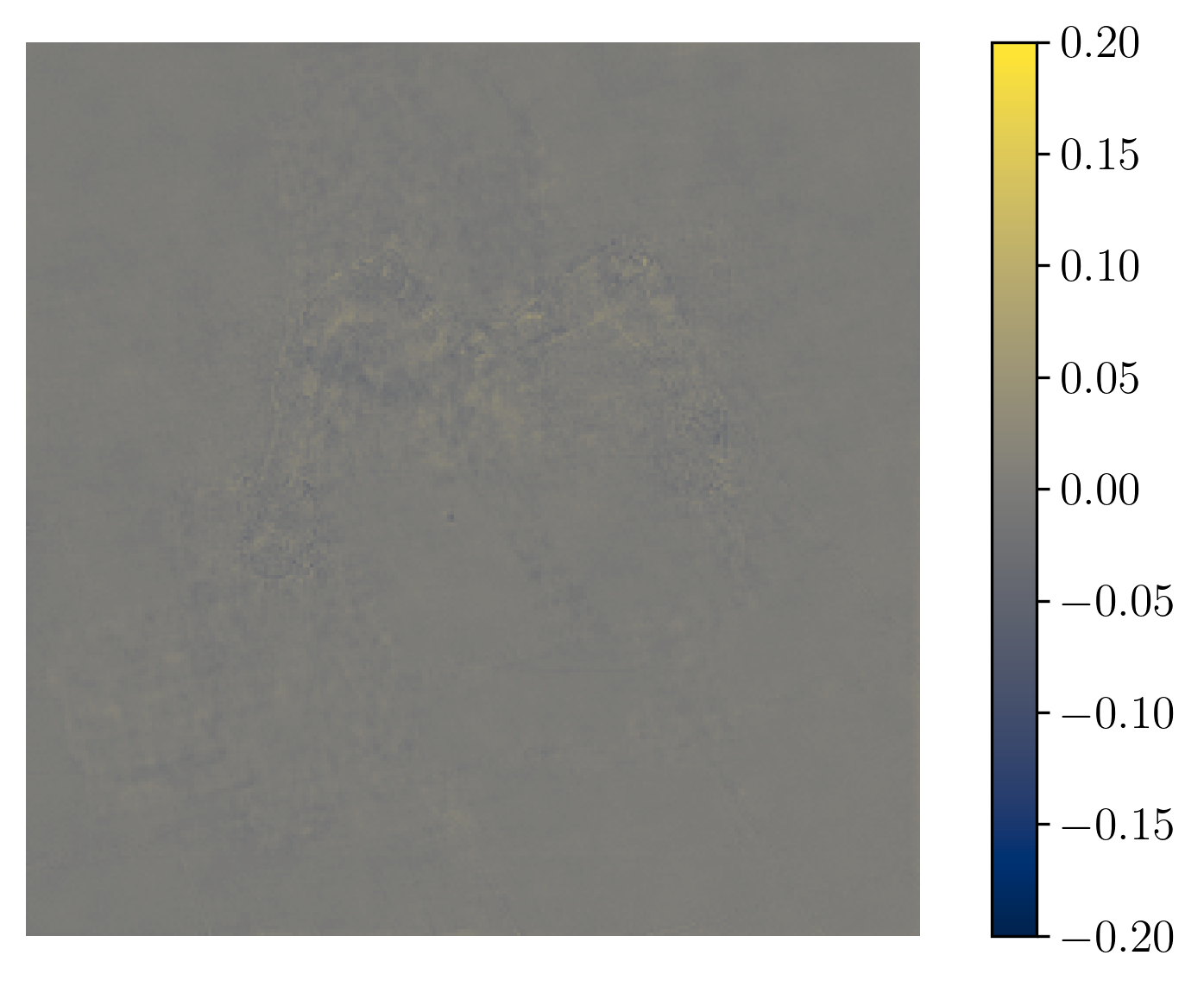}
     \end{subfigure}
    \caption{Comparison of reconstruction results for the dual robot dataset across aggregation methods. The figure displays Original Input followed by reconstructions using concatenation, summation, and attention (columns 2–4). Pixel-wise error maps (consistent scale) are presented below the reconstruction.}
    \label{fig:reconstructions}
\end{figure}

To evaluate the quality of the learned representations, we present sample reconstructions from each model in \autoref{fig:reconstructions}. These reconstructions offer a visual comparison of how effectively each approach captures and preserves input features. Differences in reconstruction quality provide insights into the models’ ability to leverage multimodal information and the impact of the chosen fusion strategies. While qualitative assessment of the reconstructions offers initial insights, subtle differences in reconstruction quality are often difficult to discern visually. To address this, we additionally provide heatmaps of the pixel-wise reconstruction error. These highlight regions where deviations from the original inputs occur, allowing for a more precise and interpretable comparison of model performance.

\subsection{Downstream Failure Detection}

\begin{figure}
     \centering
     \begin{subfigure}[b]{0.24\textwidth}
         \centering
         \includegraphics[width=\textwidth]{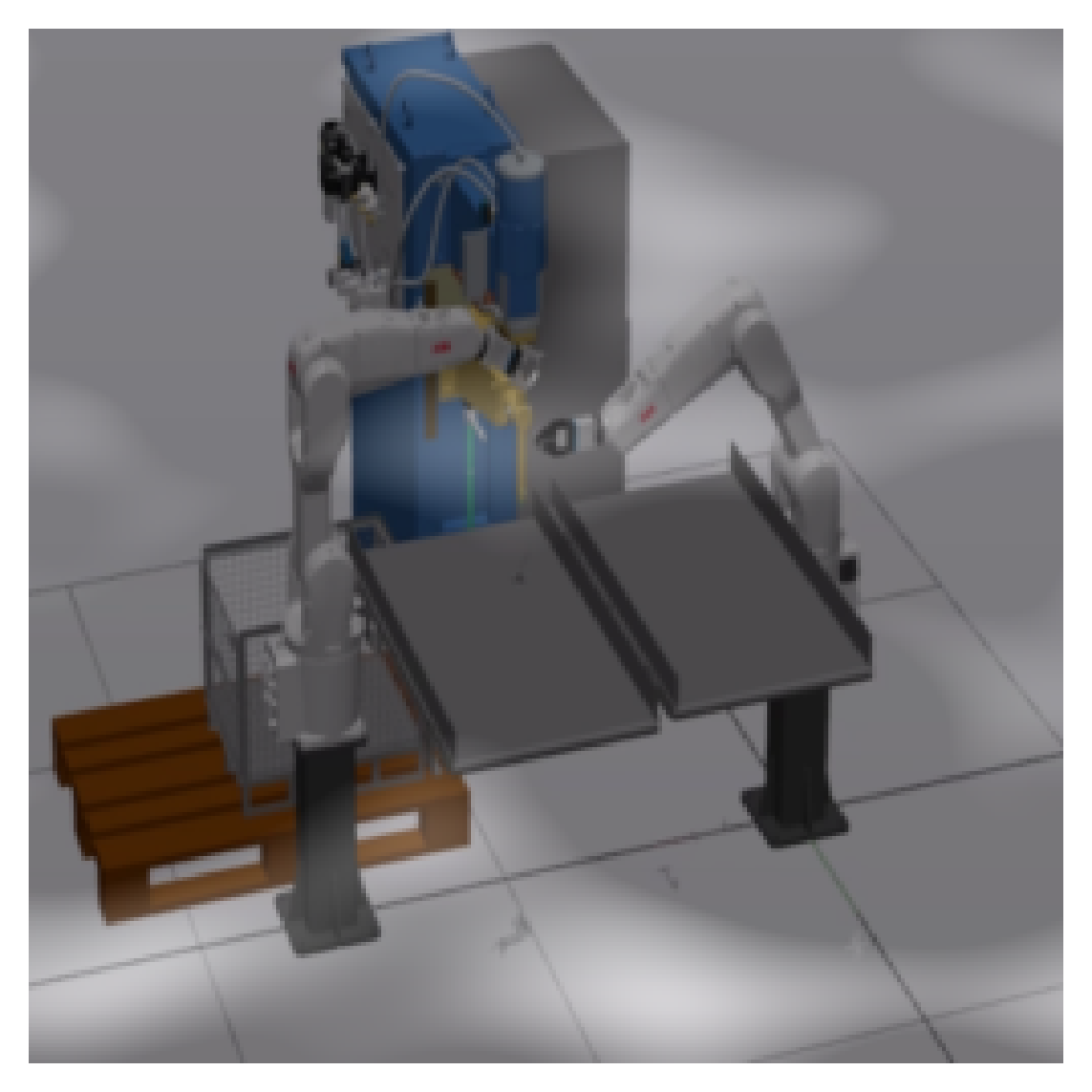}
     \end{subfigure}
     \hfill
     \begin{subfigure}[b]{0.24\textwidth}
         \centering
         \includegraphics[width=\textwidth]{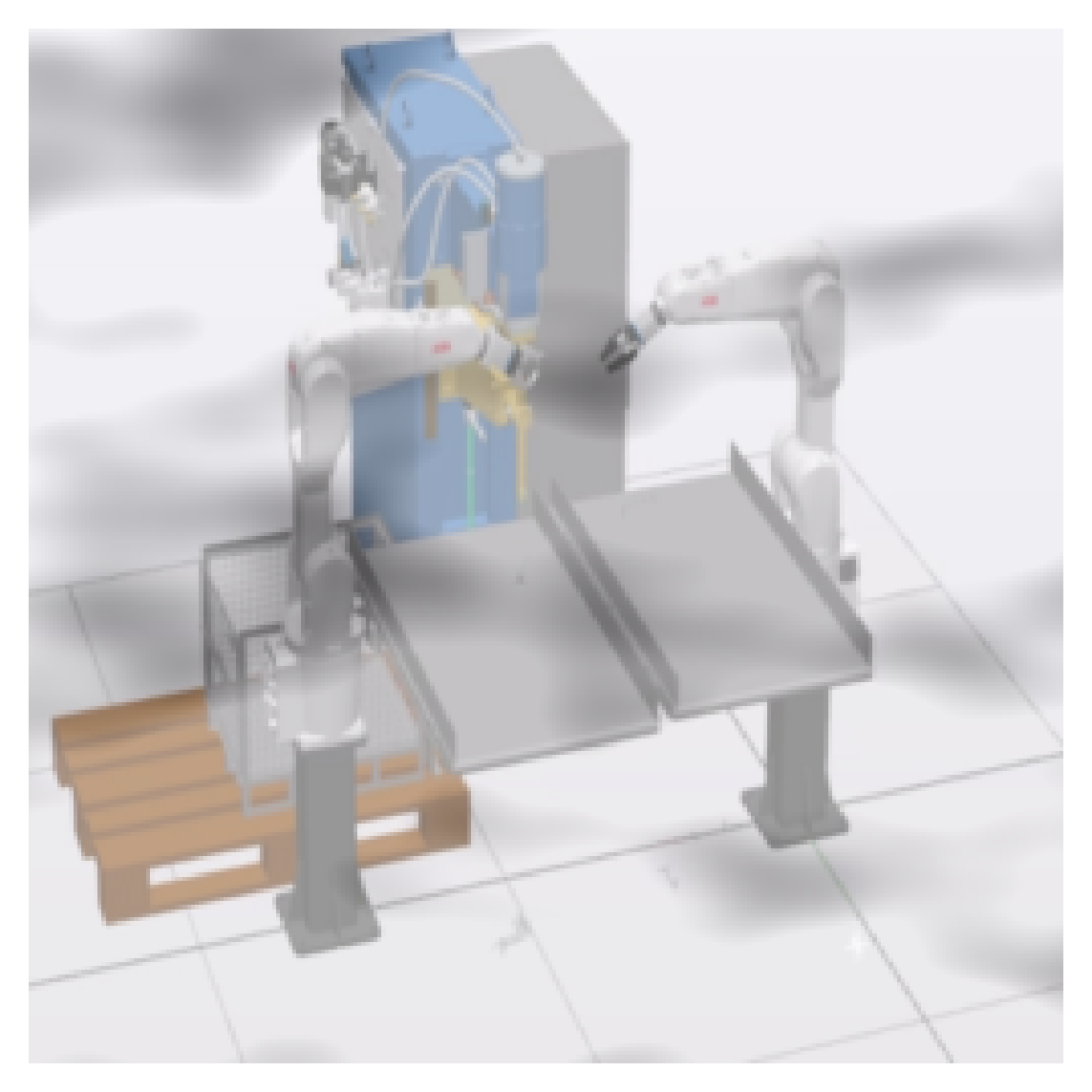}
     \end{subfigure}
     \hfill
     \begin{subfigure}[b]{0.24\textwidth}
         \centering
         \includegraphics[width=\textwidth]{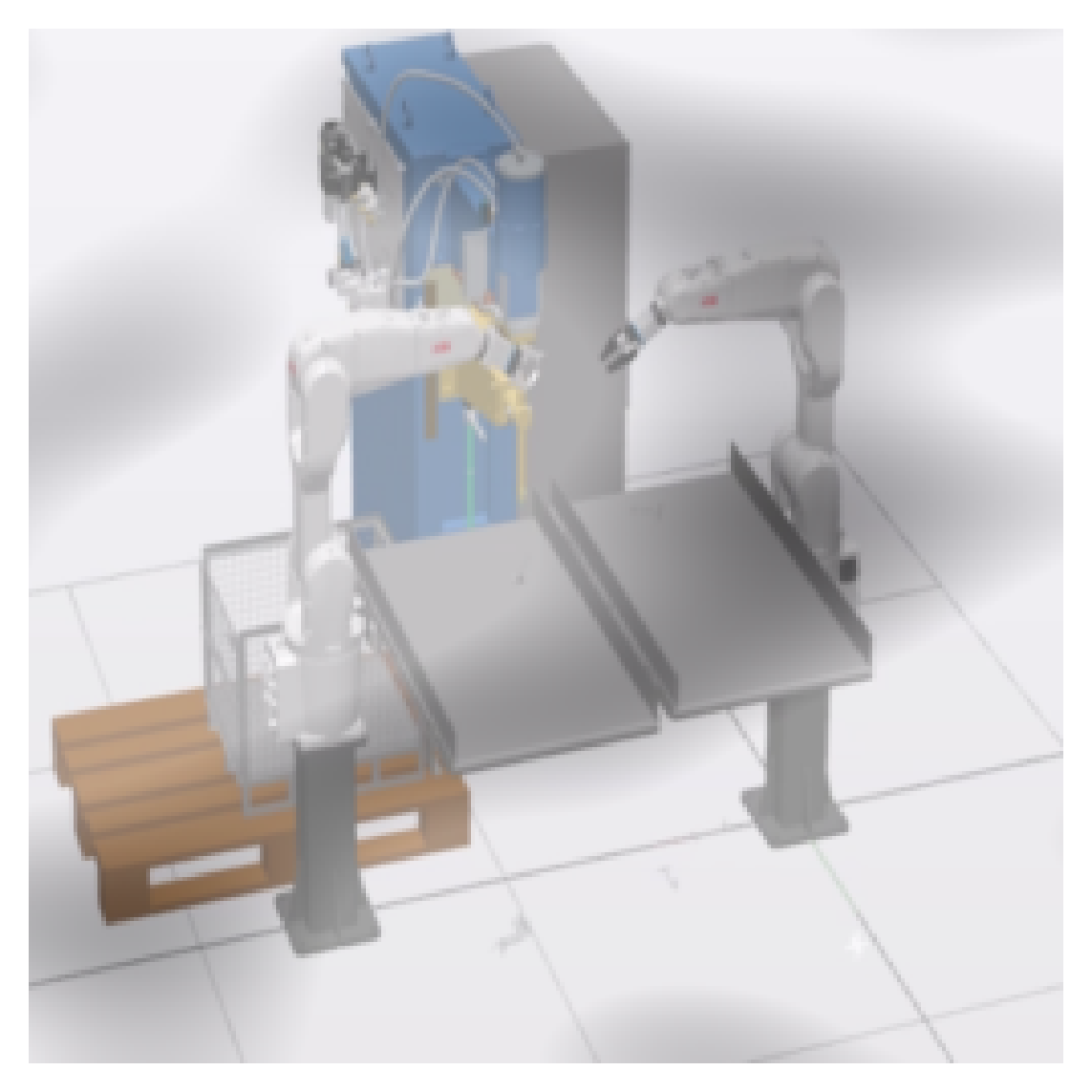}
     \end{subfigure}
    \hfill
     \begin{subfigure}[b]{0.24\textwidth}
         \centering
         \includegraphics[width=\textwidth]{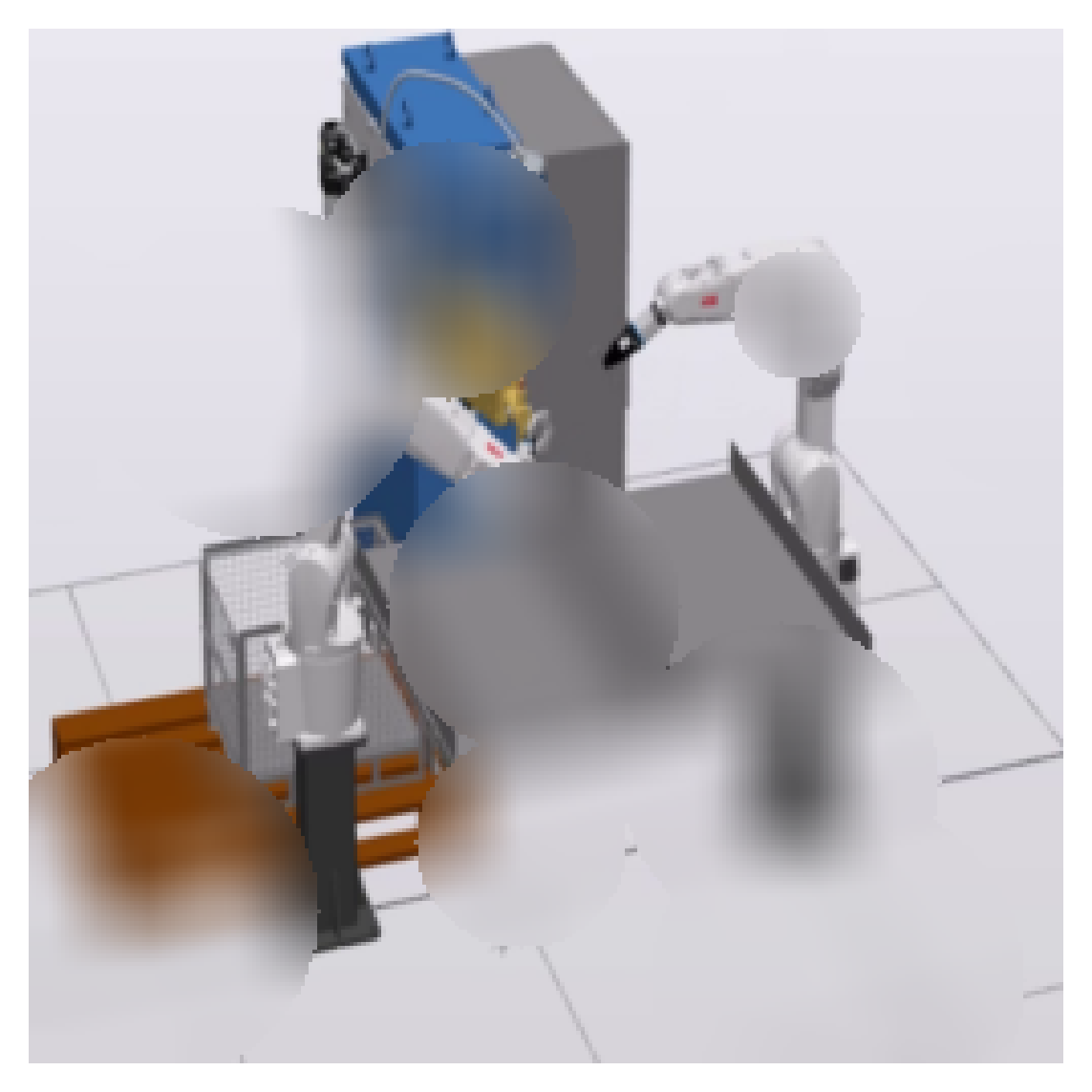}
     \end{subfigure}
          \begin{subfigure}[b]{0.24\textwidth}
         \centering
         \includegraphics[width=\textwidth]{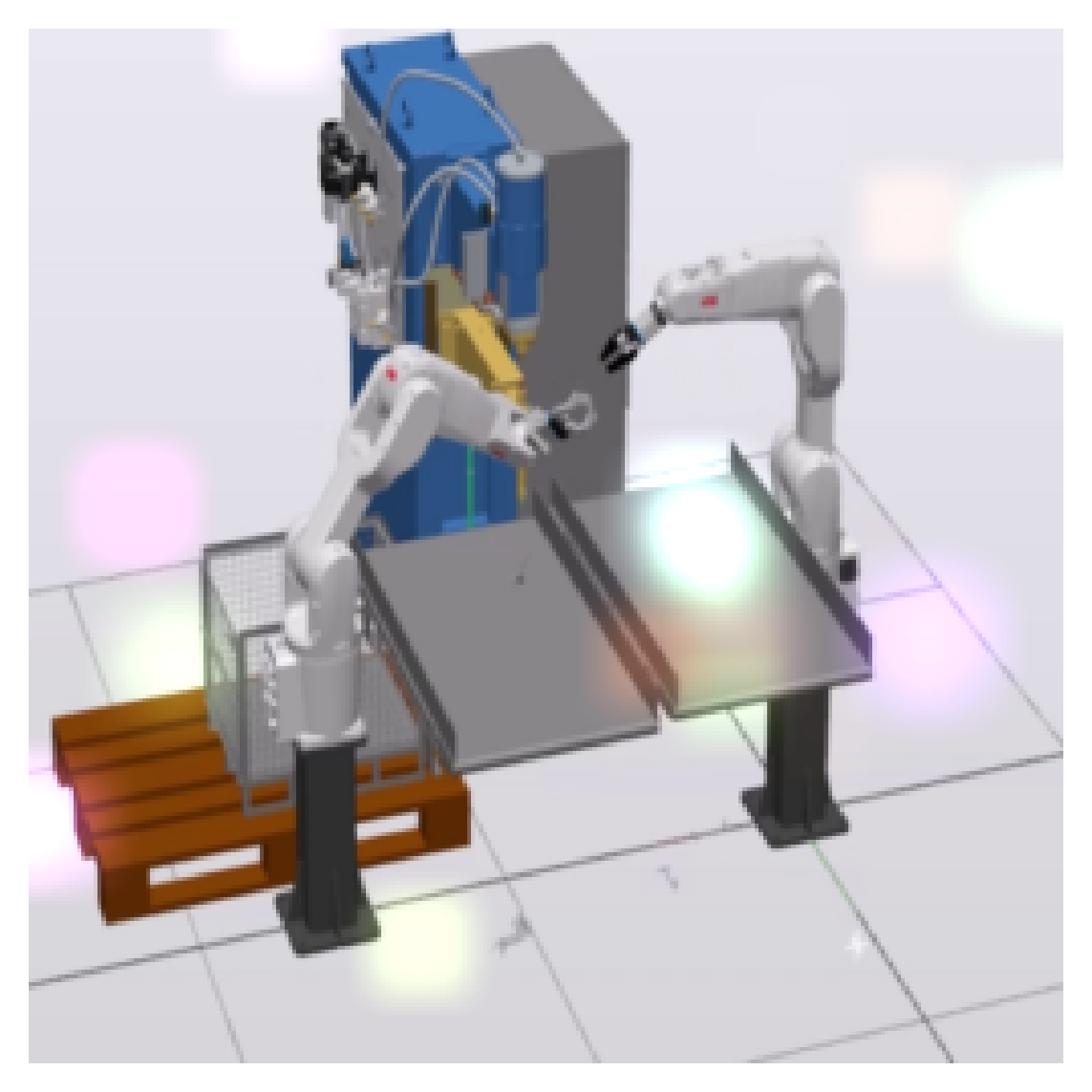}
     \end{subfigure}
     \hfill
     \begin{subfigure}[b]{0.24\textwidth}
         \centering
         \includegraphics[width=\textwidth]{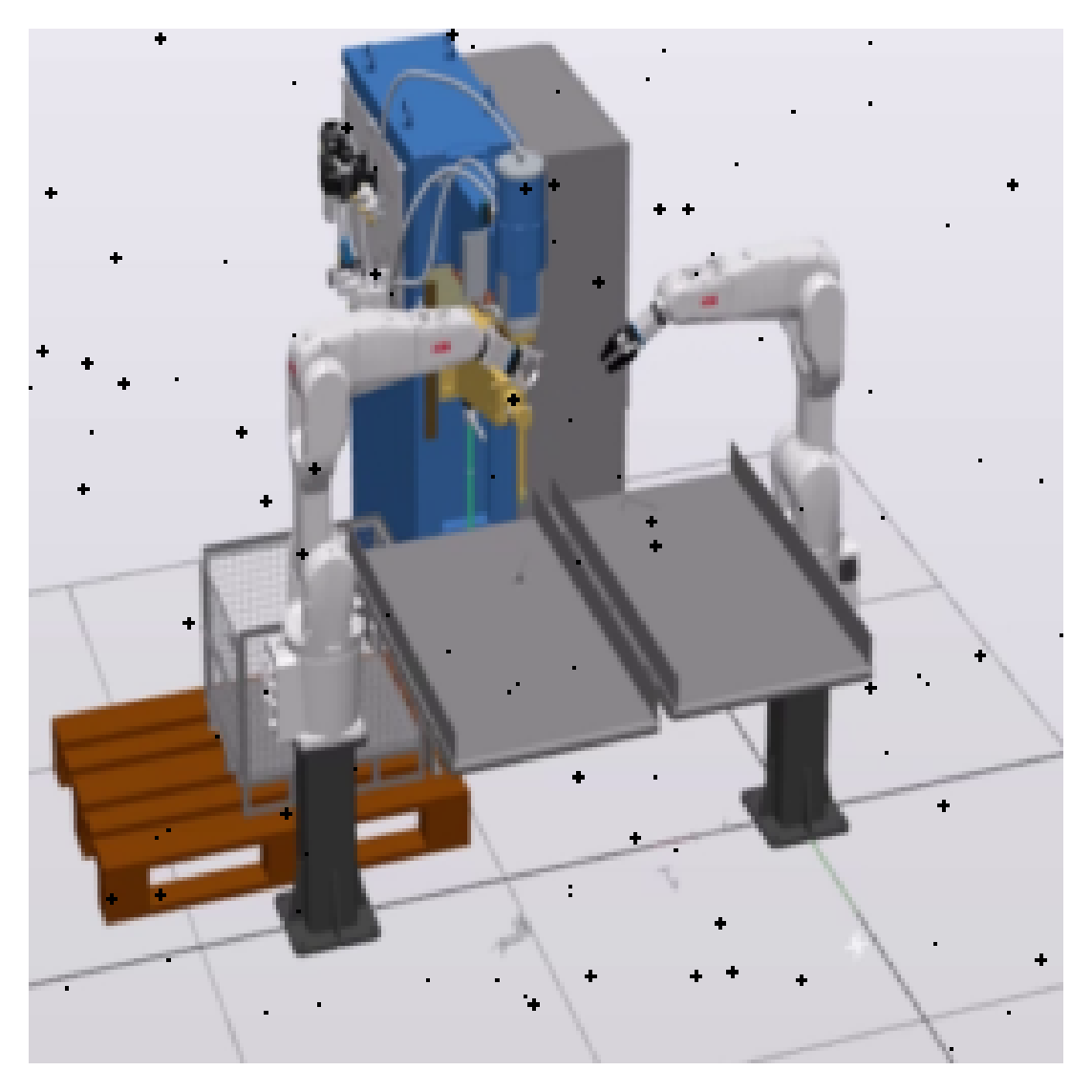}
     \end{subfigure}
     \hfill
     \begin{subfigure}[b]{0.24\textwidth}
         \centering
         \includegraphics[width=\textwidth]{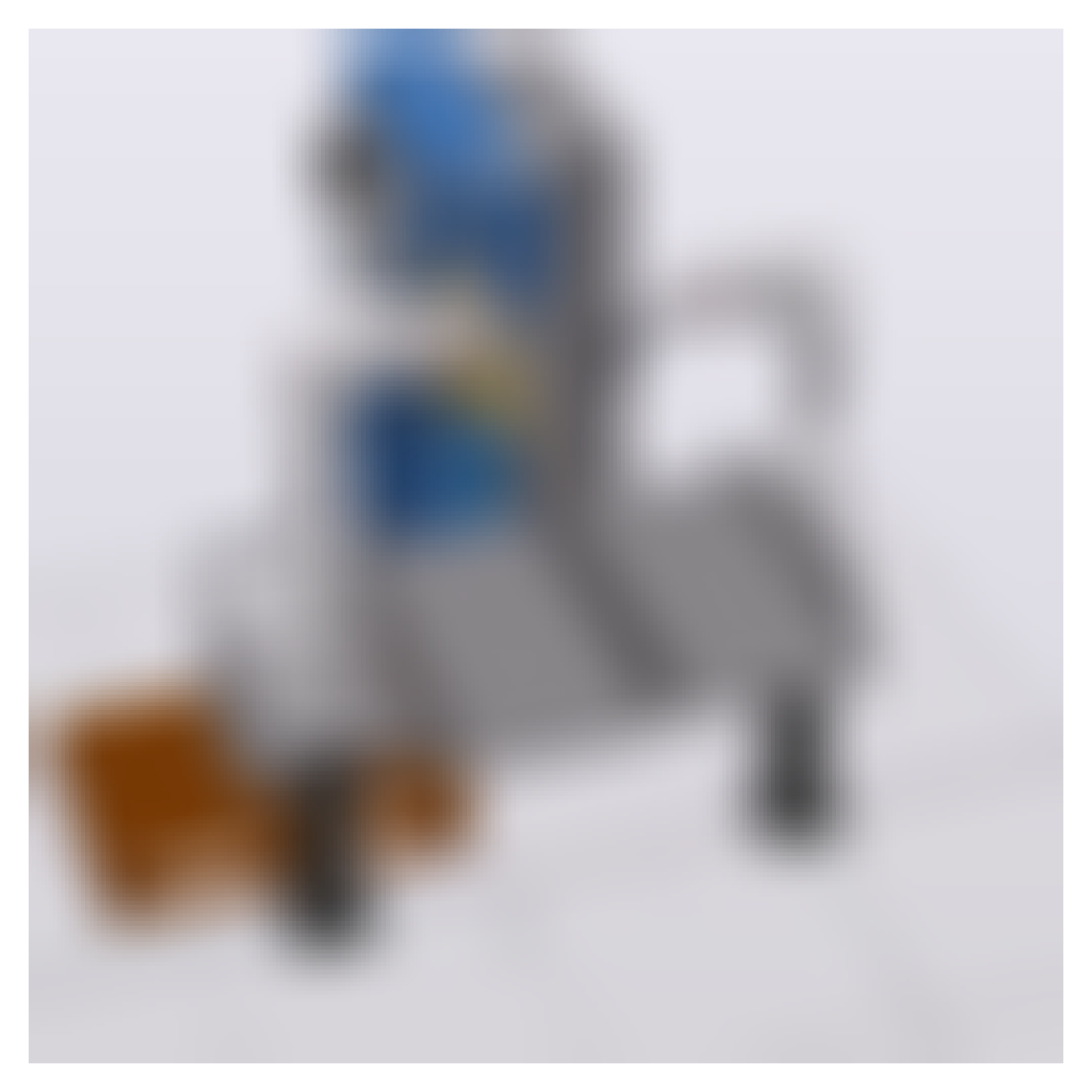}
     \end{subfigure}
    \hfill
     \begin{subfigure}[b]{0.24\textwidth}
         \centering
         \includegraphics[width=\textwidth]{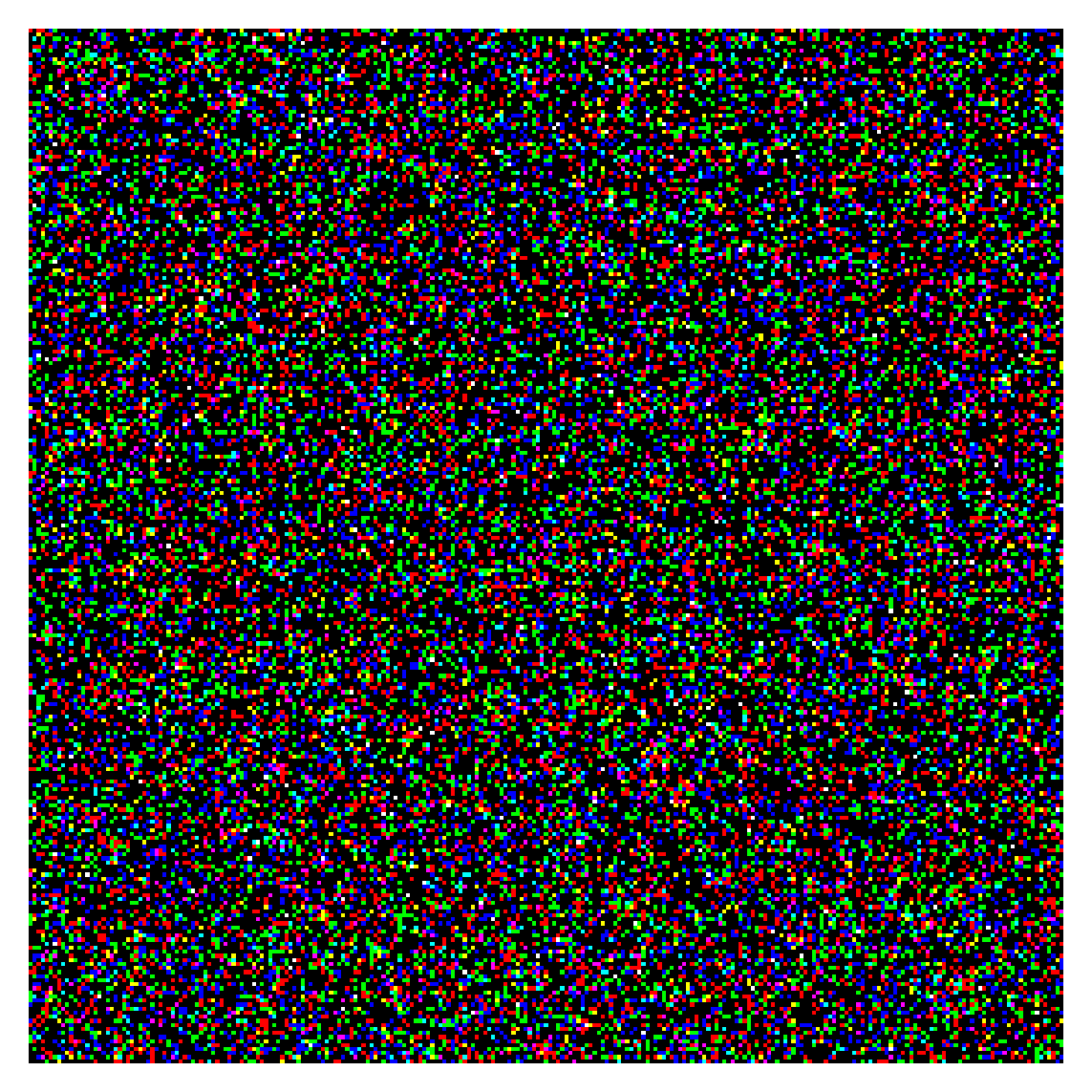}
     \end{subfigure}
    \caption{Representative examples of injected failure cases.}
    \label{fig:failures}
\end{figure}

In this section, we evaluate the downstream failure detection performance of the proposed multimodal autoencoder architectures employing different fusion strategies. Our objective is to assess how effectively each fusion approach can distinguish between normal and faulty operating conditions based on the learned latent representations. Failure conditions are injected based on \citep{altinses2023multimodal}, which defines the perturbation mechanisms and the corresponding modalities affected. In Figure~\ref{fig:failures}, we illustrate several representative examples of the image-based failure cases used in our evaluation.

\begin{figure}
     \centering
     \begin{subfigure}[b]{0.325\textwidth}
         \centering
         \includegraphics[width=\textwidth]{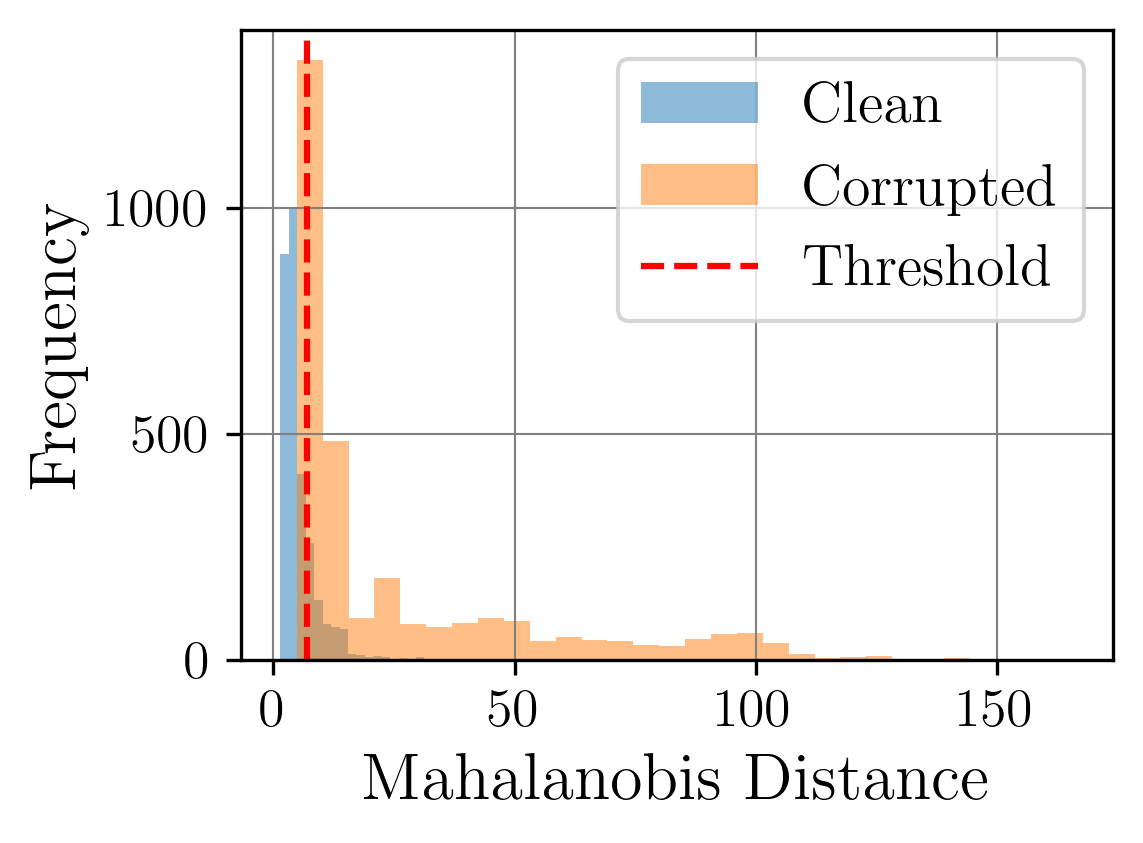}
         \caption{Summation}
     \end{subfigure}
     \hfill
     \begin{subfigure}[b]{0.325\textwidth}
         \centering
         \includegraphics[width=\textwidth]{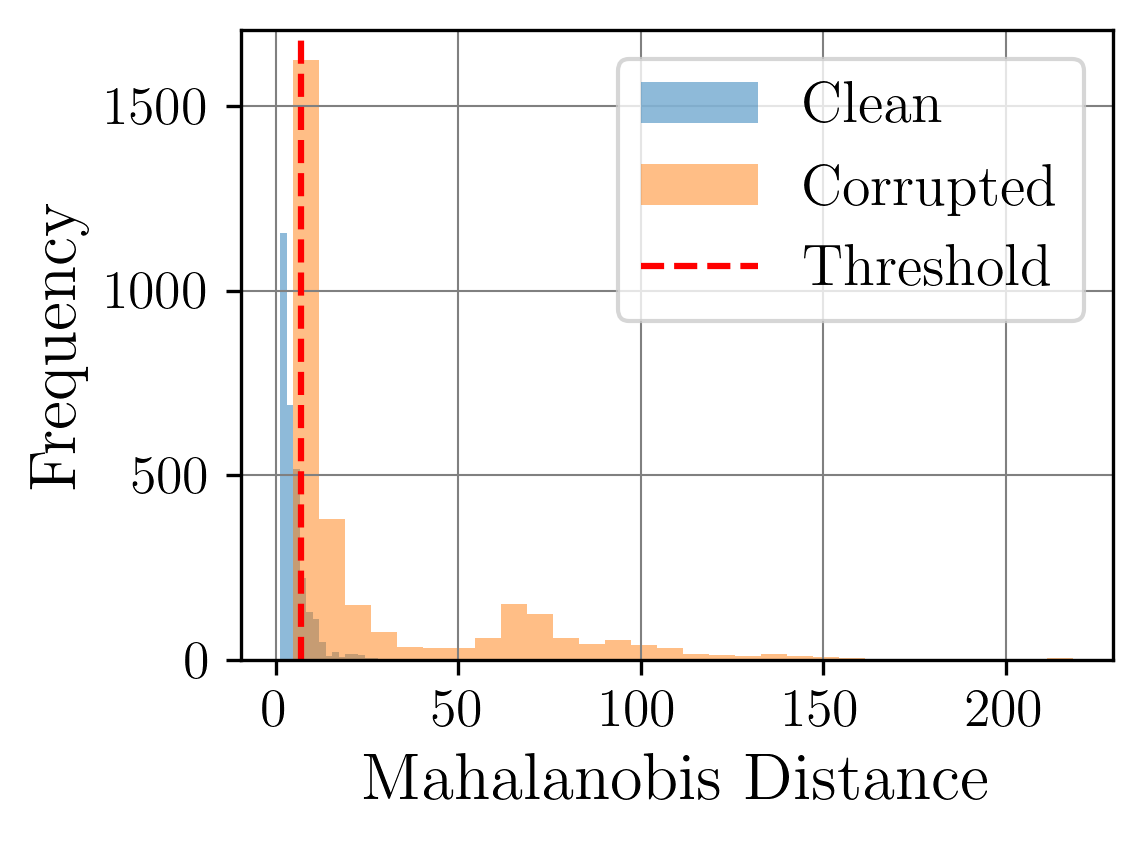}
         \caption{Concatenation}
     \end{subfigure}
     \hfill
     \begin{subfigure}[b]{0.325\textwidth}
         \centering
         \includegraphics[width=\textwidth]{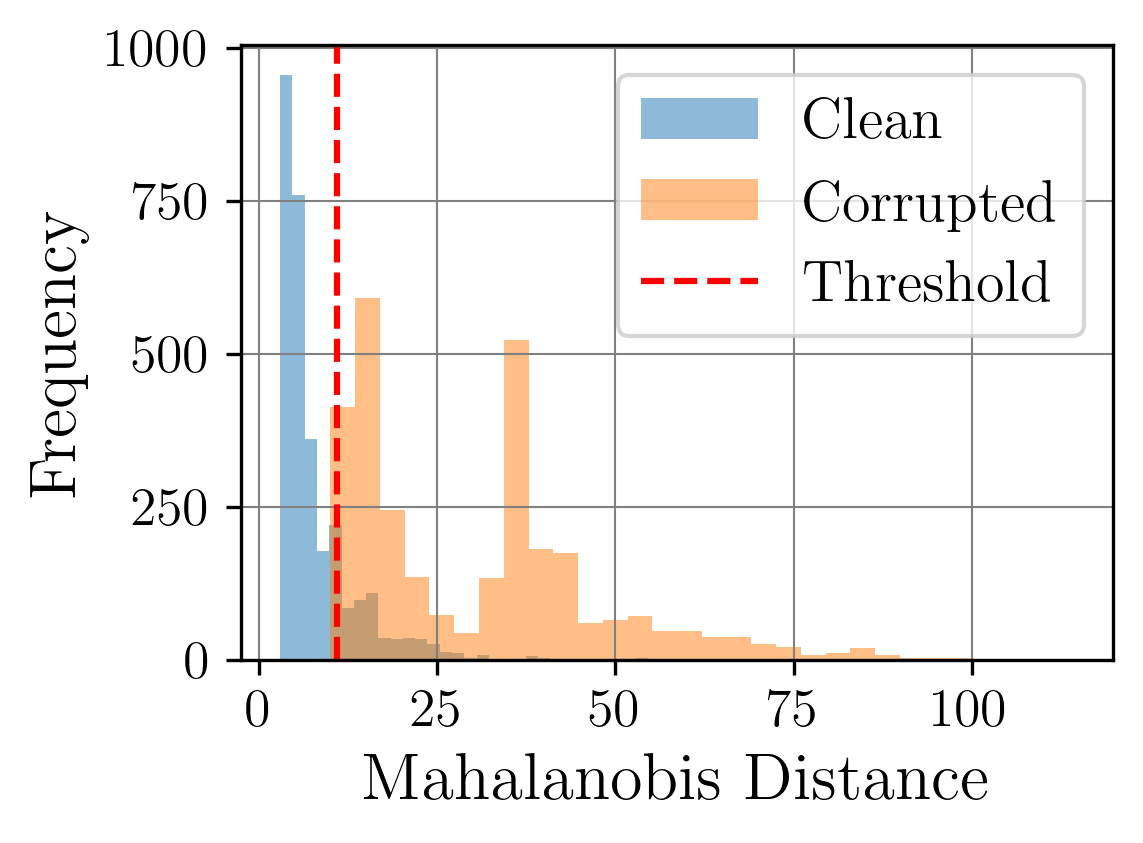}
         \caption{Attention}
     \end{subfigure}
    \caption{Mahalanobis distance distributions for clean and faulty representations across fusion strategies.}
    \label{fig:mahalanobis}
\end{figure}

For anomaly detection, we apply kernel Principal Component Analysis (kPCA) to the latent spaces generated by the autoencoders. Specifically, we compute separate latent embeddings for the clean and faulty data and project them into a lower-dimensional manifold using kPCA. The Mahalanobis distance is then used to quantify deviations between the latent representations of clean and faulty samples. By setting a detection threshold at the 95th percentile of the Mahalanobis distance distribution of the clean data, we are able to effectively separate clean from faulty latent spaces. This process and the resulting distributions are shown in Figure~\ref{fig:mahalanobis}.

\begin{figure}
     \centering
     \begin{subfigure}[b]{0.325\textwidth}
         \centering
         \includegraphics[width=\textwidth]{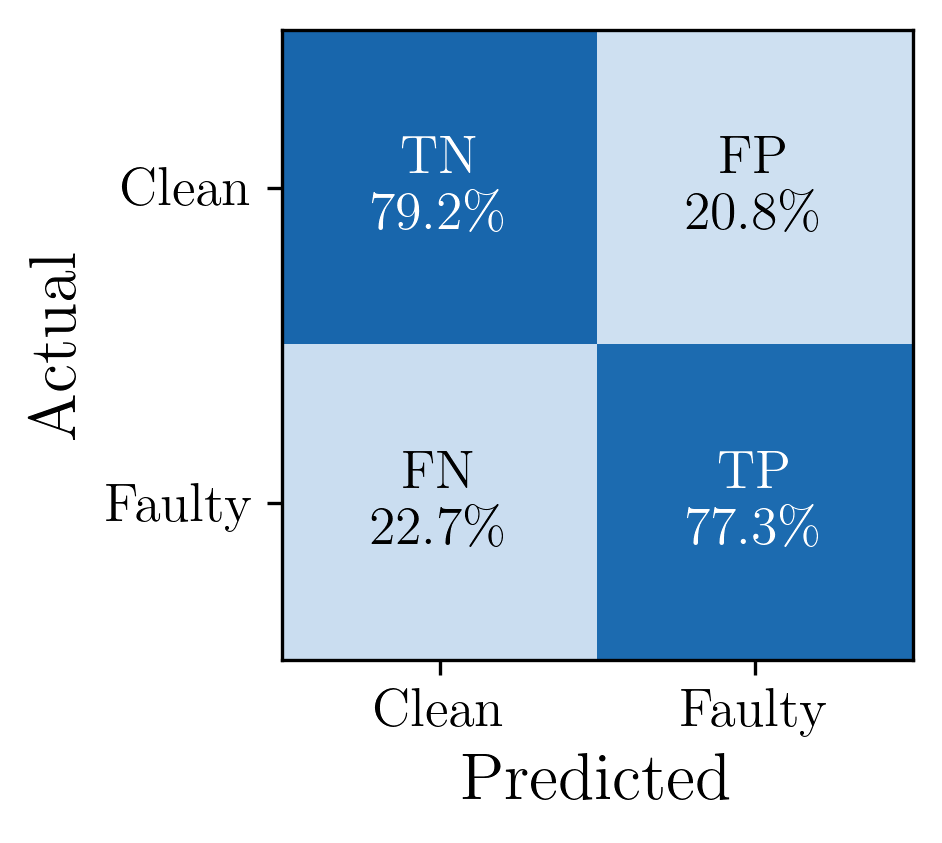}
         \caption{Summation}
     \end{subfigure}
     \hfill
     \begin{subfigure}[b]{0.325\textwidth}
         \centering
         \includegraphics[width=\textwidth]{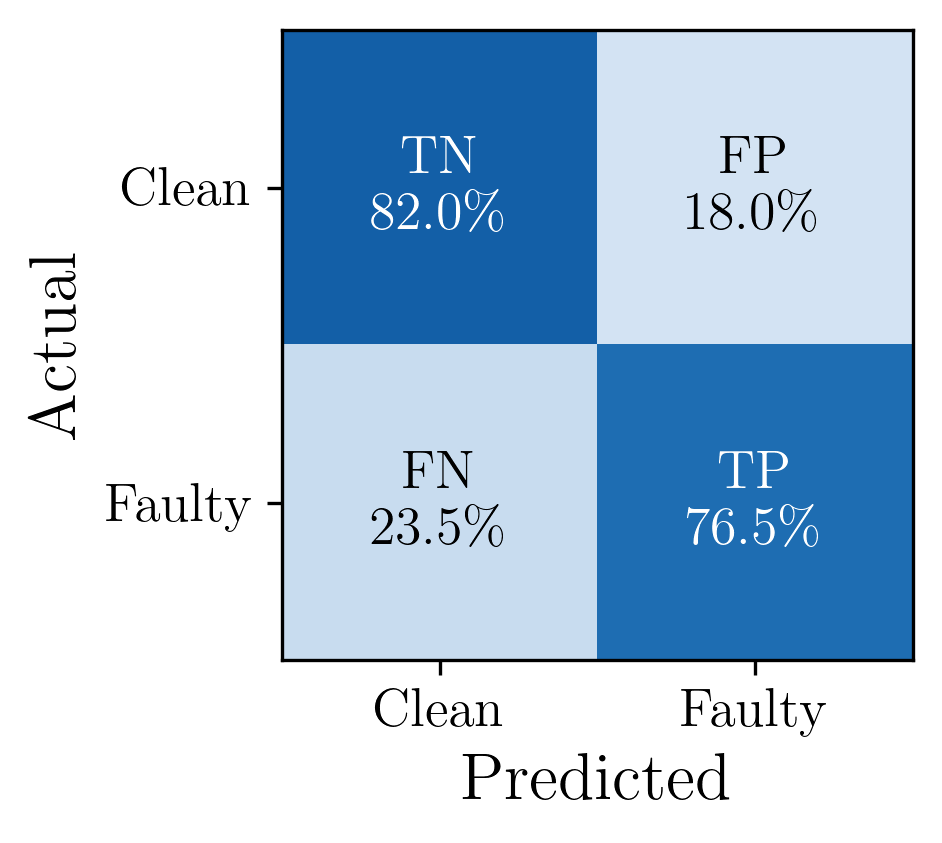}
         \caption{Concatenation}
     \end{subfigure}
     \hfill
     \begin{subfigure}[b]{0.325\textwidth}
         \centering
         \includegraphics[width=\textwidth]{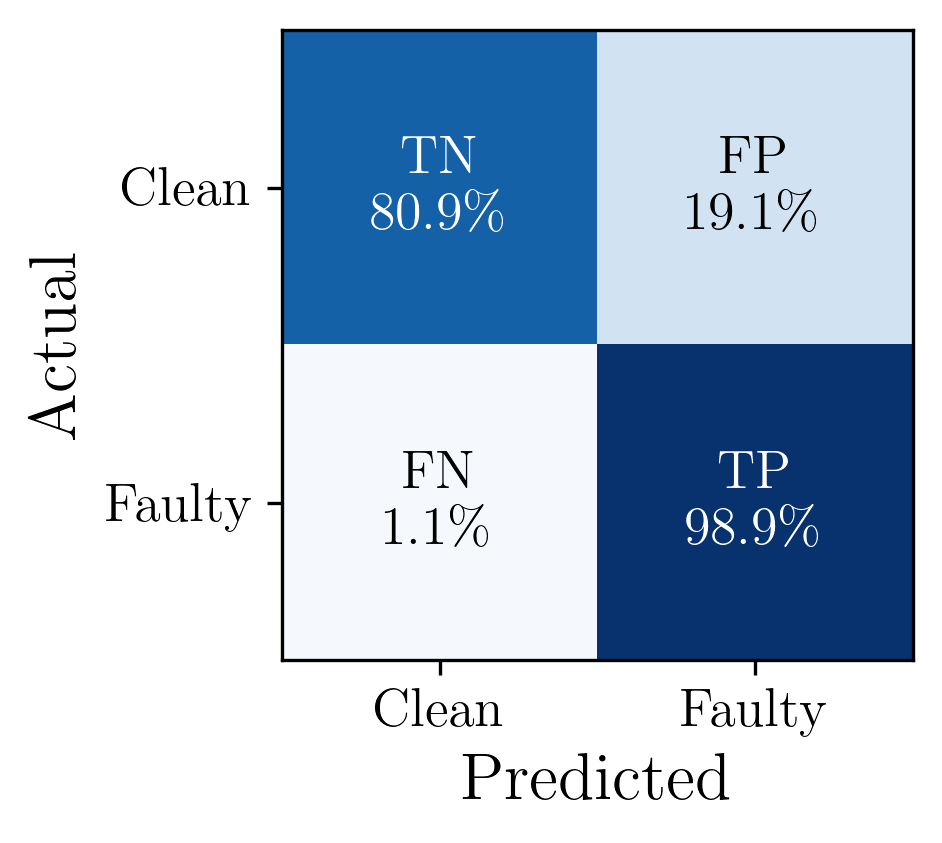}
         \caption{Attention}
     \end{subfigure}
    \caption{Confusion matrices of failure detection results for different fusion strategies.}
    \label{fig:confusion}
\end{figure}

From Figure~\ref{fig:mahalanobis}, it can be observed that the attention-regulated fusion approach exhibits a more pronounced separation between the clean and corrupted samples compared to the other fusion schemes (e.g., Summation, concatenation-based fusion). This indicates that the attention mechanism helps the model emphasize more informative modality interactions, resulting in latent representations that are more sensitive to underlying faults. To further quantify this distinction, we compute and visualize the confusion matrices corresponding to the different fusion strategies (Figure~\ref{fig:confusion}). These results clearly demonstrate that the attention-regulated fusion achieves higher true positive rates and lower false positive rates, confirming its superior capability in detecting downstream failures.

The attention-driven multimodal fusion not only enhances the feature alignment between modalities but also improves the robustness of latent representations, leading to more reliable failure detection performance in downstream tasks.

\subsection{Evaluation on Real-World Data}

In this section, we evaluate our proposed method on the RoboMNIST dataset, which provides a real-world dataset for assessing robustness and generalization beyond synthetic data. The RoboMNIST dataset consists of real sensor-captured digit images under varying illumination, pose, and background conditions, introducing additional complexity compared to the controlled synthetic datasets used in our earlier experiments. To ensure consistency and comparability, we adopt the same evaluation strategy and experimental setup as for the synthetic datasets. This includes identical training procedures, hyperparameter configurations, and performance metrics. In \autoref{tab:robotmnist_results}, we present the loss performance over 10 independent trials.

\begin{table}[htbp]
\caption{Comparison of fusion strategies based on loss metrics using the bimodal real-world RoboMNIST dataset. The results present the mean ($\mu$) $\pm$ standard deviation ($\sigma$) of the loss computed over 10 independent trials for each architecture.}
\label{tab:robotmnist_results}
\begin{tabular}{l|ccc}
\toprule
              & \multicolumn{3}{c}{Training in $10^{-2}$} \\
\midrule
Model         & Sensor Modality & Image Modality & Combined  \\
\midrule
Summation & 21.351 $\pm$ 2.043 & 0.098 $\pm$ 0.005 & 21.449 $\pm$ 2.043  \\
Concatenation & 26.945 $\pm$ 1.939 & 0.097 $\pm$ 0.007 & 27.042 $\pm$ 1.941 \\
Attention & 1.5580 $\pm$ 0.075 & 0.063 $\pm$ 0.003 & 1.6210 $\pm$ 0.075  \\
\midrule
              & \multicolumn{3}{c}{Testing in $10^{-2}$} \\
\midrule
Model         & Sensor Modality & Image Modality & Combined \\
\midrule
Summation & 25.728 $\pm$ 2.178 & 0.191 $\pm$ 0.017 & 25.919 $\pm$ 2.171 \\
Concatenation & 33.563 $\pm$ 2.296 & 0.188 $\pm$ 0.020 & 33.751 $\pm$ 2.299 \\
Attention & 2.1990 $\pm$ 0.087 & 0.152 $\pm$ 0.010 & 2.3520 $\pm$ 0.085 \\
\bottomrule
\end{tabular}
\end{table}

\begin{figure}
     \centering
     \begin{subfigure}[b]{0.24\textwidth}
         \centering
         \includegraphics[trim={0 1.cm 0 0},clip,width=\textwidth]{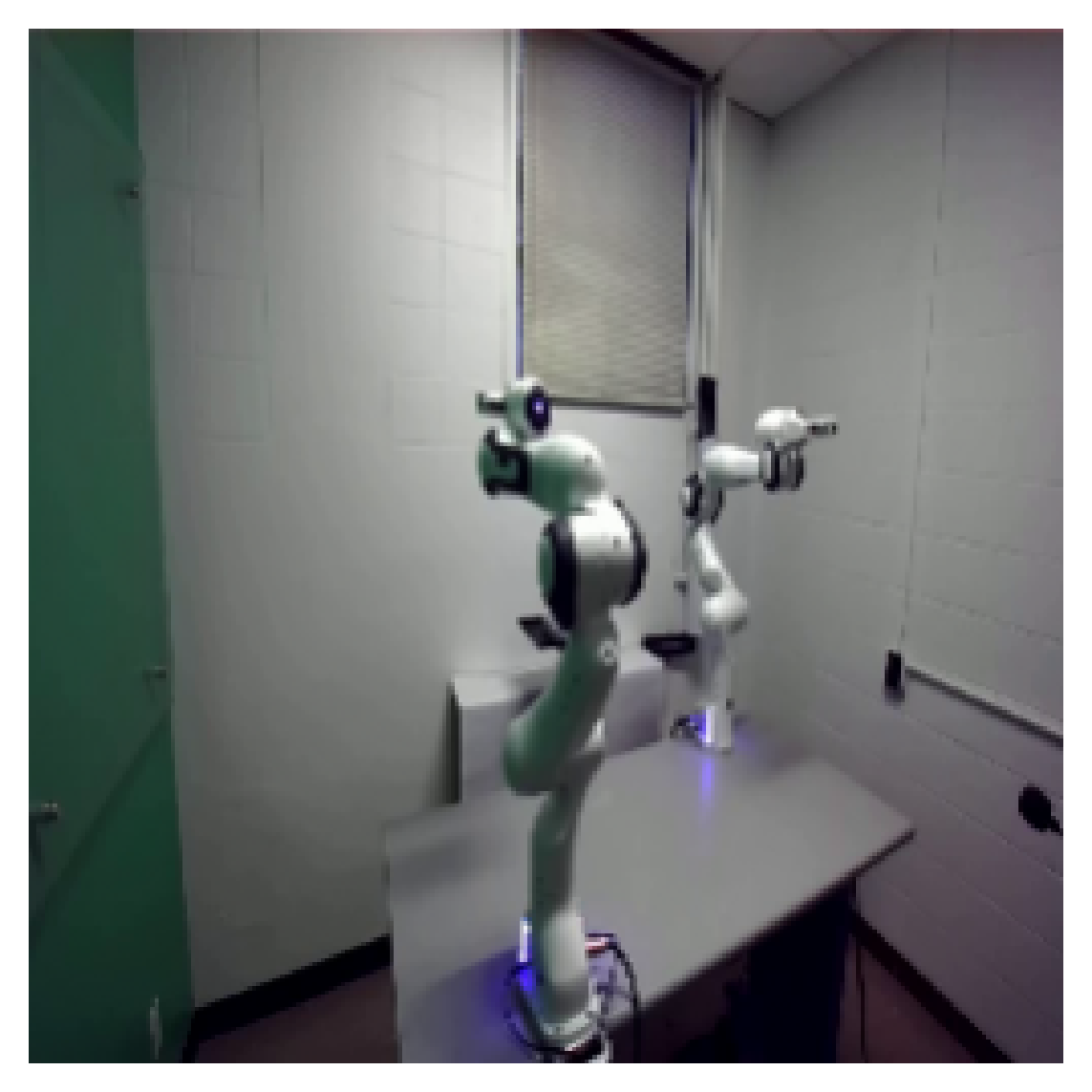}
     \end{subfigure}
     \hfill
     \begin{subfigure}[b]{0.24\textwidth}
         \centering
         \includegraphics[trim={0 1.cm 0 0},clip,width=\textwidth]{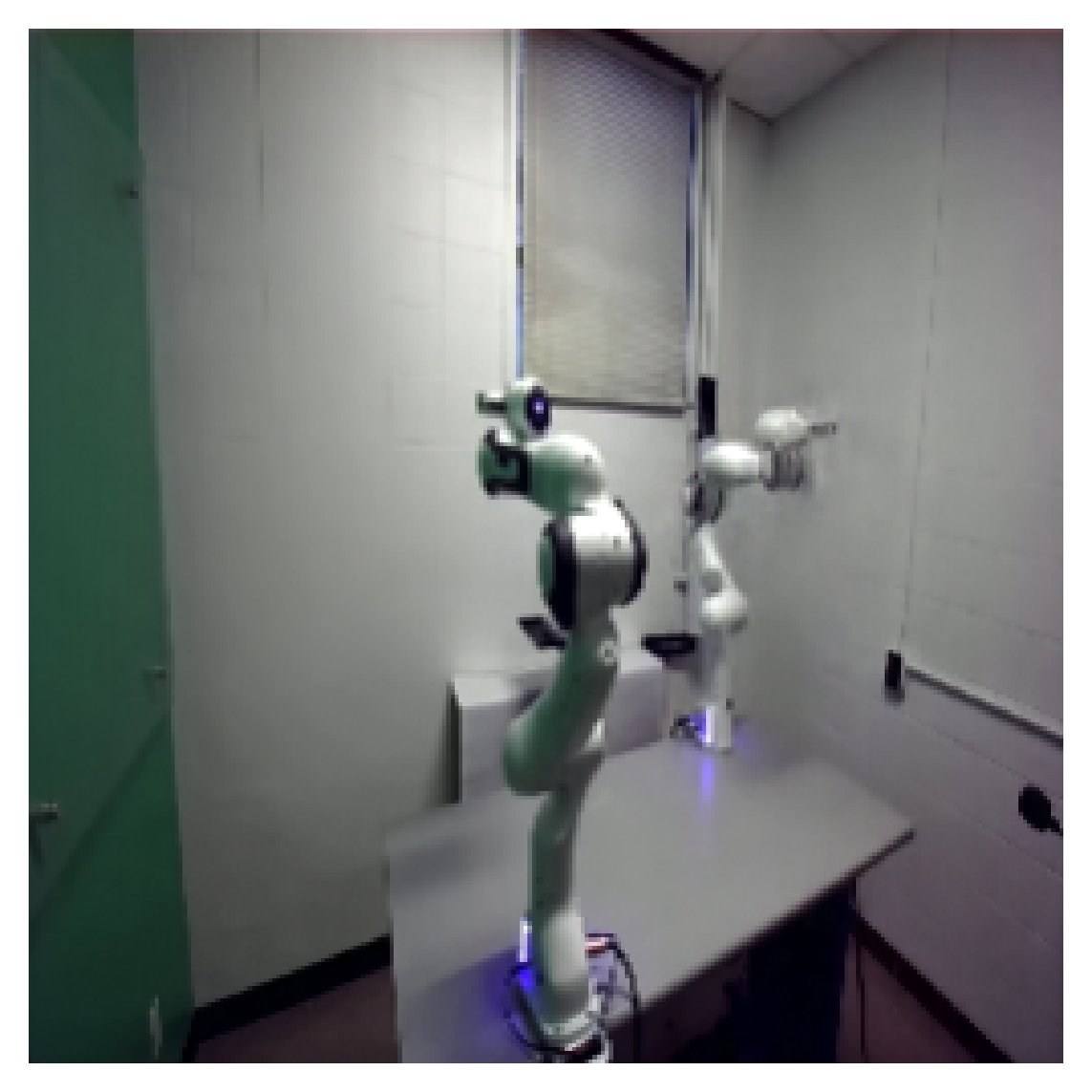}
     \end{subfigure}
     \hfill
     \begin{subfigure}[b]{0.24\textwidth}
         \centering
         \includegraphics[trim={0 1.cm 0 0},clip,width=\textwidth]{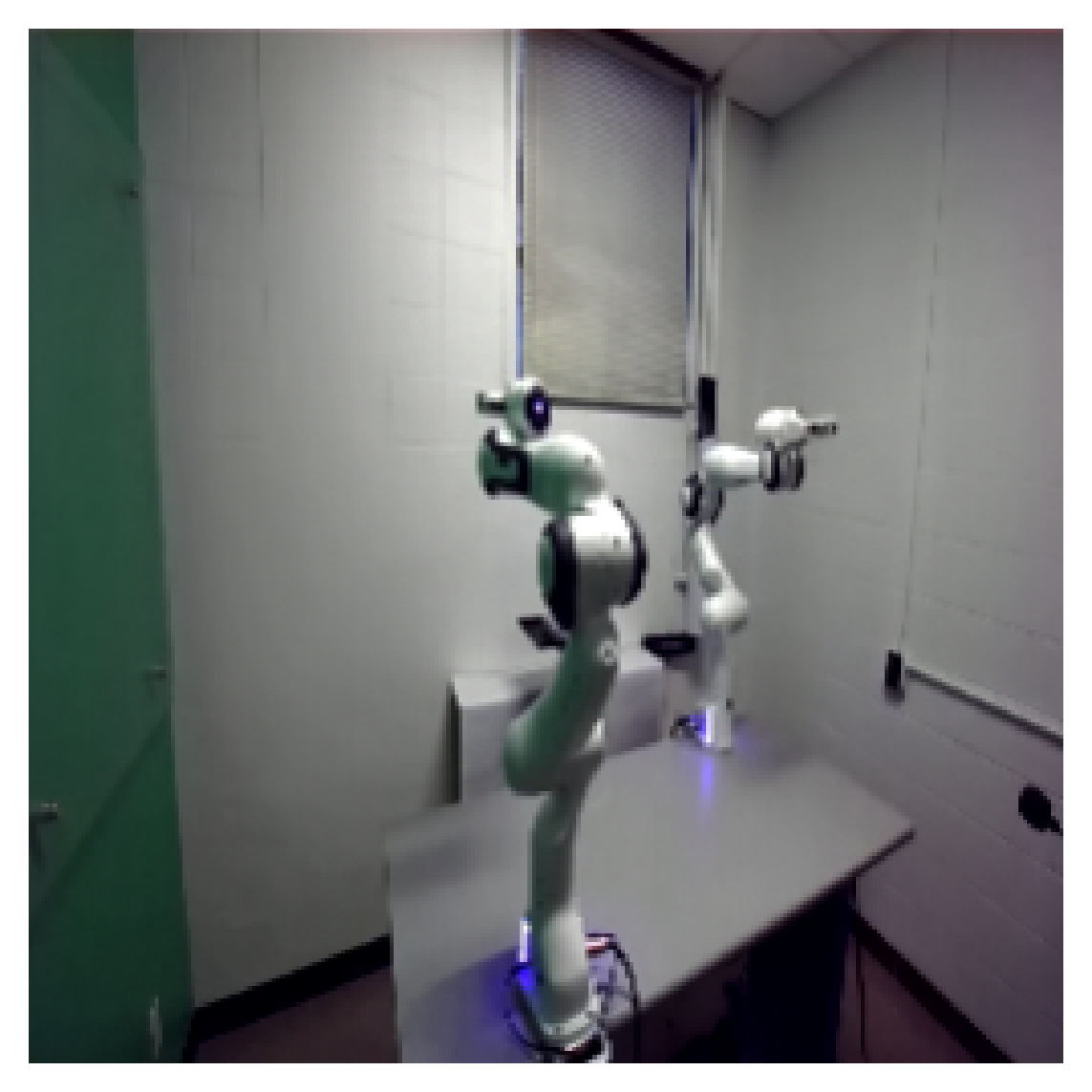}
     \end{subfigure}
    \hfill
     \begin{subfigure}[b]{0.24\textwidth}
         \centering
         \includegraphics[trim={0 1.cm 0 0},clip,width=\textwidth]{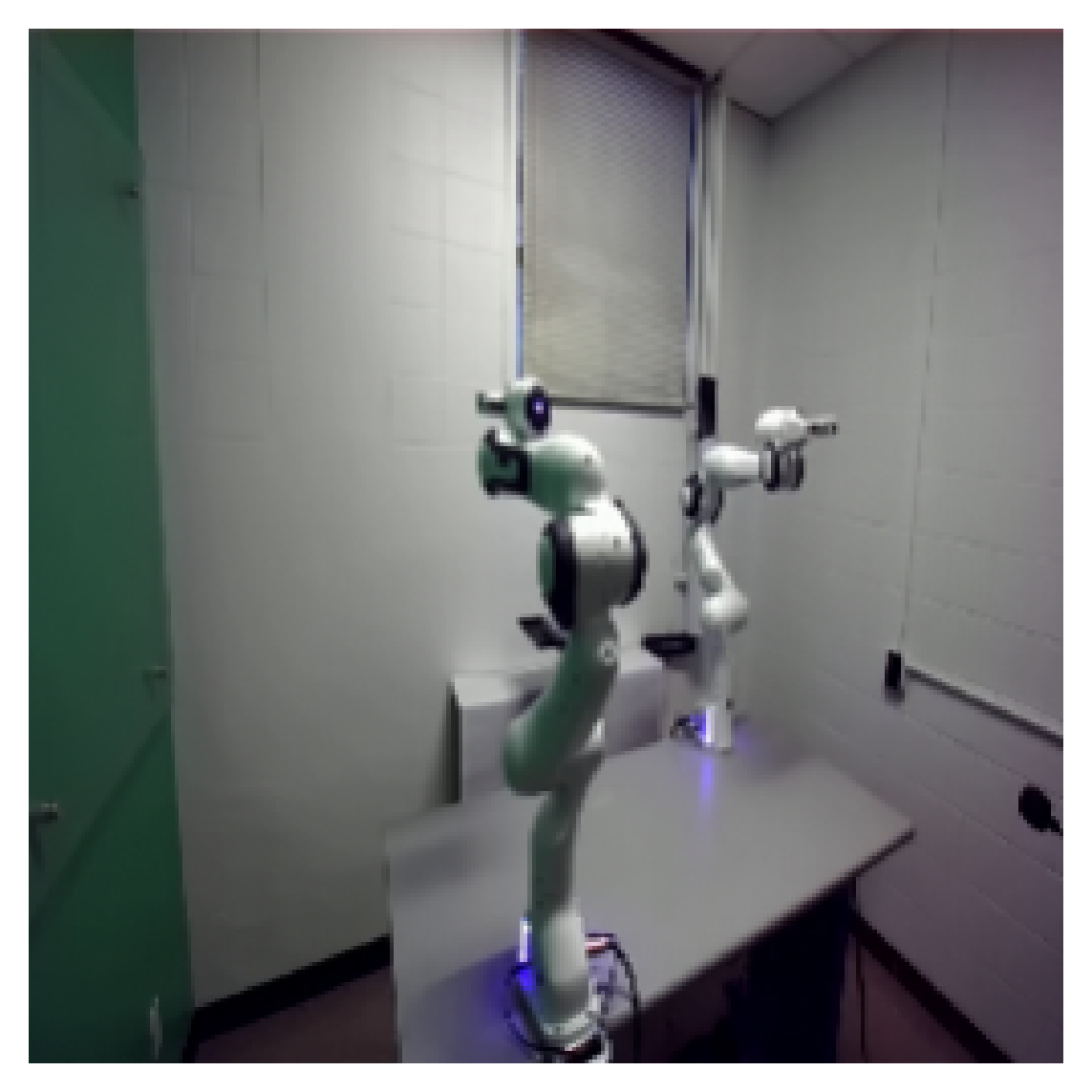}
     \end{subfigure}

          \begin{subfigure}[b]{0.24\textwidth}
         \centering
         \includegraphics[trim={0 1.cm 0 0},clip,width=\textwidth]{blank.png}
     \end{subfigure}
     \hfill
     \begin{subfigure}[b]{0.24\textwidth}
         \centering
         \includegraphics[width=\textwidth]{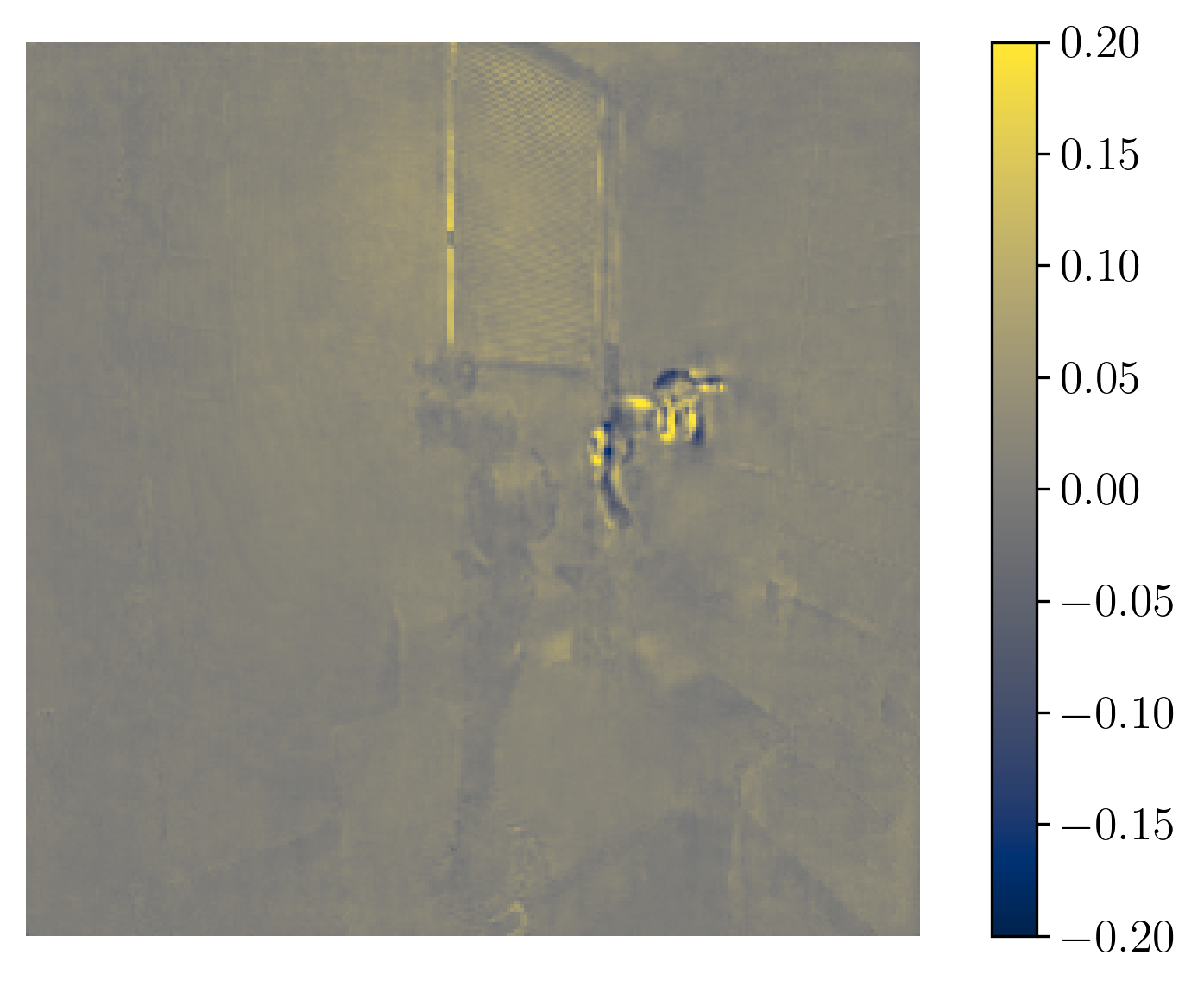}
     \end{subfigure}
     \hfill
     \begin{subfigure}[b]{0.24\textwidth}
         \centering
         \includegraphics[width=\textwidth]{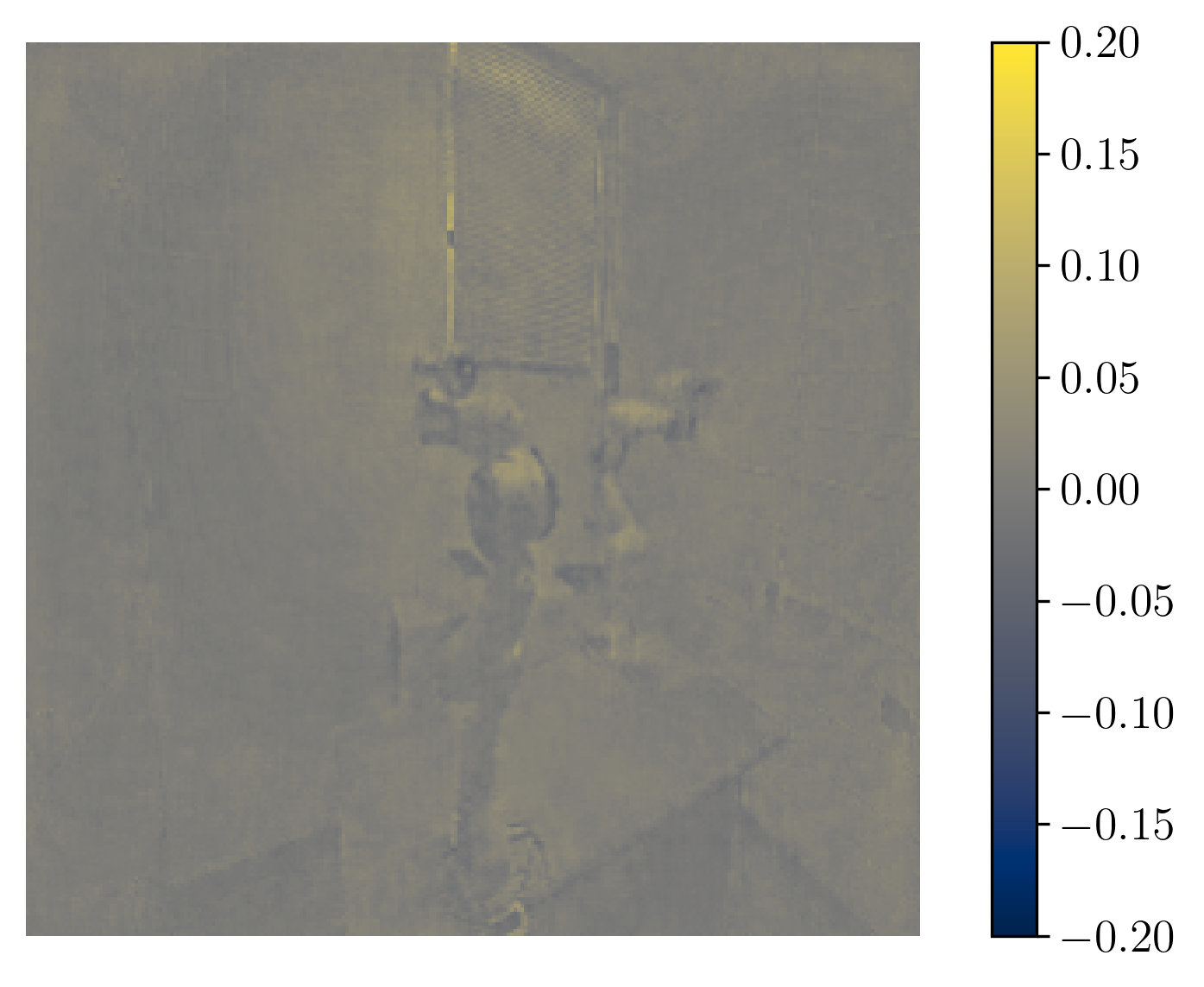}
     \end{subfigure}
    \hfill
     \begin{subfigure}[b]{0.24\textwidth}
         \centering
         \includegraphics[width=\textwidth]{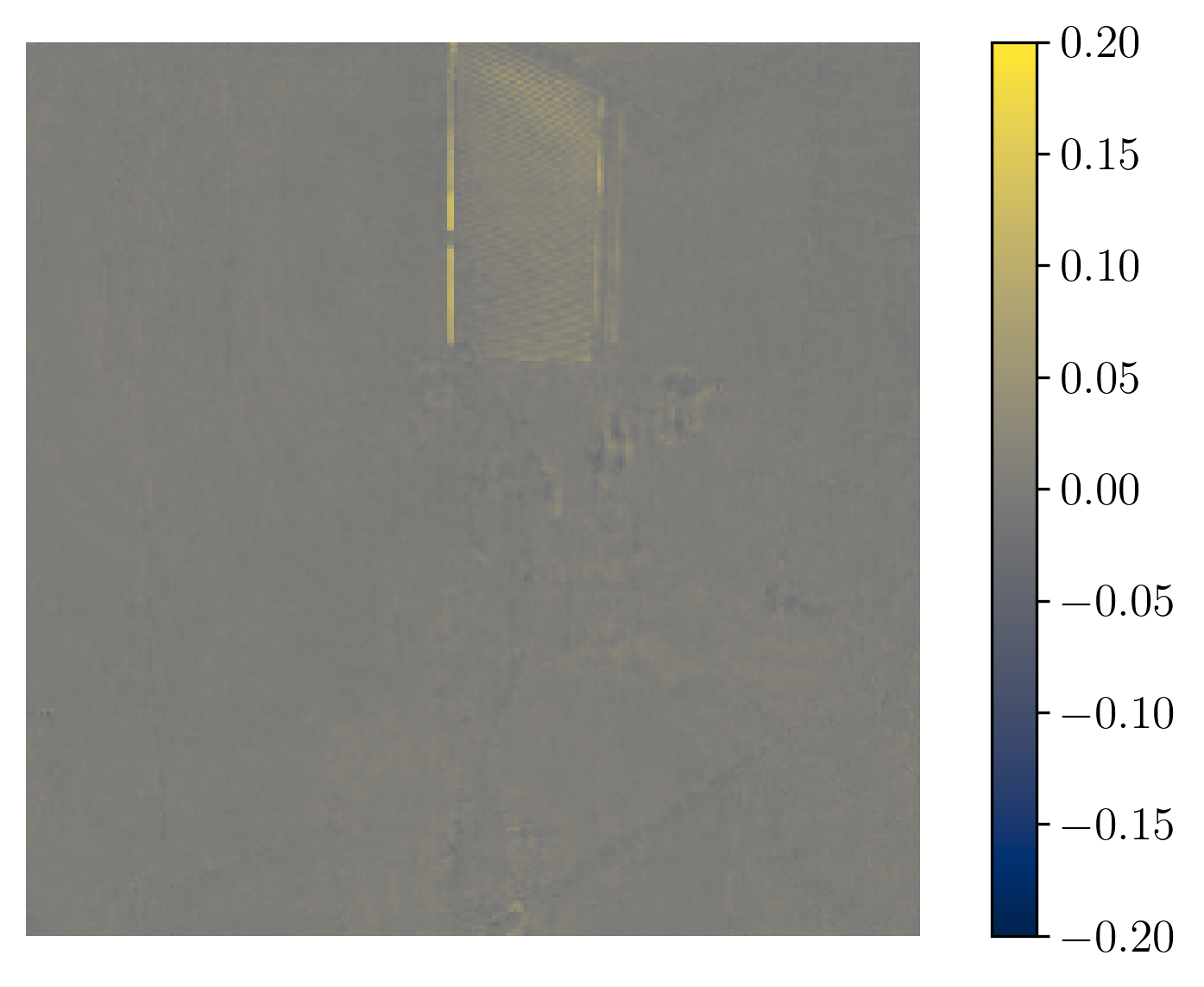}
     \end{subfigure}

          \begin{subfigure}[b]{0.24\textwidth}
         \centering
         \includegraphics[trim={0 1.cm 0 0},clip,width=\textwidth]{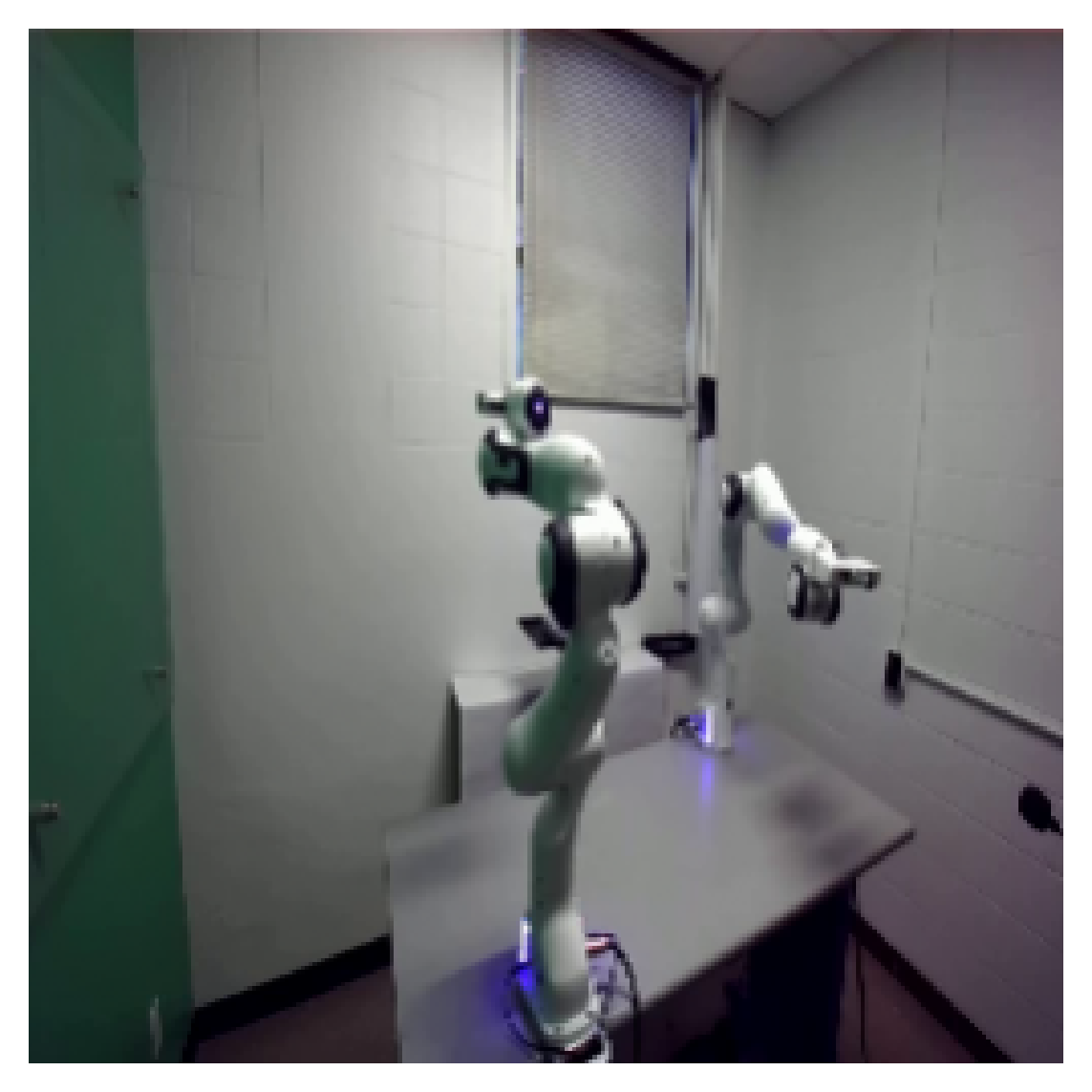}
     \end{subfigure}
     \hfill
     \begin{subfigure}[b]{0.24\textwidth}
         \centering
         \includegraphics[trim={0 1.cm 0 0},clip,width=\textwidth]{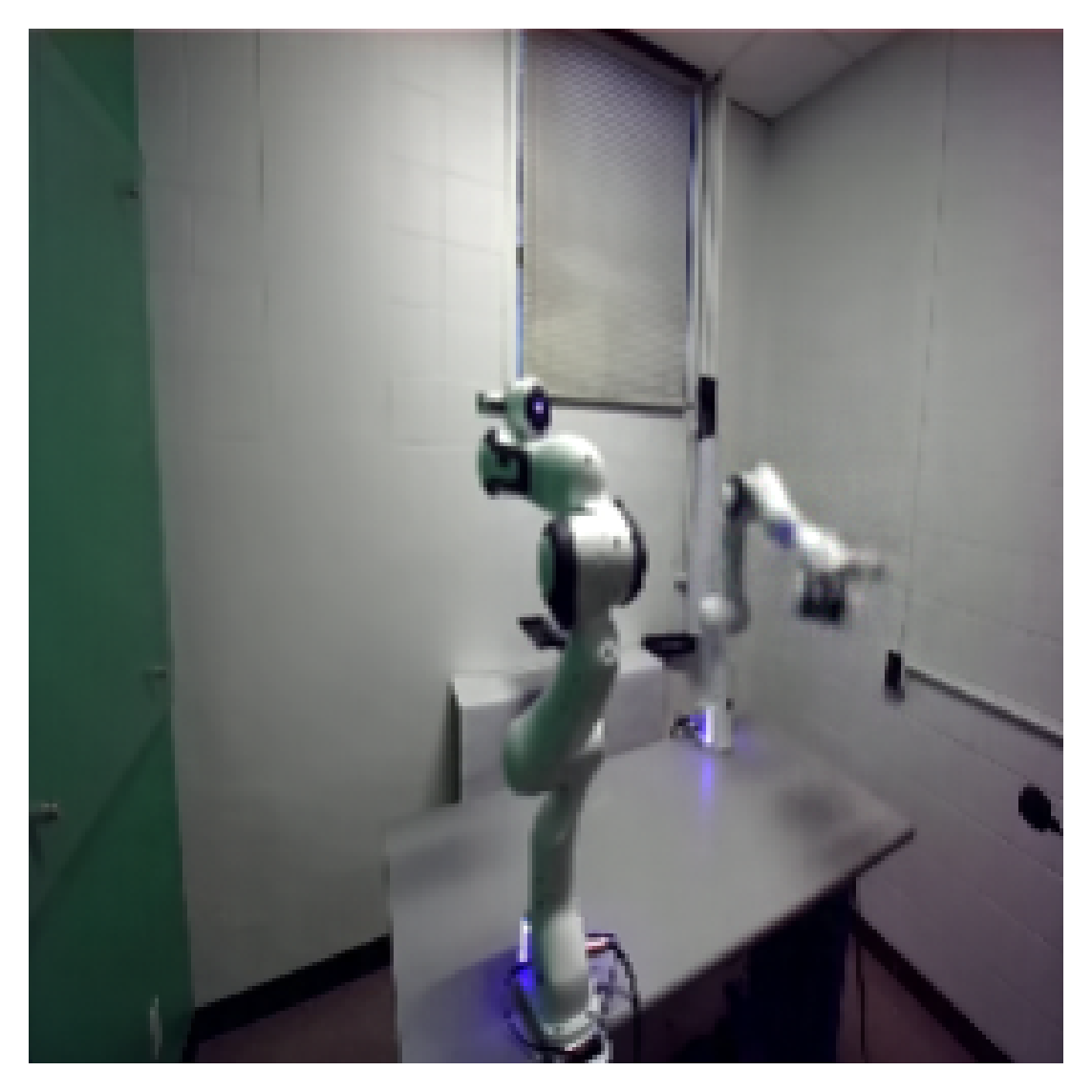}
     \end{subfigure}
     \hfill
     \begin{subfigure}[b]{0.24\textwidth}
         \centering
         \includegraphics[trim={0 1.cm 0 0},clip,width=\textwidth]{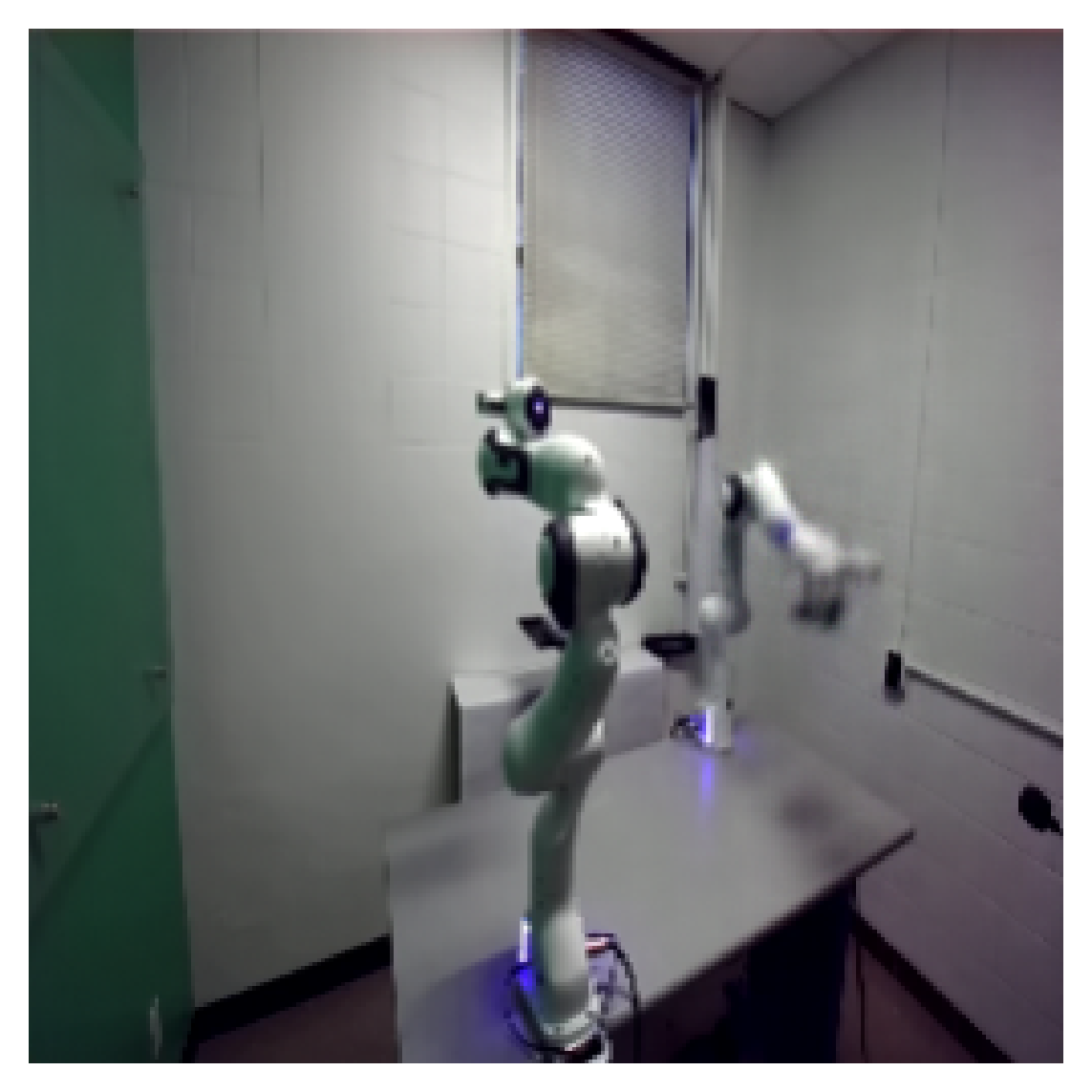}
     \end{subfigure}
    \hfill
     \begin{subfigure}[b]{0.24\textwidth}
         \centering
         \includegraphics[trim={0 1.cm 0 0},clip,width=\textwidth]{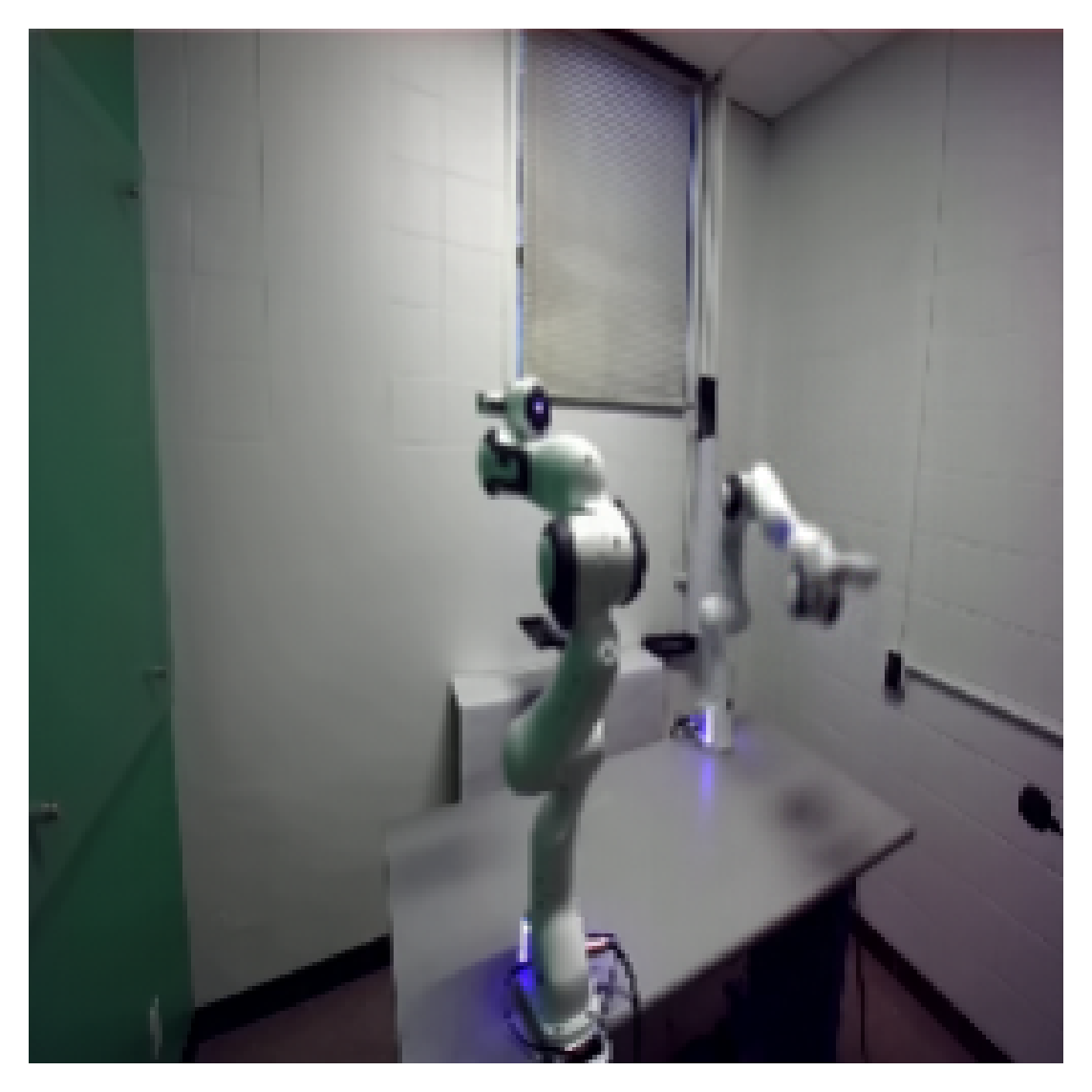}
     \end{subfigure}

          \begin{subfigure}[b]{0.24\textwidth}
         \centering
         \includegraphics[trim={0 1.cm 0 0},clip,width=\textwidth]{blank.png}
     \end{subfigure}
     \hfill
     \begin{subfigure}[b]{0.24\textwidth}
         \centering
         \includegraphics[width=\textwidth]{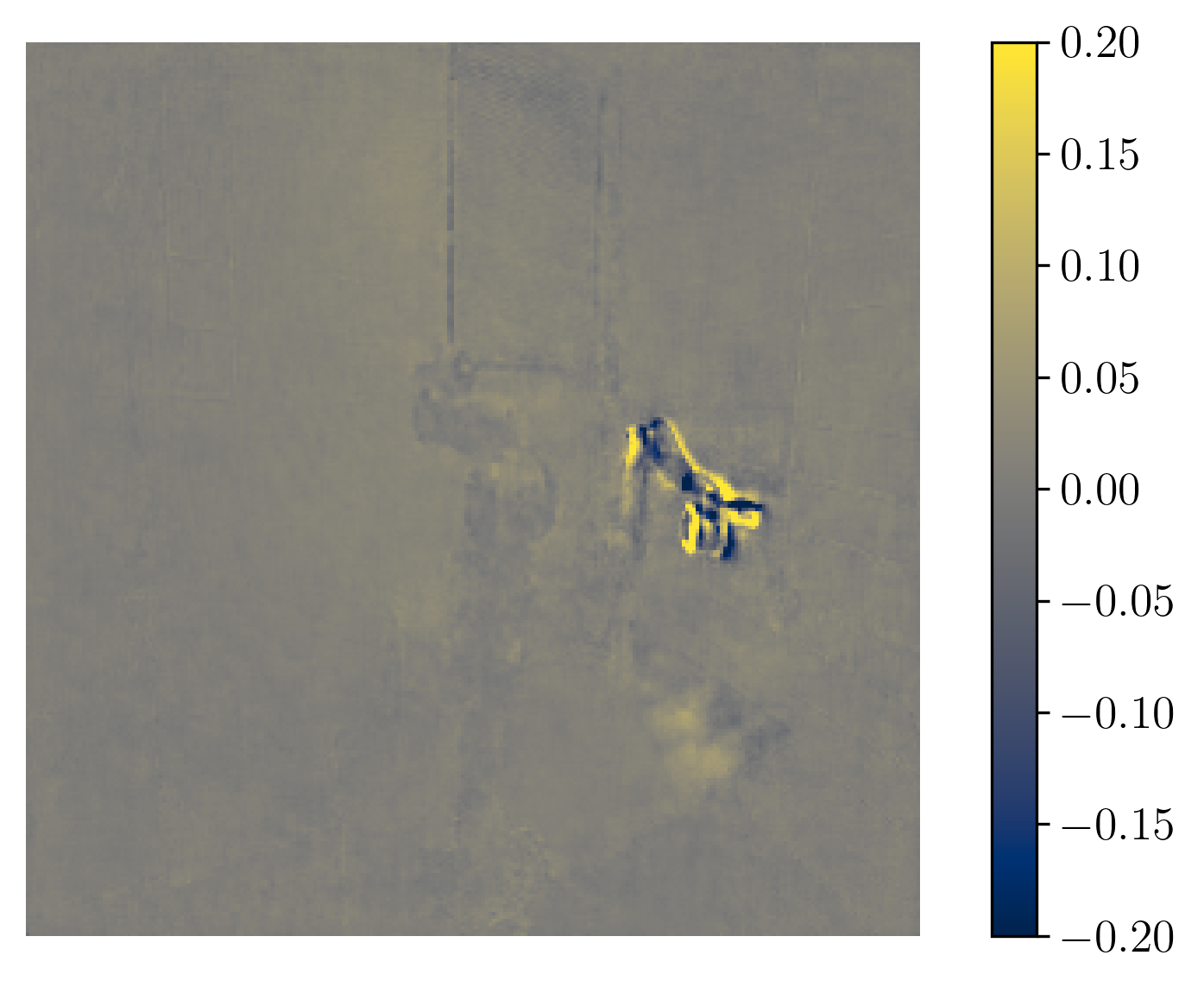}
     \end{subfigure}
     \hfill
     \begin{subfigure}[b]{0.24\textwidth}
         \centering
         \includegraphics[width=\textwidth]{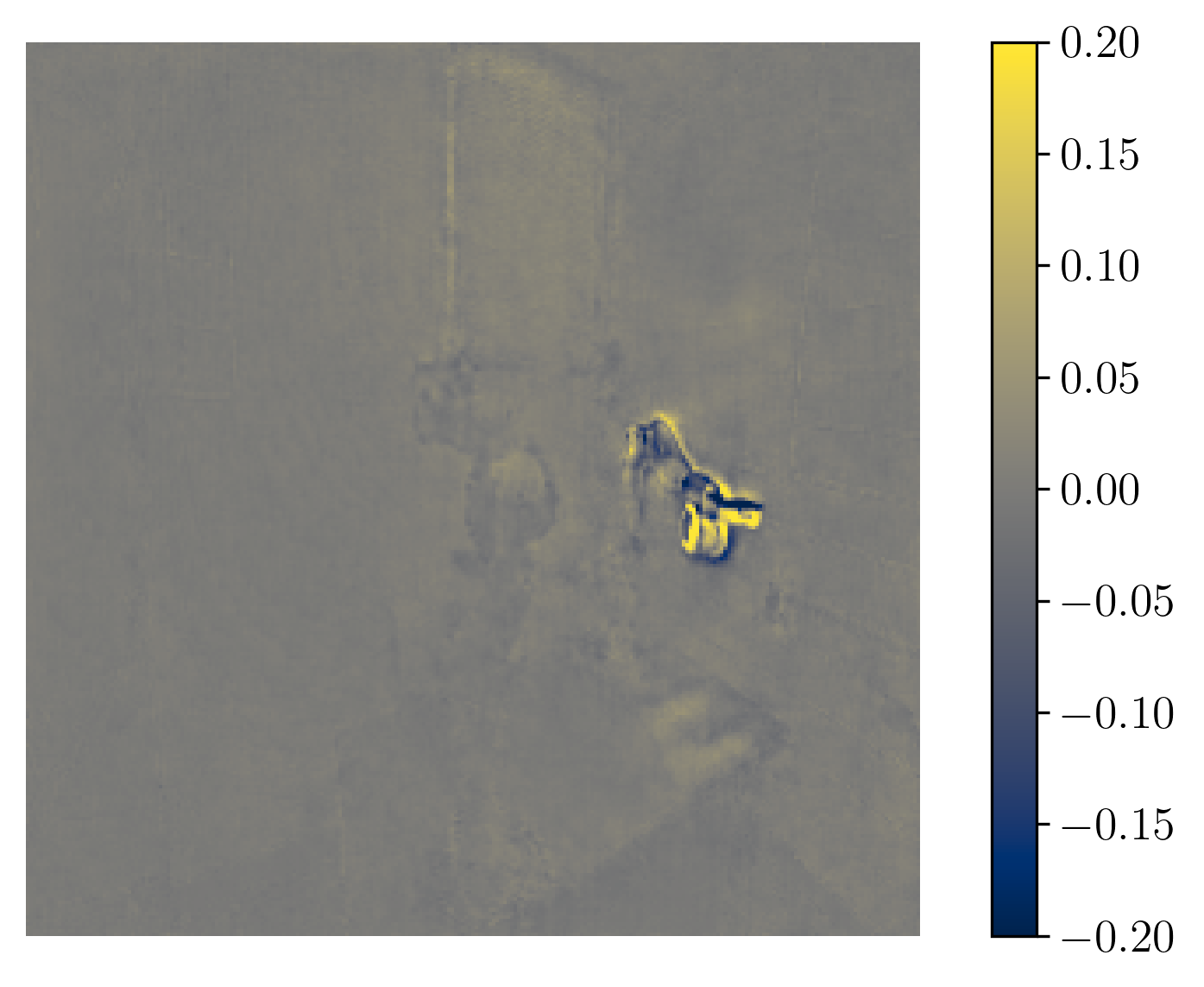}
     \end{subfigure}
    \hfill
     \begin{subfigure}[b]{0.24\textwidth}
         \centering
         \includegraphics[width=\textwidth]{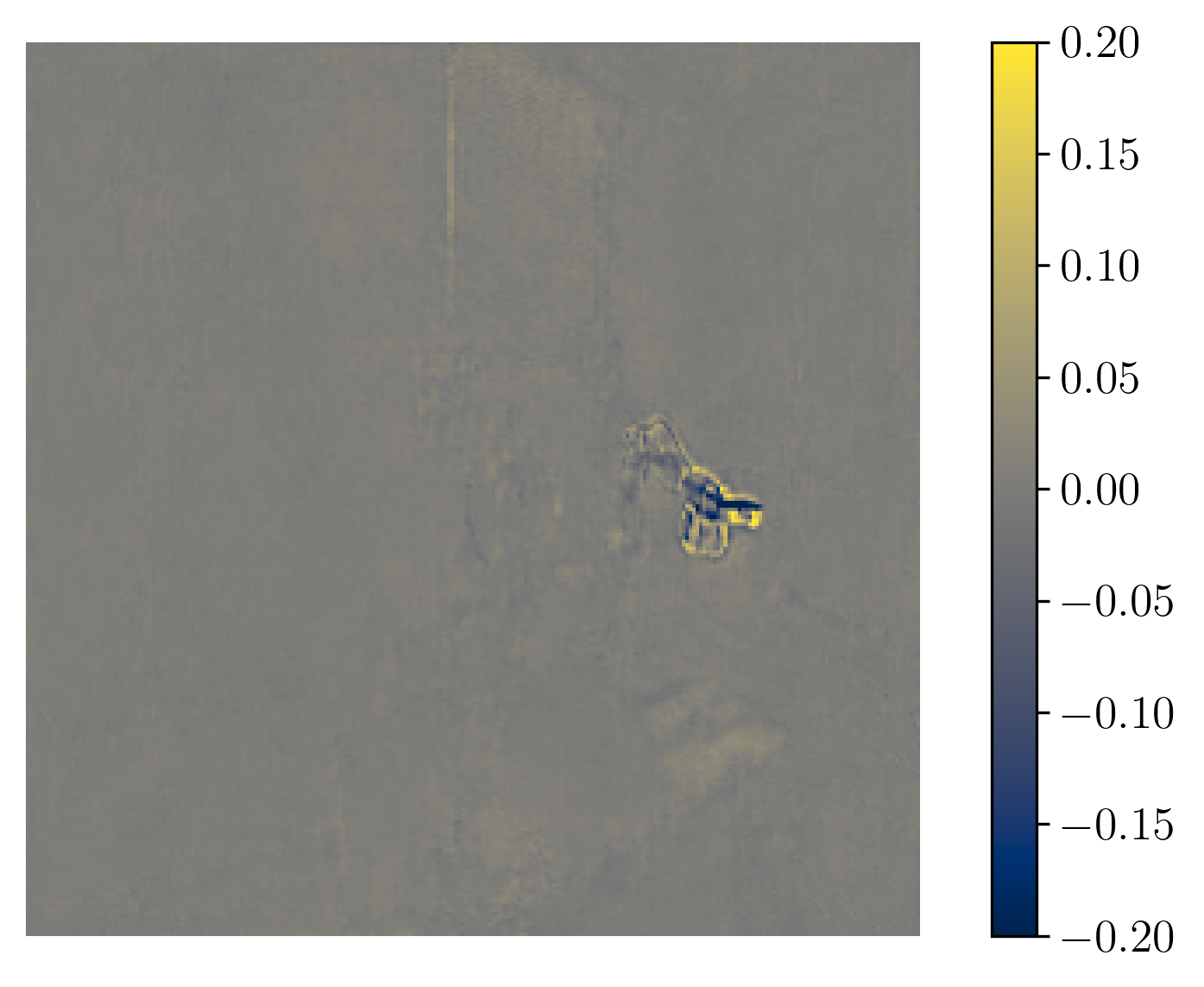}
     \end{subfigure}
    \caption{Comparison of reconstruction results for the bimodal real-world RoboMNIST dataset across aggregation methods. The figure displays Original Input followed by reconstructions using concatenation, summation, and attention (columns 2–4). Pixel-wise error maps (consistent scale) are presented below the reconstruction.}
    \label{fig:reconstructions_robomnist}
\end{figure}

Our quantitative assessment of the fusion strategies, conducted on the Real-World RoboMNIST benchmark, identifies the Attention-based architecture as the most effective method. The root of this performance gap lies in a pronounced imbalance between the input modalities. Analysis reveals that the image modality provides a consistently and exceptionally low reconstruction loss, underscoring its role as the primary source of predictive information. Conversely, the sensor modality is characterized by an extremely high loss, suggesting that its signal is either too noisy or insufficiently informative for the task at hand. The Attention mechanism's success stems from its capacity to adaptively focus on the more reliable modality. This quantitative finding is qualitatively supported by \autoref{fig:reconstructions_robomnist}, where the reconstructions from the Attention model demonstrate higher fidelity, further validating its ability to manage imbalanced and noisy multi-modal data.

All experiments presented so far were conducted in a bimodal setup, where the model jointly processed two data modalities. We now extend this framework to a third modality, incorporating wireless signal information to enrich the sensory representation. As wireless signals propagate through an environment, they interact with surrounding objects, leading to reflections, diffractions, and scattering. By leveraging WiFi Channel State Information, we can analyze the propagation characteristics of individual subcarriers from the transmitter to the receiver. This additional modality provides valuable spatial and environmental context, enabling the model to capture subtle variations in scene geometry and dynamics that are not observable through visual data alone. The results using the trimodal setup are presented in \autoref{tab:robotmnist_results_trimodal}.

\begin{table}[htbp]
\caption{Comparison of fusion strategies based on loss metrics using the trimodal real-world RoboMNIST dataset. The results present the mean ($\mu$) $\pm$ standard deviation ($\sigma$) of the loss computed over 10 independent trials for each architecture.}
\label{tab:robotmnist_results_trimodal}
\begin{tabular}{l|cccc}
\toprule
              & \multicolumn{4}{c}{Training in $10^{-2}$} \\
\midrule
Model         & Sensor Modality & Image Modality & Signal Modality & Combined  \\
\midrule
Summation & 16.676 $\pm$ 2.118 & 0.092 $\pm$ 0.004 & 33.990 $\pm$ 2.770 & 50.757 $\pm$ 4.559  \\
Concatenation & 23.448 $\pm$ 3.096 & 0.093 $\pm$ 0.006 & 54.465 $\pm$ 2.619 & 78.006 $\pm$ 4.428 \\
Attention & 1.8760 $\pm$ 0.144 & 0.073 $\pm$ 0.004 & 17.582 $\pm$ 0.198 & 19.5310 $\pm$ 0.145 \\
\midrule
              & \multicolumn{4}{c}{Testing in $10^{-2}$} \\
\midrule
Model         & Sensor Modality & Image Modality & Signal Modality & Combined \\
\midrule
Summation & 44.452 $\pm$ 4.546 & 0.191 $\pm$ 0.016 & 87.011 $\pm$ 1.310 & 131.654 $\pm$ 4.844 \\
Concatenation & 54.730 $\pm$ 9.361 & 0.191 $\pm$ 0.016 & 97.978 $\pm$ 2.182 & 152.899 $\pm$ 9.494 \\
Attention & 3.3520 $\pm$ 0.185 & 0.160 $\pm$ 0.013 & 31.698 $\pm$ 0.533 & 35.2090 $\pm$ 0.539 \\
\bottomrule
\end{tabular}
\end{table}

Across both the training and testing phases, the Attention-based fusion model consistently yields the lowest loss compared to the summation and concatenation methods, highlighting its superior capability to effectively weigh and integrate information from the different modalities. Notably, the image modality consistently exhibits the lowest individual loss contribution, while the signal modality, and to a lesser extent the sensor modality, contribute the largest magnitudes to the overall combined loss, suggesting a challenge in feature extraction or noise handling for these data streams. The drastic increase in loss when moving from single-modality to the combined-modality approach for all architectures, especially the pronounced difference in the testing losses for the summation and concatenation methods, indicates that these simpler fusion techniques struggle to generalize robustly when combining disparate information, leading to degraded performance, whereas the Attention model maintains a more controlled and effective integration of the trimodal input.

\begin{figure}
     \centering
     \begin{subfigure}[b]{0.325\textwidth}
         \centering
         \includegraphics[width=\textwidth]{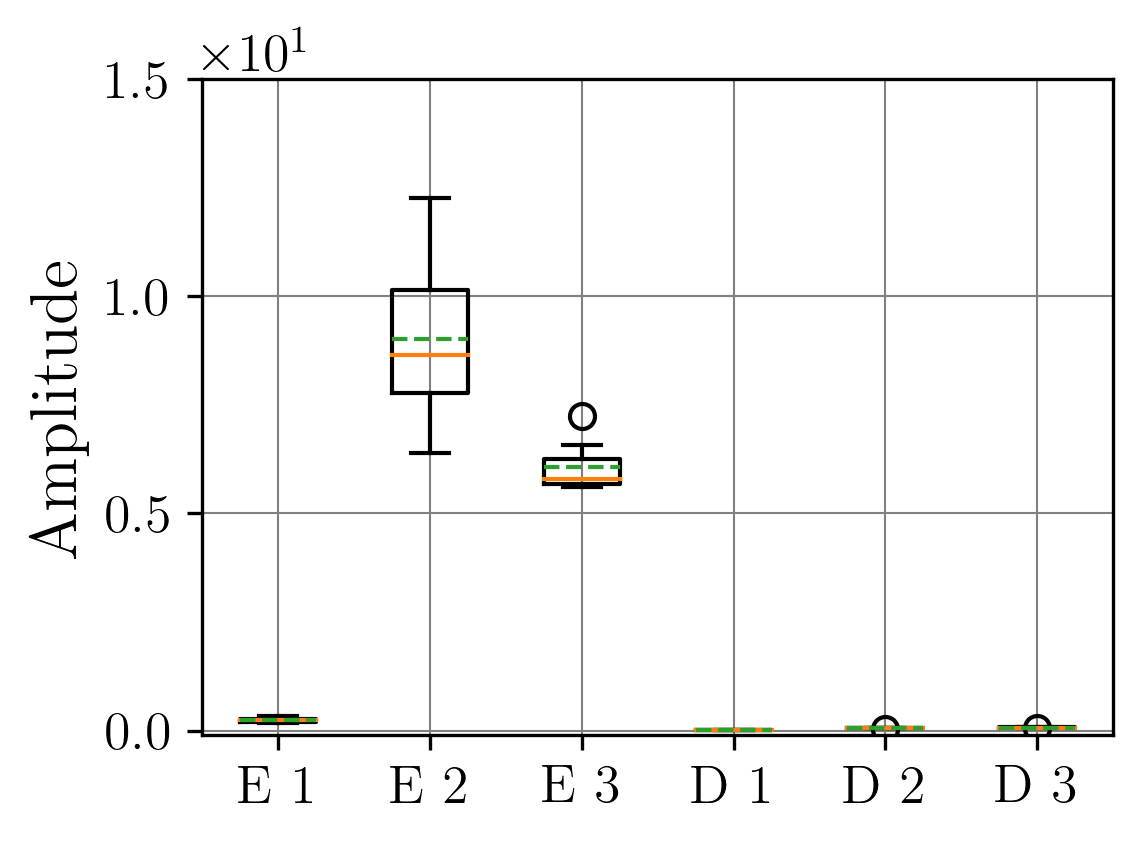}
         \caption{Summation}
     \end{subfigure}
     \hfill
     \begin{subfigure}[b]{0.325\textwidth}
         \centering
         \includegraphics[width=\textwidth]{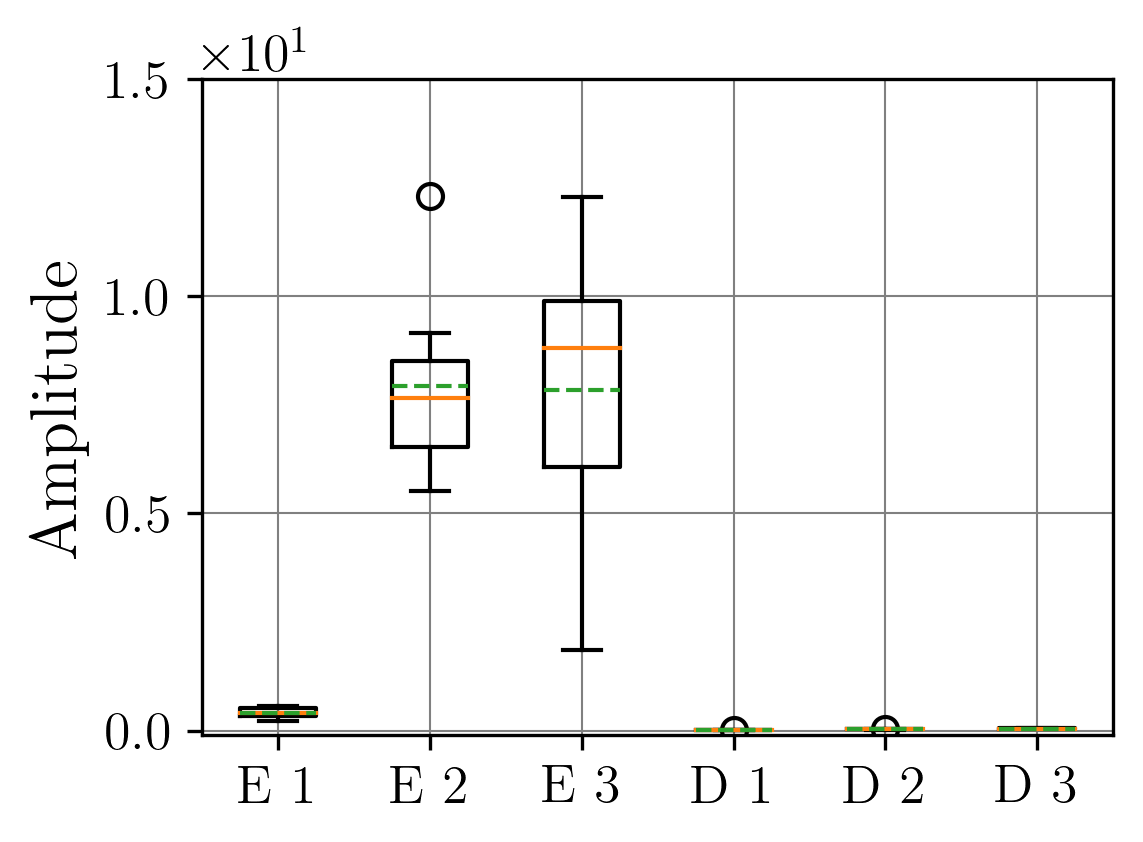}
         \caption{Concatenation}
     \end{subfigure}
     \hfill
     \begin{subfigure}[b]{0.325\textwidth}
         \centering
         \includegraphics[width=\textwidth]{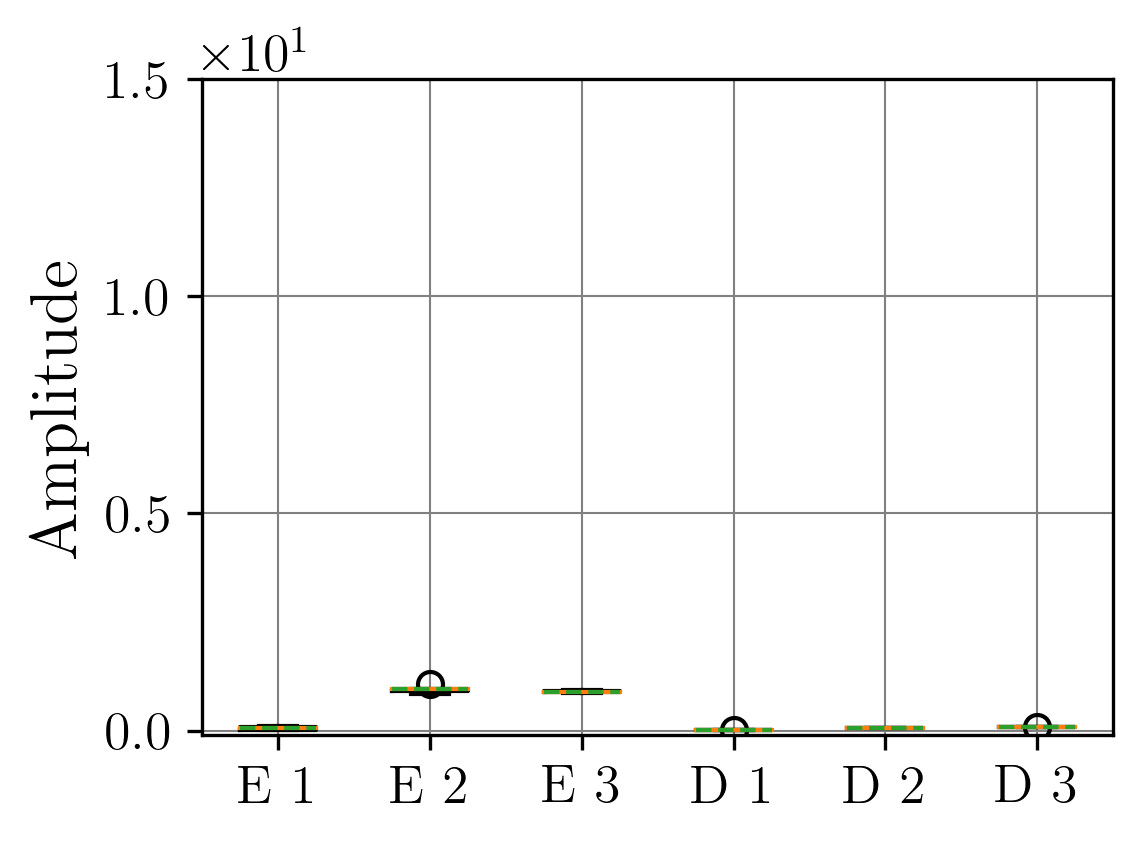}
         \caption{Attention}
     \end{subfigure}
    \caption{Boxplots of the estimated Lipschitz constants across 10 trials of each submodel on the real-world RoboMNIST dataset, using different aggregation methods.}
    \label{fig:trimodal_lipschitz}
\end{figure}

\autoref{fig:trimodal_lipschitz} presents the behavior of the Lipschitz constant for each sub-block within the trimodal autoencoder across the different fusion methods. Consistent with findings from bimodal experiments, the attention-based fusion method yields the lowest Lipschitz constant values, indicating a more stable, less sensitive, and smoother mapping function that prevents vanishing or exploding gradients. Conversely, the Summation and Concatenation methods are likely to result in substantially higher values. This confirms that the Attention mechanism successfully regularizes the feature space integration by adaptively controlling the influence of each modality, leading to a much better conditioned combined representation even with more modalities.

\subsection{Ablations on Real-World Data}

\begin{figure}
     \centering
     \begin{subfigure}[b]{0.49\textwidth}
         \centering
         \includegraphics[width=\textwidth]{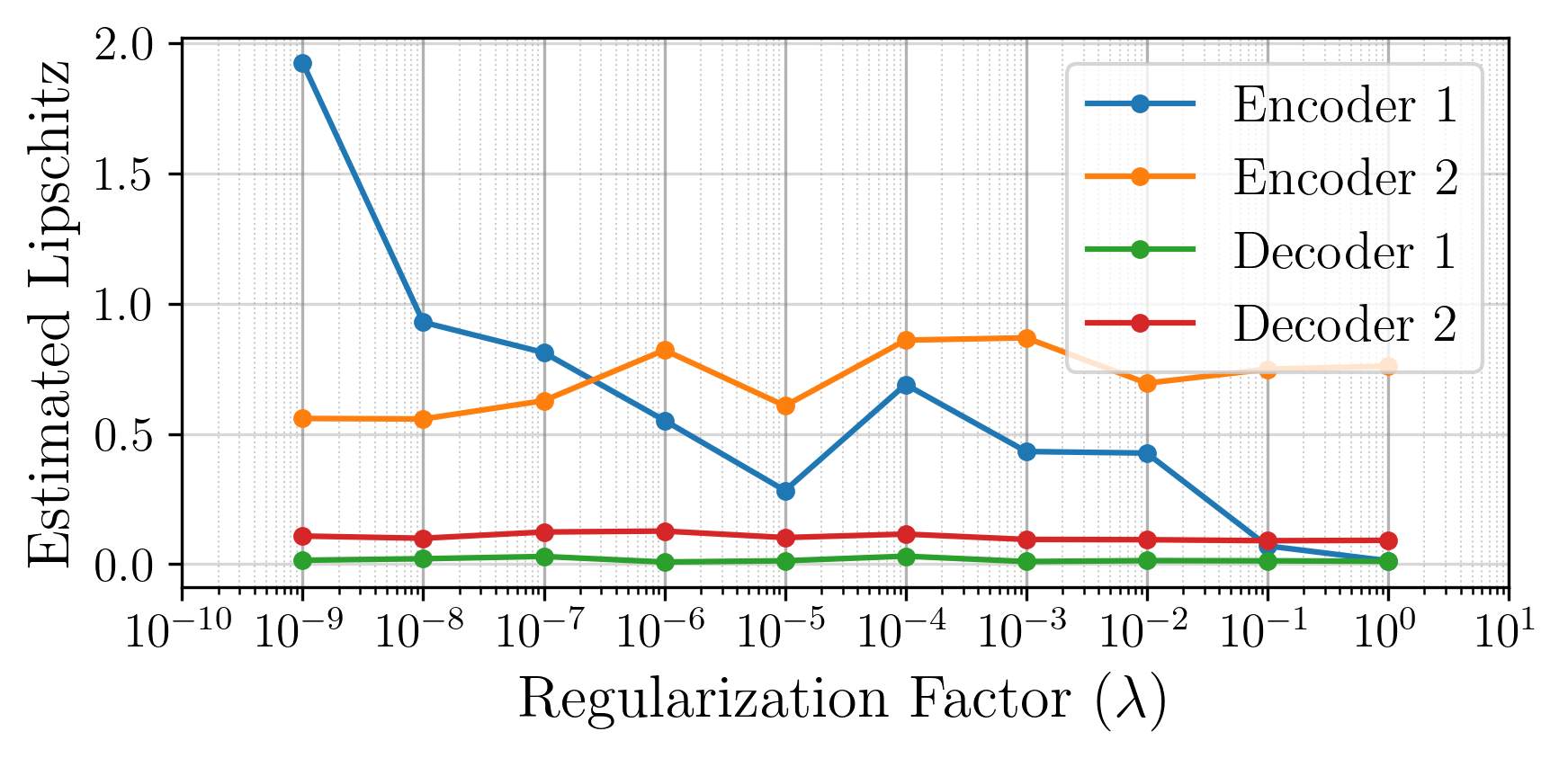}
         \caption{Lipschitz Constant}
     \end{subfigure}
     \hfill
     \begin{subfigure}[b]{0.49\textwidth}
         \centering
         \includegraphics[width=\textwidth]{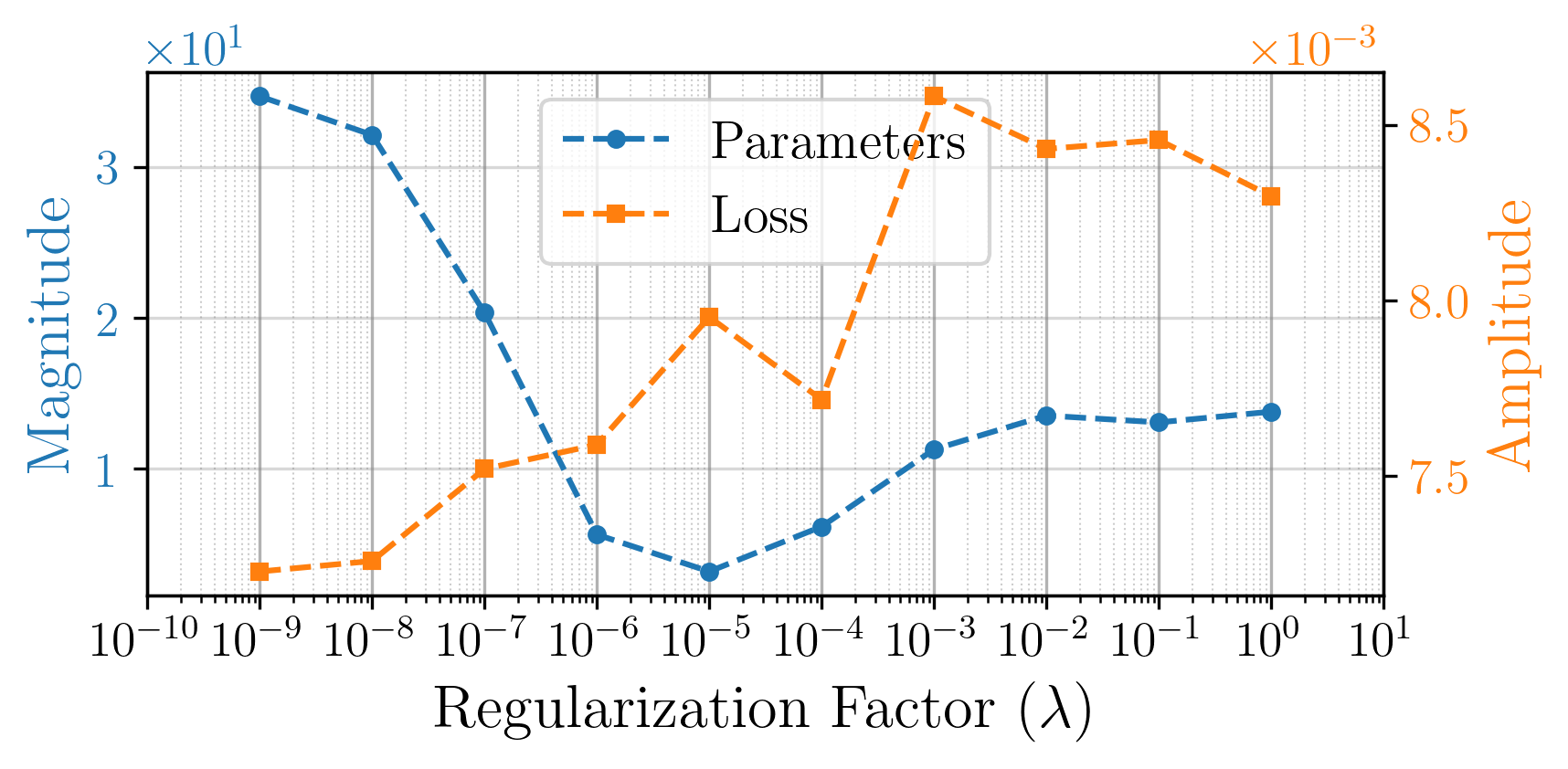}
         \caption{Performance}
     \end{subfigure}
    \caption{Ablation study evaluating the relationship between the regularization factor on the logarithmic axis from $10^{-9}$ to $10^0$) and two critical network metrics: the Lipschitz constant (left) and network properties (right).}
    \label{fig:ablations__real}
\end{figure}

To assess the impact of regularization strength, we conducted ablation studies varying the regularization factor ($\lambda$). Specifically, we evaluated two key network properties as a function of $\lambda$: the Lipschitz constant (a measure of model stability/robustness) and the magnitude of the network's parameter set. Both metrics are presented in \autoref{fig:ablations__real}, plotted across a logarithmic scale spanning nine orders of magnitude, from $10^{-9}$ to $10^0$.

The ablation study reveals the critical trade-off between model complexity and stability imposed by the regularization factor ($\lambda$) across a logarithmic scale ($10^{-9}$ to $10^0$). For the parameter set, $\lambda$ exerts a strong, inverse, non-linear control: the magnitude of the parameters remains high for $\lambda < 10^{-6}$ and then undergoes a steep decline until saturating at a minimal level. Concurrently, the Lipschitz constant starts at a high, unstable value, rapidly decreasing and stabilizing as the parameter constraints become effective. This demonstrates that regularization effectively controls the model's sensitivity. The optimal region for regularization, balancing both the necessary complexity reduction and achieving maximum stability (minimum Lipschitz constant), is identified around $\lambda \in [10^{-6}, 10^{-5}, 10^{-4}]$, where the network achieves high robustness without fully collapsing its weight structure and performance.

%%%%%%%%%%%%%%%%%%%%%%%%%%%%%%%%%%%%%%%%%%%%%%%%%%%%%%%%%%%%%%%%%%%%%%%%%%%%%%%%%%%
%%%%%%%%%%%%%%%%%%%%%%%%%%%%%%%%%%%%%%%%%%%%%%%%%%%%%%%%%%%%%%%%%%%%%%%%%%%%%%%%%%%
%%%%%%%%%%%%%%%%%%%%%%%%%%%%%%%%%%%%%%%%%%%%%%%%%%%%%%%%%%%%%%%%%%%%%%%%%%%%%%%%%%%
%%%%%%%%%%%%%%%%%%%%%%%%%%%%%%%%%%%%%%%%%%%%%%%%%%%%%%%%%%%%%%%%%%%%%%%%%%%%%%%%%%%

\section{Conclusion}\label{sec:conclusion}

In this work, we presented a comprehensive analysis of Lipschitz constants for various aggregation methods in multimodal autoencoders, leading to the development of a novel Lipschitz-guided attention-based fusion approach for multimodal learning. Our method incorporates scaling and Lipschitz regularization to control training dynamics and ensure stability. We provided both theoretical guarantees and empirical validation across four industrial datasets of varying complexity. The experimental results consistently demonstrate that our proposed approach outperforms classical aggregation methods in terms of performance metrics, training stability, and output consistency.

Several promising research directions emerge from this work to enhance both theoretical foundations and practical deployment. First, investigating adaptive Lipschitz bounds could automatically tune stability constraints based on modality-specific characteristics, potentially improving performance on heterogeneous data streams. Secondly, systematically evaluate the method's robustness to missing or corrupted modalities and improve its reliability in real-world applications where sensor failures or data dropouts occur. These extensions would collectively advance the practical applicability of our approach in industrial environments while preserving its theoretical stability guarantees.

%%%%%%%%%%%%%%%%%%%%%%%%%%%%%%%%%%%%%%%%%%%%%%%%%%%%%%%%%%%%%%%%%%%%%%%%%%%%%%%%%%%
%%%%%%%%%%%%%%%%%%%%%%%%%%%%%%%%%%%%%%%%%%%%%%%%%%%%%%%%%%%%%%%%%%%%%%%%%%%%%%%%%%%
%%%%%%%%%%%%%%%%%%%%%%%%%%%%%%%%%%%%%%%%%%%%%%%%%%%%%%%%%%%%%%%%%%%%%%%%%%%%%%%%%%%
%%%%%%%%%%%%%%%%%%%%%%%%%%%%%%%%%%%%%%%%%%%%%%%%%%%%%%%%%%%%%%%%%%%%%%%%%%%%%%%%%%%

\bibliographystyle{unsrt}  
\bibliography{references} 

@article{bertolini2021machine,
  title={Machine Learning for industrial applications: A comprehensive literature review},
  author={Bertolini, Massimo and Mezzogori, Davide and Neroni, Mattia and Zammori, Francesco},
  journal={Expert Systems with Applications},
  volume={175},
  pages={114820},
  year={2021},
  publisher={Elsevier}
}

@article{emmanouilidis2019enabling,
  title={Enabling the human in the loop: Linked data and knowledge in industrial cyber-physical systems},
  author={Emmanouilidis, Christos and Pistofidis, Petros and Bertoncelj, Luka and Katsouros, Vassilis and Fournaris, Apostolos and Koulamas, Christos and Ruiz-Carcel, Cristobal},
  journal={Annual reviews in control},
  volume={47},
  pages={249--265},
  year={2019},
  publisher={Elsevier}
}

@inproceedings{altinses2023deep,
  title={Deep Multimodal Fusion with Corrupted Spatio-Temporal Data Using Fuzzy Regularization},
  author={Altinses, Diyar and Schwung, Andreas},
  booktitle={IECON 2023-49th Annual Conference of the IEEE Industrial Electronics Society},
  pages={1--7},
  year={2023},
  organization={IEEE}
}

@inproceedings{zeng2019deep,
  title={Deep surface normal estimation with hierarchical rgb-d fusion},
  author={Zeng, Jin and Tong, Yanfeng and Huang, Yunmu and Yan, Qiong and Sun, Wenxiu and Chen, Jing and Wang, Yongtian},
  booktitle={Proceedings of the IEEE/CVF conference on computer vision and pattern recognition},
  pages={6153--6162},
  year={2019}
}

@inproceedings{hazirbas2017fusenet,
  title={Fusenet: Incorporating depth into semantic segmentation via fusion-based cnn architecture},
  author={Hazirbas, Caner and Ma, Lingni and Domokos, Csaba and Cremers, Daniel},
  booktitle={Computer Vision--ACCV 2016: 13th Asian Conference on Computer Vision, Taipei, Taiwan, November 20-24, 2016, Revised Selected Papers, Part I 13},
  pages={213--228},
  year={2017},
  organization={Springer}
}

@article{nagrani2021attention,
  title={Attention bottlenecks for multimodal fusion},
  author={Nagrani, Arsha and Yang, Shan and Arnab, Anurag and Jansen, Aren and Schmid, Cordelia and Sun, Chen},
  journal={Advances in neural information processing systems},
  volume={34},
  pages={14200--14213},
  year={2021}
}

@article{marucheck2011product,
  title={Product safety and security in the global supply chain: Issues, challenges and research opportunities},
  author={Marucheck, Ann and Greis, Noel and Mena, Carlos and Cai, Linning},
  journal={Journal of operations management},
  volume={29},
  number={7-8},
  pages={707--720},
  year={2011},
  publisher={Elsevier}
}

@article{gourisaria2021application,
  title={Application of machine learning in industry 4.0},
  author={Gourisaria, Mahendra Kumar and Agrawal, Rakshit and Harshvardhan, GM and Pandey, Manjusha and Rautaray, Siddharth Swarup},
  journal={Machine learning: Theoretical foundations and practical applications},
  pages={57--87},
  year={2021},
  publisher={Springer}
}

@article{behzad2025robomnist,
  title={Robomnist: A multimodal dataset for multi-robot activity recognition using wifi sensing, video, and audio},
  author={Behzad, Kian and Zandi, Rojin and Motamedi, Elaheh and Salehinejad, Hojjat and Siami, Milad},
  journal={Scientific Data},
  volume={12},
  number={1},
  pages={326},
  year={2025},
  publisher={Nature Publishing Group UK London}
}

@article{gouk2021regularisation,
  title={Regularisation of neural networks by enforcing lipschitz continuity},
  author={Gouk, Henry and Frank, Eibe and Pfahringer, Bernhard and Cree, Michael J},
  journal={Machine Learning},
  volume={110},
  pages={393--416},
  year={2021},
  publisher={Springer}
}

@article{pauli2021training,
  title={Training robust neural networks using Lipschitz bounds},
  author={Pauli, Patricia and Koch, Anne and Berberich, Julian and Kohler, Paul and Allg{\"o}wer, Frank},
  journal={IEEE Control Systems Letters},
  volume={6},
  pages={121--126},
  year={2021},
  publisher={IEEE}
}

@inproceedings{patrini2017making,
  title={Making deep neural networks robust to label noise: A loss correction approach},
  author={Patrini, Giorgio and Rozza, Alessandro and Krishna Menon, Aditya and Nock, Richard and Qu, Lizhen},
  booktitle={Proceedings of the IEEE conference on computer vision and pattern recognition},
  pages={1944--1952},
  year={2017}
}

@inproceedings{kim2020lipschitz,
  title={Lipschitz continuous autoencoders in application to anomaly detection},
  author={Kim, Young-geun and Kwon, Yongchan and Chang, Hyunwoong and Paik, Myunghee Cho},
  booktitle={International Conference on Artificial Intelligence and Statistics},
  pages={2507--2517},
  year={2020},
  organization={PMLR}
}

@inproceedings{carlini2017towards,
  title={Towards evaluating the robustness of neural networks},
  author={Carlini, Nicholas and Wagner, David},
  booktitle={2017 ieee symposium on security and privacy (sp)},
  pages={39--57},
  year={2017},
  organization={Ieee}
}

@inproceedings{zhou2019lipschitz,
  title={Lipschitz generative adversarial nets},
  author={Zhou, Zhiming and Liang, Jiadong and Song, Yuxuan and Yu, Lantao and Wang, Hongwei and Zhang, Weinan and Yu, Yong and Zhang, Zhihua},
  booktitle={International Conference on Machine Learning},
  pages={7584--7593},
  year={2019},
  organization={PMLR}
}

@article{vaswani2017attention,
  title={Attention is all you need},
  author={Vaswani, Ashish and Shazeer, Noam and Parmar, Niki and Uszkoreit, Jakob and Jones, Llion and Gomez, Aidan N and Kaiser, {\L}ukasz and Polosukhin, Illia},
  journal={Advances in neural information processing systems},
  volume={30},
  year={2017}
}

@inproceedings{ngiam2011multimodal,
  title={Multimodal deep learning},
  author={Ngiam, Jiquan and Khosla, Aditya and Kim, Mingyu and Nam, Juhan and Lee, Honglak and Ng, Andrew Y},
  booktitle={Proceedings of the 28th international conference on machine learning (ICML-11)},
  pages={689--696},
  year={2011}
}

@article{ramachandram2017deep,
  title={Deep multimodal learning: A survey on recent advances and trends},
  author={Ramachandram, Dhanesh and Taylor, Graham W},
  journal={IEEE signal processing magazine},
  volume={34},
  number={6},
  pages={96--108},
  year={2017},
  publisher={IEEE}
}

@article{liu2018towards,
  title={Towards robust human-robot collaborative manufacturing: Multimodal fusion},
  author={Liu, Hongyi and Fang, Tongtong and Zhou, Tianyu and Wang, Lihui},
  journal={IEEE Access},
  volume={6},
  pages={74762--74771},
  year={2018},
  publisher={IEEE}
}

@article{brena2020choosing,
  title={Choosing the best sensor fusion method: A machine-learning approach},
  author={Brena, Ramon F and Aguileta, Antonio A and Trejo, Luis A and Molino-Minero-Re, Erik and Mayora, Oscar},
  journal={Sensors},
  volume={20},
  number={8},
  pages={2350},
  year={2020},
  publisher={MDPI}
}

@article{ma2018deep,
  title={Deep coupling autoencoder for fault diagnosis with multimodal sensory data},
  author={Ma, Meng and Sun, Chuang and Chen, Xuefeng},
  journal={IEEE Transactions on Industrial Informatics},
  volume={14},
  number={3},
  pages={1137--1145},
  year={2018},
  publisher={IEEE}
}

@article{liu2021comparing,
  title={Comparing recognition performance and robustness of multimodal deep learning models for multimodal emotion recognition},
  author={Liu, Wei and Qiu, Jie-Lin and Zheng, Wei-Long and Lu, Bao-Liang},
  journal={IEEE Transactions on Cognitive and Developmental Systems},
  volume={14},
  number={2},
  pages={715--729},
  year={2021},
  publisher={IEEE}
}

@article{srivastava2012multimodal,
  title={Multimodal learning with deep boltzmann machines},
  author={Srivastava, Nitish and Salakhutdinov, Russ R},
  journal={Advances in neural information processing systems},
  volume={25},
  year={2012}
}

@inproceedings{yang2019aligning,
  title={Aligning latent spaces for 3d hand pose estimation},
  author={Yang, Linlin and Li, Shile and Lee, Dongheui and Yao, Angela},
  booktitle={Proceedings of the IEEE/CVF international conference on computer vision},
  pages={2335--2343},
  year={2019}
}

@article{suzuki2016joint,
  title={Joint multimodal learning with deep generative models},
  author={Suzuki, Masahiro and Nakayama, Kotaro and Matsuo, Yutaka},
  journal={arXiv preprint arXiv:1611.01891},
  year={2016}
}

@article{bouchacourt2018multi, 
title={Multi-Level Variational Autoencoder: Learning Disentangled Representations From Grouped Observations}, 
volume={32}, 
journal={Proceedings of the AAAI Conference on Artificial Intelligence}, 
author={Bouchacourt, Diane and Tomioka, Ryota and Nowozin, Sebastian}, 
year={2018}, 
month={Apr.} }

@inproceedings{kiela2014learning,
  title={Learning image embeddings using convolutional neural networks for improved multi-modal semantics},
  author={Kiela, Douwe and Bottou, L{\'e}on},
  booktitle={Proceedings of the 2014 Conference on empirical methods in natural language processing (EMNLP)},
  pages={36--45},
  year={2014}
}

@INPROCEEDINGS{9190246,
  author={Gadzicki, Konrad and Khamsehashari, Razieh and Zetzsche, Christoph},
  booktitle={2020 IEEE 23rd International Conference on Information Fusion (FUSION)}, 
  title={Early vs Late Fusion in Multimodal Convolutional Neural Networks}, 
  year={2020},
  volume={},
  number={},
  pages={1-6},
  doi={10.23919/FUSION45008.2020.9190246}}

@article{kullu2022deep,
  title={A deep-learning-based multi-modal sensor fusion approach for detection of equipment faults},
  author={Kullu, Omer and Cinar, Eyup},
  journal={Machines},
  volume={10},
  number={11},
  pages={1105},
  year={2022},
  publisher={MDPI}
}

@article{atrey2010multimodal,
  title={Multimodal fusion for multimedia analysis: a survey},
  author={Atrey, Pradeep K and Hossain, M Anwar and El Saddik, Abdulmotaleb and Kankanhalli, Mohan S},
  journal={Multimedia systems},
  volume={16},
  pages={345--379},
  year={2010},
  publisher={Springer}
}

@article{baltruvsaitis2018multimodal,
  title={Multimodal machine learning: A survey and taxonomy},
  author={Baltru{\v{s}}aitis, Tadas and Ahuja, Chaitanya and Morency, Louis-Philippe},
  journal={IEEE transactions on pattern analysis and machine intelligence},
  volume={41},
  number={2},
  pages={423--443},
  year={2018},
  publisher={IEEE}
}

@article{barua2023systematic,
  title={A Systematic Literature Review on Multimodal Machine Learning: Applications, Challenges, Gaps and Future Directions},
  author={Barua, Arnab and Ahmed, Mobyen Uddin and Begum, Shahina},
  journal={IEEE Access},
  year={2023},
  publisher={IEEE}
}

@inproceedings{altinses2023multimodal,
  title={Multimodal Synthetic Dataset Balancing: a Framework for Realistic and Balanced Training Data Generation in Industrial Settings},
  author={Altinses, Diyar and Schwung, Andreas},
  booktitle={IECON 2023-49th Annual Conference of the IEEE Industrial Electronics Society},
  pages={1--7},
  year={2023},
  organization={IEEE}
}

@article{altinses2024benchmarking,
  title={Performance Benchmarking of Multimodal Data-Driven Approaches in Industrial Settings},
  author={Altinses, Diyar and Andreas, Schwung},
  journal={Machine Learning with Applications},
  volume={},
  pages={},
  year={2025},
  publisher={Elsevier}
}

@article{virmaux2018lipschitz,
  title={Lipschitz regularity of deep neural networks: analysis and efficient estimation},
  author={Virmaux, Aladin and Scaman, Kevin},
  journal={Advances in Neural Information Processing Systems},
  volume={31},
  year={2018}
}

@article{zhang2021understanding,
  title={Understanding deep learning (still) requires rethinking generalization},
  author={Zhang, Chiyuan and Bengio, Samy and Hardt, Moritz and Recht, Benjamin and Vinyals, Oriol},
  journal={Communications of the ACM},
  volume={64},
  number={3},
  pages={107--115},
  year={2021},
  publisher={ACM New York, NY, USA}
}

@article{patel2024gradient,
  title={Gradient Descent in the Absence of Global Lipschitz Continuity of the Gradients},
  author={Patel, Vivak and Berahas, Albert S},
  journal={SIAM Journal on Mathematics of Data Science},
  volume={6},
  number={3},
  pages={602--626},
  year={2024},
  publisher={SIAM}
}

@inproceedings{mai2021stability,
  title={Stability and convergence of stochastic gradient clipping: Beyond lipschitz continuity and smoothness},
  author={Mai, Vien V and Johansson, Mikael},
  booktitle={International Conference on Machine Learning},
  pages={7325--7335},
  year={2021},
  organization={PMLR}
}

% \newpage

% \bio{images/authors/Diyar_Altinses.jpg}
% Diyar Altinses received the M.Sc. degree from the South Westphalia University of Applied Sciences, Soest, Germany, in 2021, and also holds a M.Sc. from the University of Southern Manchester, UK. He is currently pursuing a Ph.D. at South Westphalia University of Applied Sciences, where he also works as a Research Assistant in the Department of Automation Technology and Learning Systems. His research interests include deep multimodal neural networks, multimodal optimization frameworks, multimodal representation alignment, and industrial failure correction.
% \endbio

% \vspace{1cm}

% \bio{images/authors/Schwung.jpg}
% Andreas Schwung received the Ph.D. degree in electrical engineering from Technische Universität Darmstadt, Darmstadt, Germany, in 2011. From 2011 to 2015, he was a Technical Project Manager for automation and control with MAN Diesel \& Turbo SE, Oberhausen, Germany. Since 2015, he has been a professor of automation technology at the South Westphalia University of Applied Sciences, Soest, Germany. His research interests include networked automation systems, self-learning control and optimization, and intelligent data analytics with applications in manufacturing, process industry, and electromobility.
% \endbio

\newpage
\appendix
\newgeometry{left=25mm, right=25mm, top=25mm, bottom=25mm}

\section{Derivatives of multimodal autoencoder architectures}

Before we delve into the Lipschitz properties of multimodal autoencoder architectures, we first analyze the derivatives with respect to any element of the model. This results in building upon \autoref{lemma:derivative_decoder} and \autoref{lemma:derivative_encoder}.

\subsection{Proof of \autoref{lemma:derivative_decoder}}

\begin{proof}
We consider the gradient of the loss function
\begin{align}
    \mathcal{L}(\mathbf{x}) = \sum_{i=1}^M \left\| \mathbf{x}^{(i)} - D^{(i)}(\mathbf{u}) \right\|^2
\end{align}
with respect to the parameters $\theta_D^{(k)}$ of the $k$-th decoder. The model processes inputs from $M$ different modalities, where each input $\mathbf{x}^{(i)} \in \mathbb{R}^p$. For each modality $i$, there is a corresponding encoder $E^{(i)}: \mathbb{R}^p \to \mathbb{R}^{q_i}$ with parameters $\theta_E^{(i)}$, and a decoder $D^{(i)}: \mathbb{R}^{q} \to \mathbb{R}^p$ with parameters $\theta_D^{(i)}$. The latent representations produced by the individual encoders are aggregated into a single joint representation vector
\begin{align}
    \mathbf{u} = \biguplus_i\left( E^{(1)}(\mathbf{x}^{(1)}), \dotsc, E^{(M)}(\mathbf{x}^{(M)}) \right),
\end{align}
where $q$ denotes the total dimension of this aggregated vector.

To compute the gradient of $\mathcal{L}$ with respect to $\theta_D^{(k)}$, we observe that only the $k$-th term in the summation depends directly on these parameters. Additionally, the aggregated latent vector $\mathbf{u}$ does not depend on $\theta_D^{(k)}$, since it is constructed solely from the encoder outputs. Hence, the relevant partial derivative simplifies to:
\begin{align}
    \frac{\partial \mathcal{L}(\mathbf{x})}{\partial \theta_D^{(k)}} = \frac{\partial}{\partial \theta_D^{(k)}} \left\| \mathbf{x}^{(k)} - D^{(k)}(\mathbf{u}) \right\|^2.
\end{align}
Using the standard derivative rule for squared Euclidean norms, we obtain:
\begin{align}
    \frac{\partial}{\partial \theta_D^{(k)}} \left\| \mathbf{x}^{(k)} - D^{(k)}(\mathbf{u}) \right\|^2 = -2 \cdot \left( \mathbf{x}^{(k)} - D^{(k)}(\mathbf{u}) \right)^\top \cdot \frac{\partial D^{(k)}(\mathbf{u})}{\partial \theta_D^{(k)}}.
\end{align}
\end{proof}

%%%-------------------------------------------------------------------------------
%%%-------------------------------------------------------------------------------
%%%-------------------------------------------------------------------------------
%%%-------------------------------------------------------------------------------

\subsection{Proof of \autoref{lemma:derivative_encoder}}

\begin{proof}
We now derive the gradient of the loss function with respect to the encoder parameters $\theta_E^{(k)}$, following the same structure as in the decoder case. The loss function is defined as $\mathcal{L}(\mathbf{x}) = \sum_{i=1}^M \left\| \mathbf{x}^{(i)} - D^{(i)}(\mathbf{u}) \right\|^2$, where the modalities $\mathbf{x}^{(1)}, \dotsc, \mathbf{x}^{(M)} \in \mathbb{R}^p$ are encoded via $E^{(i)}(\cdot; \theta_E^{(i)}): \mathbb{R}^p \rightarrow \mathbb{R}^{q_i}$, decoded via $D^{(i)}(\cdot; \theta_D^{(i)}): \mathbb{R}^q \rightarrow \mathbb{R}^p$, and aggregated into the latent vector $\mathbf{u} = \biguplus_i(E^{(1)}(\mathbf{x}^{(1)}), \dotsc, E^{(M)}(\mathbf{x}^{(M)})) \in \mathbb{R}^q$. The goal is to compute $\frac{\partial \mathcal{L}(\mathbf{x})}{\partial \theta_E^{(k)}}$.

Since the loss function depends on the latent vector $\mathbf{u}$, which in turn depends on all encoders, the parameters of encoder $k$ influence the reconstruction loss for all modalities. Therefore, we expand the derivative across all terms:
\begin{align}
    \frac{\partial \mathcal{L}(\mathbf{x})}{\partial \theta_E^{(k)}}& = \sum_{i=1}^M \frac{\partial}{\partial \theta_E^{(k)}} \left\| \mathbf{x}^{(i)} - D^{(i)}(\mathbf{u}) \right\|^2\\
    &= \sum_{i=1}^M \frac{\partial}{\partial \theta_E^{(k)}} \left( \left( \mathbf{x}^{(i)} - D^{(i)}(\mathbf{u}) \right)^\top \left( \mathbf{x}^{(i)} - D^{(i)}(\mathbf{u}) \right) \right) 
\end{align}
As $\mathbf{u}$, and hence $D^{(i)}(\mathbf{u})$, depends on $\theta_E^{(k)}$, we apply the chain rule:
\begin{align}
    \frac{\partial}{\partial \theta_E^{(k)}} \left\| \mathbf{x}^{(i)} - D^{(i)}(\mathbf{u}) \right\|^2 = -2 \cdot \left( \mathbf{x}^{(i)} - D^{(i)}(\mathbf{u}) \right)^\top \cdot \frac{\partial D^{(i)}(\mathbf{u})}{\partial \mathbf{u}} \cdot \frac{\partial \mathbf{u}}{\partial \theta_E^{(k)}}
\end{align}
This leads to the overall expression:
\begin{align}
    \frac{\partial \mathcal{L}(\mathbf{x})}{\partial \theta_E^{(k)}} = -2 \sum_{i=1}^M \left( \mathbf{x}^{(i)} - D^{(i)}(\mathbf{u}) \right)^\top \cdot \frac{\partial D^{(i)}(\mathbf{u})}{\partial \mathbf{u}} \cdot \frac{\partial \mathbf{u}}{\partial \theta_E^{(k)}}
\end{align}
where, $\frac{\partial D^{(i)}(\mathbf{u})}{\partial \mathbf{u}}$ is the derivative of the decoder function with respect to the latent representation $\mathbf{u}$ and $\frac{\partial \mathbf{u}}{\partial \theta_E^{(k)}}$ is the derivative of the aggregation function with respect to the encoder parameters $\theta_E^{(k)}$.

\end{proof}

%%%-------------------------------------------------------------------------------
%%%-------------------------------------------------------------------------------
%%%-------------------------------------------------------------------------------
%%%-------------------------------------------------------------------------------

\subsection{Proof of \autoref{lemma:derivative_concat}}

\begin{proof}
   The concatenation can be written as:  
   \begin{align}
       \mathbf{u} = \begin{bmatrix} E^{(1)}(\mathbf{x}^{(1)}) \\ \vdots \\ E^{(k)}(\mathbf{x}^{(k)}) \\ \vdots \\ E^{(M)}(\mathbf{x}^{(M)}) \end{bmatrix}, \quad \mathbf{u} \in \mathbb{R}^{\sum_{n=1}^M D_n}.
   \end{align}    
   The derivative \(\frac{\partial \mathbf{u}}{\partial \theta_E^{(k)}}\) is a block matrix where only the \(k\)-th block depends on \(\theta_E^{(k)}\):  
   \begin{align}
       \frac{\partial \mathbf{u}}{\partial \theta_E^{(k)}} = \begin{bmatrix} \frac{\partial E^{(1)}}{\partial \theta_E^{(k)}} \\ \vdots \\ \frac{\partial E^{(k)}}{\partial \theta_E^{(k)}} \\ \vdots \\ \frac{\partial E^{(M)}}{\partial \theta_E^{(k)}} \end{bmatrix}.
   \end{align}  
   For all \(n \neq k\), \(E^{(n)}\) does not depend on \(\theta_E^{(k)}\), so \(\frac{\partial E^{(n)}}{\partial \theta_E^{(k)}} = \mathbf{0}_{D_n \times |\theta_E^{(k)}|}\).   
   \begin{align}
       \frac{\partial \mathbf{u}}{\partial \theta_E^{(k)}} = \begin{bmatrix} \mathbf{0}_{D_1} \\ \vdots \\ \frac{\partial E^{(k)}}{\partial \theta_E^{(k)}} \\ \vdots \\ \mathbf{0}_{D_M} \end{bmatrix},
   \end{align}
   which matches the stated lemma.  

Dimensionality Verification:
\begin{enumerate}
    \item Output \(\mathbf{u}\): Dimension \(\sum_{n=1}^M D_n\).  
    \item Parameter \(\theta_E^{(k)}\): Dimension \(|\theta_E^{(k)}|\).  
    \item Jacobian \(\frac{\partial \mathbf{u}}{\partial \theta_E^{(k)}}\): Shape \(\left(\sum_{n=1}^M D_n\right) \times |\theta_E^{(k)}|\), with non-zero block \(D_k \times |\theta_E^{(k)}|\).  
\end{enumerate}

The derivative is a block-sparse matrix where only the \(k\)-th encoder’s Jacobian appears, analogous to the sum case but preserving concatenation structure.  
\end{proof}

%%%-------------------------------------------------------------------------------
%%%-------------------------------------------------------------------------------
%%%-------------------------------------------------------------------------------
%%%-------------------------------------------------------------------------------

\section{Lipschitz continuity of multimodal autoencoder architectures}

In this section, we provide a detailed mathematical proof for the Lipschitz constants. As with the derivative analysis, we begin with the decoder case.

\subsection{Proof of \autoref{theorem:lipschitz_multimodal_autoencoder_D}}

\begin{proof}
The objective is to determine a constant $L_D^{(k)} \in \mathbb{R}_+$ such that for any pair of inputs $\mathbf{x}_1$ and $\mathbf{x}_2$, the following Lipschitz condition holds for the gradient of the loss function with respect to the decoder parameters $\theta_D^{(k)}$:
\begin{align}
    \left\| \frac{\partial \mathcal{L}(\mathbf{x}_1)}{\partial \theta_D^{(k)}} - \frac{\partial \mathcal{L}(\mathbf{x}_2)}{\partial \theta_D^{(k)}} \right\| \leq L_D^{(k)} \cdot \left\| \mathbf{x}_1 - \mathbf{x}_2 \right\|
\end{align}

To analyze this, we start by expressing the difference between the gradients for the two inputs:
\begin{align}
\Delta &= \frac{\partial \mathcal{L}(\mathbf{x}_1)}{\partial \theta_D^{(k)}} - \frac{\partial \mathcal{L}(\mathbf{x}_2)}{\partial \theta_D^{(k)}} \\
&= -2 \cdot \left[ \left( \mathbf{x}_1^{(k)} - D^{(k)}(\mathbf{u}_1) \right)^\top \cdot \frac{\partial D^{(k)}(\mathbf{u}_1)}{\partial \theta_D^{(k)}} - \left( \mathbf{x}_2^{(k)} - D^{(k)}(\mathbf{u}_2) \right)^\top \cdot \frac{\partial D^{(k)}(\mathbf{u}_2)}{\partial \theta_D^{(k)}} \right]
\end{align}

This difference can be decomposed into three parts using intermediate terms:
\begin{align}
    \Delta &= -2 \cdot \left[ A - B \right], \text{ mit:}\\
    A &= \left( \mathbf{x}_1^{(k)} - D^{(k)}(\mathbf{u}_1) \right)^\top \cdot \frac{\partial D^{(k)}(\mathbf{u}_1)}{\partial \theta_D^{(k)}}\\
    B &= \left( \mathbf{x}_2^{(k)} - D^{(k)}(\mathbf{u}_2) \right)^\top \cdot \frac{\partial D^{(k)}(\mathbf{u}_2)}{\partial \theta_D^{(k)}}
\end{align}

By applying the triangle inequality, the norm of the gradient difference can be bounded as:
\begin{align}
    \left\| \Delta \right\| \leq 2 \cdot  \underbrace{\left\| \left( \mathbf{x}_1^{(k)} - \mathbf{x}_2^{(k)} \right)^\top \cdot \frac{\partial D^{(k)}(\mathbf{u}_1)}{\partial \theta_D^{(k)}} \right\|}_{(I)}
    &+ 2 \cdot\underbrace{\left\| \left( D^{(k)}(\mathbf{u}_1) - D^{(k)}(\mathbf{u}_2) \right)^\top \cdot \frac{\partial D^{(k)}(\mathbf{u}_1)}{\partial \theta_D^{(k)}} \right\|}_{(II)}\\
    &+ 2 \cdot\underbrace{\left\| \left( \mathbf{x}_2^{(k)} - D^{(k)}(\mathbf{u}_2) \right)^\top \cdot \left( \frac{\partial D^{(k)}(\mathbf{u}_1)}{\partial \theta_D^{(k)}} - \frac{\partial D^{(k)}(\mathbf{u}_2)}{\partial \theta_D^{(k)}} \right) \right\|}_{(III)} \nonumber
\end{align}

To further bound these terms, we make the following assumptions:
\begin{enumerate}
    \item $\left\| \frac{\partial D^{(k)}(\mathbf{u})}{\partial \theta_D^{(k)}} \right\| \leq B_{\text{grad}}^{(k)}$:  ensures the gradients of the decoder are uniformly bounded.
    \item $\left\| D^{(k)}(\mathbf{u}_1) - D^{(k)}(\mathbf{u}_2) \right\| \leq L_{D^{(k)}}^{(\text{func})} \cdot \left\| \mathbf{u}_1 - \mathbf{u}_2 \right\|$: Lipschitz continuity of the decoder function
    \item $\left\| \frac{\partial D^{(k)}(\mathbf{u}_1)}{\partial \theta_D^{(k)}} - \frac{\partial D^{(k)}(\mathbf{u}_2)}{\partial \theta_D^{(k)}} \right\| \leq L_{D^{(k)}}^{(\text{grad})} \cdot \left\| \mathbf{u}_1 - \mathbf{u}_2 \right\|$: Lipschitz continuity of the decoder gradients
    \item $\left\| \mathbf{x}^{(k)} \right\| \leq C$: Bounded input norm
\end{enumerate}

Since the latent representation $\mathbf{u}$ is obtained by aggregating the outputs of the encoders, and each encoder is Lipschitz-continuous, the aggregation itself is also Lipschitz-continuous. Letting
\begin{align}
    L_E := \sup_{\mathbf{x}_1, \mathbf{x}_2} \frac{\|\mathbf{u}_1 - \mathbf{u}_2\|}{\|\mathbf{x}_1 - \mathbf{x}_2\|},\\
    \left\| \mathbf{u}_1 - \mathbf{u}_2 \right\| \leq L_{\text{agg}} \cdot L_{E^{(i)}}^{(\text{func})} \cdot \left\| \mathbf{x}_1^{(i)} - \mathbf{x}_2^{(i)} \right\|,
\end{align}

Finally, combining all parts yields a Lipschitz constant for the gradient:
\begin{align}
    L_D^{(k)} = 2 \cdot \left[ B_{\text{grad}}^{(k)} + B_{\text{grad}}^{(k)} \cdot L_{D^{(k)}}^{(\text{func})} + C \cdot L_{D^{(k)}}^{(\text{grad})} \right] \cdot L_{\text{agg}} \cdot L_{E^{(i)}}^{(\text{func})}
\end{align}
where $B_{\text{grad}}^{(k)}$ is an upper bound on the decoder gradient norm, $L_{D^{(k)}}^{(\text{func})} = \prod_{\ell=1}^{L_D^{(k)}} \| W_\ell^{(k)} \| \cdot L_{\sigma_\ell}$ is the Lipschitz constant of the decoder function, $L_{D^{(k)}}^{(\text{grad})}$ is the Lipschitz constant of the decoder gradient, $C$ is a bound on the input norm $\| \mathbf{x}^{(k)} \|$, $L_{\text{agg}}$ is the Lipschitz constant of the aggregation method (e.g., concatenation, averaging), and $L_{E^{(i)}}^{(\text{func})} = \prod_{\ell=1}^{L_E^{(i)}} \| W_\ell^{(i)} \| \cdot L_{\sigma_\ell}$ is the Lipschitz constant of encoder $i$.
\end{proof}

%%%-------------------------------------------------------------------------------
%%%-------------------------------------------------------------------------------
%%%-------------------------------------------------------------------------------
%%%-------------------------------------------------------------------------------

\subsection{Proof of \autoref{theorem:lipschitz_multimodal_autoencoder_E}}

\begin{proof}
We want to show that the gradient mapping \( \mathbf{x} \mapsto \frac{\partial \mathcal{L}(\mathbf{x})}{\partial \theta_E^{(k)}} \) is Lipschitz continuous, i.e., that there exists a constant \( L_E^{(k)} > 0 \) such that
\begin{align}
    \left\| \frac{\partial \mathcal{L}(\mathbf{x}_1)}{\partial \theta_E^{(k)}} - \frac{\partial \mathcal{L}(\mathbf{x}_2)}{\partial \theta_E^{(k)}} \right\| \leq L_E^{(k)} \cdot \| \mathbf{x}_1 - \mathbf{x}_2 \|
\end{align}
for all \( \mathbf{x}_1, \mathbf{x}_2 \in \mathbb{R}^{Mp} \).

Let us define
\begin{align}
    \Delta := \frac{\partial \mathcal{L}(\mathbf{x}_1)}{\partial \theta_E^{(k)}} - \frac{\partial \mathcal{L}(\mathbf{x}_2)}{\partial \theta_E^{(k)}}.
\end{align}

Using the expression for the gradient:
\begin{align}
\Delta
&= -2 \sum_{i=1}^M \Big[
\left( \mathbf{x}_1^{(i)} - D^{(i)}(\mathbf{u}_1) \right)^\top \cdot \frac{\partial D^{(i)}(\mathbf{u}_1)}{\partial \mathbf{u}} \cdot \frac{\partial \mathbf{u}_1}{\partial \theta_E^{(k)}} \\
&\quad - \left( \mathbf{x}_2^{(i)} - D^{(i)}(\mathbf{u}_2) \right)^\top \cdot \frac{\partial D^{(i)}(\mathbf{u}_2)}{\partial \mathbf{u}} \cdot \frac{\partial \mathbf{u}_2}{\partial \theta_E^{(k)}}
\Big].
\end{align}

We add and subtract intermediate terms to make the difference explicit:
\begin{align}
\Delta
&= -2 \sum_{i=1}^M \Big[
\underbrace{\left( \mathbf{x}_1^{(i)} - \mathbf{x}_2^{(i)} \right)^\top \cdot \frac{\partial D^{(i)}(\mathbf{u}_1)}{\partial \mathbf{u}} \cdot \frac{\partial \mathbf{u}_1}{\partial \theta_E^{(k)}}}_{(A)} \\
&\quad + \underbrace{\left( D^{(i)}(\mathbf{u}_2) - D^{(i)}(\mathbf{u}_1) \right)^\top \cdot \frac{\partial D^{(i)}(\mathbf{u}_1)}{\partial \mathbf{u}} \cdot \frac{\partial \mathbf{u}_1}{\partial \theta_E^{(k)}}}_{(B)} \\
&\quad + \underbrace{\left( \mathbf{x}_2^{(i)} - D^{(i)}(\mathbf{u}_2) \right)^\top \cdot \left( \frac{\partial D^{(i)}(\mathbf{u}_1)}{\partial \mathbf{u}} - \frac{\partial D^{(i)}(\mathbf{u}_2)}{\partial \mathbf{u}} \right) \cdot \frac{\partial \mathbf{u}_1}{\partial \theta_E^{(k)}}}_{(C)} \\
&\quad + \underbrace{\left( \mathbf{x}_2^{(i)} - D^{(i)}(\mathbf{u}_2) \right)^\top \cdot \frac{\partial D^{(i)}(\mathbf{u}_2)}{\partial \mathbf{u}} \cdot \left( \frac{\partial \mathbf{u}_1}{\partial \theta_E^{(k)}} - \frac{\partial \mathbf{u}_2}{\partial \theta_E^{(k)}} \right)}_{(D)}
\Big].
\end{align}

Each of these quantities can be upper-bounded in terms of the distance \( \| \mathbf{x}_1 - \mathbf{x}_2 \| \) by assuming Lipschitz continuity of the respective components. Assuming the following bounds:
\begin{enumerate}
    \item $\left\| \frac{\partial D^{(i)}(\mathbf{u})}{\partial \mathbf{u}} \right\| \leq L_{D^{(i)}}^{(\text{func})}$: The Lipschitz constant of the decoder function $D^{(i)}$ with respect to its input $\mathbf{u}$. This measures how sensitively the output of decoder $i$ changes as its input latent representation changes.
    \item $\left\| D^{(i)}(\mathbf{u}_1) - D^{(i)}(\mathbf{u}_2) \right\| \leq L_{D^{(i)}}^{(\text{func})} \cdot \| \mathbf{u}_1 - \mathbf{u}_2 \|$
    \item $\left\| \frac{\partial D^{(i)}(\mathbf{u}_1)}{\partial \mathbf{u}} - \frac{\partial D^{(i)}(\mathbf{u}_2)}{\partial \mathbf{u}} \right\| \leq L_{D^{(i)}}^{(\text{grad})} \cdot \| \mathbf{u}_1 - \mathbf{u}_2 \|$: The Lipschitz constant of the gradients of the decoder $D^{(i)}$ with respect to its input.
    \item $\left\| \frac{\partial \mathbf{u}_1}{\partial \theta_E^{(k)}} \right\| \leq B_{\text{agg}}^{(k)}$: An upper bound on the norm of the derivative of the latent code $\mathbf{u}$ with respect to the $k$-th encoder parameter
    \item $\left\| \frac{\partial \mathbf{u}_1}{\partial \theta_E^{(k)}} - \frac{\partial \mathbf{u}_2}{\partial \theta_E^{(k)}} \right\| \leq L_{\text{agg}}^{(\text{grad}, k)} \cdot \left\| \mathbf{x}_1 - \mathbf{x}_2 \right\|$: The Lipschitz constant of the encoder's aggregation function
    \item Bounded input norm: $\| \mathbf{x}^{(i)} \| \leq C$
\end{enumerate}

Putting all together, we obtain the bound:
\begin{align}
    L_E^{(k)} = 2 \sum_{i=1}^M \left[L_{D^{(i)}}^{(\text{func})} \cdot B_{\text{agg}}^{(k)} +\left( L_{D^{(i)}}^{(\text{func})} \right)^2 \cdot B_{\text{agg}}^{(k)} \cdot L_{\text{agg}}^{(\text{func})} + C \cdot L_{D^{(i)}}^{(\text{grad})} \cdot B_{\text{agg}}^{(k)} + C \cdot L_{D^{(i)}}^{(\text{func})} \cdot L_{\text{agg}}^{(\text{grad},k)} \right]
\end{align}

This defines the Lipschitz constant \( L_E^{(k)} \) for the gradient of \( \mathcal{L} \) with respect to the encoder parameters \( \theta_E^{(k)} \). Hence, there exists a constant \( L_E^{(k)} > 0 \) such that $\left\| \frac{\partial \mathcal{L}(\mathbf{x}_1)}{\partial \theta_E^{(k)}} - \frac{\partial \mathcal{L}(\mathbf{x}_2)}{\partial \theta_E^{(k)}} \right\| \leq L_E^{(k)} \cdot \| \mathbf{x}_1 - \mathbf{x}_2 \|.$
\end{proof}

%%%-------------------------------------------------------------------------------
%%%-------------------------------------------------------------------------------
%%%-------------------------------------------------------------------------------
%%%-------------------------------------------------------------------------------

\subsection{Proof of \autoref{theorem: lipschitz_aggregation}}

\begin{proof}

Let us consider $n$ encoder functions $E^{(i)} : \mathbb{R}^d \to \mathbb{R}^{m_i}$, each composed of $L_E^{(i)}$ layers. The functional Lipschitz constant of each encoder is given by
\begin{align}
    L_{E^{(i)}}^{(\text{func})} = \prod_{\ell=1}^{L_E^{(i)}} \| W_\ell^{(i)} \| \cdot L_{\sigma_\ell},
\end{align}
where $\| W_\ell^{(i)} \|$ denotes the operator norm (e.g., spectral norm) of the weight matrix at layer $\ell$, and $L_{\sigma_\ell}$ is the Lipschitz constant of the activation function used at that layer.

We are interested in comparing the overall Lipschitz constants of two compositions of these encoders.
\begin{enumerate}
    \item Concatenation: Define the function $E_{\text{concat}} : \mathbb{R}^d \to \mathbb{R}^{\sum_i m_i}$ by $E_{\text{concat}}(x) = [E^{(1)}(x), E^{(2)}(x), \dots, E^{(n)}(x)].$
    \item Summation: Assume for simplicity that each encoder has the same output dimensionality $m$, and define the function $E_{\text{sum}} : \mathbb{R}^d \to \mathbb{R}^m$ by $E_{\text{sum}}(x) = \sum_{i=1}^n E^{(i)}(x).$
\end{enumerate}

The concatenation of vector-valued functions corresponds to a function whose Lipschitz constant is the Euclidean norm of the individual Lipschitz constants. That is,
\begin{align}
    L_{\text{concat}} = \left( \sum_{i=1}^n \left( L_{E^{(i)}}^{(\text{func})} \right)^2 \right)^{1/2}.
\end{align}

By the subadditivity of the Lipschitz constant over sums of functions, we have
\begin{align}
    L_{\text{sum}} = \left\| \sum_{i=1}^n E^{(i)} \right\|_{\text{Lip}} \leq \sum_{i=1}^n \left\| E^{(i)} \right\|_{\text{Lip}} = \sum_{i=1}^n L_{E^{(i)}}^{(\text{func})}.
\end{align}

To compare the two, note that by the Cauchy–Schwarz inequality:
\begin{align}
    \left( \sum_{i=1}^n \left( L_{E^{(i)}}^{(\text{func})} \right)^2 \right)^{1/2} \leq \sum_{i=1}^n L_{E^{(i)}}^{(\text{func})}.
\end{align}
Therefore,
\begin{align}
    L_{\text{concat}} \leq L_{\text{sum}}.
\end{align}
Equality holds if and only if all but one of the Lipschitz constants $L_{E^{(i)}}^{(\text{func})}$ are zero, which is a degenerate case.

\end{proof}

The Lipschitz constant characterizes the worst-case sensitivity of a function to perturbations in its parameters. A higher Lipschitz constant implies greater variability in the output in response to small parameter changes, which can affect optimization stability and generalization. The concatenation structure provides better local stability, as each block is isolated and contributes independently. The summation structure, by aggregating multiple contributions into a single output, leads to accumulated sensitivity. This can amplify noise or parameter drift, particularly when the number of terms $M$ is large or when the norms $\| \mathbf{x}_k \|$ vary widely. Thus, in contexts such as neural network architecture design or structured regression, concatenated outputs may offer better Lipschitz control, which is favorable for robustness, especially in adversarial or regularized settings.

\section{Lipschitz Analysis of Multimodal Attention}

In this section, we analyze the mathematical properties of multimodal attention, focusing on its differentiability and Lipschitz continuity. We begin by examining the derivatives of the attention mechanism, which provide insights into its sensitivity to input variations. Next, we investigate the Lipschitz constant of the attention function, which quantifies its stability under perturbations. Finally, we derive the Lipschitz constant of the attention gradients, ensuring robustness in optimization and training dynamics.

%%%-------------------------------------------------------------------------------
%%%-------------------------------------------------------------------------------
%%%-------------------------------------------------------------------------------
%%%-------------------------------------------------------------------------------

\subsection{Proof of \autoref{lemma:attention_gradients}}

\begin{proof}
Fix \(k\in\{1,\dots,n\}\). For each output block \(z_i:=\alpha_i\mathbf v_i\in\mathbb{R}^d\) we compute the partial derivative with respect to \(\mathbf v_k\). By the product rule (chain rule),
\[
\frac{\partial z_i}{\partial \mathbf v_k}
= \frac{\partial(\alpha_i\mathbf v_i)}{\partial\mathbf v_k}
= \alpha_i \frac{\partial \mathbf v_i}{\partial\mathbf v_k} \;+\; \left(\frac{\partial \alpha_i}{\partial\mathbf v_k}\right)\mathbf v_i^\top,
\]
where \(\dfrac{\partial \mathbf v_i}{\partial\mathbf v_k}\) denotes the matrix derivative of the vector \(\mathbf v_i\) with respect to \(\mathbf v_k\). Thus:
\begin{enumerate}
    \item If \(i=k\), \(\dfrac{\partial \mathbf v_i}{\partial\mathbf v_k}=I_d\), so the first term becomes \(\alpha_i I_d\).
    \item If \(i\ne k\), \(\dfrac{\partial \mathbf v_i}{\partial\mathbf v_k}=0\), and the first term vanishes.
\end{enumerate}
Hence for all \(i\),
\[
B_{i,k} = \alpha_i\delta_{ik} I_d \;+\; \Big(\frac{\partial\alpha_i}{\partial\mathbf v_k}\Big)\mathbf v_i^\top,
\]
where \(\delta_{ik}\) is the Kronecker delta.

It remains to compute \(\dfrac{\partial\alpha_i}{\partial\mathbf v_k}\). By definition \(\alpha_i=\tfrac{1}{n-1}\sum_{j\ne i} a_{ij}\) with \(a_{ij}=(\mathbf W_i\mathbf v_i)^\top(\mathbf W_j\mathbf v_j)\). Differentiate \(a_{ij}\) with respect to \(\mathbf v_k\). There are three cases:

\begin{enumerate}
  \item If \(k=i\), then for each \(j\ne i\),
  \[
  \frac{\partial a_{ij}}{\partial\mathbf v_i}
  = \frac{\partial}{\partial\mathbf v_i}\big((\mathbf W_i\mathbf v_i)^\top(\mathbf W_j\mathbf v_j)\big)
  = (\mathbf W_i)^\top(\mathbf W_j\mathbf v_j),
  \]
  since \(\mathbf v_j\) is independent of \(\mathbf v_i\). Summing over \(j\ne i\) and dividing by \(n-1\) yields
  \[
  \frac{\partial\alpha_i}{\partial\mathbf v_i}
  = \frac{1}{n-1}\sum_{\substack{j=1\\j\ne i}}^n (\mathbf W_i)^\top(\mathbf W_j\mathbf v_j).
  \]
  \item If \(k\ne i\), then only the single term \(a_{ik}\) depends on \(\mathbf v_k\). For that term
  \[
  \frac{\partial a_{ik}}{\partial\mathbf v_k}
  = \frac{\partial}{\partial\mathbf v_k}\big((\mathbf W_i\mathbf v_i)^\top(\mathbf W_k\mathbf v_k)\big)
  = (\mathbf W_k)^\top(\mathbf W_i\mathbf v_i).
  \]
  Therefore,
  \[
  \frac{\partial\alpha_i}{\partial\mathbf v_k}
  = \frac{1}{n-1}(\mathbf W_k)^\top(\mathbf W_i\mathbf v_i).
  \]
  \item If \(k\) is neither \(i\) nor equal to any index appearing in a term (not possible here), the derivative would be zero, but the two cases above exhaust all possibilities for \(a_{ij}\).
\end{enumerate}

Substituting these expressions for \(\partial\alpha_i/\partial\mathbf v_k\) into
\(B_{i,k} = \alpha_i\delta_{ik} I_d + (\partial\alpha_i/\partial\mathbf v_k)\mathbf v_i^\top\)
yields exactly the stated block formulas:
\[
B_{i,k} =
\begin{cases}
\alpha_i I_d \;+\; \Big(\dfrac{1}{n-1}\sum_{j\ne i} (\mathbf W_i)^\top(\mathbf W_j\mathbf v_j)\Big)\mathbf v_i^\top,
& i=k,\\[8pt]
\dfrac{1}{n-1}\big((\mathbf W_k)^\top(\mathbf W_i\mathbf v_i)\big)\mathbf v_i^\top,
& i\ne k.
\end{cases}
\]
Stacking the blocks \(B_{i,k}\) for \(i=1,\dots,n\) gives the full Jacobian \(\dfrac{\partial\Phi}{\partial\mathbf v_k}\). This concludes the proof. %Finally, when \(n=2\) the averages have denominator \(n-1=1\) and the sums reduce to the single pairwise term, so the block formulas reduce to the two-block Jacobians stated in the bimodal lemma.
\end{proof}

%%%-------------------------------------------------------------------------------
%%%-------------------------------------------------------------------------------
%%%-------------------------------------------------------------------------------
%%%-------------------------------------------------------------------------------

\subsection{Proof of \autoref{theorem:lipschitz_attention}}

\begin{proof}
    Fix two tuples \(\mathbf{V}=(\mathbf{v}_1,\dots,\mathbf{v}_n)\) and
    \(\tilde{\mathbf{V}}=(\tilde{\mathbf{v}}_1,\dots,\tilde{\mathbf{v}}_n)\) in \(\mathcal K\).
    For brevity denote \(\Delta \mathbf{v}_i := \mathbf{v}_i-\tilde{\mathbf{v}}_i\) and
    \(\Delta a_{ij} := a_{ij}(\mathbf{V}) - a_{ij}(\tilde{\mathbf{V}})\),
    \(\Delta \alpha_i := \alpha_i(\mathbf{V}) - \alpha_i(\tilde{\mathbf{V}})\).
    Also write \(S := \sum_{k=1}^n \|\Delta\mathbf{v}_k\|\) and recall
    \( \|\mathbf{v}_i\|,\|\tilde{\mathbf{v}}_i\| \le R \) for all \(i\).

    \paragraph{Step 1: elementary bounds on magnitudes.}
    For any \(i,j\) we have
    \[
        |a_{ij}(\mathbf{V})| = |(\mathbf{W}_i\mathbf{v}_i)^\top(\mathbf{W}_j\mathbf{v}_j)|
        \le \|\mathbf{W}_i\|\,\|\mathbf{v}_i\| \;\|\mathbf{W}_j\|\,\|\mathbf{v}_j\|
        \le M^2 R^2.
    \]
    Hence \(|\alpha_i|\le M^2 R^2\) for every \(i\).

    \paragraph{Step 2: bound the change of pairwise scores.}
    Observe the exact identity
    \[
        \Delta a_{ij}
        = (\mathbf{W}_i\mathbf{v}_i)^\top(\mathbf{W}_j\mathbf{v}_j) - (\mathbf{W}_i\tilde{\mathbf{v}}_i)^\top(\mathbf{W}_j\tilde{\mathbf{v}}_j)
        = (\mathbf{W}_i\Delta\mathbf{v}_i)^\top(\mathbf{W}_j\mathbf{v}_j)
          + (\mathbf{W}_i\tilde{\mathbf{v}}_i)^\top(\mathbf{W}_j\Delta\mathbf{v}_j).
    \]
    Taking norms and using operator norm bounds gives
    \[
        |\Delta a_{ij}|
        \le \|\mathbf{W}_i\Delta\mathbf{v}_i\|\,\|\mathbf{W}_j\mathbf{v}_j\|
           + \|\mathbf{W}_i\tilde{\mathbf{v}}_i\|\,\|\mathbf{W}_j\Delta\mathbf{v}_j\|
        \le M^2 R \,\|\Delta\mathbf{v}_i\| + M^2 R \,\|\Delta\mathbf{v}_j\|.
    \]

    \paragraph{Step 3: bound the change of averaged coefficients.}
    Recall \(\alpha_i = \frac{1}{n-1}\sum_{j\ne i} a_{ij}\). Therefore
    \[
        |\Delta\alpha_i|
        \le \frac{1}{n-1} \sum_{j\ne i} |\Delta a_{ij}|
        \le \frac{1}{n-1}\sum_{j\ne i} \big( M^2 R \,\|\Delta\mathbf{v}_i\| + M^2 R \,\|\Delta\mathbf{v}_j\| \big).
    \]
    Summing the first term over \(j\ne i\) yields \((n-1) M^2 R \|\Delta\mathbf{v}_i\|\); summing the second term gives \(M^2 R \sum_{j\ne i}\|\Delta\mathbf{v}_j\|\). Thus
    \[
        |\Delta\alpha_i| \le M^2 R \,\|\Delta\mathbf{v}_i\| + \frac{M^2 R}{n-1}\sum_{j\ne i}\|\Delta\mathbf{v}_j\|
        \le M^2 R \,\|\Delta\mathbf{v}_i\| + \frac{M^2 R}{n-1} S.
    \]

    \paragraph{Step 4: perturbation of each fused block.}
    For each modality \(i\) the fused output block is \(z_i := \alpha_i \mathbf{v}_i\).
    Hence
    \[
        \Delta z_i = \alpha_i \Delta\mathbf{v}_i + (\Delta\alpha_i)\tilde{\mathbf{v}}_i,
    \]
    and thus, using the bounds \(|\alpha_i|\le M^2 R^2\) and \(\|\tilde{\mathbf{v}}_i\|\le R\),
    \begin{align*}
        \|\Delta z_i\|
        &\le |\alpha_i| \,\|\Delta\mathbf{v}_i\| + |\Delta\alpha_i| \,\|\tilde{\mathbf{v}}_i\| \\
        &\le M^2 R^2 \|\Delta\mathbf{v}_i\|
           + R\Big( M^2 R \,\|\Delta\mathbf{v}_i\| + \frac{M^2 R}{n-1} S \Big) \\
        &= M^2 R^2 \big(2\|\Delta\mathbf{v}_i\| + \tfrac{1}{n-1} S\big).
    \end{align*}

    \paragraph{Step 5: combine the block bounds to a global bound.}
    Let \(s_i := \|\Delta\mathbf{v}_i\|\). Then by the previous inequality
    \[
        \|\Delta z_i\| \le M^2 R^2\big(2 s_i + \tfrac{1}{n-1} S\big).
    \]
    Squaring and summing over \(i\) gives
    \[
        \|\Phi(\mathbf{V})-\Phi(\tilde{\mathbf{V}})\|^2
        = \sum_{i=1}^n \|\Delta z_i\|^2
        \le M^4 R^4 \sum_{i=1}^n \big(2 s_i + \tfrac{1}{n-1}S\big)^2.
    \]
    Use the elementary inequality \((a+b)^2 \le 2a^2 + 2b^2\) to obtain
    \[
        \sum_{i=1}^n \big(2 s_i + \tfrac{1}{n-1}S\big)^2
        \le 2\sum_{i=1}^n (2s_i)^2 + 2\sum_{i=1}^n \big(\tfrac{1}{n-1}S\big)^2
        = 8\sum_{i=1}^n s_i^2 + 2n\frac{S^2}{(n-1)^2}.
    \]
    Since \(S^2 = (\sum_i s_i)^2 \le n\sum_i s_i^2\) we have
    \[
        2n\frac{S^2}{(n-1)^2} \le 2n\frac{n\sum_i s_i^2}{(n-1)^2}
        = 2\frac{n^2}{(n-1)^2}\sum_{i=1}^n s_i^2.
    \]
    For \(n\ge 2\) the factor \(\dfrac{n^2}{(n-1)^2}\le 4\) (the maximum occurs at \(n=2\)), hence
    \[
        \sum_{i=1}^n \big(2 s_i + \tfrac{1}{n-1}S\big)^2
        \le \Big(8 + 2\cdot 4\Big)\sum_{i=1}^n s_i^2
        = 16\sum_{i=1}^n s_i^2.
    \]
    Therefore
    \[
        \|\Phi(\mathbf{V})-\Phi(\tilde{\mathbf{V}})\|^2
        \le M^4 R^4 \cdot 16 \sum_{i=1}^n s_i^2
        = (4 M^2 R^2)^2 \big\|(\Delta\mathbf{v}_1,\dots,\Delta\mathbf{v}_n)\big\|^2.
    \]
    Taking square roots yields the stated Lipschitz bound
    \[
        \|\Phi(\mathbf{V})-\Phi(\tilde{\mathbf{V}})\|
        \le 4\,M^2 R^2 \,\big\|(\Delta\mathbf{v}_1,\dots,\Delta\mathbf{v}_n)\big\|.
    \]
    This completes the proof.
\end{proof}

%%%-------------------------------------------------------------------------------
%%%-------------------------------------------------------------------------------
%%%-------------------------------------------------------------------------------
%%%-------------------------------------------------------------------------------

\subsection{Proof of \autoref{theorem:attention_lipschitz_gradients}}

\begin{proof}
From \autoref{lemma:attention_gradients}, each Jacobian block of $\Phi$ has the form
\[
B_{i,k} = 
\begin{cases}
\displaystyle
\alpha_i I_d + \Big(\frac{1}{n-1}\sum_{j\neq i} (\mathbf{W}_i)^\top (\mathbf{W}_j \mathbf{v}_j)\Big)\mathbf{v}_i^\top, & i=k,\\[10pt]
\displaystyle
\frac{1}{n-1}\big((\mathbf{W}_k)^\top (\mathbf{W}_i \mathbf{v}_i)\big)\mathbf{v}_i^\top, & i \neq k.
\end{cases}
\]
Differentiating $B_{i,k}$ with respect to any input vector $\mathbf{v}_\ell$ yields second-order terms that involve products of the weight matrices and one input vector. The highest-order dependency arises from terms where the derivative acts on $\mathbf{v}_j$ or $\mathbf{v}_i$ inside the bilinear products. For instance,
\[
\frac{\partial}{\partial \mathbf{v}_\ell}\Big((\mathbf{W}_i)^\top (\mathbf{W}_j \mathbf{v}_j)\mathbf{v}_i^\top\Big)
\;\sim\;
(\mathbf{W}_i)^\top \mathbf{W}_j \, \mathbf{v}_i^\top
\quad \text{or} \quad
(\mathbf{W}_i)^\top (\mathbf{W}_j \mathbf{v}_j) I_d,
\]
each of which has an operator norm bounded by
\[
\big\|(\mathbf{W}_i)^\top \mathbf{W}_j\big\|\,\|\mathbf{v}_i\|
\;\leq\;
\|\mathbf{W}_i\|\,\|\mathbf{W}_j\|\,\|\mathbf{v}_i\|
\;\leq\;
M^2 R.
\]
Because every derivative of a Jacobian block with respect to a vector variable yields at most one additional multiplication by a weight matrix, the second-order derivative tensor has entries that scale as \(M^3 R\). Summing across \(n-1\) possible pairwise terms contributes a linear dependence on \(n\). Therefore, the spectral norm of the difference between two Jacobians satisfies
\[
\big\|\nabla \Phi(\mathbf v) - \nabla \Phi(\mathbf v')\big\|
\;\leq\;
C_n M^3 R \, \big\|(\mathbf v_1,\dots,\mathbf v_n) - (\mathbf v'_1,\dots,\mathbf v'_n)\big\|,
\]
establishing the claimed Lipschitz bound.
\end{proof}

\end{document}